\documentclass[11pt, oneside]{Thesis} 

\graphicspath{{Figures/}} 

\usepackage[comma, sort&compress]{natbib} 
\usepackage{pdfpages}

\usepackage{amsmath}
\usepackage{lingmacros}
\usepackage{xcolor}
\usepackage{colortbl}
\usepackage{multirow}
\usepackage{bm}
\usepackage{diagbox}
\usepackage{subcaption}
\usepackage{pgfplots}
\usepackage{enumitem}

\usepackage{graphicx}
\usepackage{booktabs}
\usepackage{tabularx}
\usepackage{etoolbox,siunitx}
\usepackage{xspace}
\usepackage{xifthen}
\usepackage{makecell}
\usepackage{mathtools}
\usepackage{relsize}
\usepackage[stable]{footmisc}
\usepackage[linesnumbered,ruled,vlined]{algorithm2e}

\usepackage{dsttr}
\usepackage{cleveref}
\usepackage[T1]{fontenc}

\xspaceaddexceptions{\_}

\newcommand\colorhref[2]{\href{#2}{\color{#1}{\texttt{#2}}}}
\newcommand\colortexthref[3]{\href{#2}{\color{#1}{\texttt{#3}}}}

\robustify{\bfseries}

\hypersetup{urlcolor=black, colorlinks=true} 

\DeclarePairedDelimiterX{\infdivx}[2]{(}{)}{%
  #1\;\delimsize|\delimsize|\;#2%
}
\newcommand{\kld}[2]{\ensuremath{KL\infdivx{#1}{#2}}\xspace}

\DeclareMathOperator{\score}{score}
\DeclareMathOperator{\softmax}{softmax}
\DeclareMathOperator{\sigmoid}{sigmoid}

\SetCommentSty{commentfont}

\DeclarePairedDelimiter\braces{\lbrace}{\rbrace}
\DeclarePairedDelimiter\brackets{[}{]}

\newcommand\CUI{\textsc{CUI}\xspace}
\newcommand\CUIs{\textsc{CUI}s\xspace}
\newcommand\OOD{\textsc{OOD}\xspace}
\newcommand\IND{\textsc{IND}\xspace}

\newcommand\IVAs{\textsc{IVA}s\xspace}
\newcommand\ELIZA{\textsc{ELIZA}\xspace}
\newcommand\ASR{\textsc{ASR}\xspace}
\newcommand\TTS{\textsc{TTS}\xspace}
\newcommand\NLU{\textsc{NLU}\xspace}
\newcommand\NER{\textsc{NER}\xspace}
\newcommand\SLU{\textsc{SLU}\xspace}
\newcommand\DM{\textsc{DM}\xspace}
\newcommand\DST{\textsc{DST}\xspace}
\newcommand\NLG{\textsc{NLG}\xspace}
\newcommand\SL{\textsc{SL}\xspace}
\newcommand\RL{\textsc{RL}\xspace}
\newcommand\NLP{\textsc{NLP}\xspace}
\newcommand\HCI{\textsc{HCI}\xspace}
\newcommand\MRC{\textsc{MRC}\xspace}
\newcommand\QA{\textsc{QA}\xspace}
\newcommand\NMT{\textsc{NMT}\xspace}
\newcommand\CV{\textsc{CV}\xspace}

\newcommand\LM{\textsc{LM}\xspace}
\newcommand\MLM{\textsc{MLM}\xspace}
\newcommand\NSP{\textsc{NSP}\xspace}

\newcommand\MLE{\textsc{MLE}\xspace}
\newcommand\MSE{\textsc{MSE}\xspace}
\newcommand\NLL{\textsc{NLL}\xspace}
\newcommand\BPTT{\textsc{BPTT}\xspace}
\newcommand\ELBO{\textsc{ELBO}\xspace}
\newcommand\KL{\textsc{KL}\xspace}

\newcommand\HMMs{\textsc{HMM}s\xspace}
\newcommand\CRF{\textsc{CRF}\xspace}
\newcommand\CRFs{\textsc{CRF}s\xspace}

\newcommand\SVMs{\textsc{SVM}s\xspace}
\newcommand\LDA{\textsc{LDA}\xspace}
\newcommand\VW{\textsc{VW}\xspace}

\newcommand\MDP{\textsc{MDP}\xspace}
\newcommand\MDPs{\textsc{MDP}s\xspace}

\newcommand\POMDPs{\textsc{POMDP}s\xspace}
\newcommand\DP{\textsc{DP}\xspace}
\newcommand\TD{\textsc{TD}\xspace}
\newcommand\SARSA{\textsc{SARSA}\xspace}

\newcommand\ReLU{\textsc{ReLU}\xspace}
\newcommand\MLP{\textsc{MLP}\xspace}
\newcommand\MLPs{\textsc{MLP}s\xspace}
\newcommand\LSTM{\textsc{LSTM}\xspace}
\newcommand\LSTMs{\textsc{LSTM}s\xspace}
\newcommand\MTLSTM{\textsc{MT}-\LSTM}
\newcommand\GRU{\textsc{GRU}\xspace}
\newcommand\RNN{\textsc{RNN}\xspace}
\newcommand\RNNs{\textsc{RNN}s\xspace}
\newcommand\CNN{\textsc{CNN}\xspace}
\newcommand\CNNs{\textsc{CNN}s\xspace}
\renewcommand\AE{\textsc{AE}\xspace}
\newcommand\VAE{\textsc{VAE}\xspace}
\newcommand\VAEs{\textsc{VAE}s\xspace}
\newcommand\DIVAE{\textsc{DI}-\VAE}
\newcommand\DIVST{\textsc{DI}-\textsc{VST}\xspace}
\newcommand\LAED{\textsc{LAED}\xspace}

\newcommand\GANs{\textsc{GAN}s\xspace}
\newcommand\WordToVec{\textsc{Word2Vec}\xspace}
\newcommand\CBOW{\textsc{CBoW}\xspace}
\newcommand\GloVe{\textsc{GloVe}\xspace}
\newcommand\ELMo{\textsc{ELM}o\xspace}
\newcommand\ULMFiT{\textsc{ULMFiT}\xspace}
\newcommand{\GPT}[1][]{\textsc{GPT}\ifthenelse{\isempty{#1}}{}{-#1}\xspace}
\newcommand\DialoGPT{\textsc{Dialo}\GPT}
\newcommand\BERT{\textsc{BERT}\xspace}
\newcommand\RoBERTa{\textsc{RoBERT}a\xspace}
\newcommand\SpanBERT{\textsc{Span}\BERT}
\newcommand\DistilBERT{\textsc{Distil}\BERT}

\newcommand\TransferTransfo{\textsc{TransferTransfo}\xspace}

\newcommand\ImageNet{\textsc{ImageNet}\xspace}
\newcommand\ResNet{\textsc{ResNet}\xspace}
\newcommand\VGG{\textsc{VGG}\xspace}

\newcommand\SGD{\textsc{SGD}\xspace}

\newcommand\RNNLM{\textsc{RNNLM}\xspace}
\newcommand\SeqToSeq{\textsc{Seq2Seq}\xspace}
\newcommand\DSSM{\textsc{DSSM}\xspace}
\newcommand\QALSTM{\QA-\LSTM}
\newcommand\PSM{\textsc{PSM}\xspace}
\newcommand\PSMs{\textsc{PSM}s\xspace}

\newcommand\TFIDF{\textsc{TF-IDF}\xspace}
\newcommand\BM{\textsc{BM25}\xspace}

\newcommand\API{\textsc{API}\xspace}
\newcommand\APIs{\textsc{API}s\xspace}
\newcommand\KB{\textsc{KB}\xspace}
\newcommand\POS{\textsc{POS}\xspace}

\newcommand\KVRet{\textsc{KVRet}\xspace}
\newcommand\HCN{\textsc{HCN}\xspace}
\newcommand\HCNs{\textsc{HCN}s\xspace}
\newcommand\HHCN{\textsc{HHCN}\xspace}
\newcommand\VHCN{\textsc{VHCN}\xspace}
\newcommand\AEHCN{\AE-\HCN}

\newcommand{\memnn}[0]{\textsc{MemN2N}\xspace}
\newcommand{\dylan}[0]{\textsc{DyLan}\xspace}
\newcommand{\memnns}[0]{\textsc{MemN2N}s\xspace}
\newcommand\babble{\textsc{BABBLE}\xspace}
\newcommand\DS{\textsc{DS}\xspace}
\newcommand\TTR{\textsc{TTR}\xspace}
\newcommand\DSTTR{\DS-\TTR}
\newcommand\bAbI{bAbI\xspace}
\newcommand\bAbIplus{bAbI+\xspace}
\newcommand\SMD{\textsc{SMD}\xspace}
\newcommand\SWDA{\textsc{SWDA}\xspace}

\newcommand\DSTC[1][]{\mbox{\textsc{DSTC}\ifthenelse{\isempty{#1}}{}{-#1}}\xspace}
\newcommand\dstc[1][]{\mbox{\textsc{dstc}\ifthenelse{\isempty{#1}}{}{-#1}}\xspace}
\newcommand\ConvAI{\textsc{ConvAI2}\xspace}
\newcommand\metalwoz{\textsc{MetaLWOz}\xspace}
\newcommand\multiwoz{\textsc{MultiWOZ}\xspace}

\newcommand\METEOR{\textsc{METEOR}\xspace}
\newcommand\BLEU[1][]{\textsc{BLEU}\ifthenelse{\isempty{#1}}{}{-#1}\xspace}
\newcommand\CIDEr{\textsc{CIDE}r\xspace}
\newcommand\ROUGEL{\textsc{ROUGE-L}\xspace}
\newcommand\HRED{\textsc{HRED}\xspace}

\newcommand\NLGEval{\textsc{NLGE}val\xspace}
\newcommand\ZSDG{\textsc{ZSDG}\xspace}

\newcommand\WOz{\textsc{WOz}\xspace}
\newcommand\AMT{\textsc{AMT}\xspace}
\newcommand\ParlAI{\textsc{ParlAI}\xspace}

\newcommand\SPFT{\textsc{SP}+\textsc{FT}\xspace}
\newcommand{\diktnet}[0]{\textsc{DiKTNet}\xspace}
\newcommand\GRTr{\textsc{GRTr}\xspace}

\newcommand\OOV{\textsc{OOV}\xspace}

\title{\ttitle} 

\pgfplotsset{compat=1.7}

\begin{document}

\frontmatter 

\setstretch{1.3} 

\fancyhead{} 
\rhead{\thepage} 
\lhead{} 

\pagestyle{fancy} 

\newcommand{\HRule}{\rule{\linewidth}{0.5mm}} 

\hypersetup{pdftitle={\ttitle}}
\hypersetup{pdfsubject=\subjectname}
\hypersetup{pdfauthor=\authornames}
\hypersetup{pdfkeywords=\keywordnames}


\begin{titlepage}
\begin{center}

\textsc{\LARGE \univname}\\[1.5cm] 
\textsc{\Large PhD Thesis}\\[0.5cm] 

\HRule \\[0.4cm] 
{\huge \bfseries \ttitle}\\[0.4cm] 
\HRule \\[1.5cm] 
 
\begin{minipage}{0.4\textwidth}
\begin{flushleft} \large
\emph{Author:}\\
\href{http://shalyminov.com}{\authornames} 
\end{flushleft}
\end{minipage}
\begin{minipage}{0.4\textwidth}
\begin{flushright} \large
\emph{Supervisors:} \\
\href{https://www.hw.ac.uk/schools/mathematical-computer-sciences/staff-directory/oliver-lemon.htm}{\supname} 

\href{https://www.hw.ac.uk/schools/mathematical-computer-sciences/staff-directory/arash-eshghi.htm}{\secondsupname} 
\end{flushright}
\end{minipage}\\[3cm]
 
\large \textit{A thesis submitted in fulfilment of the requirements\\ for the degree of \degreename}\\[0.3cm] 
\textit{in the}\\[0.4cm]

\deptname\\[2cm] 
 
{\large \today}\\[1cm] 
\includegraphics[width=6cm]{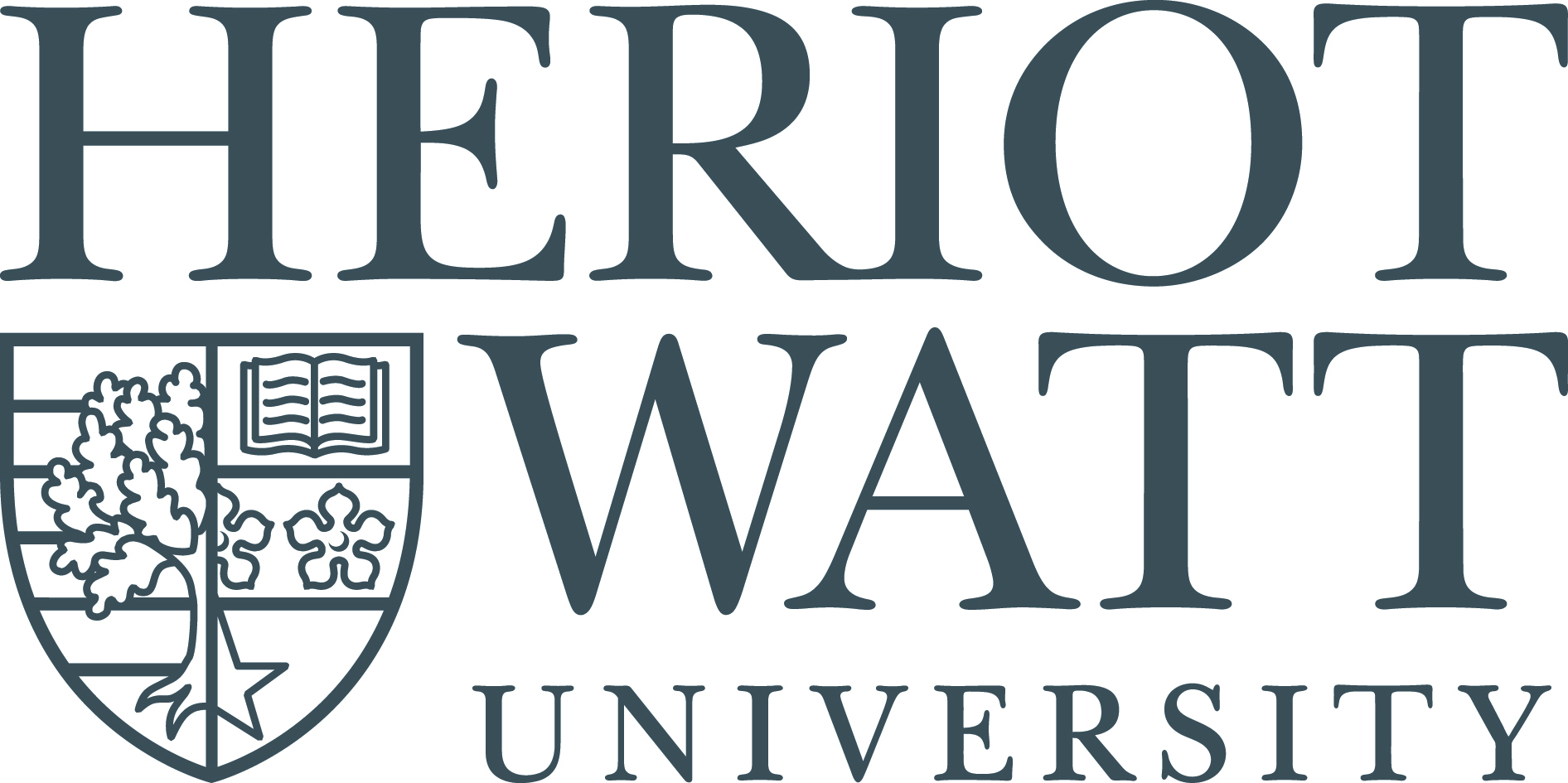} 

\vfill
\end{center}

\end{titlepage}


\Declaration{


I, \authornames, declare that this thesis titled, '\ttitle' and the work presented in it is my own. I confirm that this work submitted for assessment is my own and is
  expressed in my own words. Any uses made within it of the works of
  other authors in any form (e.g., ideas, equations, figures, text,
  tables, programs) are properly acknowledged at any point of their
  use. A list of the references employed is included.

 \vspace{2cm} 
Signed:\\
\rule[1em]{25em}{0.5pt} 
 
Date:\\
\rule[1em]{25em}{0.5pt} 
}

\clearpage 


\pagestyle{empty} 

\null\vfill 

\textit{This page is intentionally left blank.}


\vfill\vfill\vfill\vfill\vfill\vfill\null 

\clearpage 


\addtotoc{Abstract} 


{\huge{\textit{Abstract}} \par}

Conversational User Interface (\CUI) has become ubiquitous in everyday life, in consumer-focused
products like Siri and Alexa or more business-oriented customer support automation solutions.
Deep learning underlies many recent breakthroughs in dialogue systems but requires very large
amounts of training data, often annotated by experts~--- and this dramatically increases the cost of
deploying such systems in production setups and reduces their flexibility as software products.
Trained with smaller data, these methods end up severely lacking robustness to various phenomena of
spoken language (e.g. disfluencies), out-of-domain input, and often just have too little
generalisation power to other tasks and domains.

In this thesis, we address the above issues by introducing a series of methods for bootstrapping
robust dialogue systems from minimal data. Firstly, we study two orthogonal approaches to dialogue:
a linguistically informed model (\dylan) and a machine learning-based one (\memnn)~--- from the data
efficiency perspective, i.e. their potential to generalise from minimal data and robustness to
natural spontaneous input. We outline the steps to obtain data-efficient solutions with either
approach and proceed with the neural models for the rest of the thesis.

We then introduce the core contributions of this thesis, two data-efficient models for dialogue
response generation: the Dialogue Knowledge Transfer Network (\diktnet) based on transferable latent
dialogue representations, and the Generative-Retrieval Transformer (\GRTr) combining response
generation logic with a retrieval mechanism as the fallback. \GRTr ranked first at the Dialog System
Technology Challenge 8 Fast Domain Adaptation task.

Next, we the problem of training robust neural models from minimal data. As such, we look at
robustness to disfluencies and propose a multitask \LSTM-based model for domain-general disfluency
detection. We then go on to explore robustness to anomalous, or out-of-domain (\OOD) input. We
address this problem by (1) presenting Turn Dropout, a data-augmentation technique facilitating
training for anomalous input only using in-domain data, and (2) introducing \VHCN and \AEHCN,
autoencoder-augmented models for efficient training with turn dropout based on the Hybrid Code
Networks (\HCN) model family.

With all the above work addressing goal-oriented dialogue, our final contribution in this
thesis focuses on social dialogue where the main objective is maintaining natural, coherent, and
engaging conversation for as long as possible. We introduce a neural model for response ranking in
social conversation used in Alana, the 3rd place winner in the Amazon Alexa Prize 2017 and 2018. For
our model, we employ a novel technique of predicting the dialogue length as the main objective for
ranking. We show that this approach matches the performance of its counterpart based on the
conventional, human rating-based objective~--- and surpasses it given more raw dialogue transcripts,
thus reducing the dependence on costly and cumbersome dialogue annotations.

%

\clearpage 


\setstretch{1.3} 

\acknowledgements{
I am extremely grateful to my supervisors Prof Oliver Lemon and Dr Arash Eshghi: starting with the
very opportunity to join Heriot-Watt University that they gave me, I have had their continued
support throughout my PhD. Under their thoughtful advice, I was fortunate enough to have my efforts
constantly guided while being given the freedom to pursue things that I am really passionate about.

A major part of this thesis was done during my internships at Microsoft Research, and I want to
thank Dr Sungjin Lee (later with Amazon Alexa AI), Dr Hannes Schulz, Dr Alessandro Sordoni, and Adam
Atkinson for the excellent mentorship I was receiving from them, as well as for the fruitful
collaborations we have started together that are still ongoing.

I am very grateful to Dr Julian Hough for offering extensive consultation and constructive feedback
on our disfluency detection study in Ch. \ref{Chapter6}. A special thanks goes to
Dr Ondřej Dušek for his thoughtful advices on deep learning models at the very start of my PhD and
for being a great mentor and collaborator in our study of response ranking for social dialogue
in Ch. \ref{Chapter8}.

It was an honour to be part of the Amazon Alexa Prize team Alana and work alongside its members,
fellow PhD students: Shubham Agarwal, Amanda Cercas Curry, Ioannis Papaioannou, Jose Part,
Karin Sevegnani, Alessandro Suglia, Xinnuo Xu, Yanchao Yu (later an alumnus, PhD)~--- as well
as the faculty advisors: Prof Verena Rieser and Dr Ioannis Konstas.

I was extremely fortunate to be part of Heriot-Watt University's Interaction Lab with its
atmosphere of passion to change the ways people communicate with machines. I am thankful to the
iLab members and alumni with whom I had a pleasure to converse about conversation:
Prof Helen Hastie, Dr Emanuele Bastianelli, Dr Frank Broz, Dr Christian Dondrup,
Dr Ioannis Efstathiou, Dr David Howcroft, Dr Simon Keizer, Dr Xingkun Liu,
Dr Katrin Lohan, Dr José David Águas Lopes, Dr Jekaterina Novikova, Dr Andrea Vanzo,
Angus Addlesee, Miltiadis Marios Katsakioris, and Weronika Sieińska.

I express deep appreciation to Heriot-Watt University for supporting me with the James Watt
Scholarship which was vital for this work to happen. I acknowledge Alexa Prize 2017 and 2018 grants
from Amazon  which contributed to my conference travel funding. Research in Ch. \ref{Chapter3}
received financial support from the EPSRC project BABBLE (grant EP/M01553X/1).

A huge thanks goes to Louise Bingham, Tim Burns, Derek Davis, Christine McBride, Claire Porter,
and Peter Scott for helping me at the organisationally challenging times of my PhD.

Finally, I am endlessly grateful to my parents Viktor and Tamara, as well as my sister Elena for
their unconditional love and support throughout this journey.
}
\clearpage 



\pagestyle{fancy} 

\lhead{\emph{Contents}} 
\tableofcontents 

\lhead{\emph{List of Figures}} 
\listoffigures 

\lhead{\emph{List of Tables}} 
\listoftables 


\clearpage 

\setstretch{1.5} 

\lhead{\emph{Abbreviations}} 
\listofsymbols{ll} 
{
{\bf AE-HCN} & Autoencoder Hybrid Code Network \\
{\bf AMT} & Amazon Mechanical Turk \\
{\bf API} & Application Programming Interface \\
{\bf ASR} & Automatic Speech Recognition \\

{\bf BERT} & Bidirectional Encoder Representations from Transformers \\
{\bf BLEU} & Bilingual Evaluation Understudy \\
{\bf BPR} & Batch Prior Regularisation \\
{\bf BPTT} & Backpropagation Through Time \\

{\bf CIDEr} & Consensus-based Image Description Evaluation \\
{\bf CNN} & Convolutional Neural Network \\
{\bf CRF} & Conditional Random Field \\
{\bf CUI} & Conversational User Interface \\
{\bf CV} & Computer Vision \\

{\bf DiKTNet} & Dialogue Knowledge Transfer Network \\
{\bf DI-VAE} & Discrete Information Variational Autoencoder \\
{\bf DI-VST} & Discrete Information Variational Skip-Thought \\
{\bf DP} & Dynamic Programming \\
{\bf DS} & Dynamic Syntax \\
{\bf DSSM} & Deep Semantic Similarity Model \\
{\bf DST} & Dialogue State Tracker \\
{\bf DSTC} & Dialog State Tracking Challenge, Dialog System Technology Challenge \\

{\bf ELBO} & Evidence Lower Bound \\
{\bf ELMo} & Embeddings from Language Models \\

{\bf FT} & FastText \\

{\bf GAN} & Generative Adversarial Network \\
{\bf GloVe} & Global Vectors for Word Representation \\
{\bf GPT} & Generative Pretrained Transformer \\
{\bf GRTr} & Generative-Retrieval Transformer \\
{\bf GRU} & Gated Recurrent Unit \\

{\bf HCN} & Hybrid Code Network \\
{\bf HMM} & Hidden Markov Model \\
{\bf HRED} & Hierarchical Recurrent Encoder-Decoder \\

{\bf IDF} & Inverted Document Frequency \\
{\bf IVA} & Intelligent Virtual Assistant \\

{\bf KB} & Knowledge Base \\

{\bf OOD} & Out-of-Domain \\

{\bf VHCN} & Variational Hybrid Code Network \\

{\bf LAED} & Latent Action Encoder-Decoder \\
{\bf LDA} & Latent Dirichlet Allocation \\
{\bf LSTM} & Long Short-Term Memory \\

{\bf MDP} & Markov Decision Process \\
{\bf MemN2N} & End-to-End Memory NEtwork \\
{\bf MetaLWOz} & Meta-Learning Wizard-of-Oz \\
{\bf METEOR} & Metric for Evaluation of Translation with Explicit Ordering \\
{\bf MLE} & Maximum Likelihood Estimate \\
{\bf MLP} & Multilayer Perceptron \\
{\bf MLM} & Masked Language Modelling \\
{\bf MRC} & Machine Reading Comprehension \\
{\bf MSE} & Mean Squared Error \\

{\bf NER} & Named Entity Recognition \\
{\bf NLG} & Natural Language Generation \\
{\bf NLL} & Negative Log-Likelihood \\
{\bf NLP} & Natural Language Processing \\
{\bf NLTK} & Natural Language Toolkit \\
{\bf NLU} & Natural Language Understanding \\
{\bf NMT} & Neural Machine Translation \\
{\bf NP} & Noun Phrase \\
{\bf NSP} & Next Sentence Prediction \\

{\bf POMDP} & Partially Observable Markov Decision Process \\
{\bf POS} & Part of Speech \\
{\bf PP} & Prepositional Phrase \\

{\bf QA} & Question Answering \\

{\bf ReLU} & Rectified Linear Unit \\
{\bf RL} & Reinforcement Learning \\
{\bf RNN} & Recurrent Neural Network \\
{\bf RNNLM} & Recurrent Neural Network Language Model \\
{\bf ROUGE} & Recall-Oriented Understudy for Gisting Evaluation \\
{\bf RT} & Record Type \\

{\bf Seq2Seq} & Sequence-to-Sequence \\
{\bf SGD} & Stochastic Gradient Descent \\
{\bf SMD} & Stanford Multi-Domain (dataset)\\
{\bf SP} & SentencePiece \\
{\bf SVM} & Support Vector Machine \\
{\bf SWDA} & Switchboard Dialog Acts \\

{\bf TD} & Temporal Difference \\
{\bf TF} & Term Frequency \\
{\bf TTR} & Type Theory With Records \\
{\bf TTS} & Text-to-Speech \\

{\bf ULMFiT} & Universal Language Model Fine-Tuning \\

{\bf VAE} & Variational Autoencoder \\
{\bf VP} & Verb Phrase \\

{\bf WOZ} & Wizard-of-Oz \\

{\bf ZSDG} & Zero-Shot Dialogue Generation \\
}


\clearpage 





\clearpage 






\setstretch{1.3} 

\pagestyle{empty} 

\dedicatory{\textit{This page is intentionally left blank}} 

\addtocontents{toc}{\vspace{2em}} 


\mainmatter 

\pagestyle{fancy} 



\chapter{Introduction} 

\label{Chapter1} 

\lhead{Chapter 1. \emph{Introduction}} 


According to the \cite{merriam_webster_dialogue}, dialogue is ``a conversation between two or more
persons, also a similar exchange between a person and something else (such as a computer)''.
Conversation is the most versatile and efficient way for humans to communicate: a debate, a
negotiation, or a friendly chit-chat~--- dialogue is the main means for all kinds of interaction
between people. But it is not only other people whom one would want to interact with naturally:
communicating with machines in a similar intuitive way has been a major goal for researchers and
engineers around the world. And having past decades of Human-Computer-Interaction (\HCI) research
and several generations of human-computer interfaces, we are now on the brink of being able to
leverage the empowering potential of computers just by talking to them. Specifically, advances in
the fields of Speech Processing and Natural Language Processing (\NLP) made it possible to enable
interaction with an ever growing amount of devices and services using natural language~---
interfaces of this kind are called Conversational User Interfaces (\CUIs), or dialogue systems.

\begin{figure}
    \caption{Enterprise dialogue ecosystem\footnotemark}
    \centering
    \includegraphics[width=1.0\textwidth]{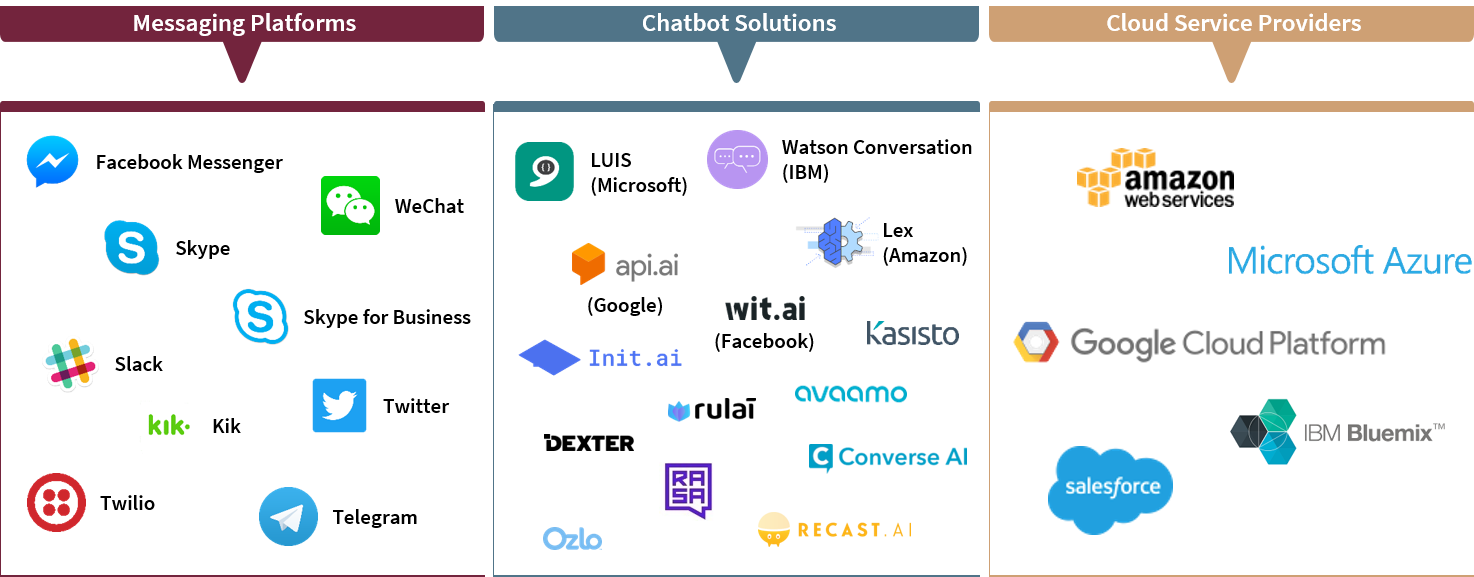}
    \label{fig:dialogue_ecosystem}
\end{figure}

Currently, dialogue systems are already ubiquitous in everyday life:

\begin{itemize}[label=---]
    \item \CUI is now enabled by most of the personal devices, e.g. Apple Siri, Amazon Alexa,
    Google Assistant, Microsoft Cortana, with their functionality constantly growing,
    \item Enterprises, especially in banking, healthcare, and retail spheres are deploying CUI
    solutions for automating call centres, customer support websites, and sales assistance at
    online marketplaces,
    \item Due to popular demand, a wide range of \CUI building platforms are now thriving on the
    market, e.g. Google Dialogflow (former Api.ai), Facebook Wit.ai, Rasa, Microsoft Bot Framework,
    Baidu DuerOS, Amazon Lex, and Apple Siri platforms for developers. See Figure
    \ref{fig:dialogue_ecosystem} for a visualisation of the variety of components and services in
    the ecosystem of dialogue solutions for enterprise.
\end{itemize}

The\footnotetext{Image source: {\tt seekpng.com}} market for Intelligent Virtual Assistants (\IVAs)
and related products was valued at \$2.2 billion in 2018 and expected to grow 10-fold by 2025 thus
reaching \$25.63 billion\footnotemark.
\footnotetext{Information from GrandView Research, accessed December 2019}

\section{Dialogue Systems (CUIs)}

Although \CUI gained wide adoption in the very recent years, conversational, or dialogue systems
have a long history. The first computer programs to support Natural Language interaction appeared
decades ago~--- the first one that gained wide public recognition was \ELIZA developed in 1966 by
Joseph Weizenbaum at MIT \citep{Weizenbaum66}. Programmed with a set of simple rules, it was tasked
to maintain dialogue with the user close to what can be expected at a psychotherapy session. The
system did not have any capabilities of ``understanding'' the user's words to any significant
degree~--- nor was it its goal. However, in its behaviour it appeared sufficiently close to a human,
and thus it was one of the first programs to attempt the Turing Test which evaluates machine's
ability to simulate human behaviour so that it cannot be distinguished from a real human
\citep{10.1093/mind/LIX.236.433}.

Since then, a lot of research and engineering effort has been put into creating conversational
systems of ever-increasing functionality~--- however early systems were constrained by the
limitations of the rule-based logic which was never robust and flexible enough for natural language.
This has changed with the spread of machine learning techniques in the NLP field. Based on
statistical analysis of real datasets rather than handcrafted rules, machine learning greatly
improved dialogue systems' robustness to the aspects of spoken language as well as enabled Automatic
Speech Recognition (\ASR) systems to attain a practical performance level. This led to the creation
of the first systems with voice interfaces~--- e.g. ESPIRIT SUNDIAL \citep{peckham-1991-speech} was
one of the first spoken dialogue systems: it worked over the phone line and provided air travel
information for the users. Nevertheless, its language coverage was not very wide, and apart from the
state-of-the art \ASR for that time, its dialogue behaviour was quite inflexible as it was based on
a set of rules. However, around that time dialogue systems started gaining interest in the
enterprise sphere due to their potential in optimising business processes.

Having attained sufficient performance with speech recognition (i.e. converting an utterance from
audio signal into its textual form), dialogue systems research was then focused on understanding the
more high-level concepts contained in the utterance, i.e. extracting \textit{slots} (types of
entities in the context of the dialogue task, e.g. `to', `from' for the taxi booking task) with
their values, as well as detecting the overall user's \textit{intent} in the utterance. This problem
is called Natural Language Understanding (\NLU) which we are going to cover in Section \ref{ch2:nlu}.
Machine learning approaches brought significant advances to \NLU, thus making conversational
interfaces able to execute a wide range of voice commands in different domains. But still, the \NLU
logic did not cover temporal structure of conversation and the presence of the dialogue context
reducing those interfaces to 1-shot interactions for the most part (some basic support of dialogue
was handled by handcrafted rules).

The direction towards bringing full dialogue support to \CUIs started with the introduction of
Dialogue Manager (\DM), a component in the dialogue system architecture
\citep{2001:TCH:567363.567365} maintaining the conversational context in the form of a set of slots
and their values provided by the user so far and taking the next dialogue action based on it (DM
will be discussed in detail in Section \ref{ch2:dst}). Furthermore, with the highly time-distributed
nature of the conversation and extremely sparse feedback (success or failure of a dialogue can only
be determined upon the end of the conversation), Supervised Learning (\SL) techniques~--- i.e.,
those requiring reference output for each input to train the model~--- which were normally used for
\ASR or \NLU could not be directly applied to dialogue management, and the attention of the research
community shifted to Reinforcement Learning (\RL) methods (\citealp{DBLP:series/tanlp/RieserL11};
\citealp{DBLP:journals/csl/WilliamsY07})~--- those representing the model as and \textit{agent} and
training it from interactions with an \textit{environment}, simulated or real-world one. However,
due to the problems of scalability of early \RL models and their optimisation techniques to the
real-world problems, as well as the need for an excess amount of natural interactions for better
training performance, Reinforcement Learning was mainly seen in academic research and
proof-of-concept projects.

With the latest revolution in machine learning~--- the availability of massive datasets and the
computational power to process them~--- a technique called deep learning caused a major shift in
dialogue systems research, as well as the rest of the NLP field. As such, neural networks with
multiple layers of neurons (thus \textit{deep} models) trained in a unified way based on the
backpropagation algorithm \citep{10.5555/65669.104451} were able to learn complex patterns in raw
data, their performance increasing with the amount of training examples. Specifically, one notable
deep learning-based breakthrough was the neural conversation model of
\cite{DBLP:journals/corr/VinyalsL15}~--- a dialogue system with a single underlying neural model
trained from raw transcripts, without the use of any explicit complex features or domain knowledge.
Existing approaches to dialogue system components originally powered with `classic' \SL and \RL
algorithms also benefited from incorporating deep learning techniques
(\citealp{henderson-etal-2014-word}; \citealp{wen-etal-2015-semantically};
\citealp{DBLP:conf/iwsds/Cuayahuitl16,Li16_2}).
The complex patterns in the data that deep learning models were able to learn~--- e.g. properties of
words, sentences, and paragraphs~--- largely replaced the need for modular architectures, by
efficiently learning all the steps of conversation in a unified, \textit{end-to-end} fashion
\citep{Serban:2016:BED:3016387.3016435}.

Data has therefore become key in dialogue systems development, and the means to collect high-quality
real dialogue examples came into focus \citep{DBLP:conf/emnlp/MillerFBBFLPW17}. However, this
extremely high data consumption makes deep learning models not flexible enough for the growth pace
of the field~--- particularly limiting is the human effort in data collection and manual annotation
which is both time-consuming and expensive. Moreover, these two steps have to be re-visited every
time the product requirements get corrected or new functionality is desired.
Therefore, it is now of key priority to develop methods for training robust and practical neural
dialogue models with small amounts of data~--- e.g. a few example dialogues. This will keep dialogue
systems being highly maintainable software products, and the corresponding development cycles rapid.

\section{The Need for Data-Efficient Dialogue Models}

Recent advances in dialogue modelling resulted in a massive industrial adoption of deep learning
techniques. Initially originated as open-ended academic research and trained/evaluated on static
large datasets, those techniques have to undergo a substantial adaptation in order to fit the
industrial demands.

The first drastic difference between experimental testbeds and real-world software products is that
the latter are much more dynamic and flexible. In order to stay up to date with new feature requests
and tweaks for better user experience, any well-maintained product is continuously modified
throughout its lifecycle. Having the core components trained directly on massive datasets, while
giving them a certain level of coverage and making them generalisable to some extent, leaves the
resulting models static and inflexible to the variety of target domains and usage aspects of the
end products~--- ultimately resulting in insufficient maintainability of the latter. Apart from that,
requires sensible timespans required by the large-scale training restrict the models' fitness for
fast-paced development cycles. Lastly, the amount of annotations required by large-scale supervised
training results in a significant financial overhead of such development strategy as well.

The above concerns can be mitigated by adopting the \textit{data-efficient} approach to the
development of neural models, i.e. enabling training from a limited amount of \textit{seed} data.
Data-efficient training assumes a series of specific problems to be solved~--- for dialogue systems,
those are as follows. Firstly, given that deep learning methods are greatly prone to overfitting,
it is of key importance for the model to attain a sufficient generalisation level outside its seed
dataset. Secondly, there is a major mismatch between the specifics of written language of openly
available data used for large-scale training of \NLP models (e.g. internet news articles, Wikipedia
documents or books) and spoken language of dialogue. The mismatch spans from differences in
vocabulary and syntax to various incremental phenomena of spontaneous speech like pauses,
self-corrections, restarts and other disfluencies (\citealp{Hough14};
\citealp{DBLP:journals/topics/HealeyMEH18}). Under the end-to-end deep learning framework, the
conventional solution to that would be to have all the relevant speech phenomena covered in the
training data, but for low-resource training, this assumption does not hold. Finally, every dialogue
system, especially the ones that provide a conversational interface to an underlying Application
Programming Interface (\API) or service, has the boundaries of its domain, and when working with a
large user audience, it often happens that the system gets unexpected, or anomalous input outside
its designated domain. Given that most such systems are trained for maximum accuracy within the
domain, it is important to have a means to guarantee predictable system's behaviour outside it. And
expanding the training dataset with out-of-domain data in order to train for coverage is hardly
possible since all the variety of different domains and anomalous cases is very challenging to list.
It is especially critical for the systems trained from minimal data, where any additional data
collection comes at an especially high cost.

\section{Problems Addressed in the Thesis}

As outlined above, in this thesis we are going to address the following problems:

\begin{enumerate}[label=\textbf{\arabic*.}]
    \item {\bf Dependence on large amounts of training data with annotations.} We are going to
    develop methods enabling the training of goal-oriented and chat-oriented dialogue systems
    (as well as their key components) with minimum human effort in terms of training data
    collection and annotation;
    \item {\bf The lack of coverage of the diverse spoken language phenomena.} Conversational corpora
    collected specifically for dialogue model development do not normally represent the aspects of
    spontaneous spoken language, therefore the models trained on those are not ready for immediate
    usage in real-world settings. We will investigate ways to improve their generalisation potential
    to such input without the increase in the amount of the required training data;
    \item {\bf Insufficient robustness of neural dialogue models to out-of-domain input.}
    As is the case with training from small data, neural dialogue models are prone to overfitting
    to the training datasets and thus often lack robustness to anomalous out-of-domain user input
    which leads to unexpected dialogue behaviour and thus considerably limits the usage of such
    models in mission-critical production environments. We are going to improve the dialogue
    systems' robustness to out-of-domain input~--- again, without the increase in the amount of
    training data, i.e. only using the available in-domain data.
\end{enumerate}

\section{Contributions and Thesis Structure}

The above problem statement defines the structure of this work: the thesis consists of a series of
contributions towards practical data efficiency of machine learning-based dialogue systems.
The thesis is organised as follows.

In Chapter \ref{Chapter2} following the Introduction, we give a brief overview of dialogue system
architectures~--- particularly, how the conventional modular architecture was transformed into the
purely data-driven architecture under the influence of deep learning methods. We cover the two major
types of purely data-driven architectures: response generation systems and response retrieval
systems. We then outline the current state of transfer learning which is the basis for
data-efficient methods in machine learning overall. After that, we give a brief background on
linguistically informed models of dialogue~--- specifically those based on the dialogue grammars.
We then discuss the problems of robustness of dialogue systems~--- especially those trained from
minimal data~--- to the unseen phenomena in the input, i.e. spoken disfluencies and out-of-domain
user queries. We conclude the chapter with a discussion of the dialogue datasets widely used in the 
field, as well as the aspects of dialogue data collection.

In Chapter \ref{Chapter3}, we conduct a motivational study of two fundamentally different
approaches to dialogue: a linguistically informed model based on the semantic parser \dylan and a
neural response retrieval model End-to-End Memory Network (\memnn)~--- specifically, how they work
in the setup of (1) natural input data containing spoken disfluencies and (2) limited training data.
For (1), we introduce \bAbIplus, an extension to Facebook AI Research's bAbI Dialog Task 1 dataset
augmented with spoken disfluencies. We outline the possible steps for obtaining the practical
data-efficient solutions with either type of models. The work in this chapter was published at EMNLP
2017 \citep[][the author's contribution is the implementation of the Memory Network model, the
\bAbIplus data augmentation technique together with the corresponding dataset, the semantic user
simulator, and the design and execution of the experiments]{DBLP:conf/emnlp/EshghiSL17} and SemDial
2017 \citep{Shalyminov.etal17}.

In Chapter \ref{Chapter4}, we focus on the problem of bootstrapping a goal-oriented dialogue system
from minimal data, i.e. in a \textit{few-shot} setup. Using the intuition that goal-oriented
dialogue has a unified, domain-agnostic latent structure (e.g. sequences of dialogue acts which are
sequences of characteristic words and phrases), we introduce the Dialogue Knowledge Transfer
Network (\diktnet), a model that represents that in the form of latent Variational Autoencoder codes
learned from a large multi-domain source of dialogues, \metalwoz.
The model outperforms the previous best approach on its target Stanford Multi-Domain (\SMD) dataset.
The work in this chapter was published at EMNLP 2019 \citep{DBLP:conf/emnlp/ShalyminovLEL19} and
SigDial 2019 \citep{shalyminov-EtAl:2019:W19-59}.

In Chapter \ref{Chapter5}, we continue with the problem of training dialogue systems from minimal
data as part of the Dialog System Technology Challenge (\DSTC) 8 Fast Domain Adaptation task.
We propose the hybrid Generative-Retrieval Transformer (\GRTr). The model maintains a high
diversity level by using sampling-based decoding and alternates between generation and retrieval
from the support sets, part of the \DSTC[8] data split. \GRTr is the winning entry at the \DSTC[8]
Fast Domain Adaptation task as evaluated with human judges. The work in this chapter was published
at the DSTC8@AAAI 2020 \citep{DBLP:journals/corr/abs-2003-01680} and ICASSP 2020 \citep{9053599}.

In Chapter \ref{Chapter6}, we focus on the robustness of dialogue models to real-world variations of
the user's input~--- specifically, spoken disfluencies (e.g. pauses, restarts, self-corrections). We
propose a multitask \LSTM-based model for domain-general disfluency detection improving upon the
previous best model for incremental disfluency detection on open-domain Switchboard Dialog Acts
(\SWDA) dataset and showing superior generalisation to \bAbIplus. The work was presented at SemDial
2018 \citep{Shalyminov.etal18}.

In Chapter \ref{Chapter7}, we go on to explore robustness to a different kind of anomalous input~---
\textit{out-of-domain} (\OOD) utterances. In the low-resource setting, it is cumbersome to collect
real \OOD data for robustness training because of the great variety of possible domains and tasks.
We address this problem by (1) presenting turn dropout, a data-augmentation technique allowing
training for anomalous input only using in-domain data, and (2) introducing \AEHCN, a Hybrid Code
Network-based dialogue management model with an autoencoder for robust training using turn dropout.
\AEHCN improves upon the standard \HCN on \bAbI Task 6 (\DSTC[2]) and the Google Multi-Domain
Dialogues Dataset. The work from this chapter was presented at CAI@NIPS 2018
\citep{DBLP:journals/corr/abs-1811-12148} and ICASSP 2019 \citep[][the author's contribution is
the \bAbIplus data augmentation procedure and the augmented \bAbI Task 6 dataset, as well as the
turn dropout training technique]{LeeShalyminov2019}.

In Chapter \ref{Chapter8}, we turn our attention to open-domain chat-oriented dialogue and its
data-efficiency aspects. Our problem here is response ranking in an open-domain ensemble-based
dialogue system, which competed in the Amazon Alexa Prize. We introduce the neural
response ranker used in Alana, the 3rd place winner in the Amazon Alexa Prize 2017 and 2018.
We explore two alternative supervision signals, dialogue rating and length, and show that the
length-based model matches the performance of its rating-based counterpart and surpasses it given
more unannotated training data, thus reducing the dependence on costly and cumbersome dialogue
annotations. The work in this chapter was published at SCAI@EMNLP 2018
\citep{DBLP:conf/emnlp/ShalyminovDL18} and Amazon Alexa Prize
Proceedings \citep[][the author's contribution is the neural ranking model and the data-efficient
response ranking study]{Curry.etal2018}.

Having introduced a series of data-efficient methods for neural dialogue systems, in Chapter
\ref{Chapter9} we conclude the thesis with an outline of future work left to be done in order to
facilitate the adoption of these techniques in large-scale \CUIs.

\chapter{Background and Motivation}

\label{Chapter2} 

\lhead{Chapter 2. \emph{Background and Motivation}} 

Dialogue systems can be categorised by their purpose (goal-oriented and chat-oriented), modality of
interaction (spoken, text-based, visual, multimodal) or the underlying model architecture
(rule-based, data-driven), but on a high level they all have a similar structure. Over the years of
research and development, this structure has undergone a series of transformations caused by the key
breakthroughs affecting dialogue modelling. The most recent and influential of these transformations
came with the adoption of machine learning techniques, especially deep learning. Following is a
description of the dialogue system architecture and how it changed having become machine
learning-centric. The structure of this chapter is as follows. Firstly, we are going to discuss the
conventional dialogue system architecture (Section \ref{ch2:conventional}) and how it then
transformed into the fully data-driven architectures (Section \ref{ch2:data_driven}). Specifically,
we will cover the two principal approaches that fully data-driven systems follow, i.e.
\textit{response retrieval} models (Section \ref{ch2:retrieval}) and \textit{response generation}
models (Section \ref{ch2:generation}), with a subsequent overview of the key techniques defining
the most widely-used advances in dialogue response generation (Section \ref{ch2:gen_techniques}). In
Section \ref{ch2:transfer_learning}, we will give a background on transfer learning, a
general-purpose machine learning technique lying in the core of practical data efficiency, and the
corresponding models for dialogue and NLP tasks. Having briefly covered the state of machine
learning-based dialogue modelling, we then describe an alternative, linguistically informed approach
to dialogue in Section \ref{ch2:linguistic}. In Section \ref{ch2:generalisation_robustness}, we will
discuss the generalisation and robustness issues of the systems trained from minimal data. As such,
we will cover robustness to disfluencies in the spoken language~--- and how proper processing of
those phenomena is necessary for dialogue understanding. We will then discuss the problem of
robustness to \textit{out-of-domain} (\OOD) input, i.e. user queries that a closed-domain dialogue
system is unable to process correctly~--- this is especially important in the settings where no real
\OOD data is available during training. We will conclude the chapter with Section
\ref{ch2:data_collection}, a brief overview of dialogue datasets and how data collection stage is
integrated into the dialogue system pipelines and frameworks.

\section{Conventional Dialogue System Architecture}
\label{ch2:conventional}

\begin{figure}
  \centering
  \includegraphics[width=1.0\textwidth]{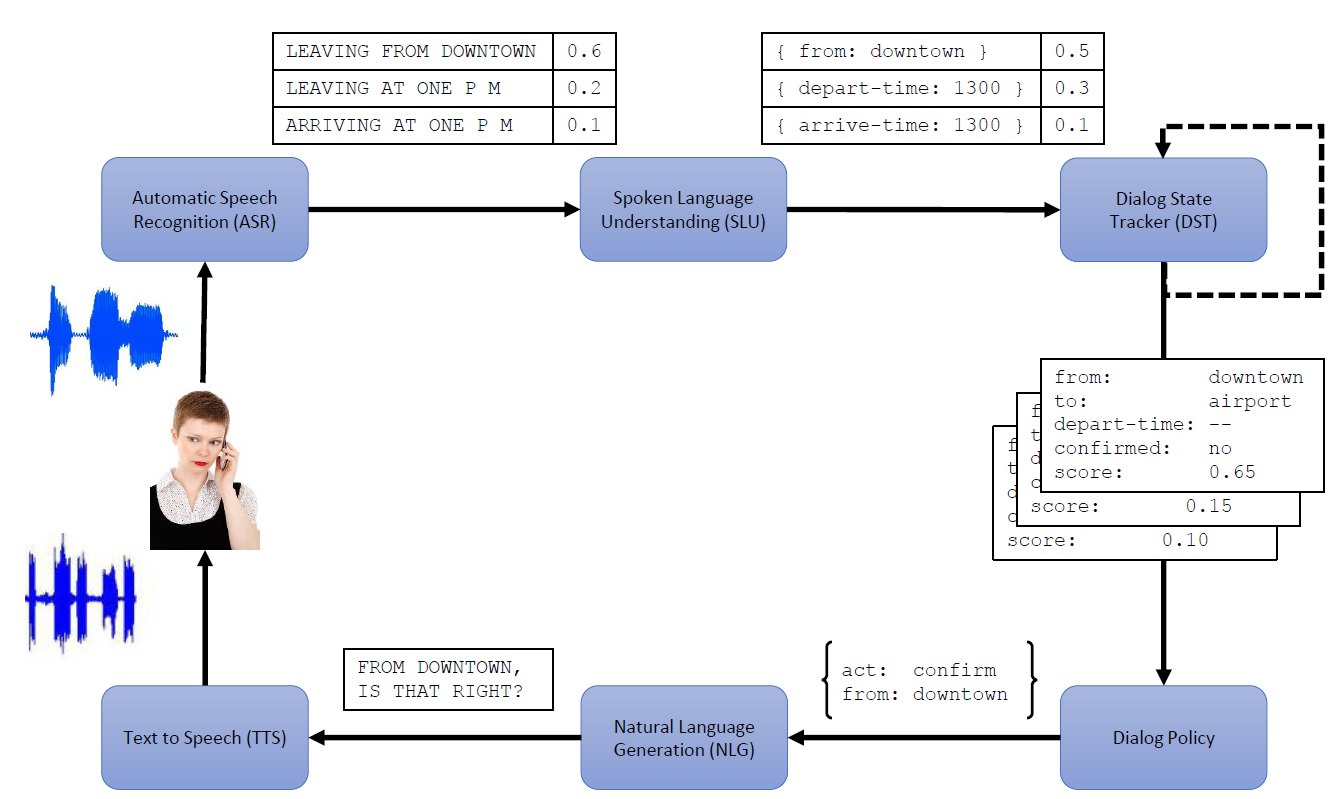}
  \caption[Conventional dialogue system architecture]{Conventional dialogue system architecture
  \citep{DBLP:journals/dad/WilliamsRH16a}}
  \label{fig:architecture}
\end{figure}

Maintaining fluent and coherent dialogue assumes having solved a series of underlying problems in
speech recognition/synthesis, language understanding, and action planning.
The dialogue system architecture that emerged historically reflects that in its modular structure.
It is normally referred to as the \textit{conventional architecture}
(\citealp{DBLP:conf/interspeech/Young10}; \citealp{DBLP:journals/dad/WilliamsRH16a}) and is shown in
Figure \ref{fig:architecture}.

The conventional architecture covers the widest range of dialogue systems on a high level, including
spoken and text-based ones~--- as such, it contains 2 components specific to voice-based
interaction, namely \ASR decoding audio signals from the microphone into text, and Text-to-Speech
(\TTS), synthesising the sound from the system's textual response. Advances in these two systems are
key to the recent wide spread of personal voice assistants like Apple Siri, Amazon Alexa, and Google
Assistant~--- and the consequent development of a large-scale market for those systems. However,
audio signal processing is out of this thesis's scope. In this work, we will be focusing on the
`core' dialogue system logic that works with the user's input in the textual form~--- either typed
in directly or already decoded by the \ASR, and produces a textual response as its final output, for
displaying it on the screen or feeding it into the \TTS. Throughout the description of the
conventional architecture, we'll be mainly talking about goal-oriented dialogue which it corresponds
to most. Later in Chapter \ref{Chapter8}, we'll turn to open-domain chat-oriented dialogue as part
of discussing the more recent versions of the architecture. 

\subsection{Natural Language Understanding (\NLU)}
\label{ch2:nlu}

\NLU (also referred to as Spoken Language Understanding, or \SLU) is the first subsystem of
the conventional pipeline whose function is the extraction of relevant information from the user's
input and incorporating it into the system's internal state. \NLU performs this extraction on the
turn level, i.e. from a single user's utterance.

Historically, dialogue follows the \textit{frame semantics} convention for formally describing
meaning (\citealp{dinarelli-etal-2009-annotating} provide a comprehensive up-to-date example).
Under this notation, every situation is represented as an attribute-value frame (or form).
In dialogue, a form describes a specific user's goal (also referred to as \textit{intent}), e.g.
booking a flight or searching for a restaurant by the attributes (called slots).
For the flight booking task used in the diagram, the complete form contains \texttt{from},
\texttt{to}, \texttt{depart-time}, and \texttt{confirmed} slots. Therefore, goal-oriented dialogue
can be represented as the form-filling process, with \NLU responsible for extracting information
relevant to it from a single utterance. Named Entity Recognition (\NER), usually considered as a
general \NLP task, can be performed by the \NLU as well. \NER addresses the extraction of a
domain-agnostic set of entities such as persons, organisations, locations or timestamps~--- see for
example the approach of \cite{DBLP:conf/acl/FinkelGM05}. Finally, as dialogue systems can handle
multiple user intents with the corresponding set of tasks (e.g. `set alarm', `put on music',
`search web', `chit-chat'), another type of processing performed at this stage is user intent
detection. Intent detection is especially important since task-specific dialogue logic can be
implemented as a completely independent subsystem~--- e.g. as of early 2020, Amazon Alexa contains
more than 80,000 skills\footnote{Information from Voicebot.ai} (dialogue `applications' within the
Alexa platform) which are implemented mostly by the independent developers.

Slot value extraction is the problem that was traditionally approached using a linguistically
informed method, i.e. semantic parsing relying on large-scale grammars built by linguists~--- e.g.
the CMU Phoenix system consisted of about 13,000 rules (\citealp{tur2011spoken};
\citealp{10.5555/1170742.1170864}). More recently though, the slot value extraction task shifted to
the machine learning framework and was treated as a sequence labelling problem. Thus, `classic'
machine learning approaches like Hidden Markov Models (\HMMs, \citealp{wang2005spoken};
\citealp{young_hvs}), Conditional Random Fields (\CRFs, \citealp{laffertyCrf};
\citealp{10.3115/1073445.1073473}), and Support Vector Machines (\SVMs,
\citealp{Mairesse09spokenlanguage}) were used in \NLU predominantly. User intent detection was
also initially approached using heuristic methods like keyword detection or regular expression match,
but later was treated as a classification task, with the corresponding classification models like
those mentioned above or efficient methods of ensembling them, e.g. \textit{boosting}
\citep{boostexter}.

Among the more recent approaches to \NLU, \cite{DBLP:conf/acl/ENCS19} proposed a neural model for
joint slot-value extraction and intent detection. The part of the model predicting slot values is a
combination of a Long Short-Term Memory cell (\LSTM, \citealp{DBLP:journals/neco/HochreiterS97})
producing the latent states and a \CRF on top making the actual predictions, which itself has been
widely used recently for sequence tagging tasks.

\subsection{Dialogue State Tracking (\DST)}
\label{ch2:dst}

The \DST subsystem incorporates the information obtained from the NLU into the system's internal
state (usually based on the form being filled) which it maintains throughout the dialogue. In early
approaches, \DST was implemented as a finite state machine maintaining the form with the new partial
results coming from the \NLU, and resolving any conflicts and updates caused by user's intention
change or the overall ambiguity of the dialogue process. However, the actual degree of ambiguity in
the real-world dialogue and fluidity of conversation with real users also made \DST research shift to
machine learning-based methods. As such, the annual Dialog State Tracking Challenge (\DSTC)
\citep{DBLP:journals/dad/WilliamsRH16a} was organised with the goal of advancing state-of-the-art in
dialogue research. A number of approaches for \DST emerged from \DSTC: Recurrent Neural Networks
(\RNNs) for incremental word-by-word tracking (\citealp{Henderson14}; \citealp{ZilkaJurcicek15}),
\RNN-based domain-adaptive state tracking \citep{DBLP:conf/acl/MrksicSTGSVWY15}, and Convolutional
Neural Networks (\CNNs) for multi-language tracking \citep{Shi16}. Since the 6th edition, with the
expansion of the overall dialogue systems area, \DSTC became known as Dialog System Technology
Challenge, with the corresponding widening of its coverage to several tracks (e.g. `Multi-Domain
Task Completion Challenge', `Noetic End-to-End Response Selection', and `Visual Scene-Aware
Dialogue'). As observed in \DSTC[8] results, recent state-of-the-art approaches to DST tend to
combine state tracking with the techniques from Machine Reading Comprehension, (\MRC)
\citep{DBLP:journals/corr/abs-1912-09297} for better out-of-domain/zero-shot performance.

\subsection{Dialogue Policy}

Policy is the key component in the conventional architecture: given the accumulated system state by
the \DST, it makes a decision of which action to take next. Actions can be interlocutory (e.g.
confirmation, information request, greeting) or functional, e.g. calls to the underlying APIs or
databases. The dialogue policy is often combined with the state tracker into the Dialogue
Manager (\DM) component, and in the early approaches, \DM was implemented as a finite state machine
whose logic was to scan through the dialogue form and enquire about the next unfilled slot or, when
all the necessary information is collected, issue an API call or switch to the corresponding
business logic.

At later stages though, it became clear that this deterministic process is not adequate to the
complexity of real-world conversations, and the following two assumptions were made: (1) the
information in the dialogue state is not perfectly certain nor complete, and (2) the dialogue policy
should optimise an objective function defining the overall dialogue success. Thus, a policy
essentially was then considered as a planning under uncertainty problem. In goal-oriented dialogue,
the planning objective is normally the number of turns taken to reach the user's goal, whereas in
open-domain chat-oriented dialogue it could be formulated via the user's engagement during the
conversation and the overall satisfaction at the end of it \citep{ram_conversational_2017}.

\begin{figure}
  \centering
  \includegraphics[width=0.6\textwidth]{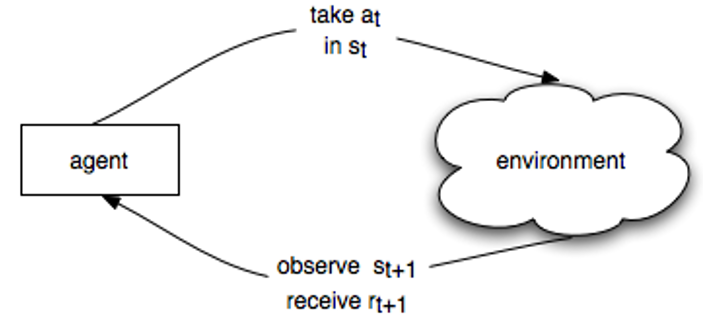}
  \caption[Dialogue as a Markov Decision Process]{Dialogue as a Markov Decision Process
  \citep{DBLP:series/tanlp/RieserL11}}
  \label{fig:dialogue_mdp}
\end{figure}

In planning under uncertainty, the problems distributed in time and with a sparse delayed reward
signal are modelled as Markov Decision Processes (\MDPs, \citealp{DBLP:series/tanlp/RieserL11}; 
\citealp{DBLP:journals/jair/SinghLKW02}) or Partially Observable \MDPs (\POMDPs,
\citealp{Young13}; \citealp{DBLP:journals/csl/WilliamsY07}), with RL as the optimisation framework.
Naturally, an \MDP is defined as a tuple $<S, A, \mathcal{T}, \mathcal{R}>$~--- the state space, the
action space, the transition function between states, and the reward function. The
dialogue system thus becomes an \textit{agent} communicating with the \textit{environment} (user) by
making observations, performing \textit{actions} and receiving the corresponding \textit{rewards}.
The form maintained by the \DST becomes the agent's internal state, and the agent's goal is to go
from the initial state (empty form) to the final state (form filled out, API call issued,
information presented to the user) via an optimal (or sufficiently close to such) series of actions
defined by the learned policy~--- the procedure is visualised in Figure \ref{fig:dialogue_mdp}.

In turn, the \RL framework accounts for the time-distributed nature of the dialogue as well as
sparse, delayed reward. Specifically, in \RL we operate on the \textit{value} of each state~--- that
is, the expected cumulative (final) reward that the agent will get starting at a certain state $s$
at the timestep $t$ and following the optimal policy $\pi$:

\begin{equation}
  V^\pi(s) = \mathbb{E}_\pi (\mathcal{R} \mid s_t = s)
\end{equation}

The values for each state are obtained via solving Bellman equations
\citep{DBLP:journals/tit/BellmanK57} in a Dynamic Programming (\DP) way, or by using
simulation-based methods, e.g. Temporal Difference (\TD). TD-based methods such as Q-learning or
\SARSA are more widely used as they do not require knowing the dynamics of the model, i.e. all the
state transition probabilities.

Successful application of \TD learning assumes having the source of interactions for the agent, e.g.
a simulated environment or an embodiment for real-world interaction. In some cases, especially
videogames (\citealp{Mnih13}; \citealp{DBLP:journals/corr/abs-1912-06680}), it is feasible to obtain
lots of real-world interactions to run a sufficient amount of training episodes. In dialogue however,
such data may only come from the interactions with real users which requires an existing deployment
of a prototype system. Therefore, a lot of research effort was made to develop efficient user
simulators for `bootstrapping' the initial policy (\citealp{DBLP:series/tanlp/RieserL11};
\citealp{DBLP:conf/emnlp/ShiQWY19}), but in practice, training an efficient simulator is as hard as
training the final dialogue system.

More recent approaches to Dialogue Management use supervised learning techniques such as Recurrent
Neural Network-based (\RNN) Hybrid Code Networks, or \HCNs \citep{DBLP:conf/acl/WilliamsAZ17} and
End-to-End Memory Networks, or \memnns (\citealp{Sukhbaatar15}; \citealp{DBLP:conf/iclr/BordesBW17}):
trained with long enough context, these models approximate long-term optimal behaviour well enough
for practical usability in relatively short conversations. 

\begin{figure}
  \centering
  \includegraphics[width=1.0\textwidth]{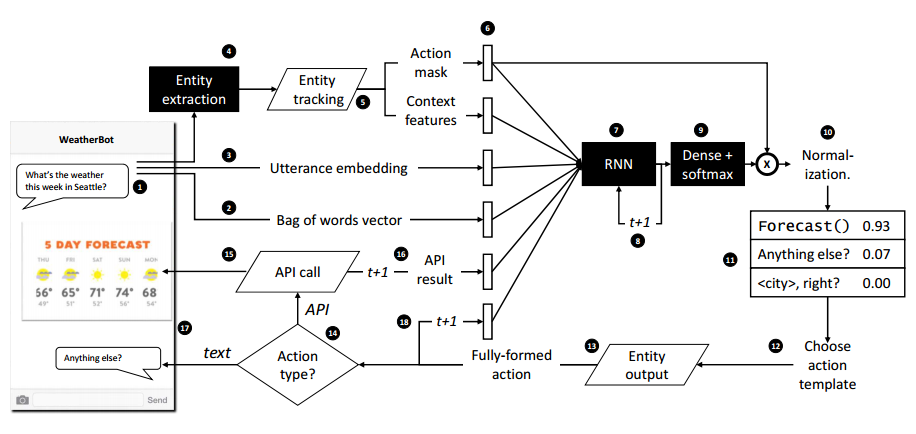}
  \caption[Hybrid Code Network architecture]{Hybrid Code Network architecture
  \citep{DBLP:conf/acl/WilliamsAZ17}. The black parts are trainable components (with entity
  extraction being an independent NLU engine); trapezoids denote the parts of in-domain logic
  implemented by the engineers.}
  \label{fig:hcn}
\end{figure}

The Hybrid Code Network is shown in Figure \ref{fig:hcn}~--- it is a recent example of a neural
network-based dialogue management model. The overall system architecture follows the modular
approach as it has separate components for entity extraction (for which it used the LUIS system by
\citealp{williams2015fast}) and template-based Natural Language Generation (we'll cover this
subsystem in the next section). \HCN only focuses on state tracking and action selection, and it is
aimed at training from minimal amounts of data for the direct use in products. To ensure the
stability in real-world interactions in the setting of minimal training data, \HCN introduces the
concept of \textit{action masks} which may be considered as expert rules incorporated into a machine
learning-based model. Action masks, the binary masks applied at the \DM's output, prohibit the
system from issuing infeasible actions at the critical points in the dialogue, e.g. issuing an API
call of bank transfer before confirming the recipient account with the user. These actions have to
be hand-crafted by the domain experts and incorporated into the training pipeline. Overall, this
hybrid architecture allows to alternate the emphasis between the tight, handcrafted control over the
system's behaviour via action masks and the flexibility of learning from examples. The authors
presented a 2-stage approach to train an \HCN: at the first stage, the model learns to mimic the
training dialogue examples in a supervised way, and at the second stage, the system can be further
fine-tuned autonomously, from the interactions with the users in a reinforcement learning fashion.
For that, the authors used a \textit{policy gradient} approach \citep{DBLP:journals/ml/Williams92}
with the following gradient update:


\begin{equation}
  w \leftarrow w + \alpha \left( \sum_t \nabla_w \log \pi (a_t \mid h_t; w) \right) (G - b)
\end{equation}

In the above formula, the gradient is applied to the policy \LSTM $\pi$ producing a distribution
over actions $a$ at the timestep $t$ given the dialogue history $h_t$. The error is the difference
between $G$, the return of the dialogue (the expected discounted sum of rewards), and $b$, baseline
average return set heuristically. This switch from supervised to reinforcement learning (referred to
as \textit{continuous learning}) of the same exact model proved to be the most widely-used way to
incorporate reinforcement learning into the dialogue system training setups
(\citealp{DBLP:conf/sigdial/SuBUGY17}; \citealp{DBLP:journals/corr/SuGMRUVWY16a}).

\HCN lies in the core of Microsoft Conversation Learner \citep{DBLP:conf/acl/ShuklaLSKLMPPG20}, a
tool for rapid prototyping of goal-oriented dialogue systems from example conversations which
assumes data-efficient training. We are going to re-visit the model in Chapter \ref{Chapter7} to
explore the problem of handling out-of-domain user input unseen during training in the setting where
no training examples like that are available.


\subsection{Natural Language Generation (\NLG)}
\label{ch2:nlg}
The last processing stage in the conventional, modular dialogue system architecture is the NLG.
At this stage, with the system's action chosen by the policy, the surface form of the corresponding
utterance is generated. As shown in Figure \ref{fig:architecture}, the information coming from the
policy is the dialogue act identifier itself as well as some of its attributes. The reason this
information is passed into \NLG is that traditionally, response generation is considered a 2-stage
process: firstly, a \textit{delexicalised} template of the final utterance is generated or selected
(i.e., with all the case-specific slot values replaced by filler tokens like
\texttt{$<$slot\_name$>$}). After that, the template gets lexicalised back into its final surface
form using the extra information from the policy. One of the most widely used notations for such
frames is CUED Standard Dialogue Acts \citep{cued}.

In early, template-based approaches to \NLG, templates were stored explicitly, and the system would
just pick a random one corresponding to a certain dialogue act
\citep[e.g.][]{DBLP:conf/interspeech/RudnickyTCTSLXO99}. Later on, \NLG was considered as an
optimisation problem (similarly to the components earlier in the pipeline discussed above), inspired
by the advances in artificial intelligence and planning under uncertainty. \NLG's optimisation
objectives were those of the information presentation problem, i.e. maintaining the users' focus,
speeding up information exchange, and improving the overall task success rate. For example, an
adaptive \NLG component can learn how to present a database lookup result with 1, 5, or 50
results~--- that is, whether to go one by one or limit the output to the top-3, or announce the
number of results and just display the top result. As every dialogue assumes making a series of such
information presentation decisions, with the feedback coming at the very end, it is intuitive to
approach this problem within the \RL framework. There has been a number of works proposing \RL
methods for \NLG, e.g. \cite{Rieser09,DBLP:conf/inlg/DethlefsC10}. 
They mainly focused on learning the 1st generation stage, i.e. response planning, while still
heavily relying on rule-based surface realisation, thus being limited in the overall output's
flexibility. \cite{Dusek15} used a hybrid approach for planning: the $A^*$ algorithm for the
syntactic dependency tree construction, with perceptron-based pruning (their surface realisation
stage for constructing sentence plans in the shape of syntactic dependency tree was also rule-based).

\begin{figure}
  \centering
  \includegraphics[width=0.8\textwidth]{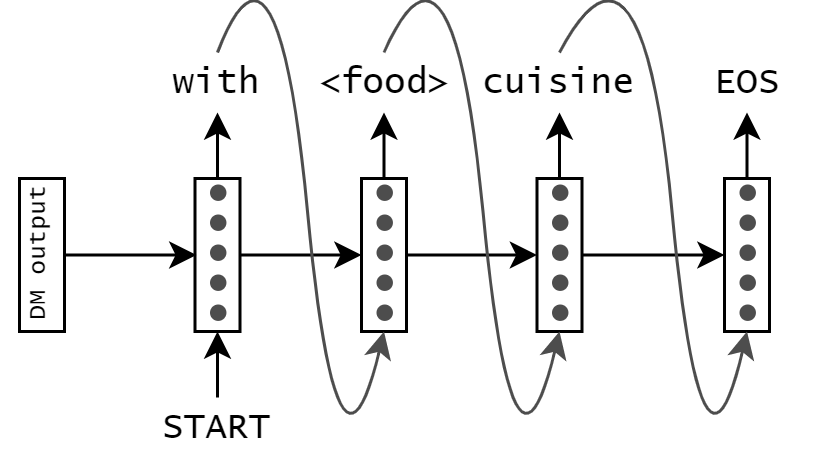}
  \caption{\RNN-based Natural Language Generation}
  \label{fig:rnn_nlg}
\end{figure}

The most important transformation to statistical \NLG happened with the coming of deep learning-based
methods: recurrent neural networks (\RNN, \LSTM) allowed to streamline the process by directly
generating sentences word-by-word in a language model way, that is generating the next word given
the context, e.g. \RNNLM \citep{DBLP:conf/interspeech/MikolovKBCK10}~--- with the dialogue act
information stored in the latent network state. A basic \RNN-based \NLG model is shown in Figure
\ref{fig:rnn_nlg}. The model generates output words by maintaining its hidden state $h$ and
updating it with every input token as follows (the equations below correspond to the particular
model known as Elman \RNN, \citealp{DBLP:journals/cogsci/Elman90}):

\begin{equation}
  h_t = \sigma_h \left( W_h x_t + U_h h_{t-1} + b_h \right)
\end{equation}%
\begin{equation}
  y_t = \sigma_y \left( W_y h_t + b_y \right)
\end{equation}

where $x_t$ is the input at time step $t$ (in our case, an encoded token), $y_t$ is the model's
corresponding output, $h_{t-1}$ is the model's previous state, $W_h, W_y, U_h, b_h, b_y$ are the
model's trainable weights, and $\sigma_h, \sigma_y$ are activation functions. Usually, $\sigmoid$
activation is used for the hidden state and $\softmax$ is used for the output. The initial state
$h_0$ of the model in the picture is the \DM output (normally a dialogue act). This information can
also be passed into the network at every step, e.g. as in \cite{DBLP:conf/sigdial/WenGKMSVY15}. Note
that in the picture, the model's input $x_t$ is its previous output $y_{t-1}$. This is the setup
such models operate in at inference time, however at training time, $x_t$ may as well be the ground
truth tokens (an approach referred to as teacher forcing).

In practice, basic \RNN networks were quickly replaced by \LSTMs
\citep{DBLP:journals/neco/HochreiterS97} which improve upon \RNNs in training stability via using
trainable {\it input, output, and forget} gates directly controlling what information to store in
the model's state and what to explicitly forget. In addition, the approach of
\cite{wen-etal-2015-semantically} combined two \RNN cells, one being a regular \LSTM for input
tokens, and the other being a special `lightweight' Dialogue Act cell.

The Sequence-to-Sequence (\SeqToSeq) neural architecture (\citealp{DBLP:journals/corr/VinyalsL15};
\citealp{DBLP:conf/nips/SutskeverVL14}; \citealp{DBLP:conf/emnlp/ChoMGBBSB14}, to be discussed in
detail later in Section \ref{ch2:generation}) also allowed to both encode the dialogue act into a
latent representation and then decode the output in a unified token-by-token process.
\cite{DBLP:conf/acl/DusekJ16} used this approach to generate both syntax trees for the further
surface realisation and the fully realised output sentences.

The encoder-decoder \SeqToSeq architecture represented the fully data-driven approach to \NLG which
assumes having a single streamlined training/prediction procedure with no intermediate stages and
supervision. This approach was adopted with other dialogue system components, e.g. dialogue managers
went through a similar transformation from having 2 separate subsystems, \DST and Policy, to a
unified architecture like the \HCN model discussed above. Eventually, the entire dialogue system
pipeline became fully data-driven. In the next section, we are going to discuss
the key types of dialogue system architectures that emerged as part of the fully data-driven
\citep{gao2019neural}, or corpus-based \citep{DBLP:books/lib/JurafskyM19} approach to dialogue
modelling.

\section{Fully Data-Driven Architectures}
\label{ch2:data_driven}

`Classic' machine learning methods transformed the dialogue systems field so that the most efficient
approaches shifted from analytical methods like expert rules, grammars, and ontologies towards
data-driven techniques, e.g. \CRFs for \NLU or reinforcement learning over (PO)\MDPs for dialogue
management. This resulted in more flexible and adaptive systems, and also in a shift of focus
in their development towards, firstly, collecting data (corpora of hundreds of dialogues were
normally used at that point) and secondly, tasking experts with feature engineering for the machine
learning models instead of directly writing the rules for the system. This reliance on feature
engineering instead of task-specific expert knowledge was the main factor contributing to the
overall success and efficiency of the machine learning approach.

With the arrival of large, internet-scale datasets and computational power to train machine learning
models large enough to make use of this data, the next generation of machine learning~--- deep
learning~--- had started. Characterised by using multi-layer (or deep) model architectures and the
unified technique for training them (predominantly variants of Stochastic Gradient Descent, \SGD%
\citealp[,][]{10.1007/978-3-7908-2604-3_16}), deep learning approaches transformed the machine
learning framework in the following fundamental way. With enough training data, deep neural networks
were able to approximate non-linear relations between the input variables, thus learning
features~--- often hierarchical~--- without the need for manual engineering
\citep{Lecun98gradient-basedlearning}. This opened the possibility to train models for the target
tasks directly from raw data, with the underlying latent features learned automatically and more
efficiently. One of the first and most notable examples of training a neural network from raw data
was the approach of \cite{Mnih13} featuring an agent for playing Atari games trained with Deep \RL
directly from the pixels of the game screen. This advance also had a massive impact on \NLP: the
benefits of training from more data with less annotations were experienced in machine translation,
question answering, document summarisation. Dialogue systems research was one of those areas
transformed under the influence of deep learning methods. As such, the models became less modular,
with the main focus on collecting a large amount of conversations and training the the entire model
(or its core part) on it. We are going to describe several kinds of such models below.

\subsection{Response Retrieval Models}
\label{ch2:retrieval}

Producing conversational utterances may be considered a response selection task, when a dialogue
system works similarly to a search engine: indeed, searching in a collection of documents for those
relevant to a user's query is analogous to searching for utterances given user's input (possibly
together with the dialogue context). Dialogue models working in this way are called
\textit{response retrieval} models. They have a search database (or \textit{index}) of responses or
full conversations, and given a new dialogue context at the input, they retrieve from the index the
best candidate given a certain optimality criterion. This criterion can be the similarity between
the context and the response, e.g. \TFIDF or Okapi \BM \citep{DBLP:books/daglib/0021593}~--- or some
more advanced learned objective functions.

\begin{figure}
  \centering
  \includegraphics[width=1.0\textwidth]{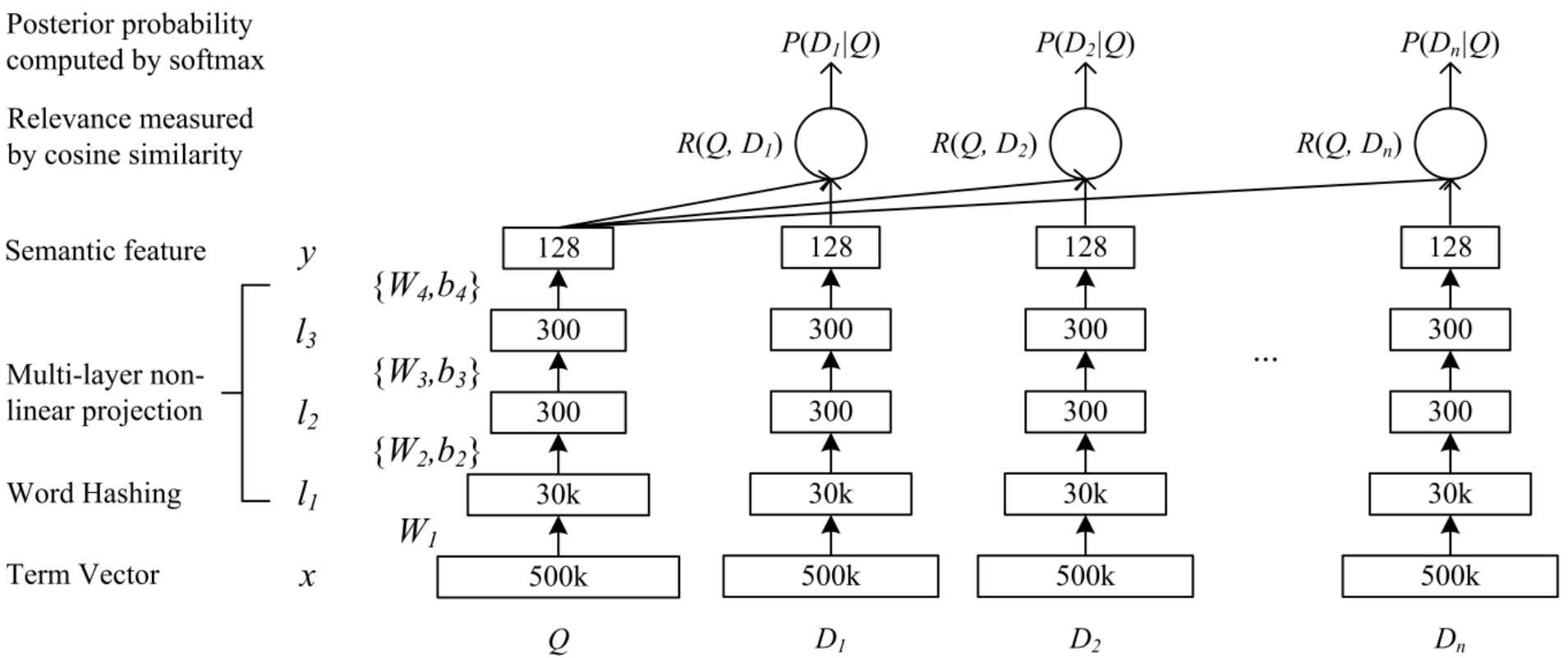}
  \caption[The \DSSM model architecture for web search]{The \DSSM model architecture for web search
  \citep{DBLP:conf/cikm/HuangHGDAH13}. Semantic representation of indexed documents $D_i$ as well as
  the user's query $Q$ is obtained with multiple trainable non-linear projection layers. The
  relevance scores are obtained via cosine similarity of query/document encoding pairs, then turned
  into a probability distribution over documents.}
  \label{fig:dssm}
\end{figure}

Retrieval models are widely used in \textit{chat-oriented} dialogue where the objective is to
maintain the conversation and keep the user engaged and entertained. Naturally, chat-oriented
systems can benefit from conversational data openly available on Internet, e.g. discussion forums
\citep{DBLP:journals/corr/abs-2001-08435}, movie subtitles \citep{DBLP:conf/lrec/LisonT16} or post
threads on social networks \citep{sordoni-etal-2015-neural}. One example of such architecture is the
Deep Semantic Similarity Model (\DSSM) \citep{DBLP:conf/cikm/HuangHGDAH13} which was originally
developed for document re-ranking in web search. The \DSSM architecture is visualised in Figure
\ref{fig:dssm}.

\DSSM is a deep feed-forward neural architecture: the input, a bag-of-words term vector $x$
(corresponding to a search query $Q$ or to a document $D_i,\ i = 1, \ldots, n$ in the search
database) is fed through a series of non-linear layers $l_1, \ldots, l_n$ with the trainable weight \&
bias parameters $W_1, \ldots, W_n$ and $b_1, \ldots, b_n$, respectively~--- ultimately resulting in
a deep semantic representation $y$ of the input. Formally, this pipeline is of the following form:

\begin{equation}
  \begin{array}{c}
    l_1 = W_1 x \\
    l_i = f\left(W_N l_{i-1} + b_i \right),\ i=2, ..., N - 1 \\
    y = f\left(W_N l_{N-1} + b_N \right) \\
  \end{array}
\end{equation}

where $f$ is the $\tanh$ activation function:
\begin{equation}
  \tanh(x) = \frac{1 - \exp(-2x)}{1 + \exp(-2x)}
\end{equation}

The semantic relevance of a document $D$ given the query $Q$ is then calculated as a cosine distance:
\begin{align}
    \begin{split}
        R(Q, D) &= \cos(y_Q, y_D) \\
        &= \frac{{y_Q}^\mathsf{T}  y_D}{\big{\|} y_Q \big{\|} \big{\|} y_D \big{\|}}
    \end{split}
\end{align}

The technique for calculating document relevance given a query in deep semantic representation was
translated into the conversational response selection framework: as such, it was used in Microsoft
Research's XiaoIce bot \citep{DBLP:journals/coling/ZhouGLS20}. \DSSM was used there as the response
re-ranker~--- that is, having retrieved the initial set of response candidates over a large
collection of conversations using simple and `fast' relevance metrics like \TFIDF or Okapi \BM,
those candidates were then re-ranked using the more fine-grained (and computationally `heavy')
model. The same 2-stage retrieval process is used in search engines, with the re-ranker constantly
pushed to work faster in order to handle a greater number of documents with real-time performance,
eventually taking place of the main ranking stage \citep{DBLP:conf/cikm/ZamaniDCLK18}.

Later on, with the development of neural architectures more suitable for textual data (predominantly
\LSTMs), new DSSM-based response selection models emerged. As such, \QALSTM
\citep{DBLP:journals/corr/TanXZ15} initially introduced for the Question Answering (\QA) task was
later used as the response ranker in the personal chatbot Replika \citep{fedorenko_avoiding_2017}
which gained massive adoption on the mobile application market.

Another widely-used family of neural retrieval architectures is Memory Networks (\textsc{MemNN}s,
\citealp{DBLP:journals/corr/WestonCB14}) which are based on the notion of explicit memory, with the
network operating a set of `cells' storing observations e.g. supporting facts for \QA or context
utterances for dialogue. Part of the architecture was also a differentiably trainable controller for
reading/writing the memory. The end-to-end variant of memory networks, \memnn \citep{Sukhbaatar15}
was the successor of \textsc{MemNN}s for the \QA task~--- but later on, it was adapted to dialogue
response selection. Therefore, we will explain the intuition behind the model for both \QA and
dialogue.

\begin{figure}
  \centering
  \includegraphics[width=1.0\textwidth]{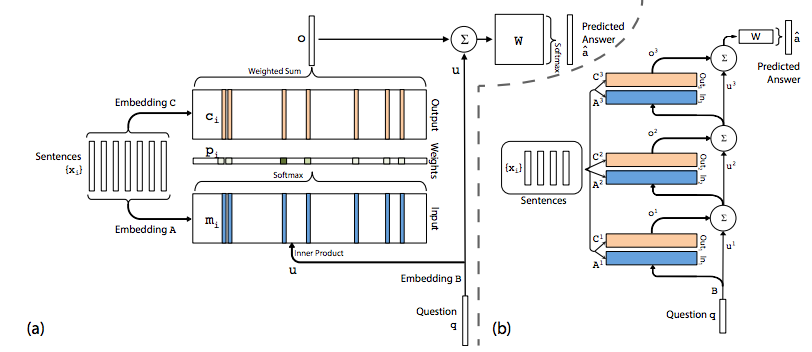}
  \caption[\memnn model architecture]{\memnn model~--- single-hop (a) and multi-hop (b)
  architectures \citep{Sukhbaatar15}}
  \label{fig:memn2n}
\end{figure}

The main \memnn architecture is shown in Figure \ref{fig:memn2n} (a)~--- its
main components are embedding matrices $A$, $B$, and $C$ providing the differentiable
representations for memories and the user's query and the output projection matrix $W$.
Specifically, input sentences (context in case of dialogue or facts in case of \QA) are represented
in the form of memories using the embedding matrix $A$. The same is done with $q$~--- user's
utterance in case of dialogue or the question in the \QA setting~--- using the matrix $B$.
Then, similarly to \DSSM and \QALSTM, the key operation is calculating the similarity between the
user's query and each of the memories:

\begin{equation}
    p_i = \softmax\left(u^\mathsf{T} m_i\right)
\end{equation}

where $\softmax(x) = \frac{1}{1 + \exp(-x)}$, $u$ is the embedded user's query, and $m_i$ are
system's memory cells.

These similarity scores implement what is originally referred to as \textit{reading from memory with
attention}. That is, the contents of the memory cells contribute to the final answer proportionally
to their attention weights $p_i$ (we'll also discuss other variants of the attention mechanism in
detail in Section \ref{ch2:gen_techniques}). Next, \memnn takes a weighted sum of the memories using
the output embedding $C$ and weights $p_i$:

\begin{equation}
  \begin{array}{c}
    c_i = {x_i} ^\mathsf{T} C \\
    o = \sum_i{p_i c_i}
  \end{array}
\end{equation}

The final operation in \memnn is similar to \DSSM or \QALSTM: a relevance metric is calculated over
the user's query $u$ together with the combined system's memory state $o$ and all the actions
available for the system $y_i$ (the set $Y$ includes all the possible responses and \API calls):

\begin{equation}
  \begin{array}{c}
    \hat{a} = \softmax\left((o + u)^\mathsf{T} \cdot W(y)\right)
  \end{array}
  \label{eq:memn2n_output}
\end{equation}

Unlike \DSSM and \QALSTM, cosine similarity is not used as the relevance metric in this case: output
matrices $W$ and $C$ serve this purpose instead (as well as $A$ and $B$, as in end-to-end training,
all the model components contribute to the final task). The crucial difference between \QA and
dialogue setups here is that in the \QA task, answers correspond to single vocabulary words, so the
projection $W$ produces a distribution over the vocabulary ids. In dialogue however, response
candidates are multi-word utterances, with different sets for training and testing.
Therefore, they cannot be predicted as ids, and are instead embedded using the matrix $W$ as shown
in Eq. \ref{eq:memn2n_output}.

Described above is a `single-hop' \memnn; the architecture however can be extended to multiple
`hops'. Such deep \memnn is visualised in Figure \ref{fig:memn2n} (b), with separate $A_i$ and $B_i$
matrices for every hop. A multi-hop model is basically a stack of base MemN2Ns connected in the
following way: the combined system/user state $o_i + u$~--- before the output projection~--- is
passed onto the $i+1$'th hop as the input, and the projection $W$ is only applied at the final hop
for producing the final answer.

The above model was used by \cite{DBLP:conf/iclr/BordesBW17} to train a goal-oriented dialogue
response retrieval system only using raw utterances. The model was used within a synthesised dataset
\bAbI Dialog Tasks, an experimental testbed designed to shed light on the complex problem of
goal-oriented dialogue management by decomposing it into several tasks of increasing complexity.
By showing quite impressive behaviour of \memnn on synthesised still challenging data (the action
set of \bAbI Dialog Tasks exceeded 4,000), \citeauthor{DBLP:conf/iclr/BordesBW17} demonstrated that
goal-oriented dialogue which has traditionally been considered complicated, mission-critical task
and relied on an extensive pipeline of components like language understanding, state tracking,
response planning (all described earlier in this chapter), and integration with domain-specific
\APIs can potentially be solved with a single unified model with raw example dialogues at the input.
We are going to look into MemN2N's performance within the testbed of bAbI Dialog Tasks in Chapter
\ref{Chapter3} to see how specific surface variations of the user's input unseen at training time
can affect the model's performance.

There emerged more Memory Network-based models later on, including knowledge-based
\citep{DBLP:journals/corr/abs-1804-08204}, personalised \citep{DBLP:conf/aaai/LuoHZN019}, Key-Value
MemNNs \citep{DBLP:conf/emnlp/MillerFDKBW16}, as well as those pipelined in a `retrieve-and-refine'
architecture \citep{DBLP:conf/emnlp/WestonDM18}.

Although response retrieval models gained wide adoption in industrial applications and products
because their output is fluent and more predictable~--- dependent on the quality of the corpus of
the candidate responses~--- their flexibility is always bounded by the corpus at the same time. This
limitation is addressed in a parallel line of dialogue systems research, focused on \textit{response
generation}. Being significantly harder to train and providing less certainty about the output
fluency, models of this type are theoretically capable of generating novel utterances word-by-word
and are most extensively studied in academia. We are going to discuss them in detail below.

\subsection{Response Generation Models}
\label{ch2:generation}

Response generation approach to conversation modelling stemmed from an advancement in Neural Machine
Translation (\NMT), namely the \SeqToSeq model \citep{DBLP:conf/nips/SutskeverVL14}. This
\LSTM-based model was one of the first end-to-end approaches to machine translation, working with
raw text tokens as opposed to previous phrase-based \citep[e.g.][]{DBLP:conf/naacl/KoehnOM03} and
syntax-based \citep[e.g.][]{DBLP:conf/acl/YamadaK01} models. Apart from machine translation,
\SeqToSeq became a general-purpose model for various \NLP tasks due to its versatility: it was able
to take a sequence of raw tokens at the input, encode it in a `thought vector' and produce an output
sequence of a length independent from the input (referred to as a \textit{many-to-many} model).
Arguably the most ambitious application of \SeqToSeq was in training an open-domain conversation
model, the problem that has never been approached before with a single, purely data-driven machine
learning model \citep{DBLP:journals/corr/VinyalsL15}.

As opposed to goal-oriented dialogue with the specific objective to reach the user's goal with the
minimum amount of conversational exchanges, open-domain conversation does not have a certain notion
of the goal. Therefore, the initial objective that the dialogue \SeqToSeq was trained with is to
essentially mimic the responses seen in a large training set of conversations, e.g. a corpus of
movie subtitles (the OpenSubtitles corpus of \citealp{Tiedemann:RANLP5} was used in the original
work). More formally, this objective is expressed as the Maximum Likelihood Estimate (\MLE), of the
output tokens $y_1, ..., y_{T'}$ given the input tokens $x_1, ..., x_T$:

\begin{equation}
  p(y_1, ..., y_{T'} \mid x_1, ..., x_T) = \prod_{t=1}^{T'}{p(y_t \mid v, y_1, ..., y_{t-1})}
  \label{eq:seq2seq}
\end{equation}

The formula above reflects the \textit{encoder-decoder} nature of \SeqToSeq: the encoder component
(an \RNN-based model in the original work) processes the input sequence $x_1, ..., x_T$ word by word,
eventually producing a latent representation of the input $v$. The decoder component (also
\RNN-based) produces the output sequence $y_1, ..., y_{T'}$ word by word, at each step based on the
internal \RNN state and the last generated token. Starting with $v$ as its initial state, the
decoder updates and maintains it throughout the generation process. Crucially, the decoder's output
sequence is not limited to any fixed length~--- instead, the decoder is directly trained to produce
a special \textit{end-of-sequence} \texttt{$<$EOS$>$} symbol at which the output effectively ends
for all the downstream processing. As we noted above, in the original paper, both encoder and
decoder were recurrent networks~--- specifically, \LSTMs, but in general, the requirements to
\SeqToSeq components are as follows:

\begin{enumerate}
  \item the encoder is any `many-to-one' model, i.e. able to encode an input sequence into a vector
  (in later versions of the model, a `many-to-many' model outputting all the intermediate
  word-by-word encodings is required),
  \item the decoder is any `one-to-many' model able to generate an output sequence out of a single
  vector.
\end{enumerate}

\begin{figure}[t]
  \centering
  \includegraphics[width=1.0\textwidth]{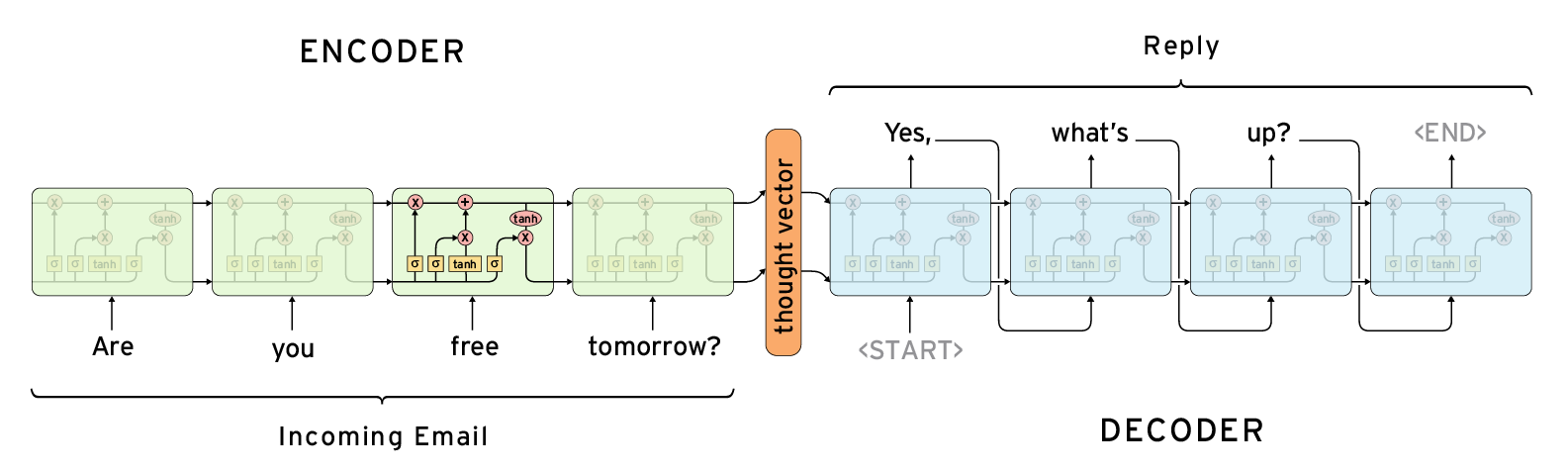}
  \caption[Sequence-to-sequence conversation model architecture]{Sequence-to-sequence conversation
  model architecture\footnote{Image source: Google AI blog}}
  \label{fig:seq2seq}
\end{figure}

A \SeqToSeq conversation model based on LSTMs is visualised in Figure \ref{fig:seq2seq}. As
mentioned above, the model is trained to `mimic' responses from the training dataset by learning
from randomly sampled context-response pairs over large conversational corpora. Initially, datasets
of movie subtitles were used for that, e.g. Cornell Movie Dialogs dataset
\citep{Danescu-Niculescu-Mizil+Lee:11a}, OpenSubtitles \citep{DBLP:conf/lrec/LisonT16}~--- as well
as threads of posts on message boards, e.g. Reddit conversations
\citep{DBLP:journals/corr/abs-2001-08435} and Twitter comments \citep{sordoni-etal-2015-neural}.

\section{Key Techniques for Dialogue Response Generation Models}
\label{ch2:gen_techniques}

The \SeqToSeq architecture was one of the most transformative advances in dialogue modelling, and it
was followed by a multitude of approaches building on top of it and improving it in various ways.
Specifically in case of dialogue, there were a number of aspects in which it could be improved as
the model did not account for the turn-taking nature of the dialogue and the overall objective of a
successful conversation still remained to be formulated better. Therefore, in this section we will
be focusing on the key improvements that made \SeqToSeq model represent dialogue more adequately.

\subsection{Hierarchical Response Generation Models}
\begin{figure}
  \centering
  \includegraphics[width=1.0\textwidth]{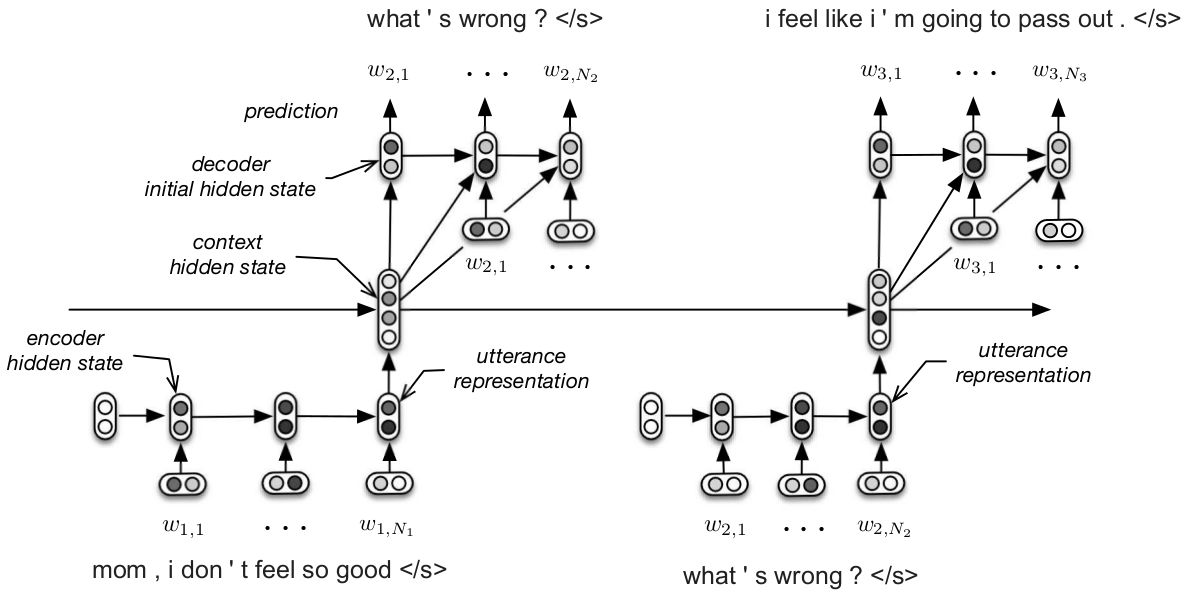}
  \caption[\HRED model architecture]{\HRED model architecture
  \citep{DBLP:journals/corr/SerbanSBCP15}}
  \label{fig:hred}
\end{figure}

One of the key improvements of the standard \SeqToSeq model addresses the turn-taking nature of
dialogue. Initially designed for machine translation, \SeqToSeq assumed passing an utterance in the
source language at the input, and producing its translation in the target language at the output.
This approach was adopted for the response generation by `flattening' the dialogue context in one
large input sequence. Although better than providing the model with no context at all, flattening it
has the obvious shortcomings that the information about the speakers gets lost, and the overall
length of the input becomes extremely high thus resulting in additional challenges with model
training and often requiring simply cutting the context quite severely.
To address this, \cite{DBLP:journals/corr/SerbanSBCP15} proposed the Hierarchical Recurrent
Encoder-Decoder (\HRED) model~--- see Figure \ref{fig:hred}.

In \HRED, the encoding procedure is 2-stage: first, every utterance of the context is encoded with
an utterance-level encoder \RNN into a dense vector. Then, these compact latent utterance
representations are encoded using another \RNN, producing a dense vector similar to the one at the
previous stage. This final dialogue context representation serves as the initial state of the
decoder, equivalent to that in the standard \SeqToSeq.

In addition to the reasons mentioned above, this hierarchical nature of the model also helps
training the entire architecture more efficiently. Backpropagation through time (\BPTT), the
technique normally used to train recurrent networks, assumes passing gradients through the \RNN's
weights word-by-word throughout the entire input sequence, thus resulting in exponentially small
gradient values propagated to the first encoding steps (also referred to as `vanishing gradient').
The hierarchical architecture reduces the length of gradient paths in the computational graphs, thus
alleviating this problem. This was especially relevant for `plain' \RNNs which widely suffer from
vanishing gradient (see our discussion of \RNNs in Section \ref{ch2:nlg}).

\subsection{Representation Learning with Autoencoders}
\label{ch2:autoencoder}

\begin{figure}
  \centering
  \includegraphics[width=0.8\textwidth]{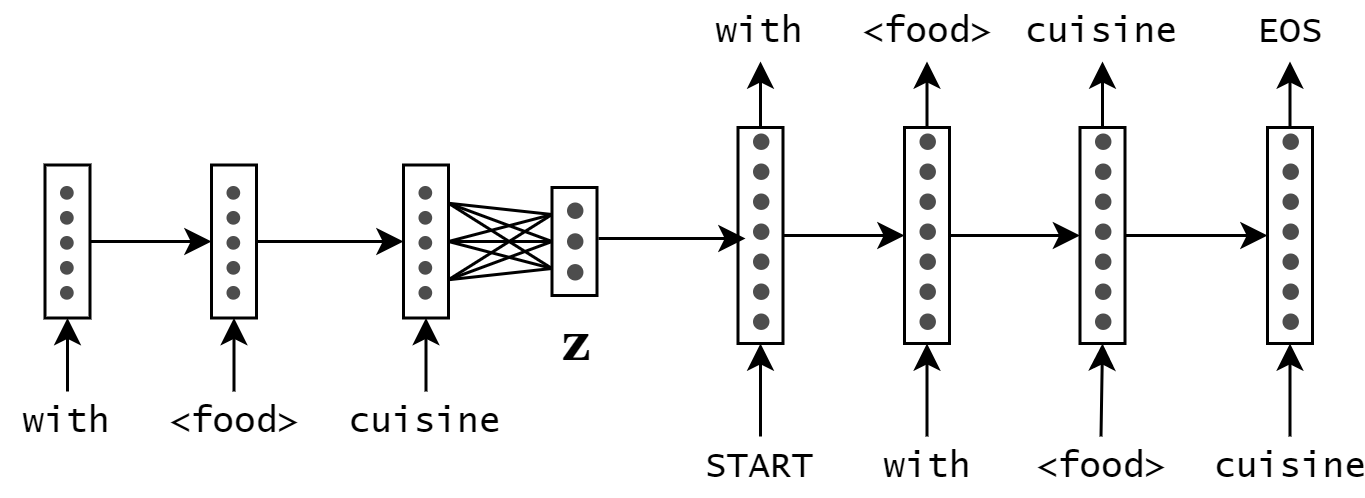}
  \caption{\RNN-based autoencoder}
  \label{fig:ae_lstm}
\end{figure}

A key aspect of modelling dialogue is obtaining an efficient latent representation of the underlying
utterances capturing their meaning and invariant to the surface variations. This
\textit{representation learning} problem is normally tackled with a type of models called
\textit{autoencoders}. Autoencoder (\AE) is a model that is trained to attempt to copy its input to
its output \citep{Goodfellow-et-al-2016}. An autoencoder normally consists of two parts: the encoder
and the decoder. The encoder produces a latent representation of the model's input, which then the
decoder uses to produce the output, which is ideally the copy of the input. Crucially in this
process, the resulting latent representation, i.e. the encoder's output (also referred to as the
\textit{bottleneck}) is of a dimensionality significantly lower than that of the input and the
reconstructed output. By learning to `compress' the data into a more compact representation (i.e. of
a reduced dimensionality) which still contains enough information for the decoder to reconstruct the
input, the autoencoder determines which `features' of the input are informative and which ones are
not and can be ignored. In Figure \ref{fig:ae_lstm} is shown an autoencoder with both the encoder
and the decoder being \RNNs, a similar model to the one we will be working with in Chapter
\ref{Chapter7}~--- in the picture, $z$ is the latent input representation that is regarded as the
main autoencoder's output.

The key question in training an autoencoder is to make its latent space \textit{continuous}. That
is, given a pair of latent vectors for the two inputs seen during training, an autoencoder with a
continuous latent space would not only be able to reconstruct those two inputs, but also to produce
meaningful reconstructions from arbitrary positions along the line between the encodings.
The type of an autoencoder directly aimed at learning continuous latent spaces is based on the
Bayesian methods and is called the Variational Autoencoder (\VAE,
\citealp{DBLP:journals/corr/KingmaB14}; \citealp{DBLP:conf/icml/RezendeMW14}). \VAE represents the
encoding $z$ as a continuous latent variable produced by the probabilistic \textit{recognition
model} (the counterpart of \AE's deterministic encoder) approximating the posterior distribution of
this latent variable given the input $q(z \mid x)$, in practice usually a diagonal Gaussian. The
actual reconstruction is then obtained by running the \textit{generation model} (the counterpart of
\AE's decoder) conditioned on a \textit{sample} from $q(z\mid x)$.

\begin{figure}
  \centering
  \includegraphics[width=0.8\textwidth]{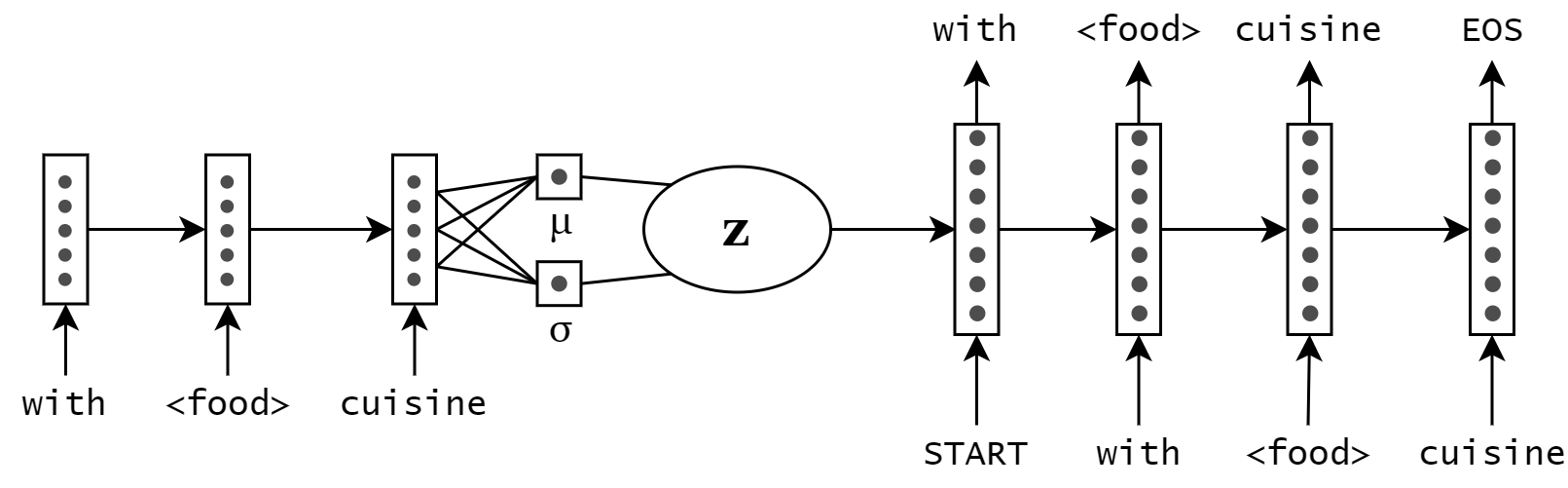}
  \caption{\RNN-based \VAE}
  \label{fig:vae_lstm}
\end{figure}

The variational counterpart to the \RNN autoencoder is shown in Figure \ref{fig:vae_lstm} ~--- the
recognition model there consists of an \RNN encoder and a linear projection layer generating
parameters $\mu$ and $\sigma$ of the $q(z \mid x)$ Gaussian, a sample from which then goes into the
generation model.

As opposed to the \MLE optimisation objective of a regular \AE, \VAE is trained to optimise the
Evidence Lower Bound (\ELBO) objective which looks as follows:

\begin{equation}
  \mathcal{L}(x) = - \kld{q(z \mid x)}{p(z)} + \mathbb{E}_{q(z \mid x)} \log p(x \mid z)
  \leq \log p(x)
  \label{eq:vae_elbo}
\end{equation}

where $p(z)$ is the prior distribution of $z$ usually set to a standard Gaussian with
$\mu = 0, \sigma = 1$, and $\log p(x)$ is the data likelihood, which \AE optimises. The first term
of the formula is Kullback-Leibler divergence between the approximate posterior and the prior:

\begin{equation}
  \kld{q}{p} = \sum_{x \in X}{q(x) \log \frac{q(x)}{p(x)}}
  \label{eq:vae_kl}
\end{equation}

where $X$ is the probability space $q$ and $p$ are defined on.

\KL divergence provides an expectation of the likelihood ratio between the two distributions (second
term) with respect to $q(x)$. The intuition behind it is measuring how much more probable the data
is under one distribution than under the other. In the \VAE optimisation objective, the \KL term
keeps the model from making $q(z \mid x)$ deterministic (which it could become if the recognition
network learned to output $\sigma$ close to $0$) and instead forces it to be closer to the prior
$p(z)$.

In practice, penalising the divergence between the prior and posterior comes with a challenge: early
into the training, the \KL term tends to reduce to zero, essentially making it identical to the prior.
This problem is known as the `vanishing \KL term' problem \citep{DBLP:conf/conll/BowmanVVDJB16}.
In the context of the \RNN-based models which are extremely prone to overfitting, the source of
stochasticity (i.e. latent variable $z$) can be overfitted as well, so it does not matter what
information is encoded in it as the network adapts itself to noise. Such a model would essentially
converge to its non-variational and severely overfitted counterpart. Therefore, the optimisation of
the \ELBO objective is highly problematic in practice.

\begin{figure}
  \centering
  \includegraphics[width=1.0\textwidth]{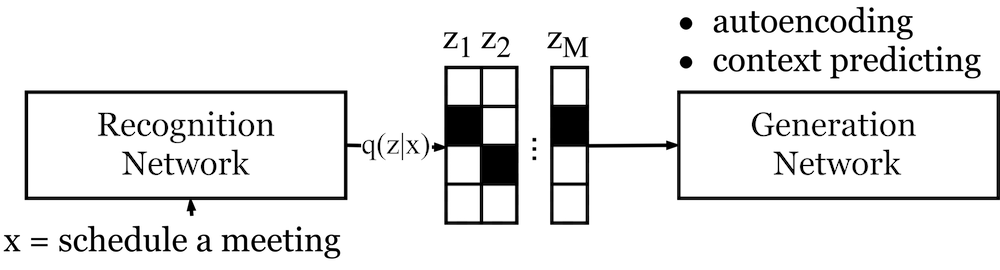}
  \caption[The \DIVAE/\DIVST models]{The \DIVAE/\DIVST models \citep{DBLP:conf/acl/EskenaziLZ18}}
  \label{fig:di_vae}
\end{figure}

In order to facilitate learning, \cite{DBLP:conf/conll/BowmanVVDJB16} propose the `\KL term
annealing' technique. \KL weight annealing imposes a weight upon the \KL term that is set to zero at
the beginning of training, effectively leaving the model with a regular \MLE objective. After a
certain number of epochs of initial fitting to data, the annealing weight starts growing up on the
interval $[0, 1]$, eventually ending up with the objective as in Eq. \ref{eq:vae_elbo}. However,
in practice, \KL annealing is itself very unstable and needs careful fine-tuning of the annealing
schedule. Therefore, more principled techniques for \ELBO optimisation were introduced later on~---
the most notable is the approach of mutual information maximisation between the input and the latent
variable \citep{DBLP:conf/aaai/ZhaoSE19} and the version of it for discrete latent variables
\citep{DBLP:conf/acl/EskenaziLZ18}. The latter is called Discrete Information \VAE (\DIVAE) and is
shown in Figure \ref{fig:di_vae}. It is a variant of a \VAE with two modifications. Firstly,
its optimisation objective accounts for the mutual information $I$ between the input and the latent
variable which is found to be implicitly discouraged in the original \VAE objective (see the authors'
derivation in Eqs.\ \ref{eq:vae_fsdg} and \ref{eq:di_vae}).

\begin{equation}
  \label{eq:vae_fsdg}
  \begin{split}
    \mathcal{L}_{VAE} & = \mathbb{E}_x\left[\mathbb{E}_{q_{\mathcal{R}}(z \mid x)}
    \left[ \log p_\mathcal{G}(x \mid z)\right]
    - \kld{q(z)}{p(z)}\right] \\
    & = \mathbb{E}_{q(z \mid x) p(x)}\left[ \log p_\mathcal{G}(x \mid z)\right] - I(Z, X)
    - \kld{q(z)}{p(z)},
  \end{split}
\end{equation}


\begin{equation}
  \label{eq:di_vae}
  \begin{split}
    \mathcal{L}_{DI\text{-}VAE} & = \mathcal{L}_{VAE} + I (Z, X) \\
    & = \mathbb{E}_{q_{\mathcal{R}}(z \mid x) p(x)}
    \left[ \log p_\mathcal{G}(x \mid z)\right] - \kld{q(z)}{p(z)}
  \end{split}
  \end{equation}

where $x$ is the input utterance, $z$ is the latent variable ($X$ and $Z$ corresponding
to their batch-wise vectors), $\mathcal{R}$ and $\mathcal{G}$ are the recognition and generation
models (implemented as \RNNs) respectively, and $q(z) = \mathbb{E}_x[q_\mathcal{R}(z \mid x)]$.

Secondly, the latent variable $z$ in \DIVAE is \textit{discrete} as opposed to the continuous
one in a regular \VAE. The discrete latent code lends itself well to interpretation and can be used
e.g. to cluster inputs by the values of the certain `bits' of their latent codes. The discrete
nature also makes the calculation of the \KL term more tractable via the Batch Prior Regularisation
technique \citep{DBLP:conf/acl/EskenaziLZ18}:


\begin{equation}
  \label{eq:bpr}
  \begin{split}
    \kld{q'(z)}{p(z)} =
    \sum_{k=1}^K{q'(z = k) \log \frac{q'(z = k)}{p(z = k)}}
  \end{split}
\end{equation}

where $K$ is the number of $z$'s possible values and $q'(z)$ is the approximation to $q(z)$ over $N$
data points:


\begin{equation}
  \label{eq:q_prime}
  \begin{split}
    q'(z) = \frac{1}{N}\sum_{n=1}^N{q_{\mathcal{R}}(z \mid x_n)}
  \end{split}
\end{equation}

In addition to \DIVAE, \cite{DBLP:conf/acl/EskenaziLZ18} introduced \DIVST, \DIVAE's counterpart
working in a Variational Skip-Thought manner \citep{DBLP:conf/naacl/HillCK16}. Specifically, it
reconstructs the input $x$'s previous ($x_p$) and next ($x_n$) context utterances with the
corresponding independent generation models $\mathcal{G}^p$ and $\mathcal{G}^n$:

\begin{equation}
\label{eq:di_vst}
\begin{split}
  \mathcal{L}_{DI\text{-}VST} & = \mathbb{E}_{q_{\mathcal{R}}(z \mid x) p(x)}
  [ \log p^n_\mathcal{G}(x_n \mid z) p^p_\mathcal{G}(x_p \mid z)] - \kld{q(z)}{p(z)}
\end{split}
\end{equation}

We are going to use \DIVAE and \DIVST later in Chapter \ref{Chapter4}, as well as empirically
compare them to the `classic' \VAE on a downstream dialogue response generation task.

\subsection{Latent Variable Models for Dialogue Response Generation}
\label{ch2:latent_variable}

\begin{figure}
  \centering
  \includegraphics[width=0.5\textwidth]{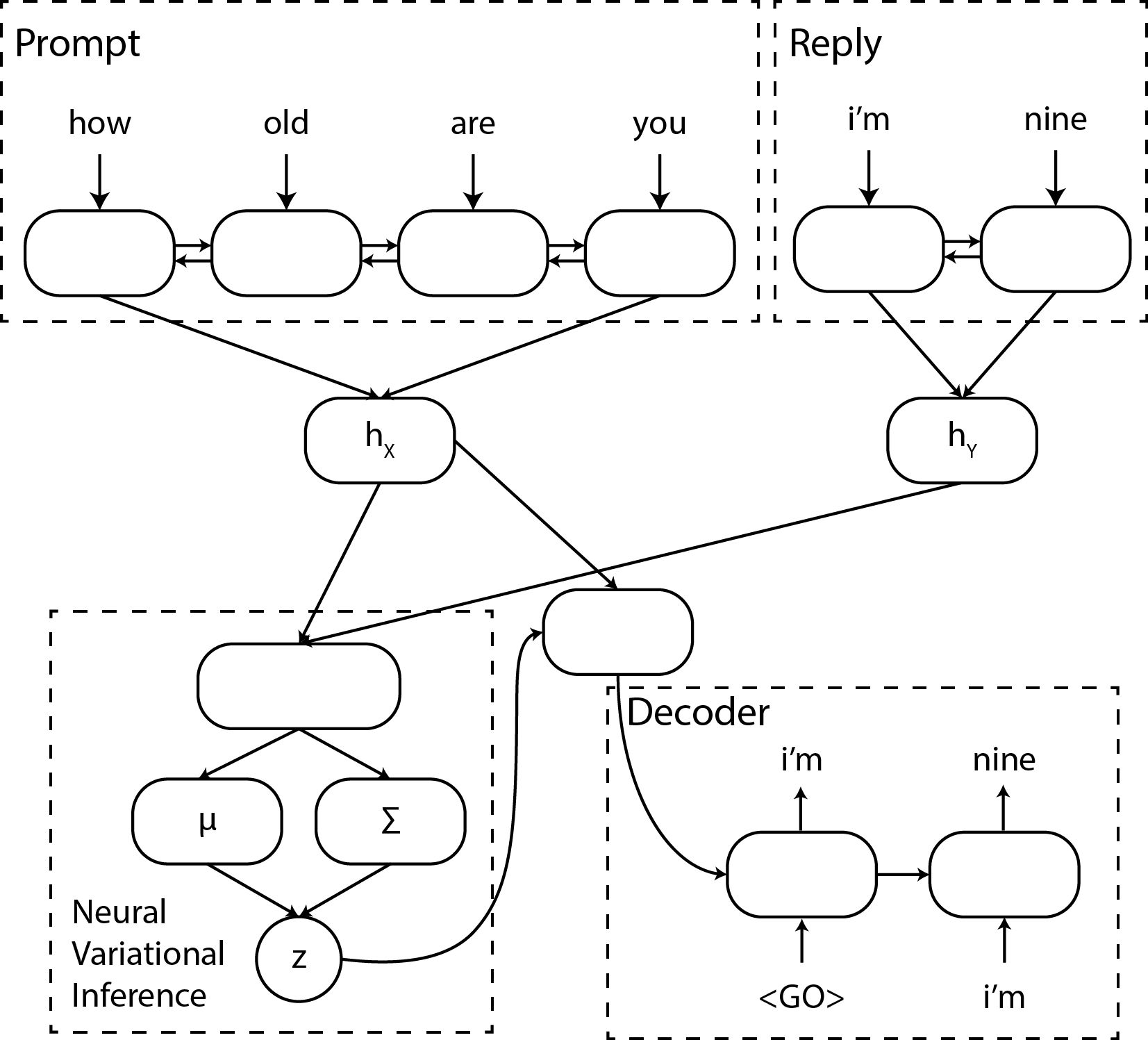}
  \caption[Latent variable \SeqToSeq model]{Latent variable \SeqToSeq model
  \citep{DBLP:conf/eacl/ClarkC17}}
  \label{fig:variational_seq2seq}
\end{figure}

Apart from resulting in a continuous latent space, the use of a latent variable has another
intuitive benefit for dialogue~--- increasing the diversity of responses. In chat-oriented dialogue,
low diversity and insufficient informativeness in a common problem with \SeqToSeq models. Given the
extremely high variance of responses given their contexts in the training corpora (more on those
later in Section \ref{ch2:data_collection}), the regular \MLE optimisation objective of such models
forces the model to learn the `corpus-average', i.e. most recurring responses in the training set
which usually are `I don't know', `I'm not sure' and the like \citep{DBLP:conf/emnlp/JiangR18}.

A series of approaches addressed that problem, e.g. \cite{DBLP:conf/naacl/LiGBGD16} used
maximisation of mutual information (MMI) between the context and response as part of the objective
function; \cite{DBLP:conf/nips/ZhangGGGLBD18} formulated it in terms of the Variational Information
Maximisation Objective (VIMO) and approached the problem under the adversarial learning framework.

A widely used technique to increase diversity of responses was using random sampling in the
\SeqToSeq decoder \citep{DBLP:conf/acl/IppolitoKSKC19}, i.e. drawing a random entry from the model's
probability distribution for every output token. But this often resulted in degraded generation
performance as the MLE objective by definition only assumes optimising for the most probable output,
and even the 2nd most probable word can be irrelevant or disfluent in most cases.

In turn, latent variable models allow to directly incorporate non-determinism into the model as well
as into the training objective. From the dialogue perspective, it means that for every dialogue
context, there can potentially be multiple correct responses, and the model will optimise for all of
them. From the Bayesian optimisation point of view, that means learning the posterior distributions
over these responses instead of point-wise optimisation in case of \MLE, as we discussed with \VAEs
previously. Such models were introduced by \cite{DBLP:conf/aaai/SerbanSLCPCB17} and
\cite{DBLP:conf/eacl/ClarkC17}~--- the latter is visualised in Figure \ref{fig:variational_seq2seq},
we are going to discuss it below.

Under this approach, the output probability gets conditioned on (in addition to the input as in
regular \MLE training):

\begin{equation}
  P(Y \mid X) = \int_z P(Y \mid z, X) P(z) dz
  \label{eq:vae_seq2seq}
\end{equation}

where $P(z) = \mathcal{N}(0, I_n)$ is usually set to a standard Gaussian prior.
This conditional probability denotes a distribution of correct responses for a given input mentioned
above, and as the value for $z$ is sampled, the specific response from the distribution gets picked
for further generation. With the presence of the system's output $Y$ different from the input $X$
(as opposed to the VAE case), the \ELBO objective takes the following form:

\begin{equation}
  \log P(Y \mid X) \geq - \kld{Q(z \mid X, Y)}{P(z)} + \mathbb{E}_{z \sim Q} \log P(Y \mid z, X)
  \label{eq:latent_model_elbo}
\end{equation}

where $Q$ is the \textit{proposal} (or \textit{variational}) distribution which is used to
approximate the posterior $P(z \mid X, Y)$ during the optimisation.

\begin{figure}
  \centering
  \includegraphics[width=1.0\textwidth]{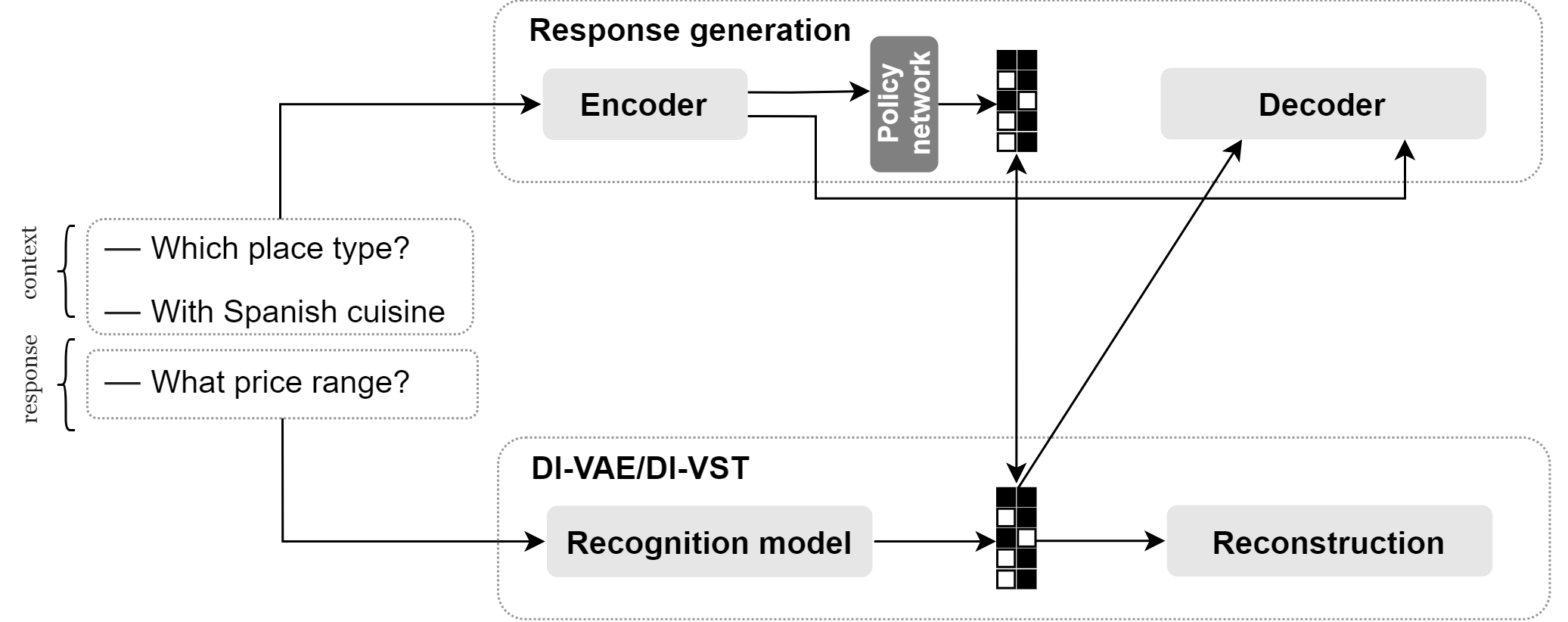}
  \caption[The \LAED architecture]{The \LAED architecture (training)}
  \label{fig:laed}
\end{figure}

The Discrete-Information \VAE technique we discussed previously in Section \ref{ch2:autoencoder} was
integrated into an encoder-decoder architecture, resulting in the Latent Action Encoder-Decoder
(\LAED) model \citep{DBLP:conf/acl/EskenaziLZ18}~--- see Figure \ref{fig:laed}.
The main part of \LAED is the encoder-decoder \SeqToSeq model which (1) encodes the dialogue context
$c$ (including the user's last turn) with a hierarchical recurrent encoder and then produces an
approximate posterior $p_\pi (z \mid c)$ of the latent variable $z$ using its `policy' feed-forward
network $\pi$, (2) decodes the system's response $y$ using samples from $p_\pi (z \mid c)$ and $c$.

Crucially though, at training time, it works in a multi-task setup with a \DIVAE/\DIVST model whose
recognition model $\mathcal{R}$ (separate from the main model's encoder) encodes the
\textit{response} and approximates a posterior $q_\mathcal{R}(z \mid y)$. It then reconstructs
either the response itself (\DIVAE), or its previous+next context utterances (\DIVST)
using its generation model, also separate from the main task's decoder. During training, the main
task's decoder uses samples from $q_\mathcal{R}(z \mid y)$ from the auxiliary \DIVAE/\DIVST instead
of $p_\pi (z \mid c)$, and the policy $\pi$ only learns to reproduce $q_\mathcal{R}(z \mid y)$ via
the \MLE objective: $\mathbb{E}_{p(y \mid c)} \left[ q_\mathcal{R}(z \mid y) \right]$.

\LAED training objective is as follows:
\begin{equation}
  \label{eq:laed}
  \begin{split}
    & \mathcal{L}_{LAED} =
    \mathbb{E}_{q_{\mathcal{R}}(z \mid y) p(c, y)}
    \left[ \log p_\pi (z \mid c) + \log p_\mathcal{F} (y \mid z, c) \right]
\end{split}
\end{equation}

where $\mathcal{F}$ is the recurrent decoder.
We are going to look at the potential of \LAED's latent codes for dialogue knowledge transfer across
datasets in Chapter \ref{Chapter4}.

\subsection{Attention Mechanism}
\label{ch2:attention}

\begin{figure}
  \centering
  \includegraphics[width=1.0\textwidth]{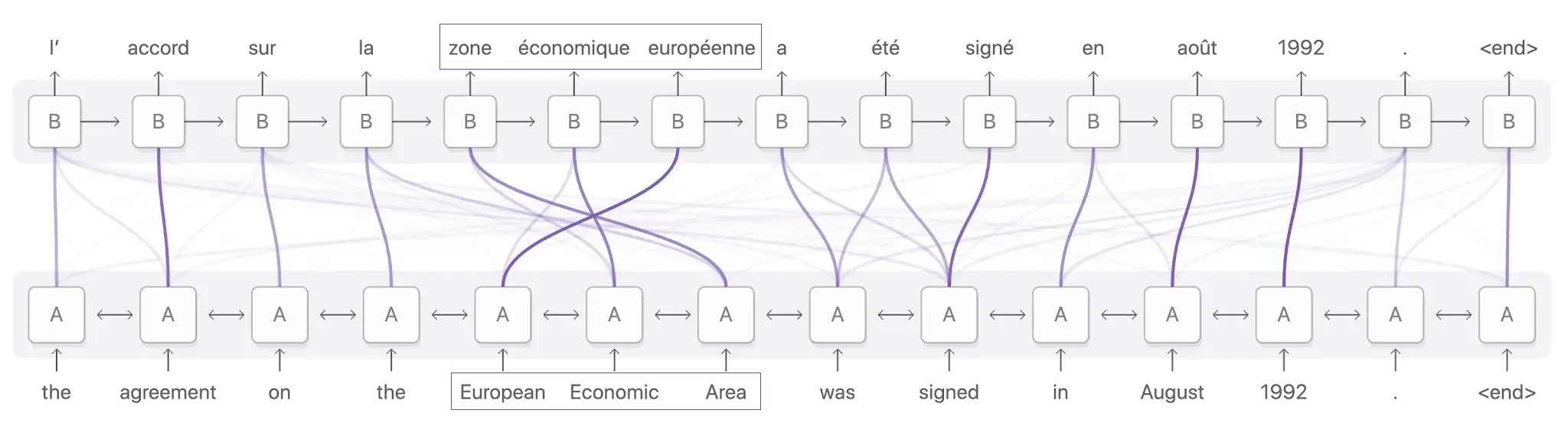}
  \caption[Visualisation of attention in machine translation]{Visualisation of attention in machine
  translation \citep{olah2016attention}}
  \label{fig:attn_vis}
\end{figure}

The most widely-used modification to the standard encoder-decoder architecture came, as the original
model itself, from machine translation: the attention mechanism introduced by
\cite{DBLP:journals/corr/BahdanauCB14} addressed the problem of the insufficient information coming
to the decoder from the final encoder state. While significantly more efficient than \RNNs,
\LSTM/\GRU models still do not produce the perfect representation of the input sequence for further
decoding from it, and more importantly, this representation is static, while intuitively the
translation takes place in segments.

For example, for generating one noun phrase the translating decoder in Figure \ref{fig:attn_vis}
\citep{olah2016attention} would have to `attend' mainly to this phrase at the input~--- that is, the
translation of the phrase `zone économique européenne' can be done without knowing its left or right
contexts.

\begin{figure}
  \centering
  \includegraphics[width=0.4\textwidth]{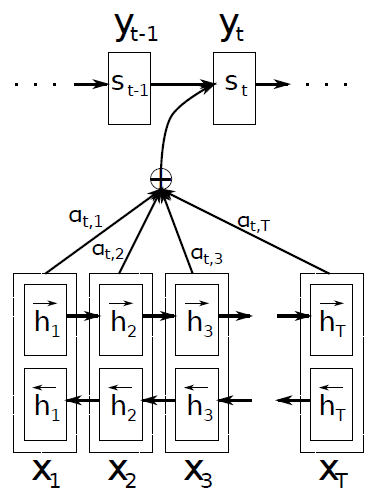}
  \caption[Attention mechanism]{Attention mechanism \citep{DBLP:journals/corr/BahdanauCB14}}
  \label{fig:attention}
\end{figure}

Attention is visualised in Figure \ref{fig:attention}: as opposed to the original \SeqToSeq model
only passing the final encoder state as the context for the decoder, this approach preserves all the
intermediate encoder states and passes them all weighted with the corresponding \textit{alignment
scores} learned as part of the end-to-end training procedure. Alignment scores $\alpha_{t,i}$ tell
how much information the decoder can infer from the encoder state $h_i$ (i.e. having encoded the
tokens $x_1, ..., x_i$) while generating the $t$th output token.


\begin{equation}
  \alpha_{t,i} = \frac{\exp(\score(s_{t-1},
  h_i))}{\sum_{i'=1}^n{\exp(\score(s_{t-1}, h_{i'})})}
  \label{eq:attn_alignment}
\end{equation}

where $s_{t-1}$ is the previous decoder state, $h$ is the encoder state, and $\score$ is a scoring
function implemented as a feed-forward neural network (among a series of other
implementations) with trainable weights $W_a$ and $v_a$:


\begin{equation}
  \score(s_t, h_i) = {v_a}^\mathsf{T} \tanh\left(W_a[s_t; h_i]\right)
  \label{eq:attn_score}
\end{equation}

At the $t$th decoding step, the decoder receives the information from all the encoder states as a
weighted mixture of the following form:


\begin{equation}
  c_t = \sum_{i=1}^n = \alpha_{t,i} h_i
  \label{eq:attn_ctx}
\end{equation}

\cite{DBLP:conf/emnlp/LuongPM15} also explore a range of similar models. Specifically, {\it global}
attention, the closest to that of \cite{Bahdanau14}, defines the alignment score for $t$th step via
$s_t$ instead of $s_{t-1}$, as well as considering several variants of the $\score$ function:


\begin{align}
  \score(s_t, h_i) = 
  \begin{cases}
    s_t^\intercal h_i & \textit{dot}\\
    s_t^\intercal W_a h_i & \textit{general}\\
    {v_a}^\mathsf{T} \tanh\left(W_a[s_t; h_i]\right) & \textit{concat}
  \end{cases}
  \label{eq:luong_score}
\end{align}

The alternative {\it local} attention, instead of calculating alignment scores of the decoder state
with all the encoder's states, uses a fixed window $[p_t - D, p_t + D]$ where $p_t$ is the position
within the encoded sequence predicted by the model, and $D$ is a constant set empirically. The
authors use 2 ways of calculating $p_t$: {\it monotonic} alignment where $p_t = t$, and {\it
predictive} alignment defined as follows:


\begin{equation}
  p_t = S \cdot \sigmoid\left(v_p^\intercal \tanh(W_p h_t)\right)
\label{eq:luong_local_p}
\end{equation}

where $W_p$ and $v_p$ are trainable parameters, and $S$ is the input sentence's length.
In addition, in order to favour alignment scores around $p_t$, the authors place a Gaussian
distribution centered around it on $\alpha$:

\begin{equation}
  \alpha_{t, i}' = \alpha_{t, i} \exp \left(-\frac{(s-p_t)^2}{2\sigma^2}\right)
\label{eq:luong_alpha}
\end{equation}

where $\sigma = \frac{D}{2}$.

Experimental evaluation on English-to-German machine translation task showed that local attention
with predictive alignment was superior to other individual approaches, however the best result they
obtained was with an ensemble of 8 models with different attention mechanisms and training aspects.

Although motivated by machine translation, attention was widely used for dialogue generation, both
chat-oriented and goal-oriented. For the latter, this mechanism was in the core of another key
technique~---\textit{copy-augmented decoding}.

\subsection{Copy-Augmented Decoding}
\label{ch2:copy_augmented}

Copy-augmented \SeqToSeq models (\citealp{DBLP:conf/eacl/ManningE17};
\citealp{DBLP:conf/sigdial/ZhaoE18}) are used for addressing rare or out-of-vocabulary (\OOV) words
or some content of the user's query that needs to be reflected in the response as well, e.g. in the
affirmation that all the details of the user's request are received, and the processing is started.

\begin{figure}
  \centering
  \includegraphics[width=1.0\textwidth]{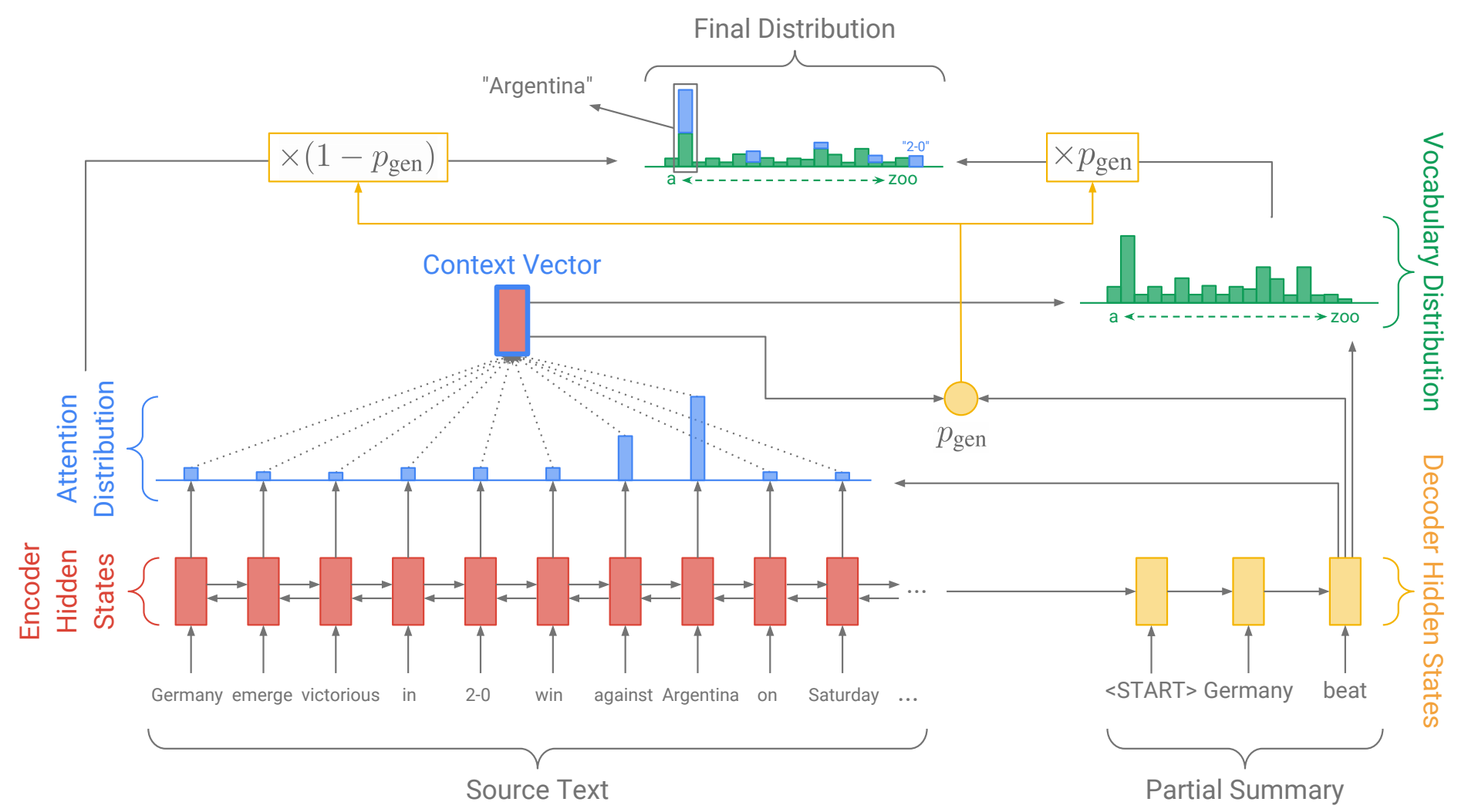}
  \caption[Pointer-Generator network architecture]{Pointer-Generator network architecture
  \citep{DBLP:conf/acl/SeeLM17}}
  \label{fig:gttp}
\end{figure}

Attention-based copying is based on the idea of \textit{pointer networks}
\citep{DBLP:conf/nips/VinyalsFJ15} which produce a permutation of the input tokens as their output
and were primarily aimed at solving the problem of sorting the input sequence and various
combinatorial optimisation problems. In dialogue as well as other \NLP tasks, hybrid models
combining the `pure generation' and the `pure pointer' mechanisms are used, and here we will focus
on Pointer-Generator Networks (\citealp{DBLP:conf/acl/SeeLM17}) shown in Figure \ref{fig:gttp}.

Pointer-generator model defines the probability of generating the word $w$ at a certain decoding
step as the following mixture:

\begin{equation}
  P(w) = p_{gen}P_{vocab}(w) + (1-p_{gen})\sum_{i:w_i=w}a_{t,i}
  \label{eq:gttp}
\end{equation}

where $P_{vocab}$ is the probability of generating the word from the decoder's vocabulary under the
conventional decoding procedure, and $a_i^t$ are the attention alignment scores of all the encoder
states wherever the word $w$ is observed at the input. Finally, the mixture parameter $p_{gen}$ is
defined as follows:

\begin{equation}
  p_{gen} = \sigma\left(w_{h^*}^T h_t^* + w_s^T s_t + w_x^T x_t + b_{ptr}\right)
  \label{eq:gttp_p_gen}
\end{equation}

where $h_t^*$ is the attention context vector (equivalent to $c_t$ in the attention derivation
above), $s_t$ is the decoder state, and $x_t$ is the decoder input~--- all at the decoding step $t$,
and  $w_{h^*}, w_s, w_x, b_{ptr}$ are trainable parameters.

In this approach, the decoder works with an `extended vocabulary' which is the union of the original
vocabulary and all the words appearing at the input. Therefore, for an \OOV word, $P_{vocab}(w) =0$,
but if it is present at the input, it can still be transferred to the output if its attention weight
is high. On the other hand, if this word is not present at the input, $\sum_{i:w_i=w}a_i^t = 0$, and
it can only be generated based on the decoder's internal state as $P_{vocab}(w)$.

\begin{figure}
  \centering
  \includegraphics[width=0.9\textwidth]{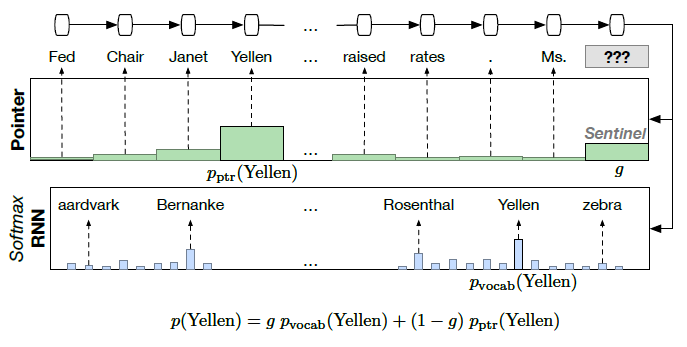}
  \caption[Pointer-Sentinel Mixture Model]{Pointer-Sentinel Mixture Model
  \citep{DBLP:conf/iclr/MerityX0S17}}
  \label{fig:psm}
\end{figure}

The particular implementation described above was used for hybrid abstractive/extractive document
summarisation, but in general, the copy-augmented decoding technique is widely used in dialogue,
semantic parsing \citep{DBLP:conf/acl/JiaL16}, and language modelling
\citep{DBLP:conf/iclr/MerityX0S17}. The latter~--- Pointer Sentinel Mixture Models (\PSMs) (see
Figure \ref{fig:psm})~--- we are going to use in Chapter \ref{Chapter4}. As seen in the figure, \PSM
extends the pointer distribution with a \textit{sentinel} element which is used to redistribute the
probability mass in case of low confidence of the model's pointer part. The intuition is as follows:
the lowest-confidence case of a `vanilla' copy model~--- a near-uniform distribution derived from
the attention scores~--- is supposed to be avoided in the \PSM by putting the most of the
probability mass on the sentinel. Then, this sentinel also working as a gating function in the
hybrid pointer-generator prediction (see $g$ in Figure \ref{fig:psm} and $p_{\text{gen}}$ in Eq.
\ref{eq:gttp})~--- will control the model's final prediction. As such, $g$ can take values in the
interval $[0, 1]$ representing the mixture parameter between the pointer and generator
distributions, so that when $g = 0$, the next word is predicted from the pointer distribution only,
and when $g = 1$, the model's confidence in pointer distribution is the lowest, and it falls back to
generator-only prediction mode.

Apart from the enhancements of the \SeqToSeq model, attention was also used as a key component of a
text representation itself~--- specifically, of a hierarchical text representation model for
document classification \citep{DBLP:conf/naacl/YangYDHSH16}. Under their approach, two attention
types were calculated~--- utterance-level attention representing the contribution of a specific word
in the overall meaning of a sentence, and document-level attention representing a similar relation
for individual sentences within the document. Later on, attention-based approach to text
representation developed into a separate technique called \textit{self-attention} which we are going
to describe below.

\subsection{Self-Attention}

The original attention mechanism brought the idea of calculating the alignment scores between
different parts of a sequential model's internal state. Self-attention introduced by
\cite{DBLP:conf/nips/VaswaniSPUJGKP17} uses the idea of alignment as the main means of representing
sequential input: instead of directional word-by-word encoding, self-attention assumes computing
alignment scores of every word to every other one at the input, thus producing a representation of
every word in the global context of the entire input sequence. More formally, the alignment scores
are calculated via 3 quantities associated to every word $x_i$: `key', `query', and `value' ($k_i$,
$q_i$, $v_i$, respectively). The model obtains those via the following 3 respective projection
matrices:

\begin{equation}
  \begin{array}{c}
    q_i = x_i^\mathsf{T} W^Q \\
    k_i = x_i^\mathsf{T} W^K \\
    v_i = x_i^\mathsf{T} W^V
  \end{array}
  \label{eq:self_attn_kqv}
\end{equation}

The mathematical formulation of self-attention is largely similar to that of the original attention
mechanism, with the main difference being that it is defined via $k_i$, $q_i$, and $v_i$.
As such, the self-attention score of how a word $x_j$ affects the target word $x_i$ is calculated as
follows: 

\begin{equation}
    \score_{i,j} = \frac{q_i \cdot k_j}{\sqrt{d_k}}
  \label{eq:self_attn_scaled_dot_product}
\end{equation}

where $d_k$ is the dimensionality of the key vectors. This operation is referred to by the authors
as \textit{scaled dot-product attention}.

The second step is similar to the original attention~--- all the self-attention scores for $x_i$ are
softmax-normalised into alignment weights:

\begin{equation}
  \alpha_{i,j} = \frac{\exp(\score_{i,j})}{\sum_{k=1}^n{\exp(\score_{i,k})}}
  \label{eq:self_attn_alpha}
\end{equation}

Finally, the final representation of a word $x_i$ is obtained as a mixture of all the input words'
values $v_j$ weighted by their alignment weights given the target word:

\begin{equation}
  z_i = \sum_{j=1}^n{\alpha_{i,j} \cdot v_j}
  \label{eq:self_attn_z}
\end{equation}

Self-attention representations of individual words $z_i$ are then fed into a feed-forward network.
This architecture is then duplicated in the form of several sub-models, with separate matrices
$W^Q$, $W^K$, and $W^V$. The final representations produced by these sub-models are referred to as
\textit{self-attention heads}. As reported in the original paper, these independently initialised
and trained heads produce different representation `subspaces' which may account for different
linguistic phenomena, e.g. anaphoric links or syntactic dependencies.

\begin{figure}[t]
  \centering
  \includegraphics[width=0.4\textwidth]{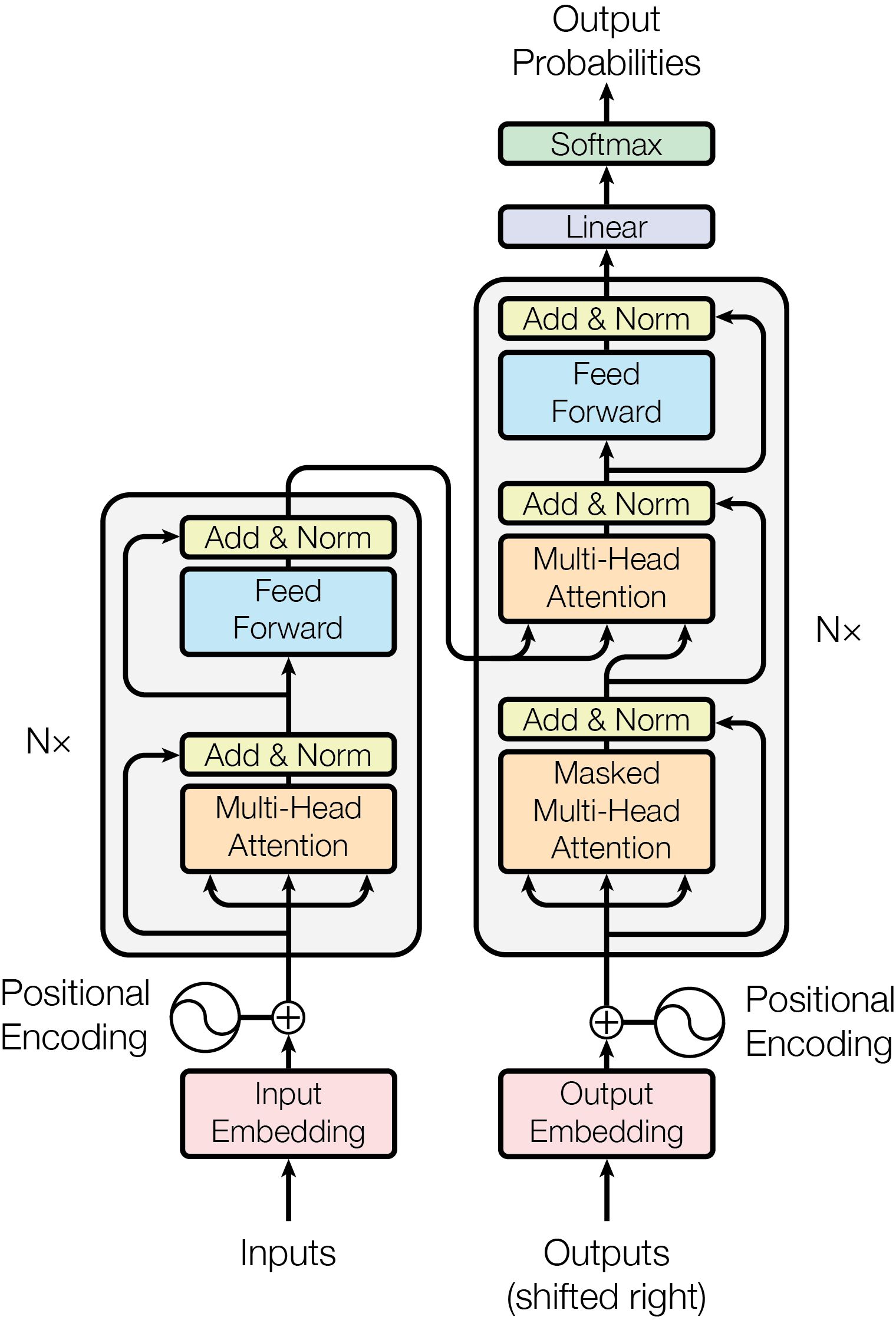}
  \caption[Transformer encoder-decoder architecture]{Transformer encoder-decoder architecture
  \citep{DBLP:conf/nips/VaswaniSPUJGKP17}}
  \label{fig:transformer}
\end{figure}

Self-attention mechanism lies in the core of the Transformer encoder-decoder model (shown in Figure
\ref{fig:transformer}). Both encoder and decoder of the Transformer use a similar logic to the one
described above, with the encoder's output serving as the input representation and decoder's output
being fed into the additional linear + softmax layers (together referred to as the `language
modelling head') used for generating the output probability distributions over the vocabulary tokens.
More specifically, the decoder generates the output words token-by-token, just like the \SeqToSeq
model, but by attending to (1) the outputs of the stack of encoders and (2) the generated sequence
up to the current timestep.

Transformers set the new state-of-the-art in a series of \NLP tasks thus largely replacing
\RNN-based architectures as the main way of producing robust text representations. Self-attention
architecture started a new generation of text models using the benefits of global-context word
representation and increased parallelism of self-attention pipeline over recurrent sequential
processing. The most notable of those models are Bidirectional Encoder Representations from
Transformers (\BERT) by \cite{DBLP:conf/naacl/DevlinCLT19} and Generative\footnote{Here,
``generative'' is used in the sense of predicting (or generating) the next word given the context,
i.e. language generation. We will also use this sense of the term ``generative'' later in Chapter
\ref{Chapter5}.} Pretrained Transformers (\GPT/\GPT[2]) by \cite{gpt2}~--- trained on massive
amounts of data and designed to be efficiently trained for a variety of downstream tasks, they
largely enabled the transfer learning paradigm in \NLP which will be discussed further.

\section{Transfer Learning in \NLP and Dialogue Modelling}
\label{ch2:transfer_learning}

\begin{figure}[t]
  \centering
  \includegraphics[width=0.8\textwidth]{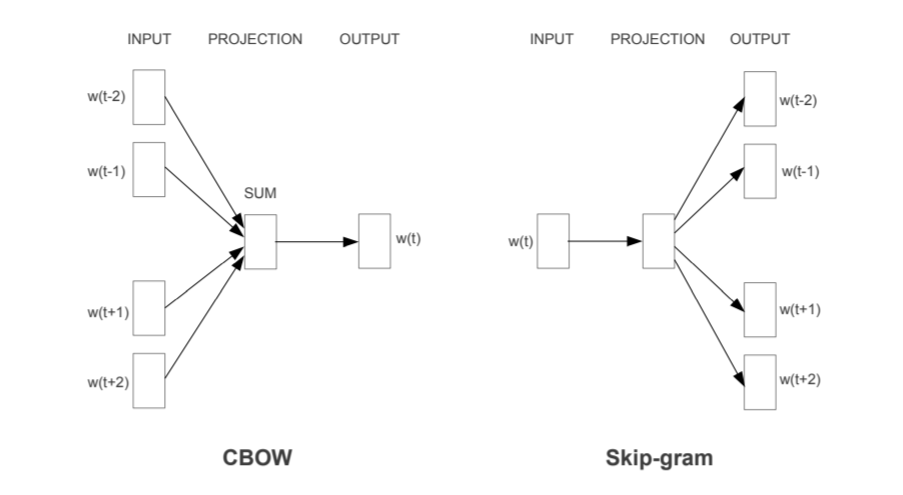}
  \caption[\WordToVec model architecture]{\WordToVec model architecture
  \citep{DBLP:conf/nips/MikolovSCCD13}}
  \label{fig:cbow_skipgram}
\end{figure}

Transfer learning is a direction in machine learning that assumes training one model for a specific
task and then re-using the knowledge it has learned, partly or fully, on another task. The initial
model is called the \textit{base model}, and it is trained from large general-purpose datasets at
the first stage called \textit{pre-training}. At the second stage~--- \textit{fine-tuning}~--- the
base model (or its core part, e.g. utterance/dialogue encoder) is further trained for the target
task, normally with additional task-specific parts introduced.

Transfer learning came to \NLP from the Computer Vision (\CV) community, where large-scale training
of image recognition models became a common practice via \ImageNet, the visual recognition challenge
and the corresponding dataset \citep{DBLP:conf/cvpr/DengDSLL009}. State-of-the-art \ImageNet models,
e.g. \ResNet \citep{DBLP:conf/cvpr/HeZRS16} and \VGG \citep{DBLP:journals/corr/SimonyanZ14a} became
widely used in the research community, and it soon became apparent that they can be adapted to a
wide range of \CV problems by fine-tuning to small in-domain datasets, eventually significantly
outperforming the corresponding models trained from scratch.

\subsection{Word Embedding Models}

\begin{figure}[t]
  \centering
  \includegraphics[width=0.8\textwidth]{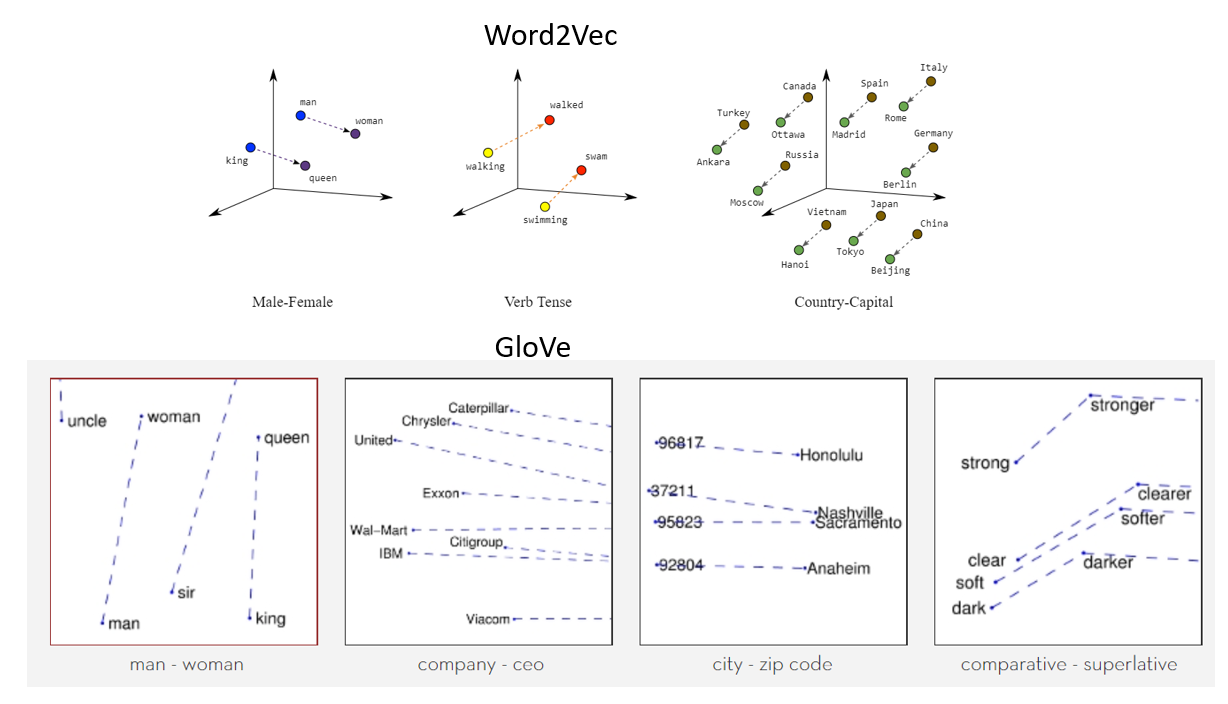}
  \caption{Example relations between \WordToVec and \GloVe vectors}
  \label{fig:w2v_relations}
\end{figure}

Similarly to \CV with \ImageNet, \NLP experienced a transformation with \textit{word embeddings}
which were the first widely transferable resources. Embeddings provide an efficient alternative to
`1-hot' representation widely used before. 1-hot representation of a word assumes having a
vocabulary-sized binary vector of zeros, with a single `one' corresponding to the word's index in
the vocabulary. Embeddings improved on that by representing words as real-valued vectors in a
low-dimensional trainable space. Word embedding training objective is normally based on modelling a
certain relationship between a word and its context (we will go into detail later)~--- this allows
them to be trained in a `self-supervised' fashion from internet-scale amounts of raw unannotated
data.

In Figure \ref{fig:cbow_skipgram} is shown \WordToVec, one of the earliest and most widely-used word
embedding models \citep{DBLP:conf/nips/MikolovSCCD13}. The 2 versions of it: Continuous Bag-of-Words
(\CBOW) and Skip-gram~--- show that it can be trained either to predict the word from its
surrounding context, or vice versa. The model's projection (or embedding) matrix that maps a 1-hot
word vector into the latent space is the main result of the training.

Another early embedding model~--- \GloVe \citep{pennington2014glove}~--- uses word co-occurrence
counts to predict the co-occurrence probability of a pair of words given their corresponding
embedding vectors~--- the training objective makes the dot-product operation over a pair of word
embedding vectors produce that probability. As multiple observations showed, both models' embedding
spaces are able to encode different linguistic relations between words, e.g. `male~--- female' or
`company~--- CEO', see Figure \ref{fig:w2v_relations}\footnote{Image credit: Renu Khandelwal
(\colorhref{blue}{http://tiny.cc/w2v\_glove\_medium})} for a visualisation.

Word representations produced by \WordToVec and \GloVe gained extremely high popularity in the \NLP
community and were used to improve the performance of models in numerous tasks. Their principal
shortcoming though was in the fact that they did not take word context while encoding it. That
results in the inherent inability to encode sequences of words (e.g. utterances and paragraphs)~---
so that workaround approaches like `mean-vector embedding' were used to represent phrases (i.e. the
embedding of the phrase is the element-wise mean of the embeddings of individual words).
Another problem is that contextless models cannot handle polysemous words like `book', `fly', 'like':
the resulting representations of such words will correspond to their most frequent sense in the
dataset. The solution to these problems came with the next generation of embeddings based on
sequence models.

\subsection{Contextual Word Embeddings}

\begin{figure}[t]
  \centering
  \includegraphics[width=0.8\textwidth]{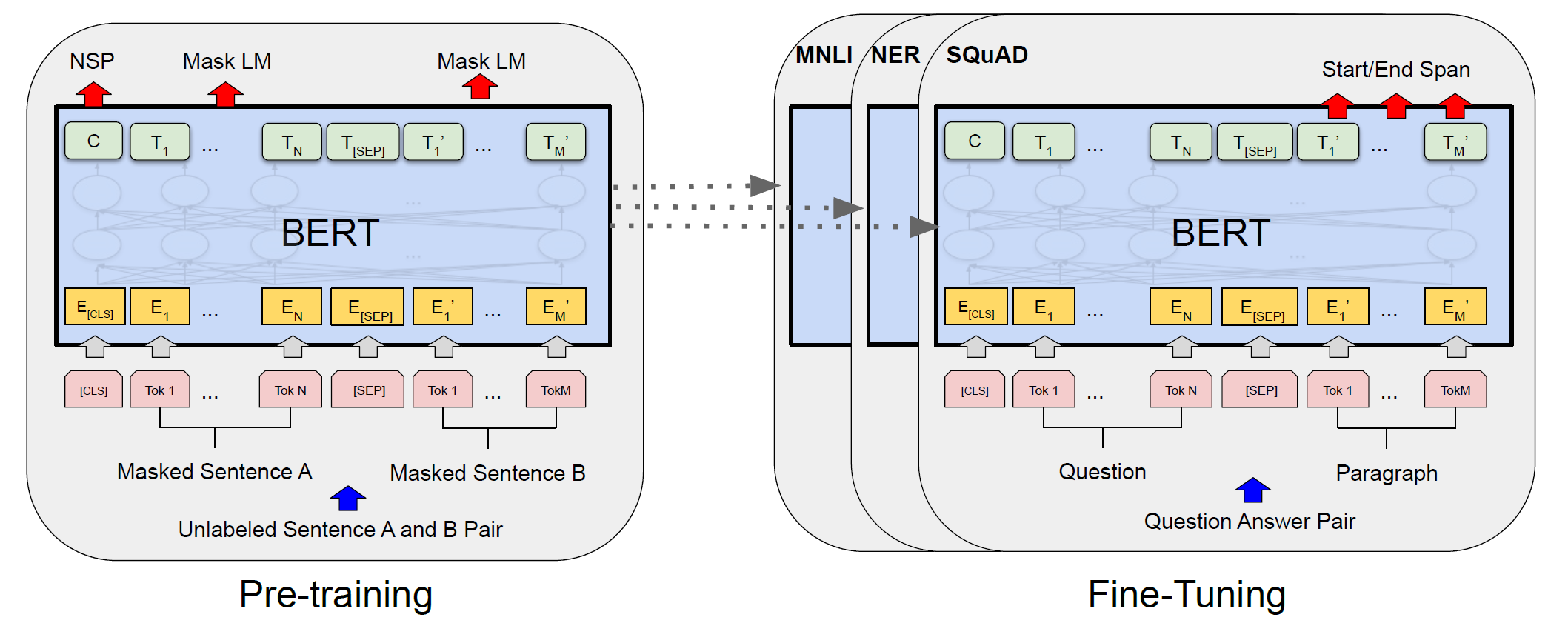}
  \caption[\BERT `pretrain-finetune' architecture]{\BERT `pretrain-finetune' architecture
  \citep{DBLP:conf/naacl/DevlinCLT19}}
  \label{fig:bert}
\end{figure}

Neural models for sequential data, e.g. \LSTMs and Transformers brought the next generation of word
embeddings. As such, there were introduced models like \ELMo \citep{DBLP:conf/naacl/PetersNIGCLZ18},
\ULMFiT \citep{DBLP:conf/acl/RuderH18}, and \BERT \cite{DBLP:conf/naacl/DevlinCLT19}. Trained as
language models, they provided what is referred to as a `contextual' representation (as opposed to
the `static' embeddings by \WordToVec or \GloVe). Both models can be efficiently used in the
downstream tasks by either replacing the `language modelling head' with a task-specific one or
training the two in a multi-task setup. 

Among all the contextual embedding models, Transformer-based ones resulted in arguably the widest
impact on the \NLP community~--- specifically, \BERT already mentioned above was designed for use in
downstream tasks in a `pretrain-finetune' fashion (shown in Figure \ref{fig:bert}). Specifically,
this versatility is achieved by (1) organising the input in 2 parts, i.e. sentence A and sentence B
(which may correspond to question/answer or context/response pairs in a downstream task) (2)
pre-training in the multitask setup using the two objectives: Language Modelling (\LM) and the Next
Sentence Prediction (\NSP). For the \LM objective, \BERT uses the notion of the Masked Language
Model (\MLM) which assumes predicting randomly chosen tokens of the input based on the full encoded
representation of the input sequence, i.e. using both the left and the right contexts. The \NSP task
makes the model learn the relation between the 2 input sentences~--- in the base case, whether
sentence B actually follows sentence A in the original text or it is a distractor. This secondary
task helps improve the model's robustness to noise (if sentence B is set to be a randomly drawn
distractor sentence) as well as make the model highly versatile in a variety of downstream
classification-like tasks, e.g. Question Answering, Natural Language Inference, response retrieval
and ranking.

Upon introduction, the original \BERT achieved state-of-the-art results on 11 \NLP tasks and was one
of the most extensively used base models for text representation in transfer learning for \NLP. More
\BERT variations followed up: for example, Robustly Optimized BERT approach (\RoBERTa%
\citealp[,][]{DBLP:journals/corr/abs-1907-11692}) trained with dynamic \MLM masking, larger batch
size, and disabled NSP task during pretraining; \SpanBERT \citep{DBLP:journals/tacl/JoshiCLWZL20}
modifying \MLM to mask and predict the entire spans of input instead of individual tokens;
\DistilBERT \citep{DBLP:journals/corr/abs-1910-01108}, a version of \BERT reduced in size by 40\%
while retaining 97\% of the original model's accuracy, obtained via knowledge distillation.

\subsection{Transfer Learning for Dialogue}

Following wide success in fundamental \NLP tasks (e.g. language modelling, semantic role labelling,
coreference resolution), transfer learning techniques started emerging in dialogue.

The problem of dialogue system's domain adaptation was posed in Dialog State Tracking Challenge 3
which was focused on adapting a goal-oriented dialogue state tracker to a new domain using a small
set of seed in-domain data \citep{DBLP:journals/dad/WilliamsRH16a}~--- that can be considered a
transfer learning task, but from today's perspective, both pre-training and fine-tuning stages were
bound to a very specific and narrow domain thus limiting the scale of the transfer. Nevertheless,
several notable methods emerged from the challenge, all \RNN-based (
\citealp{DBLP:conf/slt/HendersonTY14}; \citealp{DBLP:conf/acl/MrksicSTGSVWY15}).
Later, \cite{DBLP:conf/acl/WilliamsAZ17} introduced the \HCN model (described earlier in Section
\ref{ch2:dst}) that is designed for 2-stage training: the initial training stage from moderate
amounts of data, and the consequent fine-tuning stage in an RL setup from interactions with real
users or via an interactive human-in-the-loop process under the Conversation Learner framework
\citep{williams2017demonstration}.

More recently, a major shift in transfer learning for dialogue was brought by the Transformer model
described above. Specifically, the introduction of Transformed-based \GPT and \GPT[2] \citep{gpt2}
set the new state-of-the-art in human-like language generation. \GPT models:

\begin{itemize}[label=---]
  \item are implemented using restricted unidirectional self-attention (where a token can only
  attend to its left context, as opposed to both left and right with e.g. \BERT),
  \item consist of massive sets of parameters (the largest \GPT[2] model has 1.5 billion parameters),
  \item are pre-trained at a large scale on WebText corpus consisting of millions of documents,
  \item set the new state-of-the-art on a number of language modelling tasks, while still
  underfitting on the WebText corpus, as reported by the authors.
\end{itemize}

\GPT/\GPT[2] employ language-modelling pre-training which allows them to be re-used in a number of
generation tasks, including conversational response generation. They were rapidly adopted in the
dialogue community, and a number of `pretrain-finetune' approaches to dialogue followed, most
notable of which are (1) \TransferTransfo \citep{DBLP:journals/corr/abs-1901-08149}, the
winning submission at \ConvAI challenge on persona-based chat-oriented dialogue as per the automatic
metrics, (2) goal-oriented dialogue generation approach by \cite{DBLP:conf/emnlp/BudzianowskiV19},
and (3) \DialoGPT, an open-domain model by \cite{zhang2019dialogpt} pretrained at a large scale on
the Reddit Conversations dataset. The latter achieved human-like performance in the 1-turn Turing
test evaluation with humans. In Chapter \ref{Chapter5}, we are going to use \GPT[2] as a base model
in a `pretrain-finetune' framework for dialogue domain adaptation, in both information-seeking and
strictly task-oriented setups.

Models like \ELMo, \BERT, and \GPT[2] showed that transfer learning can be applied in \NLP as well
as in vision. A number of \NLP tasks have already experienced the benefits of transfer learning, and
it is now considered among the best practices to approach new \NLP problems via transferring those
models' knowledge in a `pretrain-finetune' fashion instead of training from scratch. In dialogue
modelling, the benefits of transfer learning are starting to emerge so that conversation models
become more human-like in naturalness, coherence, and appropriateness of their utterances~--- as
well as in goal-oriented dialogue where those models need significantly less training data to start
working with reasonable accuracy. In the next section, we are going to discuss the intuition of what
aspects of dialogue can be transferred across domains and datasets.

\subsection{Dialogue Transfer Learning Intuition: Lexical and Interactional Dialogue Similarity}
\label{ch2:dialogue_similarity}

With the variety of models and techniques of knowledge transfer presented above, it is important to
have an actual intuition of why transfer learning applies to dialogue and what exactly can be
transferred. There are two major aspects in which dialogues can vary, but nevertheless, lead to
similar meanings: interactional and lexical. Interactional similarity is analogous to syntactic
similarity~--- when two distinct sentences have effectively identical meaning~--- except that it
occurs not only at the level of a single sentence, but at the dialogue or discourse level.
Figure \ref{fig:variation} shows examples of interactional variants that lead to very similar final
contexts, in this case, that the user wants to buy an LG phone. These dialogues can be said to be
effectively similar for this domain.

Lexical similarity, on the other hand, holds among utterances, or dialogues, when different
words (or sequences of words) express meanings that are sufficiently similar in a particular domain
(see again Figure \ref{fig:variation}). Unlike syntactic or interactional ones, lexical similarity
is domain-specific~--- that is, a pair of dialogues which are equivalent sequences of dialogue acts,
each one is represented with equivalent syntactic structures, are not considered lexically similar
if one has to do with restaurant search and the other with booking movie tickets.

\begin{figure}[t]
  \begin{footnotesize}
  \centering
  \begin{tabularx}{\textwidth}{lll}\toprule
  \begin{tabular}[t]{ll}
  USR: & I would like an LG laptop\\
  &sorry uhm phone\\
  SYS: & okay.
  \end{tabular}&
  \begin{tabular}[t]{ll}
  USR: & I would like a\\ 
  &phone by LG.\\
  SYS: & sorry a what?\\
  USR: & a phone by LG.\\
  SYS: & okay.
  \end{tabular}&
  \begin{tabular}[t]{ll}
  SYS: & what would you like?\\
  USR: & an LG phone\\
  SYS: & okay.
  \end{tabular}\\\midrule
  \begin{tabular}[t]{ll}
  SYS: & what would you like?\\
  USR: & a phone\\
  SYS: & by which brand?\\
  USR: & LG\\
  SYS: & okay
  \end{tabular}&
  \begin{tabular}[t]{ll}
  SYS: & you'd like a ...?\\
  USR: & a phone\\
  SYS: & by what brand?\\
  USR: & LG.\\
  SYS: & okay
  \end{tabular}&
  \begin{tabular}[t]{ll}
  SYS: & so would you like\\
  &a computer?\\
  USR: & no, a phone.\\
  SYS: & okay. by which brand?\\
  USR: & LG.\\
  SYS: & okay.
  \end{tabular}\\\bottomrule
  
  \end{tabularx}
  \end{footnotesize}

  \caption{Interactional variations in a shopping domain}
  \label{fig:variation}
\end{figure}

The intuition behind learning dialogue representations from data that are transferable across
datasets and domains is capturing these similarities in a general `dialogue footprint'. In this way,
semantically similar dialogues with the same footprint would be clustered together~--- either by
their embedding in the latent space or by explicit meaning representations e.g. derived from a
semantic parser.

In case of linguistically informed models, \cite{Eshghi.Lemon14} developed a method similar to
\cite{Kwiatkowski.etal13} for capturing lexical similarity by creating clusters of semantic
representations based on observations that those clusters correspond to similar non-conversational
actions observed within a domain (e.g. a database query, a flight booking, or any API call).
Distributional methods could also be used for this purpose \citep{Lewis.Steedman13}.
In general, this kind of clustering is achieved when the domain-general semantics resulting from
semantic parsing is grounded in a particular domain. We note that while interactional similarity
in dialogue can be accounted for by semantic grammars or formal models of dialogue structure
(such as \DSTTR, \citealp{Eshghi.etal12} or \textsc{KoS}, \citealp{Ginzburg12}), lexical similarity
relations have to be learned from data.

\section{Linguistically Informed Models of Dialogue}
\label{ch2:linguistic}

Linguistic resources are a major source of prior knowledge for dialogue models and in NLP tasks in
general. In the setting of data-efficient training, where a model is limited in what it can learn
examples, it is especially important to incorporate prior knowledge in it, in the form of e.g. an
ontology or a grammar. In dialogue systems, there exist approaches making use of linguistic
knowledge of various kinds~--- e.g. \cite{DBLP:conf/sigdial/RamachandranR15} proposed a method to
represent and track the dialogue state via Relational Trees built on top of Knowledge Base entities
extracted from the user's utterances; social dialogue model of \cite{Curry.etal2018} benefited from
the use of an ontology and entity linking as well.

In this section, we are going to give an overview of an approach to modelling dialogue entirely
based on explicit linguistic representations~--- specifically, a formal semantic grammar~--- and
discuss its applicability to low-resource dialogue system bootstrapping.

\begin{figure}[t]
\includegraphics{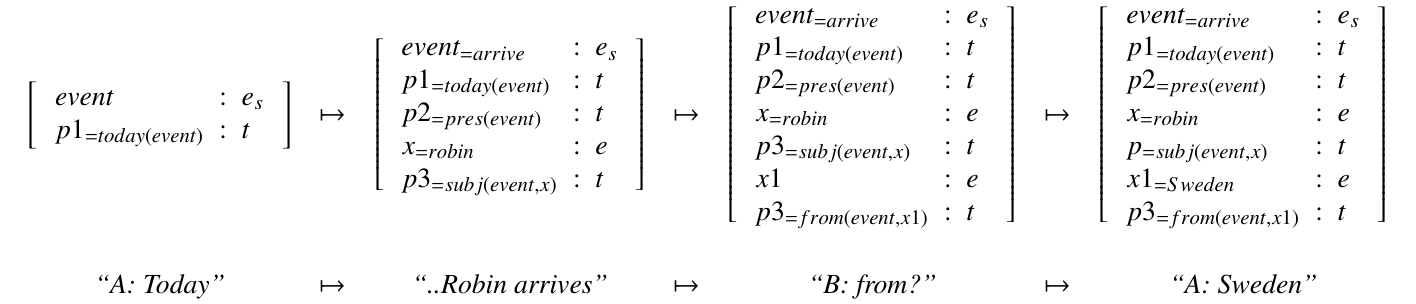}
\caption{Incremental parsing with \DSTTR/\dylan}\label{fig:subtype}
\end{figure}

\subsection{Dynamic Syntax and Type Theory with Records (\DSTTR)}
\label{dsttr}
\begin{figure}[t]
\includegraphics[width=\linewidth]{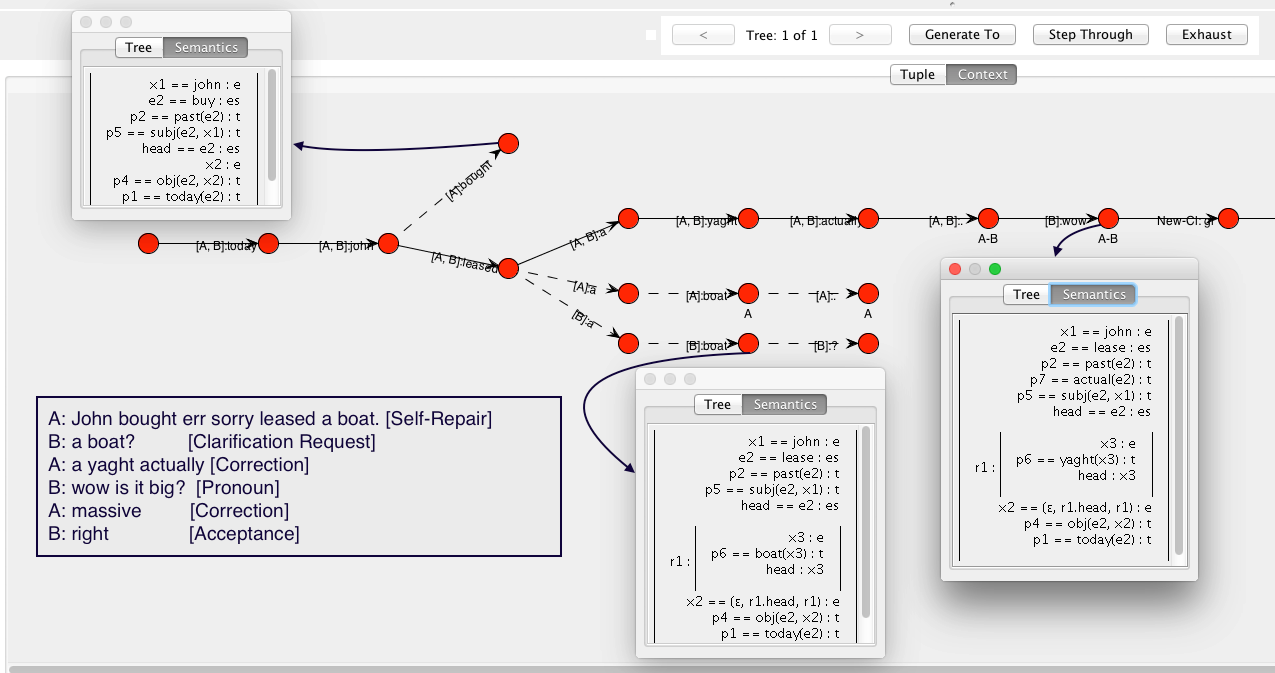}
\caption{Processing self-corrections and clarification requests with \DSTTR/\dylan}\label{fig:dag}
\end{figure}

Dynamic Syntax (DS) is an action-based, word-by-word incremental and semantic grammar formalism
(\citealp{Kempson.etal01}; \citealp{Cann.etal05a}), especially suited to the highly fragmentary and
context-dependent nature of dialogue. This formalism is implemented in the incremental semantic
parser for dialogue processing~--- \dylan\footnote{\dylan is derived from ``Dynamics of Language''}
(\citealp{Eshghi.etal11}; \citealp{Eshghi15}; \citealp{Purver.etal11}). In DS, words are conditional
actions~--- semantic updates, and dialogue is modelled as the interactive and incremental
construction of contextual and semantic representations \citep{Eshghi.etal15}~--- see Figure
\ref{fig:subtype}. The contextual representations provided by DS are fine-grained and jointly agreed
upon by the interlocutors, as a result of processing questions and answers, clarification
interaction, acceptances, self-/other-corrections, restarts, and other characteristic linguistic
phenomena in dialogue~--- see Figure~\ref{fig:dag} for an example of how self-corrections and
clarification requests are processed via a backtrack and search mechanism over the parse search
graph (\citealp{Hough11}; \citealp{Hough.Purver14}; \citealp{Eshghi.etal15}). Generation/surface
realisation in DS is defined using trial-and-error parsing (see Section~\ref{ch2:babble}), guided by
\emph{a generation goal}, i.e. the semantic representation of the utterance to be generated.
Generation thus proceeds word-by-word, similar to parsing (\citealp{Purver.etal14};
\citealp{Hough14}).

Therefore, DS allows to track not only the semantic content of the current turn while it is being
constructed word-by-word (during either parsing or generation) but also the context of the
conversation as a whole, with the grounded/agreed content of the conversation encoded as well,
see e.g. Figure~\ref{fig:encoding} (\citealp{Eshghi.etal15}; \citealp{Purver.etal10}).
Crucially for the dialogue model to be described below, the inherent incrementality of \DSTTR
together with the word-level, as well as cross-turn, parsing constraints it provides, enables
word-by-word exploration of the space of grammatical dialogues, and the semantic and contextual
representations that result from them.

Type Theory with Records (\TTR) is an extension of standard type theory used in semantics and
dialogue modelling (\citealp{Cooper05}; \citealp{Ginzburg12}). To support dialogue processing and
allow for richer representations of the dialogue context, \DS and the \TTR framework were integrated
(\citealp{Purver.etal10}; \citealp{Purver.etal11}; \citealp{Eshghi.etal12}). In \TTR, logical forms
are specified as \emph{record types} (RTs), sequences of \emph{fields} of the form
$\smttrnode{}{l&T}$ containing a label $l$ and a type $T$. RTs can be witnessed (i.e.~judged as
true) by \emph{records} of that type, where a record is a sequence of label-value pairs
$\smttrrec{}{l&v}$, and $\smttrrec{}{l&v}$ is of type $\smttrnode{}{l&T}$ in case $v$ is of type
$T$ (see Figure \ref{fig:subtype} for example record types).

A record type can be indefinitely extended, which naturally allows for representing incrementally
growing meaning representations as more words are parsed or generated. That is, the semantics of the
complete dialogue context expressed as an RT is a superset (or \textit{supertype}) of the semantics
midway in that same conversation. This is the key mechanism used in the dialogue system below.
Our linguistically informed approach in Chapter \ref{Chapter3} will be using this key \DSTTR
property as well.

\subsection{The \babble Dialogue Model}
\label{ch2:babble}

Incremental dialogue parsing described above combined with \RL is the essence of the \babble model.
Here, the parser acts as both \NLU providing semantic dialogue state representation and word-by-word
\NLG by `babbling' word sequences under the single shared grammar. \RL, in turn, works as a
trainable Dialogue Manager with word-by-word actions.

The two main resources for \babble are:
\begin{enumerate}[label=\alph*)]
  \item a \DSTTR parser $DS$~--- either learned from data \citep{Eshghi.etal13a} or constructed
  by hand~--- for incremental language processing and more generally, for tracking the context
  of the dialogue using the model of feedback of \cite{Eshghi.etal15,Eshghi15,Eshghi.etal11}; 
  \item a set $D$ of transcribed successful dialogues in the target domain. 
\end{enumerate}

In order to induce a fully incremental dialogue system from $D$, the following steps are performed:

\begin{enumerate}
  \item Automatically induce the \MDP state space, $S$, and the dialogue goal, $G_D$, from $D$;

  \item Automatically define the state encoding function $F: C \rightarrow S$, where $s \in S$
  is a binary state vector designed to extract from the current context of the dialogue,
  i.e. the semantic features observed in the example dialogues $D$; $c \in C$ is a \DS context,
  i.e. a pair of \TTR Record Types: $\langle c_{p}, c_{g}\rangle$ where $c_{p}$ is the content
  of the current clause (\textit{pending}) as it is being constructed but not necessarily fully
  grounded yet; $c_{g}$ is the content already jointly built and \textit{grounded} by the
  interlocutors (following the Dialogue Gameboard model of \citealp{Ginzburg12}).

  \item Define the \MDP action set as the $DS$ lexicon $L$ (i.e.\ actions are words);

  \item Define the reward function $R$ as reaching $G_D$, while minimising dialogue length. 
\end{enumerate}

The generated \MDP is then solved using \RL, with a standard Q-learning method: train a policy
$\pi : S \rightarrow L$, where $L$ is the \DS Lexicon, and $S$ the state space induced using $F$.
The system is trained in interaction with a semantic user simulation which is also automatically
built from the dialogue data and described in the next section.



\begin{figure}[t]
\centering
\includegraphics[width=\linewidth]{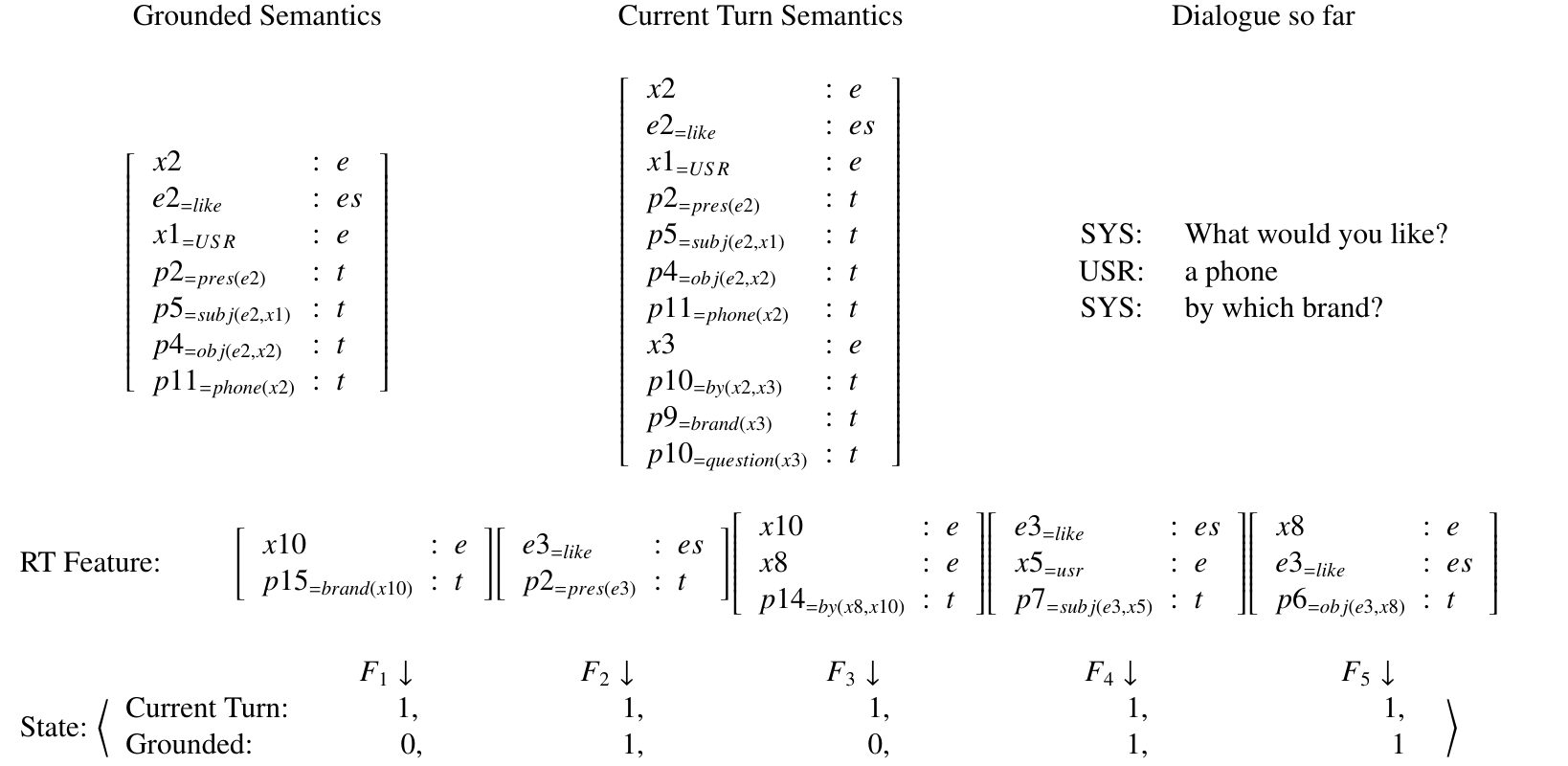}
\caption{Semantics to \MDP state encoding $F$ with RT features}
\label{fig:encoding}
\end{figure}

\textit{The state encoding function $F$}.
As shown in Figure \ref{fig:encoding} the \MDP state is a binary vector of size $2\times | \Phi |$,
i.e.\ twice the number of the RT features. The first half of the state vector contains the grounded
features (i.e.\ agreed by the participants) $\phi_i$, and the second half contains the current
semantics built incrementally in the current dialogue utterance. Formally:

\begin{equation}
  s = \langle F_1(c_{p}), \ldots, F_m(c_{p}), F_1(c_{g}), \ldots, F_m(c_{g})\rangle
\end{equation}

where $F_i(c)=1$ if $c\subtype\phi_i$, and 0 otherwise.

\subsection{Semantic User Simulation for \babble}
\label{ch2:usersim}

An \RL-based model, \babble requires a user simulator for efficient training. The simulator performs
two key tasks during training: (1) generating user turns in the right dialogue contexts;
and (2) word-by-word monitoring of the utterance so far generated by the system during exploration
(i.e.\ babbling grammatical word sequences) by the system. Both tasks use the $DS$ parser as
well the state encoding function $F$ described above.
They are thus performed based on the  \textit{semantic context} of the dialogue so far, as tracked
by $DS$. The simulator is sensitive to the order in which information is received from the user
since its context includes both the semantic features of the current turn and the history of the
conversation.

The rules required for (1) \& (2) are extracted automatically from the raw dialogue data $D$ using
$DS$ and $F$. The dialogues in $D$ are parsed and encoded using $F$ incrementally.
For (1), all the states that trigger the user into action, $s_{i}=F(c)$~--- where $c$ is a DS
context prior to any user turn~--- are recorded and mapped to what the user utters in those
contexts. For more than one training dialogue there may be more than one candidate (in the same
context/state). The rules extracted in this way will be of the form:
$$s_{trig} \rightarrow \{u_1, \ldots, u_n\}$$ where $u_i$ are user turns.

The $s_i$'s prior to the user turns also immediately follow system turns. Therefore, in order to
monitor the system's behaviour during training, one needs to check that the current state upon a
a word generated by the system subsumes (or is extendible to) one of the $s_i$. This is performed
through bitmask operation. The simulation can therefore semantically identify erroneous actions
(words) by the system. It would then terminate the learning episode and penalise the system
immediately which considerably speeds up training.

To recap, the \babble model described above involves incrementally parsing dialogues and encoding
the resulting semantics as state vectors in an \MDP, which is then used for \RL of word-level
actions for system output (i.e.\ a combined incremental \DM and \NLG module for the resulting
dialogue system). 

\section{Generalisation Power and Robustness of Dialogue Models}
\label{ch2:generalisation_robustness}

The key question in development machine learning models able to work with minimal amounts of
training data is, how well they generalise to the data unseen during training. Specific to the
dialogue systems we are going to work in this thesis, such novel properties of the data may be the
details of the dialogue task, e.g. conversations coming from a different domain or containing slot
types/values unseen in the training domains (see e.g. \citealp{DBLP:conf/acl/ZhaoZE17};
\citealp{DBLP:conf/slt/HendersonTW14}). It can also be some intrinsic property of the dataset itself,
e.g. the presence of spoken disfluencies, out-of-domain (\OOD) utterances, or just noise in
the data~--- which can all be considered anomalous input. We then say that a dialogue system able to
attain stable, consistently high performance across `clean' data and that containing anomalous
phenomena is {\it robust} to those. In this thesis, we are going to work with the following 2 types
of robustness, categorised by the specific phenomena in the input data:
\begin{itemize}[label=---]
  \item {\it robustness to disfluencies} is concerned with the surface variations in the input
  utterances appearing due to the nature of spoken language (see the next section for a detailed
  problem description). A system robust to disfluencies is expected to attain similar performance on
  `clean' data with those not present (e.g. examples collected in a controlled user study) as well
  as on more real-world conversations containing those phenomena (see a more detailed discussion in
  Section \ref{ch2:disfluency}). We will explore this problem in Chapters \ref{Chapter3} and
  \ref{Chapter6}
  \item {\it robustness to \OOD input} represents a similar system's quality, with the phenomena of
  interest being user's input turns not belonging to the system's designated domain, e.g. `put on
  my evening playlist' queried to a restaurant search system. We address this problem in a more
  specific way than the previous one, and so expect an \OOD-robust system to be able to (1)
  correctly identify anomalous inputs in the dialogues, and (2) attain a performance level on
  \OOD-containing data similar to that on purely in-domain (\IND) dialogues. That is, the system is
  supposed to produce the originally designated responses for the \IND turns as well as the special
  `fallback' response for \OOD turns signalising that it encountered anomalous input, with a minimal
  accuracy trade-off between the two (see an overview of the problem area in Section
  \ref{ch2:ood-robustness}). We will explore this problem in Chapter \ref{Chapter7}.
\end{itemize}

\subsection{Spoken Disfluencies and Data Efficiency}
\label{ch2:disfluency}


Humans process (parse and generate) language \emph{incrementally} word by word, rather than turn by
turn or sentence by sentence (\citealp{Howes.etal10}; \citealp{pickering_clifton_crocker_1999};
\citealp{Ferreira.etal04}). This leads to many characteristic phenomena in spontaneous dialogue that
are difficult to capture in traditional linguistic approaches and are still largely ignored by
dialogue system developers. These include various kinds of context-dependent fragments
(\citealp{Fernandez.Ginzburg02b}; \citealp{Fernandez06}; \citealp{Kempson.etal17}), false starts,
suggested add-ons, barge-ins, and disfluencies.

In this thesis, we are interested in the following disfluencies: pauses, hesitations, false starts,
and self-corrections~--- that are common in natural spoken dialogue. These proceed according to a
well-established general structure with three phases \citep{Shriberg94}:

\begin{center}
with $\underbrace{[\textrm{Italian}}_{reparandum}+\underbrace{\{uh\}}_{interregnum}
\underbrace{\textrm{Spanish}]}_{repair}$ \ \ cuisine
\label{repair}
\end{center}

Specific disfluency structures have been shown to serve different purposes for both the speaker and
the hearer \citep{Brennan.Schober01}~--- for example, a filled pause such as `uhm' can elicit a
completion from the interlocutor, but also serve as a turn-holding device; mid-sentence
self-corrections are utilised to deal with the speaker's own error as  early as possible, thus
minimising effort.

In dialogue systems, the detection, processing, and integration of disfluency structures is crucial
to understanding the interlocutor's intended meaning (i.e. robust \NLU), but also for coordinating
the flow of the interaction.
Like dialogue processing in general, the detection and integration of disfluencies needs to be
\emph{strongly incremental}: it needs to proceed word by word, enabling downstream processing to
begin as early as possible, leading to more efficient and more naturally interactive dialogue
systems (\citealp{Skantze.Hjalmarsson10}; \citealp{Schlangen.Skantze09}).

Furthermore, incremental disfluency detection needs to proceed with minimal latency and commit to
hypotheses as early as possible in order to avoid `jittering' in the output and having to undo the
downstream processes started based on erroneous hypotheses (\citealp{Schlangen.Skantze09};
\citealp{DBLP:conf/emnlp/HoughP14}; \citealp{DBLP:conf/interspeech/HoughS15}).

While many current data-driven dialogue systems tend to be trained end-to-end on natural data, they
do not normally take the existence of disfluencies into account.
The problem is that, taken together with the particular syntactic and semantic contexts in which
they occur, disfluencies are very sparsely distributed, which leads to a large mismatch between the
training data and actual real-world spontaneous user input to a deployed system. This suggests a
more modular, pipelined approach, where disfluencies are detected and processed by a separate,
domain-general module, and only then any resulting representations are passed on for downstream
processing. The upshot of such a modular approach would be a major advantage in generality,
robustness, and data-efficiency.

\subsubsection{Incremental Disfluency Detection Models}

Work on disfluency detection has a long history, going back to \cite{Charniak.Johnson01} who set the
challenge. One of the important dividing lines through this work is the \emph{incrementality}
aspect, i.e.\ whether disfluency structure is predicted word by word.

In the non-incremental setting, as the problem is essentially sequence tagging, neural models have
been widely used. As such, there are approaches using an encoder-decoder \SeqToSeq model with
attention \citep{DBLP:conf/coling/WangCL16} and a Stack-\LSTM model working as a buffer of a
transition-based parser (\citealp{DBLP:conf/coling/WangCL16}; \citealp{DBLP:conf/emnlp/WangCZZL17}),
the latter attaining superior results in the non-incremental setting.

Incremental, online processing of disfluencies is a more challenging task, if only because there is
much less information available for tagging, i.e. only the context on the left. In a practical
system, it also involves extra constraints and evaluation criteria such as minimal latency and
revisions to past hypotheses which lead to `jittering' in the output with all the dependent
downstream processes having to be undone, thus impeding efficiency (%
\citealp{DBLP:conf/emnlp/HoughP14}; \citealp{Purver.etal18}).

Incremental disfluency detection models include \cite{DBLP:conf/emnlp/HoughP14} who approach the
problem information-theoretically, using local surprisal/entropy measures and a pipeline of
classifiers for recognition of the various components of disfluency structure. While the model is
very effective, it leaves one desiring a simpler alternative. This was made possible after the
overall success of RNN-based models, which \cite{DBLP:conf/interspeech/HoughS15} exploit. In Chapter
\ref{Chapter6}, we will build on top of this model, as well as evaluate it further.

Disfluency detection was also addressed in a multitask fashion by \cite{DBLP:conf/eacl/SchlangenH17}
whose secondary task is utterance segmentation~--- they demonstrate that the two tasks interact and
thus are better approached jointly.

Language models have been extensively used for improving neural models' performance. For example,
\cite{DBLP:conf/naacl/PetersNIGCLZ18} showed that a pre-trained language model improves
RNN-based models' performance in a number of \NLP tasks~--- either as the main feature
representation for the downstream model, or as additional information in the form of a latent vector
in the intermediate layers of complex models. The latter way was also employed by
\cite{DBLP:conf/acl/PetersABP17} in the task of sequence labelling.

Finally, a multitask setup with language modelling as the second objective~--- the closest to our
approach in Chapter \ref{Chapter6}~--- was used by \cite{DBLP:conf/acl/Rei17} to improve the
performance of \RNN-based Name Entity Recognition. The \LM part of their model predicts both
surrounding words of an input word which is done using a bidirectional \LSTM (the forward one
predicts the next word, and the backward one predicts the previous word). In our task, we can not
make use of a backward model as we work in the word-by-word incremental fashion.

We note that there is no previous approach to multitask disfluency detection using a secondary task
as general and versatile as language modelling. Furthermore, none of the works mentioned study how
well their models {\it generalise} across datasets of different dialogue types, nor do they shed
much light on what kinds of disfluency structure are harder to detect, and why.

\subsection{Out-of-Domain Robustness and Data Efficiency}
\label{ch2:ood-robustness}

Data-driven approaches to dialogue systems development offered by the common bot building platforms
(e.g. Google Dialogflow, Amazon Alexa Skills Kit, Microsoft Bot Framework) make it possible for a
wide range of users to easily create dialogue systems with a limited amount of data in their domain
of interest (e.g. restaurant search, travel booking, city info). Most task-oriented dialogue systems
are built for a closed set of target domains, and in the setting of a low amount of in-domain
training data, this leads to overfitting of machine learning methods and unpredictable performance
outside their training sets. For a closed-domain dialogue system, it is extremely important to
maintain predictable behaviour, and any failure to detect \OOD utterances and respond with an
appropriate fallback action\footnote{See an example of the Alexa Skills Kit's built-in fallback
action on \colortexthref{blue}{http://tiny.cc/alexa_ood_intent}{Amazon Developer Blogs}.} can lead
to a frustrating user experience. In the setting of working with minimal training data, the latter
is especially relevant since there is no access to `real' \OOD examples.

There have been a set of prior approaches for OOD detection which require both {\em in-domain}
(\IND) and \OOD data (\citealp{nakano2011two}; \citealp{tur2014detecting}). However, it is a
formidable task to collect sufficient data to cover in theory an unbounded variety of \OOD
utterances. In contrast, \cite{lane2007out} introduced an in-domain verification method that
requires only \IND utterances. Later, with the rise of deep neural networks, \cite{ryu2017neural}
proposed an autoencoder-based \OOD detection method which surpasses prior approaches without access
to \OOD data. However, those approaches still have some restrictions such that there must be
multiple sub-domains to learn utterance representation, and one must set a decision threshold for
\OOD detection. This can prohibit these methods from being used for most systems that focus on a
single task. Moreover, recently, it was shown that density estimation models like autoencoders lack
stability in telling between in-distribution and out-of-distribution data
\citep{DBLP:conf/iclr/NalisnickMTGL19}~--- and a standalone autoencoder does not suffice for our
task, as we are going to demonstrate empirically in Chapter \ref{Chapter7}. There, we will focus on
studying the effect of \OOD input on goal-oriented dialogue models' performance and propose a simple
and efficient solution for improving their robustness only using \IND data.

\section{Dialogue Datasets and Data Collection}
\label{ch2:data_collection}

Deep learning methods described above perform at their best when provided with large amount of
training data. In case of the \NLP field in general, the main sources of general-purpose data are
large-scale web resources: Wikipedia, online news resources, and posts on social networks (e.g.
Reddit, Twitter). For dialogue, different kinds of datasets are used given the system type.
Chat-oriented systems aimed at eliciting human-like open-domain conversation can be trained from
large conversational (or conversation-like, e.g. comment threads on message boards) corpora. As
such, the following datasets were used for training \SeqToSeq conversation models (a wider review
can be found in \citealp{DBLP:journals/dad/SerbanLHCP18}):

\begin{itemize}[label=---]
  \item Cornell Movie Dialogs Corpus \citep{Danescu-Niculescu-Mizil+Lee:11a}~--- over 300,000 total
  utterances,
  \item OpenSubtitles \citep{DBLP:conf/lrec/LisonT16}~--- 400 million subtitle lines,
  \item Reddit conversations \citep{DBLP:journals/corr/abs-2001-08435}~--- over 3.7 billion comments,
  \item Twitter \citep{sordoni-etal-2015-neural}~--- 29 million
  `context-message-response' triples\footnote{Larger datasets can be obtained from the
  \colortexthref{blue}{https://archive.org/details/twitterstream}{Twitter Stream archives}}.
\end{itemize}

The intuition behind using movie subtitle corpora is that movie or TV series dialogues contain
everyday conversations, and given enough coverage, it's theoretically possible to obtain open-domain
chatting behaviour by mimic situations from the movies as well as learn to generalise over them to
a certain degree.

In goal-oriented dialogue, the datasets used are more domain-specific. Some of the most widely-known
are:

\begin{itemize}[label=---]
  \item `Let's Go' \citep[DSTC1,][]{DBLP:conf/sigdial/WilliamsRRB13}~---15,000 dialogues in the bus
  information domain,
  \item Cambridge restaurants dataset (DSTC 2---3\citealp[,][]{DBLP:journals/dad/WilliamsRH16a})
  with 3,000 dialogues in the domain of restaurant search,
  \item Stanford Multi-Domain (\SMD) dialogue dataset \citep{DBLP:conf/sigdial/EricKCM17} with 3,000
  dialogues in 3 goal-oriented domains, namely in-car navigation, weather information, and
  appointment scheduling,
  \item \multiwoz \citep{DBLP:conf/emnlp/BudzianowskiWTC18,eric2019multiwoz,zang2020multiwoz}~--- a
  multi-domain, multi-task goal-oriented dataset with 10,000 dialogues. Domains represented in
  \multiwoz are restaurant, hotel, taxi, police, attraction, train, and hospital,
  \item \textsc{Frames} \citep{DBLP:conf/sigdial/AsriSSZHFMS17} with 1369 dialogues in the travel
  information domain addressing complex user's goals and more advanced real-world scenarios beyond
  linear form-filling.
  \item \metalwoz~--- the dataset collected for \DSTC[8] Track 2 ``Fast Domain Adaptation''
  \citep{lee2019multi-domain}. with more than 37,000 human-human dialogues spanning the total
  of 227 tasks in 47 domains. The dialogues are collected in a way that human participants were
  assigned the role of bot or user, then given a problem domain and related specific task, and
  instructed to reach the user's goal over at least 10 dialogue turns.
\end{itemize}

\begin{figure}[t]
  \centering
  \includegraphics[width=0.8\textwidth]{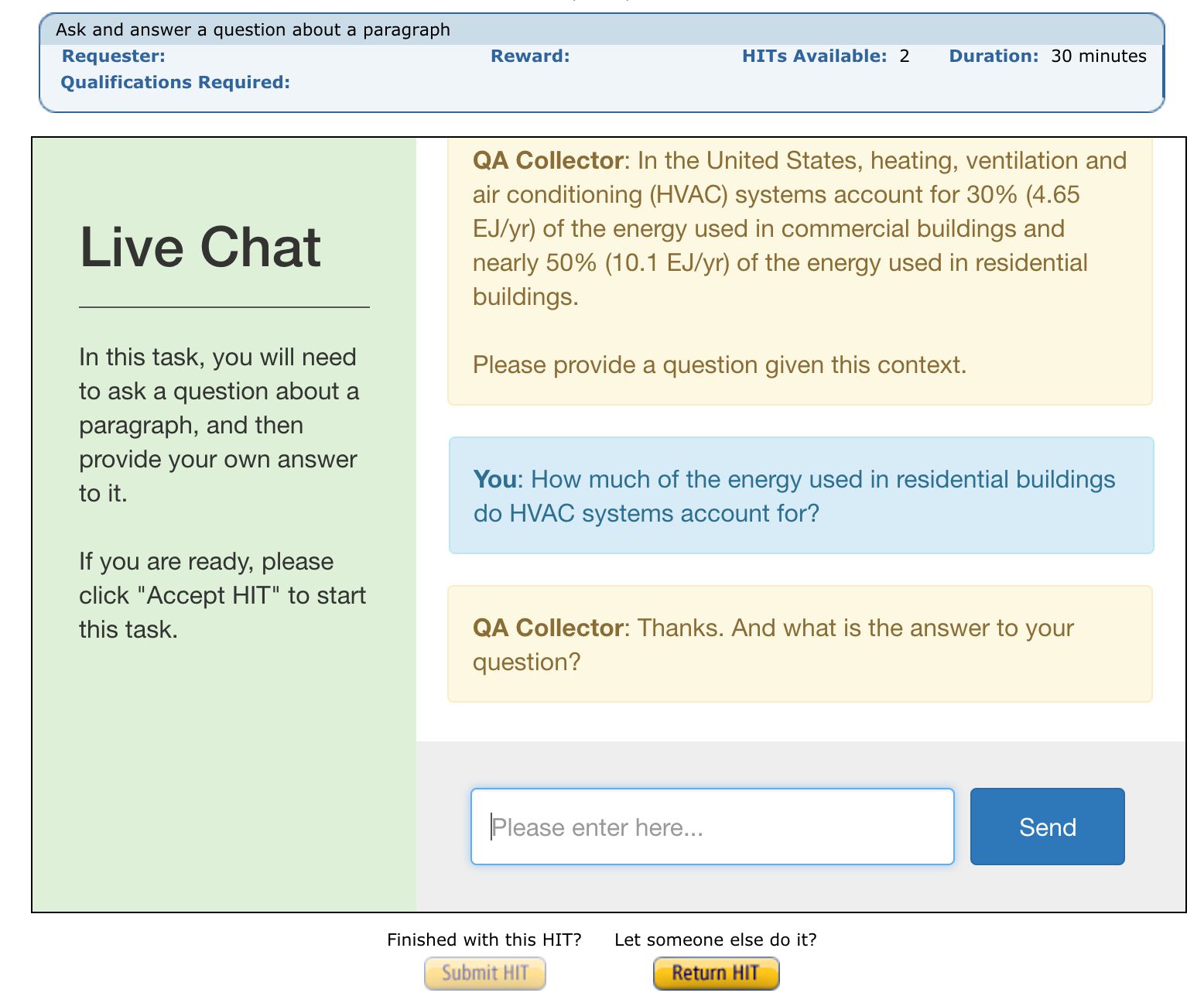}
  \caption[\ParlAI web interface for Wizard-of-Oz data collection]{\ParlAI web interface for
  Wizard-of-Oz data collection \citep{DBLP:conf/emnlp/MillerFBBFLPW17}}
  \label{fig:parlai_mturk}
\end{figure}

Apart from the openly available testbeds, the data for domain-specific dialogue scenarios is
normally collected via a technique called Wizard-of-Oz
\citep[\WOz,][]{10.5555/286013.286055}, where two humans interact with each other, one acting as the
user and the other one simulating the behaviour of the potential dialogue system. Historically, this
approach was used to conduct user experience studies, although in case of machine learning-based
dialogue systems it is used as the \textit{seed} data for training the prototype system (also
referred to as \textit{bootstrapping}) for further fine-tuning from real interactions.

With training data being the principal asset in modern dialogue system development, it has become of
key importance to incorporate data collection into the development pipeline in a principled way.
Specifically, crowdsourcing platforms like Amazon Mechanical Turk (\AMT)%
\footnote{\colorhref{blue}{https://www.mturk.com}} and \textit{Figure Eight}%
\footnote{\colorhref{blue}{https://www.figure-eight.com}} gained wide adoption for collecting
real-user data. Correspondingly, dialogue system frameworks and solutions introduced recently were
designed with \AMT integration in mind. For example, \cite{DBLP:conf/eacl/Rojas-BarahonaG17}
introduce their end-to-end trainable approach along with the \WOz framework for data collection on
\AMT. Moreover, the \ParlAI conversational platform \citep{DBLP:conf/emnlp/MillerFBBFLPW17} provides
seamless \AMT integration for \WOz data collection as one of its key features~--- the web interface
for \WOz interactions is shown in Figure \ref{fig:parlai_mturk}.

We observe that one of the key directions in conversational systems research is providing means to
collect datasets of moderate amount in a principled way for rapid prototyping or bootstrapping
dialogue systems. Still, the less are data needed for the system to perform reasonably well, the
more flexible the system gets for use outside the academic testbeds. Therefore, advancing the
training techniques for less dependence on data is of a high priority in dialogue systems research.
In the next chapter, we are going to start our study on dialogue data efficiency by comparing two
fundamentally different approaches to dialogue: linguistically informed models based on dialogue
grammars and neural response retrieval models (discussed in Sections \ref{ch2:linguistic} and
\ref{ch2:retrieval} of this chapter, respectively).

\chapter{Linguistic Knowledge or Learning from Examples: A Data Efficiency Perspective}

\label{Chapter3} 

\lhead{Chapter 3. \emph{Linguistic Knowledge or Learning from Examples}} 

We are going to start our research with an experimental study of dialogue models' generalisation
power and robustness in the low-resource setup, i.e. in the task of bootstrapping a dialogue system
from seed data. We perform the experiments in a controlled environment using the \bAbI Dialog Tasks
dataset \citep{DBLP:conf/iclr/BordesBW17} and focus our attention on two fundamentally different
types of models: a neural retrieval-based model \memnn (\cite{Sukhbaatar15}, discussed in the
previous chapter, and a linguistically informed model based on a semantic parser/generator \dylan.
We look at their performance in the limited data setup with bAbI as well as their generalisation
potential to more diverse and challenging input~--- for that, we introduce \bAbIplus\footnote{%
Available at \colorhref{blue}{https://bit.ly/babi\_plus}}, an augmented version of the \bAbI dataset
with increased surface complexity represented by simulated spoken disfluencies.


\section{Motivation}
\label{ch3:intro}

Every practical machine learning model represents a trade-off between what is \textit{learned} from
data and what is \textit{given} in the form of inductive biases. The range of possible inductive
biases spans from the fundamental ones, e.g. specific network architectures tailored to different
types of input data (\CNNs for image input, \RNNs for sequential input) to task-specific objective
functions that those networks minimise. In dialogue, as well as in \NLP in general, the main sources
of inductive biases are linguistic resources, e.g. gazetteers, ontologies, thesauri, and knowledge
bases. Now, in the setting of minimal training data, the extent of what can be learned from it is
limited very severely, so those linguistic resources and the corresponding inductive biases become
the main way to obtain a generalisable model.

In the dialogue systems area, there are currently several key problems for the practical data-driven
development of task-oriented systems, among them: (1) large amounts of dialogue data are
needed, i.e.\ thousands of examples in a domain; (2) this data is usually required to be annotated
with task-specific semantic information for the domain (e.g. various dialogue act schemes); and (3)
the resulting systems are usually trained from the data that do not properly represent many
characteristic phenomena of dialogue such as spoken disfluencies.

In overcoming issue (2), a recent advance in research on chat-oriented dialogue was the development
of end-to-end systems, in which all components are trained from textual dialogue examples, e.g.
\cite{sordoni-etal-2015-neural,DBLP:journals/corr/VinyalsL15}.
However, as \cite{DBLP:conf/iclr/BordesBW17} argued, these end-to-end methods may not transfer well
to task-based settings (where the user is trying to achieve a domain goal, such as booking a flight
or finding a restaurant, resulting in an \API call).
They then presented an end-to-end method using Memory Networks (\memnns), which achieves 100\%
performance on a testset of 1000 dialogues, after being trained on 1000 training dialogues.
This method processes dialogues turn-by-turn, and so does not have the advantages of more natural
incremental systems (\citealp{aistincremental}; \citealp{Skantze.Hjalmarsson10}); nor does it really
perform language generation, rather it is based on a retrieval model that selects from a set of
candidate system responses seen in the data.

In this chapter, we will investigate two fundamentally different approaches: (1) \memnn, a neural
retrieval-based model of \cite{Sukhbaatar15}, and (2) a linguistically informed model that uses an
incremental semantic parser/generator for dialogue based around the Dynamic Syntax grammar
formalism~--- described in Sections \ref{ch2:retrieval} and \ref{ch2:babble}, respectively.
Specifically, we explore how well these methods overcome the requirement for large amounts (i.e.
thousands of dialogues in a domain) of annotated dialogue data by putting them in a setup with up to
5 example dialogues from bAbI, Task 1. Then, in order to evaluate the systems' robustness to the
unseen but highly likely variations in the user's input, we also introduce an extended, incremental
version of the \bAbI dataset, which we call \bAbIplus (see Section \ref{ch3:babi+}): it adds some
characteristic phenomena of spoken language~--- such as mid-utterance self-corrections~--- to the
bAbI dialogues. Using this, we further experiment with the two systems and see how they are able
process this more challenging data.

\section{Experimental Framework~--- \bAbI Dialog Tasks}
\label{ch3:babi}

\begin{figure}
  \centering
  \includegraphics[width=0.85\textwidth]{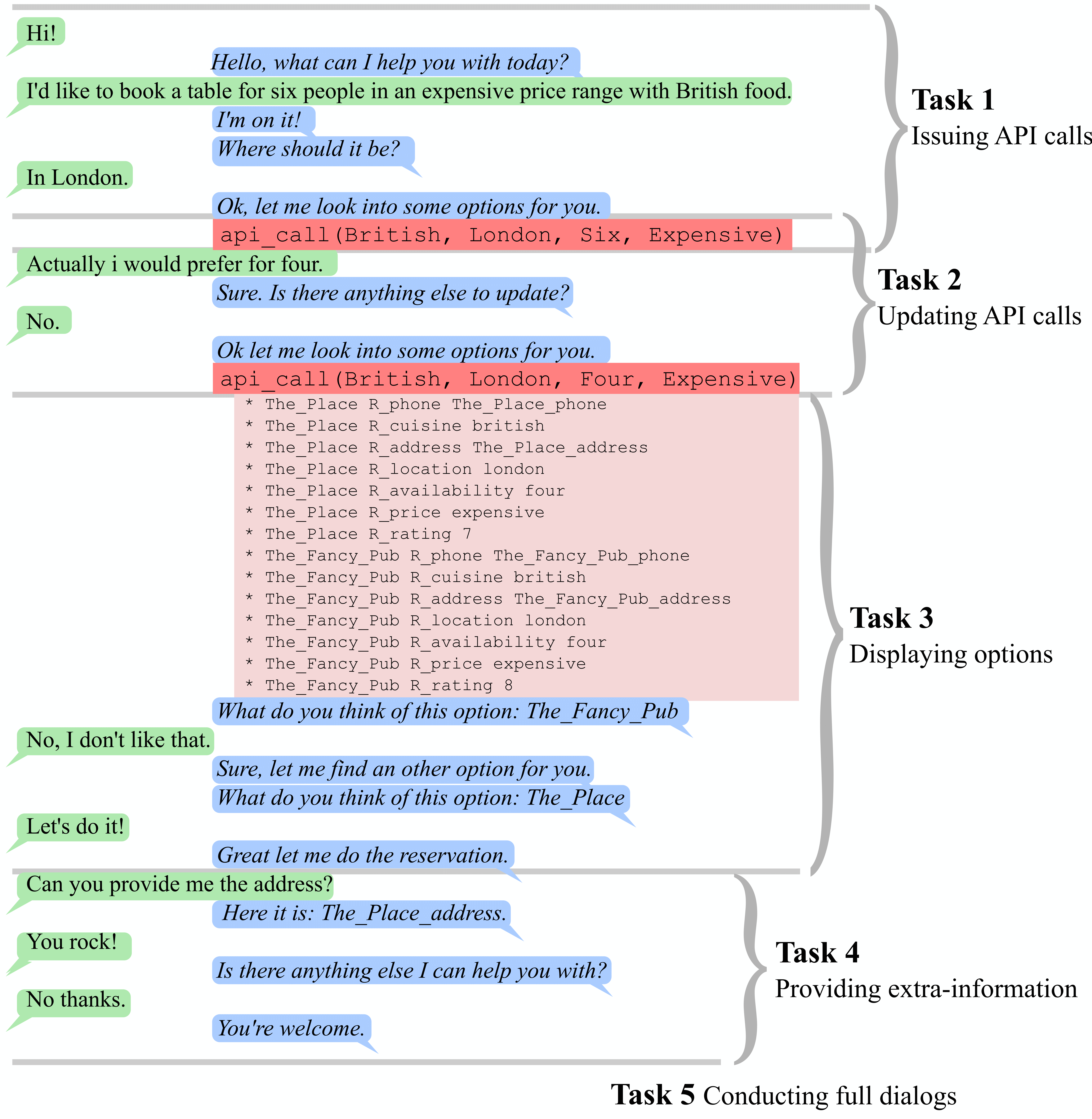}
  \caption[\bAbI Dialog Tasks structure]{\bAbI Dialog Tasks structure~--- shown are the synthetic
  Task1---Task 5 \citep{DBLP:conf/iclr/BordesBW17}}
  \label{fig:babi_dialog_tasks}
\end{figure}

In this chapter, our focus is not on building dialogue systems, but on: \textbf{(1)}
studying and quantifying the interactional and structural generalisation power of the \DSTTR grammar
formalism and that of symbolic, grammar-based approaches to language processing more
generally. We focus here on specific dialogue phenomena, such as mid-sentence self-corrections,
hesitations, and restarts (see below);
\textbf{(2)} doing the same for \citeauthor{DBLP:conf/iclr/BordesBW17}'s response retrieval model
\memnn, without the use of linguistic knowledge of any form; and \textbf{(3)} comparing (1) and (2).

In order to test and quantify the interactional and structural generalisation power of the two
models, we need contrasting dialogue datasets that control for interactional vs lexical/syntactic
variations in the input dialogues. Furthermore, to make our results comparable to the existing
approach of \cite{DBLP:conf/iclr/BordesBW17}, we need to use the same dataset that they have used.
We therefore use Facebook AI Research's \bAbI Dialog Tasks dataset
\citep{DBLP:conf/iclr/BordesBW17}. These are goal-oriented dialogues in the domain of restaurant
search. In the dataset, there are 6 tasks of increasing complexity ranging from only collecting the
user's preferences on restaurant and up to conducting full dialogues with changes in the user's goal
and providing extra information upon request~--- see Figure \ref{fig:babi_dialog_tasks} for an
illustration. The first 5 tasks are `clean' dialogues composed synthetically and they thus lack the
features of natural everyday conversations. Task 6 (not shown in the figure) is the natural
counterpart of the Task 5, containing dialogues with human users from the Dialog State Tracking
Challenge 2.

After the original \citeauthor{DBLP:conf/iclr/BordesBW17}'s result on Task 1, several studies have
shown different ways in which \memnns are outperformed: \cite{DBLP:conf/eacl/ManningE17} introduced
the Copy-Augmented Sequence-to-Sequence model that outperforms the \memnn on Task 6;
\cite{DBLP:conf/acl/WilliamsAZ17} presented Hybrid Code Networks (discussed in the previous chapter),
a combined \RNN/rule-based model trainable in a 2-stage supervised + reinforcement learning setup,
outperforming the \memnn on Tasks 5 and 6.

However, none of these studies control for {\it the type of complexity} that might result in worse
performance, and thus do not shed any light on why a particular architecture such as \memnn might
be at a disadvantage. While Task 5 dialogues have the full task complexity, conducting full
dialogues with an unfixed user goal and additional information requests, they are still composed
programmatically and contain minimal surface variation. The Task 6 dialogues on the other hand are
complex both in terms of the surface variation and the task itself.

In order to study the specific effects of incremental variations in dialogue such as conversational
disfluencies, we focus on Task 1, where in each dialogue the system asks the user about their
preferences for the properties of a restaurant, and each dialogue results in an \API call containing
values of each slot obtained (e.g.\ {\tt food-type=french})~--- the ability of predicting the \API
calls correctly thus provides a direct measure of how well a particular model can interpret the
dialogues. We would like to point out that we will be using the synthetic part of the \bAbI Dialog
Tasks dataset as a controlled experimental environment, and in the next section, we are going to
present our modifications that we apply to this dataset in a programmatic way~--- similar to how the
initial corpus was created~--- in order to simulate certain linguistic phenomena of interest.

\section{The \bAbIplus Dataset}
\label{ch3:babi+}
The original bAbI dialogues were synthesised in a way that their main source of complexity is the
dialogue goal itself, with its challengingness increasing from Task 1 to Task 5. In addition, they
also contain some basic lexical/syntactic variation, e.g. ``may i have a table with {\tt <cuisine>}
cuisine in a {\tt <price>} price range in {\tt <place>}?'', ``can you make a restaurant reservation
with {\tt <cuisine>} cuisine in a {\tt <price>} price range in {\tt <location>}?'' However, Task 1
dialogues significantly lack any simulation of incremental and interactional variations vital for
real-life dialogues. In order to obtain such variation while keeping the controllable environment
close to the laboratory conditions that bAbI offers, we created the bAbI+ dataset by systematically
transforming the original dataset's dialogues. 


\bAbIplus is an  extension of the \bAbI Task 1 dialogues with disfluent dialogue phenomena
(hesitations, restarts, and corrections~--- see below). This extension can be seen as orthogonal to
the increasing task complexity which Tasks 2---5 offer: we instead increase the complexity of
surface forms of dialogue utterances, while keeping every other aspect of the task fixed.

Our modifications model the disfluencies and communication problems in everyday spoken interaction
in real-world environments.
These variations are:
\begin{itemize}[label=\textbf{---}]

\item \textit{Hesitations}, e.g.\ as in ``we will be \texttt{uhm} eight''; 

\item \textit{Restarts}, e.g.\ ``can you make a restaurant \texttt{uhm yeah can you make a
restaurant} reservation for four people with french cuisine in a moderate price range'';

\item \textit{Corrections} affecting task-specific information~-- both short-distance ones
correcting one token, e.g.\ ``with french \texttt{oh no spanish} food'', and long-distance
NP/PP-level corrections, e.g. ``with french food \texttt{uhm sorry with spanish food}'', all within
a single user utterance, rather than across multiple turns.

\end{itemize}

The phenomena above are mixed in probabilistically (with the aim to reflect the statistics from
\citealp{Hough14}) from the fixed sets of templates to the original data\footnote{See
\colorhref{blue}{https://github.com/ishalyminov/babi\_tools}}.
The modifications affect a total of 11,336 utterances in the 3998 dialogues. Around 21\% of user
turns contain corrections, 40\% hesitations, and 5\% restarts (they are not mutually exclusive, so
that an utterance can contain up to 3 modifications). Our modifications, with respect to corrections
in particular, are more conservative than those observed in real-world data: \cite{Hough14} reports
that self-corrections appear in 20\% of all turns natural conversations from British National Corpus,
and in 40\% of turns in the Map Task, a corpus of human-human goal-oriented dialogues.

Here is part of an example dialogue in the \bAbIplus corpus showing some of the augmentations
(\colorbox{yellow}{hesitations} and \colorbox{pink}{corrections}) in the user's turns:

\begin{small}
\begin{tabular}{@{\hspace*{1.8em}}ll@{}}
\textbf{sys:}&	hello what can I help you with today?\\
\textbf{usr:}&	I'd like to book a \colorbox{yellow}{uhm yeah} I'd like to book a table\\
&in a expensive price range \colorbox{pink}{no sorry} in a cheap price range\\
\textbf{sys:}&	I'm on it. Any preference on a type of cuisine?\\
\textbf{usr:}&	with indian food \colorbox{pink}{no sorry} with spanish food\\
\textbf{sys:}&	where should it be?\\
\textbf{usr:} &	in tokyo\\
\textbf{sys:}&	how many people would be in your party?\\
\textbf{usr:}&	for \colorbox{yellow}{uhm yeah} for eight people please\\
\textbf{sys:}&	ok let me look into some options for you\\
\end{tabular}
\end{small}

\section{Models}
\label{ch3:models}

\subsection{\memnn}

We follow \citeauthor{DBLP:conf/iclr/BordesBW17}'s setup  by using a MemN2N (we took an open source
Tensorflow implementation for \bAbI \QA tasks and modified it%
\footnote{See \colorhref{blue}{https://github.com/ishalyminov/memn2n}} according to their setup~---
see details below).
For the description of the model architecture, please see Chapter \ref{Chapter2}. We're initially
interested in the single-hop configuration with which \citeauthor{DBLP:conf/iclr/BordesBW17} achieve
perfect accuracy on \bAbI Task 1.

In order to adapt the data for \memnn, we transform the dialogues into \texttt{$<$story, question,
answer$>$} triplets. The number of triplets for a single dialogue is equal to the number of the
system's turns, and in each triplet, the {\tt answer} is the current system's turn, the
{\tt question} is the user's turn preceding it, and the {\tt story} is a list of all the previous
turns from both sides. Other than that, each sentence in the {\tt story} gets 2 additional tokens:
the number of the turn, and the ID of the speaker \citep{DBLP:conf/iclr/BordesBW17}.

We also use the single embedding matrix $A$ for both input memories and the user's question; the
same matrix is used for the output memories representation~--- in that we follow
\cite{DBLP:conf/iclr/BordesBW17}, and it corresponds to the ``Adjacent'' weight tying model in
\cite{Sukhbaatar15}.

In our setup, there are no out-of-vocabulary words for the model during both training and testing,
and for both \bAbI and \bAbIplus with the maximum sentence length taking account of the increase due
to the transformations in \bAbIplus.

We train our \memnns with an \SGD optimiser for 100 epochs with a learning rate of 0.01 and a batch
size of 8~--- in this we again follow the configuration reported by \cite{DBLP:conf/iclr/BordesBW17}
to be the best for \bAbI Task 1.




\subsection{\dylan: \bAbI and \bAbIplus Setup Details}

Although \dylan's Dynamic Syntax grammar is learnable from data, the existing learned models in the
prior work (\citealp{Eshghi.etal13a}; \citealp{Eshghi.etal13b}), were induced from a corpus of
child-directed utterances, and there were some constructions as well as individual words that the
resulting lexicons did not include. We therefore extended this induced grammar manually to cover the
bAbI dataset, which, despite not being very diverse, contains a wide range of complex grammatical
constructions, such as long sequences of prepositional phrases, adjuncts, short answers
to yes/no and wh-questions, appositions of NPs, causative verbs etc~--- and all of this within and
across dialogue turns/speakers. Using \dylan, we parsed all dialogues in the \bAbI train and test
sets, as well as on the \bAbIplus corpus word-by-word, including both user and system utterances, in
context. The grammar parses 100\% of the dialogues, i.e. it does not fail on any word in any of the
dialogues.

Our aim here is to assess the ability of a \dylan-based system to generalise from small data and
compare this to the results of the \memnn in \cite{DBLP:conf/iclr/BordesBW17}. The latter is based
on the retrieval of system responses given the dialogue history up to that point. Therefore, for
direct comparison, and for simplicity of exposition, we set up an experimental testbed extending the
semantic parser in the following way: we employ the logic originally presented as the \babble user
simulation (Section \ref{ch2:usersim}), this time for the \textit{system side}, resulting in a
`system simulation'. We then use this to predict a system response, by parsing and encoding the
containing test dialogue up to the point immediately prior to the system turn. This results in a
triggering state $s_{trig}$, which is then used as the key to look up the system's response from the
rules constructed as per Section \ref{ch2:usersim}. The returned response is then parsed
word-by-word as normal, and this same process continues for the rest of the dialogue. This method
uses the full machinery of \DSTTR and our state-encoding method and will thus reflect the
generalisation properties that we are interested in.

Our overall method described in Section \ref{ch2:usersim} respects the turn ordering encountered in
the data, or more generally the order in which semantic increments are added to context. This is
because states are composed not only of the semantic features of the current turn, but also those of
the conversation history. And thus they capture \textit{the contextual boundary} at which a user
turn is being generated or a system turn monitored (e.g.\ in the \bAbI `restaurant search' domain, a
state might capture the fact that the user has already provided the cuisine type and the location of
the restaurant).

Unlike many other approaches to goal-oriented dialogue, \dylan-based approach is not based on
dialogue acts, and is word-by-word incremental. This means that there is no given/prior definition
of what sequences of words or semantic updates constitute a dialogue turn: the system needs to 
learn this (or, in general, automatically construct from the data), or else it would just go on
generating (i.e. exploring grammatical word outputs) without ever stopping. We prevent this by
having the simulator interrupt the system at the \textit{semantic} turn and clause boundaries that
are encountered in the data; and thus the behaviour of the simulator determines the system's turn
boundaries.


\section{Experiments}
\label{ch3:experiments}

\subsection{Experiment 1: Generalisation from Small Data}
\label{ch3:comparison}

We have now set out all we need to perform the first experiment. Since we are here interested in
both (1) data efficiency and (2) robustness, we use all the bAbI and \bAbIplus data~--- the train,
dev, and test sets~--- in the cross-validation setup as follows: we train the \memnn as well as
construct \dylan's semantic context-response mapping from 1---5 examples selected at random from
\textit{the longest dialogues} in \bAbI (note \bAbIplus data is never used for training in this
experiment). This process is repeated across 10 folds. The models are then tested on sets of 1000
examples selected at random, in each fold. Both the training and test sets constructed in this way
are kept constant in each fold across \dylan and \memnn. The test sets are selected either
exclusively from \bAbI or exclusively from \bAbIplus.

\subsubsection{Results: Predicting System Turns}

\begin{figure}[t]
  \includegraphics[width=\linewidth]{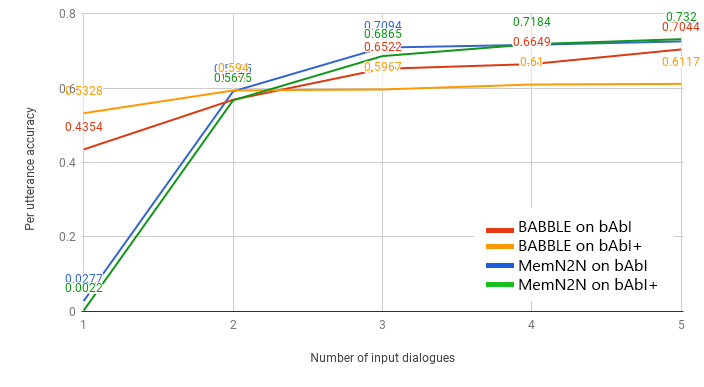}
  \caption{Few-shot performance of \dylan and \memnn}
  \label{fig:dylan_vs_memnn}
\end{figure}

Figure~\ref{fig:dylan_vs_memnn} shows per-utterance accuracies for the \dylan and \memnn models.
Per-utterance accuracy is the percentage of all system turns in the test dialogues that were
correctly predicted. The table shows that \dylan can generalise to 74\% of \bAbI and 65\% of \bAbIplus
with only 5 input dialogues from \bAbI. It also shows that \memnns can also generalise remarkably
well. Although as discussed below, this result is misleading on its own as the \memnns are very
poor at generating the final \API calls correctly on both the \bAbI and \bAbIplus data, and are thus
making too many semantic mistakes.

\subsection{Experiment 2: Semantic Accuracy}
\label{ch3:sem-acc}

The results from Experiment 1 on their own can be misleading, as correct prediction of system
responses does not in general tell us enough about how well the models are interpreting the
dialogues, or whether they are doing this with a sufficient level of granularity. To assess this,
in this second experiment, we measure the semantic accuracy of each model by looking
\emph{exclusively} at how accurately they predict the final \API calls in the \bAbI and \bAbIplus
datasets. For the \memnn model, we follow the same overall procedure as in the previous experiment:
train on \bAbI data, and test on \bAbIplus.

\subsubsection{Results: Prediction of \API Calls}

\paragraph{\dylan results.} Successful parsing of all the dialogues in the \bAbI and \bAbIplus
datasets as shown above does not mean that the semantic representations compiled for the dialogues
were in fact correct. To measure the semantic accuracy of the \DSTTR parser \dylan we
programmatically checked that the correct slot values~--- those in the \API call annotations~---
were in fact present in the semantic representations produced by the parser for each dialogue
(see Fig.~\ref{fig:subtype} for example semantic representations). We further checked that there is
no other incorrect slot value present in these representations. The results showed that the parser
has 100\% semantic accuracy on both \bAbI and \bAbIplus. This result is not surprising,
given that \DSTTR is a general model of incremental language processing, including phenomena such as
self-corrections and restarts (see \citealp{Hough14} for details of the model). 



\paragraph{\memnn results~--- small data setup.} Given just 1 to 5 training instances from bAbI as
in the previous experiment, the mean API call prediction accuracy of the \memnn model is nearly 0
on both \bAbI and \bAbIplus. This is not at all unexpected, since we see prediction of the \API calls as
an inherently \emph{generation} process, unlike the prediction of system turns which can be done on
a retrieval/look-up basis alone. For this, the model needs to observe the different word sequences
that might determine each parameter (slot) value, and observe them with sufficient frequency and
variation. This is unlike a semantic parser like \DSTTR, that produces semantic representations for
the dialogues as a result of the structural, linguistic knowledge that it embodies.

\paragraph{\memnn results~--- full data setup.} Nevertheless, we were also interested in the general
semantic robustness of the \memnn model to the transformations in \bAbIplus, i.e.\ how well does the
\memnn model interpret \bAbIplus dialogues, when trained on the full \bAbI dataset?
Does it then learn to generalise to (process) the \bAbIplus dialogues with sufficient semantic accuracy?

Our hypotheses are that (i) given the positional encoding of memory vectors in the \memnn model and
the underlying attention mechanism, it would be able to learn to process bAbI+ dialogues given that
it was trained on similar data, resulting in an insignificant drop in performance from \bAbI to
\bAbIplus data; (ii) a lot more data would be needed to learn to process the bAbI+ structures; and
(iii) if trained on \bAbI data, there would be a very significant drop in performance on \bAbIplus with
incorrect \API calls predicted as a result of incorrect weightings and total lack of opportunity to
learn the meaning of words such as ``no'' or ``sorry'' which trigger the self-corrections and
restarts.

\begin{table}[t]
\centering
\begin{tabularx}{0.8\textwidth}{@{}cSS@{}}\toprule
 \textbf{Train / test set configuration}&\textbf{Train accuracy}&\textbf{Test accuracy}\\\midrule
\bAbI / \bAbI&100&100\\
\bAbI / \bAbIplus&100&28\\
\bAbIplus / \bAbI&67&99\\
\bAbIplus / \bAbIplus&72&53\\\bottomrule
\end{tabularx}
\caption{\API call accuracy (\%) of the \memnn trained on the full dataset}
\label{tab:api_call_accuracy}
\end{table}

Table~\ref{tab:api_call_accuracy} shows that we can fully replicate the results reported in
\cite{DBLP:conf/iclr/BordesBW17}: the \memnn model can predict the \API calls with 100\% accuracy,
when trained on the \bAbI trainset and tested on the \bAbI testset. But when this same model is
tested on \bAbIplus, the accuracy drops majorly to 28\%, making any dialogue system built using this
model unusable in the face of more diverse dialogue data~--- thus confirming our hypothesis (iii).
This is further discussed below.

\subsubsection{How Much Data Is Enough Data?}

Given the results obtained so far, we are next interested in: (1) how robust \memnns are to the
surface transformations in \bAbIplus when trained on \bAbI; (2) can \memnns learn to interpret \bAbIplus
when they are in fact trained on similar data that actually contain the \bAbIplus structures~--- i.e.
when trained on \bAbIplus; and (3) if so, how much \bAbIplus data is needed for this. While (1) is a
question about generalisation properties of a model, (2) \& (3) are about potential in principle
and/or practical limitations of \memnns to learn to interpret dialogues containing, e.g.
self-corrections where utterances contain both the correct, and an incorrect (and subsequently
repaired) slot value (e.g. ``for four sorry five people''). To answer (1) we therefore train the
model on the \bAbI dataset and test on \bAbIplus; and to answer (2) \& (3), we train the model on the
\bAbIplus train set and test it on the \bAbIplus test set. Furthermore, in order to explore the impact of
the amount of training data on the model's performance, we perform the latter experiment with
varying train set size, as well as varying the hyperparameters: embedding size and the number of
hops. The extended training data is obtained in the same way as the initial \bAbIplus dataset: we go
over the same original \bAbI dialogues and keep randomly mixing in the incremental modifications.

\begin{table}[t]
\centering
\begin{tabularx}{\textwidth}{@{}SSSSS@{}}\toprule
 \textbf{Training \bAbIplus dialogues}&\textbf{Memory hops}&\textbf{Embedding size}&\textbf{Train acc.}&\textbf{Test acc.}\\\midrule
2000&2&128&72.5&57.5\\
5000&2&128&72.7&60.7\\
10,000&2&128&72.8&65.8\\
50,000&1&128&82.6&78.2\\
100,000&1&64&83.3&80.5\\\bottomrule
\end{tabularx}
\caption{\memnn \API call accuracy (\%) with extended training data}
\label{tab:memn2n_extended_data}
\end{table}

Table~\ref{tab:memn2n_extended_data} shows how \memnn performs on the same initial, fixed \bAbIplus
test set, when trained on progressively more data and up to 100,000 \bAbIplus dialogues. As \memnn's
performance on bigger data highly depends on the model's hyperparameters, in this experiment we
perform a grid search over the number of memory hops (1, 2, 3), and the embeddings dimensionality
(32, 64, 128) for each train set size independently~--- everything else is fixed as in the previous
experiment.
Only the best performing hyperparameter configuration for each of the train set sizes are included
in the table.

The results confirm hypothesis (ii) above, i.e. that \memnns are in principle able to learn to
process the incremental dialogue phenomena in \bAbIplus but that they require tens of thousands of
training instances for this: even with 100,000 dialogues, the semantic accuracy on the original test
set stands at 80.5\%.


\section{Discussion}
\label{ch3:discussion}

\subsection{\memnn Analysis}
The \memnn model was able to predict system responses remarkably well, even when trained on very
few training instances. But results from Experiment 2 above showed that this was misleading: the
\memnns were making a drastic number of semantic mistakes when interpreting the dialogues, both in
the \bAbI and \bAbIplus datasets. Even when trained on the full \bAbI dataset, the model performed
badly on \bAbIplus in terms of semantic accuracy. We diagnose these results as follows:

\textbf{Problem complexity}. The first thing to notice is that in \bAbI dialogue Task 1, the
responses are highly predictable and stay constant regardless of the actual task details
(slot values) up to the point of the final \API calls; and further, that the prediction of
\API calls is a \emph{generation} process, unlike the prediction of the system turns, which is
retrieval-based. This, in our view, explains the very large difference in \memnn performance across
the two prediction tasks.

\textbf{Model robustness to the bAbI+ transformations}. The variations introduced in  \bAbIplus are
repetitions of both content and non-content words, as well as of additional incorrect slot values.
The model was working in the same setup as \dylan, therefore none of these variations could be
treated as unknown tokens for either system. Although in the case of \memnn, some of the mixed-in
words never appeared in the training data, and \bAbIplus utterances were augmented significantly with
those words~--- so it was interesting to see how such untrained embeddings would affect the latent
memory representations inside \memnn. The resulting performance suggests that there was no
significant impact on \memnn from these variations as far as the predicting system responses was
concerned. But the incorrect slot values introduced in self-corrections affect the system's task
completion performance significantly, only appearing at the point of \API call predictions. 

It is worth noting that when trained on extended amounts on \bAbIplus data with sufficient
representation of target speech phenomena, \memnn compensates for the most of the initial
performance drop. However, it attains a reasonable level of accuracy ($>80\%$) when trained on
100,000 \bAbIplus dialogues which renders it potentially impractical for real-world tasks with more
complex data distributions. 

\subsection{\dylan Analysis}

The linguistically informed \dylan-based model we used in this chapter has the following conceptual
advantages over previous approaches to dialogue system development:

\begin{itemize}[label=---]
  \item word-by-word incremental (and thus more natural) language understanding, dialogue
  management, and generation\footnote{Applicable to the full-fledged \babble setup as described in
  Section \ref{ch2:babble}.};
  \item a complete dialogue system for a new task can be automatically induced, using only `raw'
  data~--- i.e. successful dialogue transcripts;
  \item wide-coverage, task-based dialogue systems can be built from much smaller amounts of data as
  shown in Section \ref{ch3:experiments}.
\end{itemize}



This final point bears further examination. Since it is an empirically adequate model of incremental
language processing in dialogue, the \DSTTR grammar is required to capture interactional variants
such as question-answer pairs, over- and under-answering, self- and other-corrections,
clarification, split-utterances, and ellipsis more generally. As we showed in Section
\ref{ch3:experiments}, even if most of these structures are not present in the seed example(s), the
final system is able to handle them, thus resulting in a very significant generalisation around the
original data. It can be said that the \dylan setup is a carefully tuned rule-based system, thus
perhaps rendering these results trivial. But we note that the results here are not due to ad-hoc
constructions of rules/lexicons, but due to the generality of the grammar model, and its attendant
incremental, left-to-right properties; and that the same parser can be used in other domains.
Furthermore, the ability to process self-corrections, restarts, etc. ``comes for free'', without the
need to add or posit new machinery.

The generalisation results we report above for \dylan follow entirely from the knowledge present
within the grammar as a computational model of dialogue processing and contextual update, rather
than this having been learned from data. Applying the full \RL method of \babble (Section
\ref{ch2:babble}) would have meant that the system would actually discover many interactional and
syntactic variations that are not present in \bAbI, nor in \bAbIplus.

\section{Conclusions}

In this chapter, we have evaluated the generalisation properties of a purely linguistically informed
model based on the dialogue semantic parser \dylan, and compared it to a neural response retrieval
model \memnn. We did so by putting them in a controllable environment of \bAbI Dialog Tasks and
performing a series of experiments assessing their generalisation potential to more interactional
variations in the input data (i.e., generalisation from seed examples to the full dataset), as well
as the models' robustness to the simulated spoken language phenomena, i.e. self-corrections,
hesitations, and restarts~--- represented in our \bAbIplus corpus.

Our experiments show that \memnn lacks the ability to generalise to such phenomena, and performs
poorly when confronted with such variations even within synthetic, programmatically generated
dialogue data. Our experiments further show that although this particular model is in principle able
to learn to process disfluent dialogue phenomena, it requires an impractically large amount of data
to do so. The results in this chapter therefore shed significant light on the currently ambiguous
accuracy results reported for end-to-end systems \citep{DBLP:conf/iclr/BordesBW17}.

On the other hand, experiments with \dylan show that it can process 74\% of the \bAbI Dialog Task
1 even when only exposed to 0.13\% of the data (5 dialogues); it can in addition process
65\% of \bAbIplus. That is in contrast to \memnn which was not robust to the structures we introduced
in bAbI+, even when trained on the full \bAbI dataset.

The above results give us the following insights on the two possible ways towards attaining a
practical level of data efficiency in dialogue. Firstly, an inductive bias in the form of a
linguistically informed model can be an efficient approach in an extreme case of 1-shot/few-shot
generalisation. However, the bottleneck here is the linguistic resource itself, the incremental
dialogue grammar of \dylan in our case. Parsing natural speech is generally challenging, especially
in case of freeform conversation, so producing wide-coverage grammars for this purpose can be a
notoriously hard task. The second path featuring neural models has a definite advantage here from
the Natural Language perspective: given a sufficient amount of data to represent the phenomena of
interest, models like \memnn are able to adjust and process it adequately from the target task's
perspective. Scaling to an unknown domain or a foreign language for that matter will require nothing
more than a sufficiently representative dataset. The challenge here though is how much data is
considered sufficient for that, and our controlled experiments with \memnn show that its actual
data requirements are quite far from practical. Therefore, the vital step towards real-world
applicability of such systems is to reduce the amount of required training data and annotations
closer to what can be considered practically data-efficient without the loss of the models'
generalisation potential.

In the next chapter, we are going to explore the purely data-driven approach to dialogue system
bootstrapping. That is, we will focus on learning transferable representations of dialogue capturing
lexical and interactional similarity (as discussed in Section \ref{ch2:transfer_learning}) and their
applicability for reducing the data consumption of goal-oriented dialogue systems.


\chapter{Learning Transferable Dialogue Representations} 

\label{Chapter4} 

\lhead{Chapter 4. \emph{Learning Dialogue Representations}} 


As we saw in the previous chapter, both linguistically informed and purely data-driven methods have
their potential in terms of data efficiency. The principal advantage of machine learning methods is
that they generally do not require constructing and maintaining syntactic/semantic grammars which are
very challenging to build, especially in the setting of spontaneous spoken language.
The problem with machine learning methods though is, as we also saw previously, their performance
depends to a high degree on the training data, and the overall data consumption of those models
significantly reduces their flexibility in practical setups. A possible solution to that is to use
large sources of domain-independent data in order to obtain the common language and dialogue
representation and then transfer it to the specific problem domain, with minimal fine-tuning to the
available in-domain data.
As outlined in Chapter \ref{Chapter2}, transfer learning has already proved to be a very efficient
technique in computer vision and is being actively adopted in Natural Language Processing~---
in this chapter, we are going to explore its applicability to dialogue.

Specifically, we are going to present the Dialogue Knowledge Transfer Network (or \diktnet),
a goal-oriented dialogue response generation model designed for few-shot learning, i.e.\ training
only using a small number of complete in-domain dialogues. The key underlying concept of this model
is transfer learning: \diktnet makes use of the latent text representation learned from several
sources ranging from large-scale general-purpose textual corpora to similar dialogues in the domains
different to the target one. We use the evaluation framework of \cite{DBLP:conf/sigdial/ZhaoE18} and
mainly compare our approach to theirs~--- and, similarly to them, we use the Stanford Multi-Domain
dialogues dataset \citep{DBLP:conf/sigdial/EricKCM17}.
While \citeauthor{DBLP:conf/sigdial/ZhaoE18}'s method does not require complete in-domain dialogues
and uses annotated utterances instead (and is therefore presented as ``zero-shot"), we show that our
model achieves superior performance with roughly the same amount of data (with respect to in-domain
utterances) while requiring no annotations whatsoever.

\section{Few-Shot Dialogue Generation}
\label{ch4:fsdg}

\begin{table}[t]
  \center
  \begin{minipage}[b][][b]{0.45\textwidth}
  \footnotesize
  \begin{tabularx}{\textwidth}{ll}\toprule

  {\bf Driver:}&I need to find out the time and parties\\
  &attending my optometrist appointment\\
  {\bf Car:}&I have 3 appointments scheduled,\\
  &with Alex, your sister, and Jeff.\\
  &Which are you referring to?\\
  {\bf Driver:}&I want to know about the one that Alex\\
  & is joining me at\\
  {\bf Car:}&That optometrist appointment is at 4 pm.\\
  {\bf Driver:}&Thanks\\
  {\bf Car:}&no problem\\

  \bottomrule
  \end{tabularx}
  \caption*{(a)}
  \end{minipage}%
  \hfill
  \begin{minipage}[b][][b]{0.5\textwidth}
  \footnotesize
  \begin{tabularx}{\textwidth}{@{}lXXXX@{}}\toprule
  {\bf Event}&{\bf Time}&{\bf Date}&{\bf Party}&{\bf Agenda}\\\midrule

  opt. appt&10 am&Thursday&sister&---\\
  dinner&8 pm&the 13th&Ana&---\\
  opt. appt&7 pm&the 20th&Jeff&---\\
  opt. appt&4 pm&the 13th&Alex&---\\
  ...&...&...&...&...\\
  \bottomrule
  \end{tabularx}
  \caption*{(b)}
  \end{minipage}
  \caption[Example dialogue from \SMD]{Example dialogue from \SMD (a) with the corresponding
  knowledge base snippet (b)~--- driver is the user, car is the system}
  \label{tab:smd_example}
\end{table}

We first describe the task we are addressing in this chapter, and the corresponding base model.
Specifically, here we work with dialogue data that is organised into multiple non-overlapping
domains. We say that \textit{source domains} can be used to train the model, whereas the
\textit{target domain} is mainly used for evaluation. Different data-efficient setups assume the
corresponding usage of target-domain data. As such, in zero-shot learning, the full target dialogues
are not used at all; few-shot learning assumes fine-tuning to several target dialogues and then
evaluating on the rest~--- we will be working in the latter setting.

Our domains here are in-car navigation, weather information, and appointment scheduling~--- as
represented in the \SMD dataset we are going to work with (see Section \ref{ch2:data_collection} for
a short description). In a single experiment, we work with 2 of those domains as \text{source}, and
the third one becomes the target domain. An example dialogue from \SMD is shown in Table
\ref{tab:smd_example}~--- every dialogue in the dataset comes with a snippet from the underlying
Knowledge Base working a as simulation of the domain-specific search API. The dialogue system's task
then is to (1) train on the available data from the source domains, while learning to extract
relevant data from the KB snippets along the way, (2) fine-tune to the given seed dialogues + KB
snippets in the target domain, and (3) predict responses for the rest of the target-domain
dialogues.

\section{The Base Model}

In our approach, we are building on top of \HRED~--- in particular, the copy-augmented model
\citep{DBLP:conf/iclr/MerityX0S17}. \HRED was discussed earlier in Section \ref{ch2:gen_techniques}.
Here, we are providing its Negative log-likelihood (\NLL) optimisation objective which we are going
to be building upon:


\begin{equation}
  \label{eq:hred}
  \begin{split}
  & \mathcal{L}_{\text{HRED}} = \log p_{\mathcal{F}^d}(y_\text{sys}
  \mid \mathcal{F}^e(c, x_\text{usr}))
  \end{split}
\end{equation}

where $x_\text{usr}$ is the user's query, $y_\text{sys}$ is the system's response, $c$ is the
dialogue context, and $\mathcal{F}^e$ and $\mathcal{F}^d$ are respectively the hierarchical encoder
and the decoder.

We work with goal-oriented dialogues where part of the system's task is to provide the integration
with an underlying database or API via the KB snippets, as described above. Given that such
information may contain unseen token sequences for the most part, especially in the target domain,
we use a copy mechanism in order to be able to directly transfer those tokens from the input into
the system's responses. More specifically, we represent the KB info as token sequences and
concatenate them to the dialogue context similarly to the CopyNet setup of
\cite{DBLP:conf/sigdial/EricKCM17}, the only difference is the actual copy mechanism
implementation for which we employ the Pointer-Sentinel Mixture model (%
\citealp{DBLP:conf/iclr/MerityX0S17}; \citealp{DBLP:conf/sigdial/ZhaoE18})~--- see the description
of these models in Sections \ref{ch2:copy_augmented}. Our model is thus supposed to produce
goal-oriented responses for target dialogue contexts using the soft decision~--- whether to generate
the next word from the vocabulary or to copy a token from earlier in the dialogue history (or at
some position in the knowledge base response).

\section{Dialogue Knowledge Transfer Networks}
\label{ch4:diktnet}

Transfer learning is considered the key means for efficient training with minimal data, and our
\diktnet model essentially introduces several knowledge-transfer augmentations to the base \HRED
model described above. \diktnet training is performed in two stages described below.

\subsection{Stage 1. Dialogue Representation Pre-training}
\label{ch4:laed}

\begin{figure}[t]
  \centering
  \includegraphics[width=0.7\textwidth]{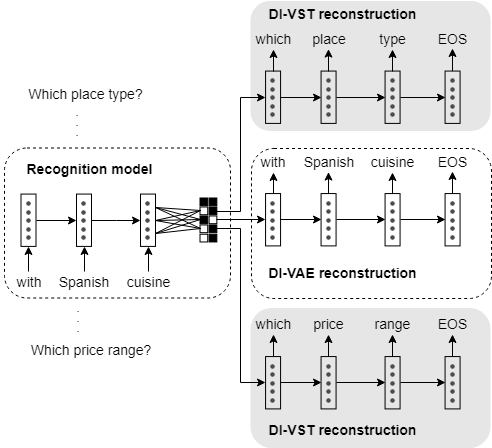}
  \caption{\DIVAE and \DIVST (\diktnet Stage 1)}
  \label{fig:diktnet_stage1}
\end{figure}

Dialogue structure~--- e.g.\ word sequences~--- is highly specific to a given domain or task, and
the meaning of conversational utterances is highly contextual, i.e.\ similar utterances may have
different meanings depending on the context. Nevertheless, there is a lot of similarity in dialogue
structure~--- i.e.\ sequences of dialogue actions~--- across domains, e.g.\ a conversation normally
starts with a mutual greeting and a question is very often followed by an answer.
Here, we propose to exploit this phenomenon in the form of learning a latent dialogue action
representation in order to better capture the dialogue structure by abstracting away from surface
forms.
Crucially, we learn such representation from \metalwoz
\citep[derived from \textit{Meta-Learning Wizard-of-Oz},][]{lee2019multi-domain}, a dataset
specifically created for the purposes of meta-learning and transfer learning and consisting of
human-human conversations in 51 unique domains~--- see the description of the dataset in Section
\ref{ch2:data_collection} and some statistics below in Section \ref{ch4:data}.

For this stage of training we use unsupervised, variational autoencoder-based (\VAE) representation
learning, and we will be particularly focusing here on a specific framework of discrete information
(DI) sentence representations of \cite{DBLP:conf/acl/EskenaziLZ18}. As will be discussed later in
Section \ref{ch4:representations}, the underlying variational autoencoding techniques facilitate
more stable training and result in more efficient representation models as compared to the more
widely-used conventional \VAE. As was introduced in Section \ref{ch2:autoencoder} \DIVAE model
differs from a classic, continuous \VAE in that it (1) implicitly promotes mutual information
between the model's input and its latent code and (2) uses Batch Prior Regularisation technique for
the calculation of the \KL divergence in the case of discrete latent codes. \DIVAE and its
skip-thought counterpart \DIVST are visualised in Figure \ref{fig:diktnet_stage1}.

In the downstream \diktnet model, we use \DIVAE autoencoder in order to obtain the representation
of the user's query: $z_\text{usr} = \text{DI-VAE}(x_\text{usr})$. \DIVST, in turn, is used to
obtain a prediction of the system's action $z_\text{sys}$ in the discretised latent form given the
user's input $x_\text{usr}$ as well as the full dialogue context $c$. For that, \DIVST autoencoder
is used as part of a hierarchical, context-aware encoder-decoder response generation model~---
following \citealp{DBLP:conf/acl/EskenaziLZ18}, we refer to it as Latent Action Encoder-Decoder, or
\LAED (discussed in Section \ref{ch2:latent_variable}), its optimisation objective is as follows:


\begin{equation}
\begin{split}
  & \mathcal{L}_{LAED} =
  \mathbb{E}_{q_{\mathcal{R}}(z_\text{sys} \mid y_\text{sys}) p(y_\text{sys}, c, x_\text{usr})}
  \left[ \log p_\pi (z_\text{sys} \mid c, x_\text{usr})
    + \log p_\mathcal{F} (y_\text{sys} \mid z_\text{sys}, c, x_\text{usr}) \right]
\end{split}
\end{equation}


where $\mathcal{F}$ is the decoder generating the system's response $y_\text{sys}$, and $\pi$ is the
`policy' feed-forward network predicting $z_\text{sys}$ from the dialogue context $c$ and the
user's last turn $x_\text{usr}$. As noted in out introduction of \LAED (Section \ref{ch2:latent_variable}),
the recognition model $q_{\mathcal{R}}(z_\text{sys} \mid y_\text{sys})$ conditioned on the gold
system's response is only used during training and is discarded at prediction time.


Our intuition behind using different models for representing the user's query and the system's
latent action follows empirical results of \cite{DBLP:conf/acl/EskenaziLZ18} who showed that \DIVAE
is better at capturing specific words of an utterance, while \DIVST represents the overall dialogue
action better. We train these two models on \metalwoz in an unsupervised way with the objectives as
described above, and use their discretised latent codes $z_\text{usr}$ and $z_\text{sys}$
respectively in the downstream model at the next stage of training.

\subsection{Stage 2. Transfer}

At this stage, we train directly for our target task, few-shot dialogue generation, and thus go back
to the model described in Section \ref{ch4:fsdg}. While the training procedure of this model
naturally assumes \textit{domain transfer}, we will provide it with more sources of textual and
dialogue knowledge of varying generality described below.

As opposed to direct domain transfer, we incorporate domain-general dialogue understanding from the
\LAED representation trained on \metalwoz at the previous stage. 
\LAED captures the background top-down dialogue structure: sequences of dialogue acts in a
cooperative conversation, latent dialogue act-induced clustering of utterances, and the overall
phrase structure of spoken utterances. We incorporate this information into the model by
conditioning \HRED's decoder on the combined latent codes from Stage 1 and refer to this model as
\HRED+Stage1. Its optimisation objective is as follows:

\begin{equation}
  \label{eq:hred_laed}
  \begin{split}
  & \mathcal{L}_{\text{HRED+Stage1}} =
  \mathbb{E}_{p(x_\text{usr},c) p (z_\text{usr} \mid x_\text{usr})
              p_\pi (z_\text{sys} \mid x_\text{usr}, c)}
  \left[ \log p_{\mathcal{F}^d} \left(y_\text{sys} \mid \left \{
            \mathcal{F}^e(x_\text{usr}, c), z_\text{usr}, z_\text{sys} \right
  \} \right) \right]
  \end{split}
\end{equation}

where $z_\text{usr}$ and $z_\text{sys}$ are respectively samples obtained from the \DIVAE
user utterance model and the \LAED system action model, and $ \left \{ \cdot \right \}$ is the
concatenation operator.

The last, most general source of knowledge we use is a pre-trained \ELMo model
\citep{DBLP:conf/naacl/PetersNIGCLZ18}. Apart from using an underlying bidirectional RNN encoder,
\ELMo captures both token-level and character-level information which is especially crucial in
understanding unseen tokens and KB items in the underrepresented target domain. The \HRED model with
\ELMo as the utterance-level encoder is referred to as \HRED+\ELMo.

Finally, \diktnet is the \HRED augmented with both \ELMo encoder and sentence representations
(\DIVAE and \LAED) from Stage 1.

\begin{figure}[t]
  \centering
  \includegraphics[width=1.0\linewidth]{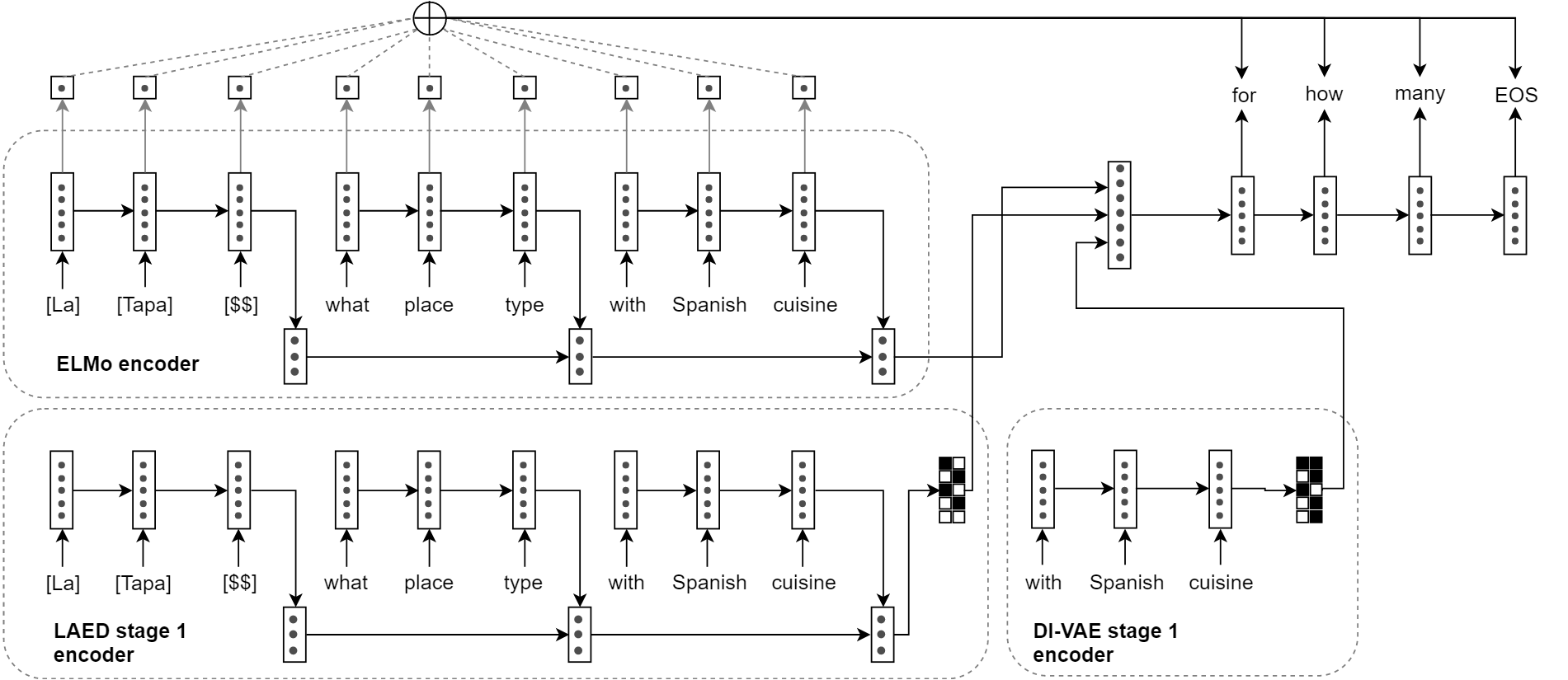}
  \caption{\diktnet Stage 2 (tokens in brackets are KB data)}
  \label{fig:diktnet_stage2}
\end{figure}

\diktnet\ is visualised in Figure \ref{fig:diktnet_stage2}. The model (as well as its variants
listed above) is implemented in PyTorch \citep{paszke2017automatic}, and the code is openly
available\footnote{\colorhref{blue}{http://tiny.cc/diktnet}}.


\section{Baselines}
\label{ch4:baselines}
We perform an exhaustive ablation study of \diktnet by comparing it to all of its variations
mentioned above: \HRED, \HRED+\ELMo, and \HRED+Stage1. In addition to that, we have the
\HRED+\VAE~--- a \HRED+Stage1 counterpart for which we use a regular, continuous \VAE behind \DIVAE
and \DIVST in order to determine the impact of discretised latent codes (see Eq. \ref{eq:vae_fsdg}
for the corresponding objective function).

Furthermore, we compare \diktnet to the previous state-of-the-art approach, Zero-Shot Dialogue
Generation \citep{DBLP:conf/sigdial/ZhaoE18}. This model did not use any complete in-domain dialogues
but instead it relied on annotated utterances in all of the domains. We use it as-is (\ZSDG),
as well its variation as follows.

We make use of \ZSDG's central idea of using \NLU-annotated in-domain utterances as `domain
descriptions' that facilitate bridging dialogue understanding across domains, but instead of using
manually annotated utterances, we employ automatic \NLU markup. Our annotations include:

\begin{itemize}[label=---]
  \item Named Entity Recognition~--- Stanford \NER model ensemble of case-sensitive and caseless
  models \citep{DBLP:conf/acl/FinkelGM05},
  \item Date/time markup~--- Stanford SUTime \citep{DBLP:conf/lrec/ChangM12},
  \item Wikidata entity linking~--- Yahoo FEL (\citealp{Blanco:WSDM2015}; \citealp{Pappu:WSDM2017}).
\end{itemize}

We serialise annotations from these sources into token sequences and make domain description tuples
out of all the utterances in the source and target domains. In this way, most of our domain
descriptions share the structure and content of the original ones.

For example, for the phrase \textit{`Will it be cloudy in Los Angeles on Thursday?'}:
\begin{itemize}[label=---]
  \item the original \ZSDG annotation is \texttt{"request \#goal cloudy \#location Los Angeles
  \#date Thursday"},
  \item our \NLU annotation is \texttt{"LOCATION Los Angeles DATE Thursday"}.
\end{itemize}

We have two models in this setup, with (\NLU\_ZSDG+Stage1) and without the use of Stage 1
representations (\NLU\_ZSDG) respectively.

\section{Datasets}
\label{ch4:data}

For the latent representation learning, we use \metalwoz described in Section
\ref{ch2:data_collection}. All the domains available in the \metalwoz dataset are listed in Table
\ref{tab:maluuba_domains} of Appendix \ref{AppendixA}, and some example dialogues can be found there
in Section \ref{a:metalwoz_examples}.

Our target dataset is the Stanford Multi-Domain (\SMD) dialogues corpus
\citep{DBLP:conf/sigdial/EricKCM17} described in Section \ref{ch2:data_collection}. \SMD contains
human-human goal-oriented dialogues in three domains, with a simulation of the underlying search
\API: each dialogue in \SMD comes with a knowledge base snippet representing the result of
implicitly querying the \API, with the KB schema (i.e. columns names and the data types) being
specific to each domain. Although sharing some common features (the setting of an intelligent
in-car assistant and the use of the underlying \KB), the dialogues differ significantly across
domains which makes the domain transfer sufficiently challenging. The statistics of the are shown in
Tables \ref{tab:smd} and \ref{tab:maluuba}, respectively.

In our experiments, we make sure that all the source data have no domain overlap with the target
dialogues we're evaluating on, therefore we make our training setup dynamic by excluding the
specific \metalwoz domains based on the target \SMD one, such that:
\begin{itemize}[label=---]
  \item for the Navigation target domain in \SMD, we exclude \metalwoz's Store Details domain,
  \item for Weather, we exclude Weather Check,
  \item for Schedule, we exclude Update Calendar and Appointment Reminder.
\end{itemize}


\begin{table}[t]
  \small
    \begin{minipage}[b][][b]{0.4\textwidth}
      \centering
      \begin{tabularx}{0.9\textwidth}{lr}\toprule
      &\\\midrule
      Domains&51\\
      Dialogues&40,388\\
      Avg. dialogue length&11.91\\\bottomrule
      \end{tabularx}
      \caption{\metalwoz dataset statistics}
      \label{tab:maluuba}
    \end{minipage}%
    \begin{minipage}[b][][b]{0.6\textwidth}
      \centering
      \begin{tabularx}{1.0\textwidth}{lrrr}\toprule
      &Navigation&Weather&Schedule\\\midrule
      Dialogues&800&797&828\\
      Utterances &5248&4,314&3,170\\
      Avg. dialogue length&6.56&5.41&3.83\\\bottomrule
      \end{tabularx}
      \caption{Stanford multi-domain dataset statistics (trainset)}
      \label{tab:smd}
    \end{minipage}
\end{table}

\section{Experimental Setup and Evaluation}
\label{ch4:setup}

Our few-shot setup is as follows. Given the target domain, we first train Stage-1 model(s) on the
\metalwoz data, having filtered source domains as described above. We used a \DIVAE and a
\DIVST-based \LAED, both of the size $10\times 5$.

Next, having trained and Stage-1 models, we train \diktnet on all the source domains from the \SMD
dataset without further fine-tuning of \DIVAE/\LAED. We also fine-tune to a portion of
target-domain data (thus working in a few-shot setup) by sampling the target dialogues together
with their \KB info, varying the amount of those from 1\% to 10\% of all the available target data.

For the \NLU\_\ZSDG setup, we annotated all available \SMD data and randomly selected a subset of
1000 utterances from each source domain, and 200 utterances from the target domain. For source
domains, this number amounts to roughly a quarter of all available training data~--- we chose it in
order to make use of as much annotated data as possible while keeping the domain description task
secondary. We made sure to keep under roughly the same target-domain data requirements as the
\ZSDG baseline.

For evaluation, we follow the approach of \cite{DBLP:conf/sigdial/ZhaoE18} and report \BLEU and
Entity F1 scores. Given the non-deterministic nature of our training setup, we report means and
variances of our results over 10 runs with different random seeds.

We also perform an additional evaluation of \diktnet's performance with extended amounts of target
data and compare it to the original results for the \SMD dataset reported by
\cite{DBLP:conf/sigdial/EricKCM17} upon the dataset introduction. Their model~--- Key-Value
Retrieval Network (\KVRet), a variant of copy-augmented \SeqToSeq with a separate copying mechanism
for the \KB snippets~--- was trained with all the available data for a given domain. In this
evaluation, we average \BLEU scores across all 3 SMD domains in order to be consistent with the form
that the corresponding results are presented in the original paper.

We train our models with the Adam optimiser \citep{DBLP:journals/corr/KingmaB14} with the learning
rate of $0.001$. Our hierarchical models' utterance encoder is an \LSTM cell
\citep{DBLP:journals/neco/HochreiterS97} of the size $256$, and the dialogue-level encoder is a
\GRU \citep{DBLP:conf/emnlp/ChoMGBBSB14} of the size $512$.

\section{Results and Discussion}
\label{ch4:results}

\definecolor{blue(ryb)}{rgb}{0.01, 0.28, 1.0}
\definecolor{blue-violet}{rgb}{0.54, 0.17, 0.89}

\definecolor{tangelo}{rgb}{0.98, 0.3, 0.0}

\definecolor{vividcerise}{rgb}{0.85, 0.11, 0.51}

\definecolor{canaryyellow}{rgb}{1.0, 0.94, 0.0}
\definecolor{electricgreen}{rgb}{0.0, 1.0, 0.0}

\pgfplotsset{scaled x ticks=false,compat=1.3}

\begin{figure}[t]
\centering
\begin{subfigure}[b]{0.49\textwidth}
\centering
\begin{tikzpicture}[thick,scale=0.78, every node/.style={transform shape}]
  \begin{axis}[
    xlabel={Target data ratio, \%},
    ylabel={\BLEU, \%},
    xmin=1, xmax=10,
    ymin=0.0, ymax=20,
    symbolic x coords={1, 3, 5, 10},
    xtick=data,
    ytick={0, 5, 10, 15, 20},
    legend style={font=\tiny},
    legend to name=named,
    legend columns=-1,
    legend entries={\ZSDG, \NLU\_ZSDG, \NLU\_\ZSDG+Stage1, \HRED, \HRED+VAE, \HRED+Stage1, \HRED+\ELMo, \diktnet},
    legend cell align=left,
    ymajorgrids=true,
    xmajorgrids=true,
    grid style=dotted,
    title=Mean \BLEU across domains
  ]

  \addplot[dashed, color=blue]
  coordinates {
    (1,7.30)(3,7.30)(5,7.30)(10,7.30)
  };
  \addplot[dashed, color=blue-violet]
  coordinates {
    (1,5.70)(3,5.70)(5,5.70)(10,5.70)
  };
  \addplot[dashed, color=cyan]
  coordinates {
    (1,8.30)(3,8.30)(5,8.30)(10,8.30)
  };
  \addplot[color=electricgreen]
  coordinates {
    (1,7.03)(3,8.57)(5,9.83)(10,11.57)
  };
  \addplot[color=canaryyellow]
  coordinates {
    (1,5.00)(3,7.67)(5,8.70)(10,11.17)
  };
  \addplot[color=vividcerise, mark=circle]
  coordinates {
    (1,8.37)(3,10.50)(5,12.20)(10,15.03)
  };
  \addplot[color=tangelo, mark=triangle]
  coordinates {
    (1,6.47)(3,9.37)(5,10.93)(10,13.10)
  };
  \addplot[color=red, mark=square]
  coordinates {
    (1,9.33)(3,12.40)(5,13.57)(10,16.93)
  };
  \end{axis}
\end{tikzpicture}
\end{subfigure}
\begin{subfigure}[b]{0.49\textwidth}
\centering
\begin{tikzpicture}[thick,scale=0.78, every node/.style={transform shape}]
  \begin{axis}[
    xlabel={Target data ratio, \%},
    ylabel={F1, \%},
    xmin=1, xmax=10,
    ymin=10, ymax=50,
    symbolic x coords={1, 3, 5, 10},
    xtick=data,
    ytick={10, 20, 30, 40, 50},
    ymajorgrids=true,
    xmajorgrids=true,
    grid style=dotted,
    title=Mean Entity F1 across domains
  ]

  \addplot[dashed, color=blue]
  coordinates {
    (1,27.30)(3,27.30)(5,27.30)(10,27.30)
  };
  \addplot[dashed, color=blue-violet]
  coordinates {
    (1,18.67)(3,18.67)(5,18.67)(10,18.67)
  };
  \addplot[dashed, color=cyan]
  coordinates {
    (1,18.17)(3,18.17)(5,18.17)(10,18.17)
  };
  \addplot[color=electricgreen]
  coordinates {
    (1,19.20)(3,28.80)(5,30.30)(10,32.60)
  };
  \addplot[color=canaryyellow]
  coordinates {
    (1,20.47)(3,27.53)(5,29.07)(10,32.67)
  };
  \addplot[color=vividcerise, mark=circle]
  coordinates {
    (1,23.53)(3,29.33)(5,28.03)(10,34.50)
  };
  \addplot[color=tangelo, mark=triangle]
  coordinates {
    (1,31.00)(3,32.93)(5,33.50)(10,36.37)
  };
  \addplot[color=red, mark=square]
  coordinates {
    (1,32.90)(3,33.80)(5,34.83)(10,39.20)
  };
  \end{axis}
\end{tikzpicture}
\end{subfigure}
\ref*{named}
\caption[Models' performance on the \SMD dataset]{Models' performance on the \SMD dataset. Dashed
are \ZSDG baselines that use a fixed set of annotated in-domain utterances instead of raw dialogues
(see Section \ref{ch4:baselines})}
\label{fig:diktnet_results}

\end{figure}
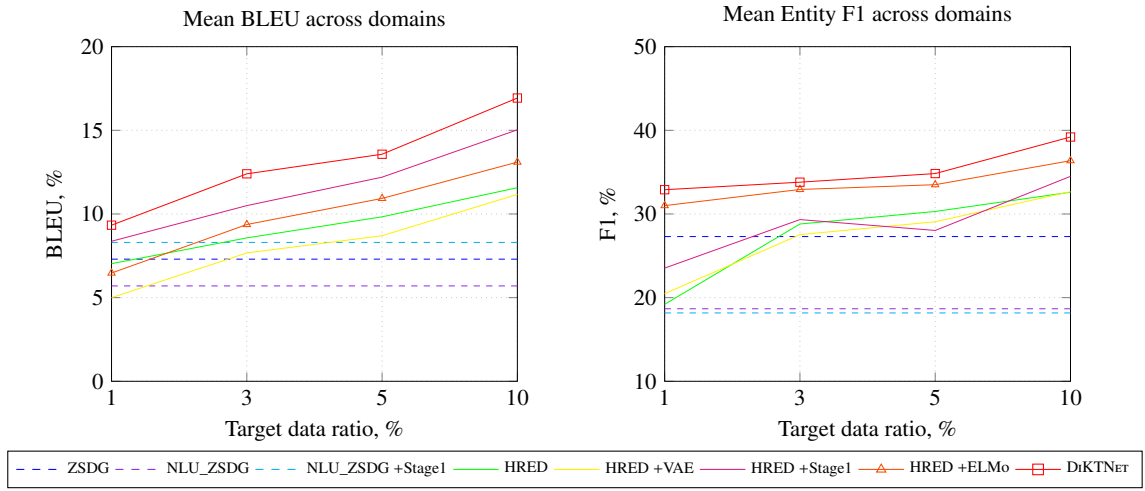

Our results are shown in Figure \ref{fig:diktnet_results}~--- see also Table
\ref{tab:diktnet_results} of the Appendix \ref{AppendixA} for a more detailed breakdown. The former
contains \BLEU and Entity F1 scores averaged over target domains, and the latter has the
corresponding values for each domain separately, showing means and variances. Our objective here is
maximum accuracy with minimum training data.

\subsection{Results for the Few-Shot Setup}
It can be seen that few-shot models with \DIVAE/\LAED representation are the best performing models
for this objective. While improvements upon \ZSDG can already be seen with simple \HRED in a
few-shot setup, the use of the Stage-1 representation and domain-general \ELMo encoding helps
significantly reduce the amount of in-domain training data needed: at 1\% of in-domain dialogues, we
see that \diktnet consistently and significantly improves upon \ZSDG in every domain. In \SMD, with
its average dialogue length of 5.25 turns, 1\% of training dialogues amounts to approximately 40
in-domain training utterances. In contrast, the \ZSDG setup used approximately 150 training
utterance-annotation pairs for each domain, including the target one, totalling about 450
\textit{annotated} utterances.

Although in our few-shot approach we use full in-domain dialogues, we end up having significantly
less in-domain training data, with the crucial difference that none of those has to be annotated for
our approach. Therefore, the method we introduced improves upon the previous best approach in both
accuracy and data-efficiency.

In turn, the results of the  \ZSDG\_\NLU setup demonstrate that single utterance annotations, if not
domain-specific and produced by human experts, do not provide as much signal as full dialogues, even
without annotations at all. Even the significant number of such annotated utterances per domain
did not make a difference in this case.

We would also like to point out that, as can be seen in the table, our results have quite high
variance~--- the main source of it is the nature of our training/evaluation setup where we average
over 10 runs with 10 different sets of seed dialogues. However, in the majority of cases with
comparable means, \diktnet has a lower variance than the alternative models at the same percentage
of seed data. And in the extreme case with 1\% target data, \diktnet improves on all the other models
in terms of both means and variances.

\subsection{Discussion of the Latent Representations}
\label{ch4:representations}

\begin{table}[t]
\small
\center
\begin{tabularx}{0.8\textwidth}{l}\toprule
Where can I go shopping?\\
Where does my friend live?\\
Where can I get Chinese food?\\
Where can I go to eat?\\
Can you please take me to a coffee house?\\
\midrule
I'd like to set a reminder for my meeting at 2pm later this month please.\\
What is the time and agenda for my meeting, and who is attending?\\ Schedule a lab appointment with
my aunt for the 7th at 1pm.\\
Schedule a calendar reminder for yoga with Jeff at 6pm on the 5th.\\
\midrule
Car I'm desiring to do some shopping: which one is it the nearest shopping ...\\ 
... center? Anything within 4 miles?\\
Get the address to my friend's house that i could get to the fastest\\ Car I need to get to my
friends house, it should be within 4 miles from here\\
\bottomrule
\end{tabularx}
\caption{Selected clusters of utterances sharing the same \DIVAE codes}
\label{tab:laed}
\end{table}

The comparison of the setups with different latent representations also gives us some insight: while
the \VAE-powered \HRED model improves on the baseline in multiple cases, it lacks generalisation
potential compared to the \DIVAE/\LAED setup. The reason for that might be the inherently more
stable training of \DIVAE/\LAED due to their modified objective function, which in turn results in
a more informative representation providing better generalisation. For instance, with the `vanilla'
\VAE setup, we immediately experienced the commonly reported vanishing \KL term problem (discussed
previously in Section \ref{ch2:latent_variable}) which effectively turned our \VAE into a
significantly overfitted \AE. With the discrete-information models~--- both \DIVAE and \DIVST~--- we
did not experience this problem even without using any techniques that are considered crucial in the
training of \VAEs \citep{DBLP:conf/conll/BowmanVVDJB16}.

In order to have a glimpse into the \DIVAE-produced clustering, in Table \ref{tab:laed} we present a
snippet of the utterance clusters sharing the same, most frequent latent codes throughout the
dataset (the clustering is obtained with \DIVAE model trained on every domain but `Store details',
i.e.\ the one for the evaluation on `Navigate' \SMD domain). From this snippet, it can be seen that
those clusters work well for domain separation, as well as capturing dialogue intents.

\subsection{Results with Extended Data}

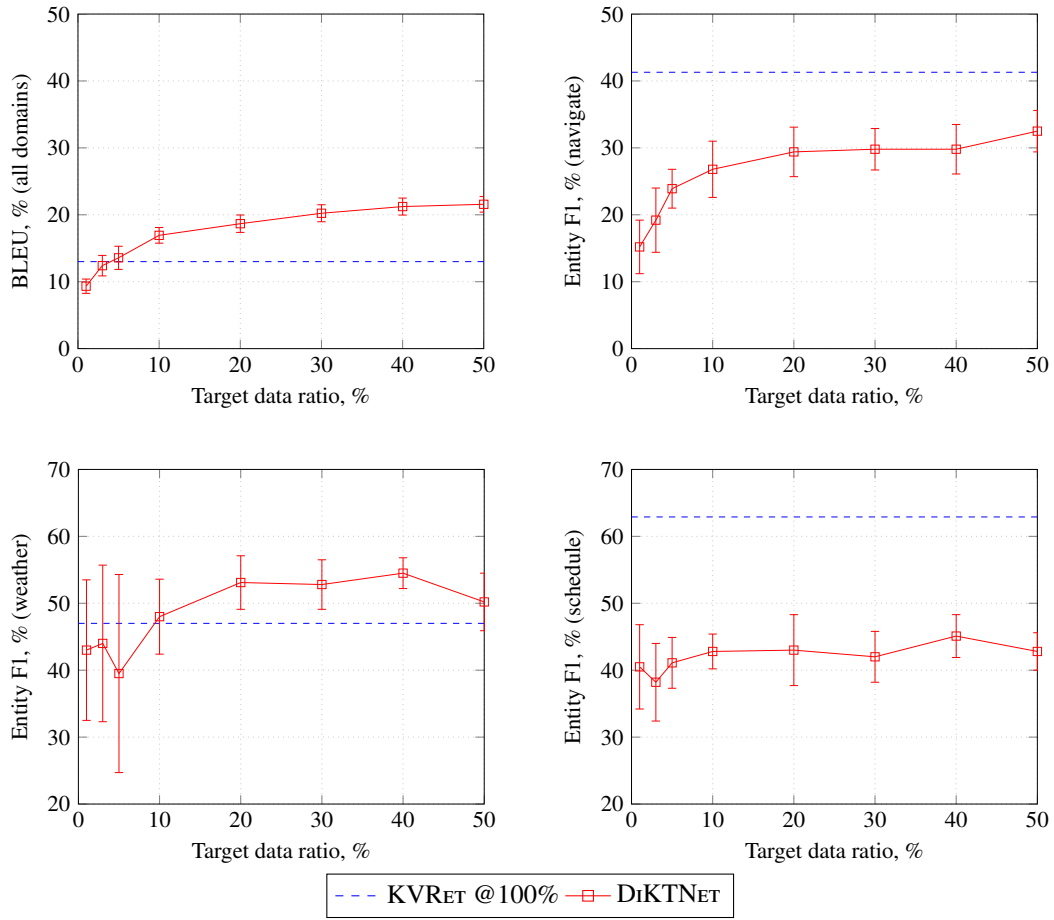
\begin{figure}
\centering
\begin{subfigure}[b]{.49\textwidth}
\centering
\pgfplotsset{scaled x ticks=false}
\begin{tikzpicture}[thick,scale=0.78, every node/.style={transform shape}]
  \begin{axis}[
    xlabel={Target data ratio, \%},
    ylabel={\BLEU, \% (all domains)},
    xmin=0, xmax=50,
    ymin=0.0, ymax=50,
    xtick={0, 10, 20, 30, 40, 50},
    ytick={0, 10, 20, 30, 40, 50},
    legend style={font=\small},
    legend to name=named_ext_data,
    legend columns=-1,
    legend entries={\KVRet@100\%, \diktnet},
    legend cell align=left,
    ymajorgrids=true,
    xmajorgrids=true,
    grid style=dotted,
  ]

  \addplot[dashed, color=blue]
  coordinates {
    (0,13.0)(3,13.0)(5,13.0)(10,13.0)(20,13.0)(30,13.0)(40,13.0)(50,13.0)
  };
  \addplot[color=red, mark=square, error bars/.cd, y dir=both, y explicit]
  coordinates {
    (1,9.33) +=(0,1.07) -= (0,1.07)
    (3,12.4) +=(0,1.53) -= (0,1.53)
    (5,13.57) +=(0,1.73) -= (0,1.73)
    (10,16.93) +=(0,1.17) -= (0,1.17)
    (20,18.67) +=(0,1.3) -= (0,1.3)
    (30,20.23) +=(0,1.27) -= (0,1.27)
    (40,21.23) +=(0,1.27) -= (0,1.27)
    (50,21.57) +=(0,1.17) -= (0,1.17)
  };
  \end{axis}
\end{tikzpicture}
\end{subfigure}
\begin{subfigure}[b]{.49\textwidth}
\centering
\pgfplotsset{scaled x ticks=false}
\begin{tikzpicture}[thick,scale=0.78, every node/.style={transform shape}]
  \begin{axis}[
    xlabel={Target data ratio, \%},
    ylabel={Entity F1, \% (navigate)},
    xmin=0, xmax=50,
    ymin=0, ymax=50,
    xtick={0, 10, 20, 30, 40, 50},
    ytick={0, 10, 20, 30, 40, 50},
    ymajorgrids=true,
    xmajorgrids=true,
    grid style=dotted,
  ]

  \addplot[dashed, color=blue]
  coordinates {
    (0,41.3)(3,41.3)(5,41.3)(10,41.3)(20,41.3)(30,41.3)(40,41.3)(50,41.3)
  };
  \addplot[color=red, mark=square, error bars/.cd, y dir=both, y explicit]
  coordinates {
    (1,15.20) +=(0,4.00) -= (0,4.00)
    (3,19.20) +=(0,4.80) -= (0,4.80)
    (5,23.90) +=(0,2.90) -= (0,2.90)
    (10,26.80) +=(0,4.20) -= (0,4.20)
    (20,29.40) +=(0,3.70) -= (0,3.70)
    (30,29.80) +=(0,3.10) -= (0,3.10)
    (40,29.80) +=(0,3.70) -= (0,3.70)
    (50,32.50) +=(0,3.10) -= (0,3.10)
  };
  \end{axis}
\end{tikzpicture}
\end{subfigure}
\vskip\baselineskip
\begin{subfigure}[b]{.49\textwidth}
\centering
\pgfplotsset{scaled x ticks=false}
\begin{tikzpicture}[thick,scale=0.78, every node/.style={transform shape}]
  \begin{axis}[
    xlabel={Target data ratio, \%},
    ylabel={Entity F1, \% (weather)},
    xmin=0, xmax=50,
    ymin=20, ymax=70,
    xtick={0, 10, 20, 30, 40, 50},
    ytick={20, 30, 40, 50, 60, 70},
    ymajorgrids=true,
    xmajorgrids=true,
    grid style=dotted,
  ]

  \addplot[dashed, color=blue]
  coordinates {
    (0,47)(3,47)(5,47)(10,47)(20,47)(30,47)(40,47)(50,47)
  };
  \addplot[color=red, mark=square, error bars/.cd, y dir=both, y explicit]
  coordinates {
    (1,43.00) +=(0,10.50) -= (0,10.50)
    (3,44.00) +=(0,11.70) -= (0,11.70)
    (5,39.50) +=(0,14.80) -= (0,14.80)
    (10,48.00) +=(0,5.60) -= (0,5.60)
    (20,53.10) +=(0,4.00) -= (0,4.00)
    (30,52.80) +=(0,3.70) -= (0,3.70)
    (40,54.50) +=(0,2.30) -= (0,2.30)
    (50,50.20) +=(0,4.30) -= (0,4.30)
  };
  \end{axis}
\end{tikzpicture}
\end{subfigure}
\begin{subfigure}[b]{.49\textwidth}
\centering
\pgfplotsset{scaled x ticks=false}
\begin{tikzpicture}[thick,scale=0.78, every node/.style={transform shape}]
  \begin{axis}[
    xlabel={Target data ratio, \%},
    ylabel={Entity F1, \% (schedule)},
    xmin=0, xmax=50,
    ymin=20, ymax=70,
    xtick={0, 10, 20, 30, 40, 50},
    ytick={20, 30, 40, 50, 60, 70},
    ymajorgrids=true,
    xmajorgrids=true,
    grid style=dotted,
  ]

  \addplot[dashed, color=blue]
  coordinates {
    (0,62.9)(3,62.9)(5,62.9)(10,62.9)(20,62.9)(30,62.9)(40,62.9)(50,62.9)
  };
  \addplot[color=red, mark=square, error bars/.cd, y dir=both, y explicit]
  coordinates {
    (1,40.50) +=(0,6.30) -= (0,6.30)
    (3,38.20) +=(0,5.80) -= (0,5.80)
    (5,41.10) +=(0,3.80) -= (0,3.80)
    (10,42.80) +=(0,2.60) -= (0,2.60)
    (20,43.00) +=(0,5.30) -= (0,5.30)
    (30,42.00) +=(0,3.80) -= (0,3.80)
    (40,45.10) +=(0,3.20) -= (0,3.20)
    (50,42.80) +=(0,2.80) -= (0,2.80)
  };
  \end{axis}
\end{tikzpicture}
\end{subfigure}
\ref*{named_ext_data}
\caption{\diktnet performance with extended amounts of target data used for training}
\label{fig:extended_data}

\end{figure}

We performed an additional experiment with extended amounts of target data (see Figure
\ref{fig:extended_data}). It showed that \diktnet, when trained with as little as 5\% of target data,
can outperform a \KVRet trained using the entire dataset. Furthermore, with 50\% of the target data,
\diktnet becomes more than twice as good as KVRet in terms of overall language generation.

However, goal-oriented metrics such as Entity F1 are more challenging to bootstrap. As such,
\diktnet outperforms \KVRet on `Weather' domain starting at 10\% of the target data, but only has a
trend on narrowing down the performance gap with KVRet on `Navigate', and certainly needs more
training data in the `Schedule' domain.

The explanation for that might be that most of the dialogue entities come from the KB snippets which
are the least represented resource in our setup. They are not available in MetaLWOz, and in \SMD,
\KB snippets share little in common across domains. Therefore, in order to increase Entity F1,
\KB information should be directly copied to the output more efficiently~--- and increasing the
robustness of the copy-augmented decoder is one of our future research directions.

\subsection{Discussion of the Evaluation Metrics}

\begin{table}[t]
\small
\center
\begin{tabularx}{\textwidth}{lllll}\toprule
\textbf{Domain}&\multicolumn{2}{c}{\textbf{Context}}&\textbf{Gold response}&\textbf{Predicted response}\\\midrule

schedule&\texttt{<usr>}&Remind me to take my pills&Ok setting your&Okay, setting \textit{a}\\
&\texttt{<sys>}&What time do you need&medicine appointment&\textit{reminder to take}\\
&&to take your pills?&for 7pm&\textit{your pills at 7 pm.}\\
&\texttt{<usr>}&I need to take my pills at 7 pm.&&\\
\midrule
navigate&\texttt{<usr>}&Find the address to a hospital&Have a good day&\textit{No problem.}\\
&\texttt{<sys>}&Stanford Express Care is&&\\
&&at 214 El Camino Real.&&\\
&\texttt{<usr>}&Thank you.&&\\
\midrule
weather&\texttt{<usr>}&What is the weather forecast&For what city would you&For what city would\\
&&for the weekend?&like to know that?&you like \textit{the week-}\\
&&&&\textit{end forecast for?}\\
\bottomrule
\end{tabularx}
\caption{\diktnet's selected responses}
\label{tab:examples}
\end{table}

We use \BLEU as one of the main evaluation metrics in this work~--- we do it in order to fully
conform with the setup of \cite{DBLP:conf/sigdial/ZhaoE18} which we base our work on.
But while being widely adopted as a general-purpose language generation metric, \BLEU might not be
sufficient in the dialogue setting (see \cite{Novikova.etal17} for a review).
Specifically, we have observed several cases where the model would produce an overall grammatical
response with the correct dialogue intent (e.g.\ ``You are welcome! Anything else?''), but \BLEU
would output a lower score for it due to word mismatch (e.g.\  ``You're welcome!'';
see more examples in Table \ref{tab:examples}). This is a general issue in dialogue model evaluation
since the variability of possible responses equivalent in meaning is very high in dialogue.
We think that putting more emphasis on the meaning of utterances, for example by incorporating
external dialogue act tagging resources in the evaluation setup which, together with general
language generation metrics like perplexity, can make for more robust evaluation criteria than word
overlap.

\section{Conclusion}
\label{ch4:future}

In this chapter, we have introduced \diktnet, a model achieving strong dialogue response generation
performance in a few-shot setup, without using any annotated data. By transferring latent dialogue
knowledge from multiple sources of varying generality, we obtained a model with superior
generalisation to underrepresented domains. Specifically, we showed that our few-shot approach
improves upon the previous best model on the Stanford Multi-Domain dataset while being more
data-efficient, by requiring significantly less data none of which has to be annotated.

In the next chapter, we will continue our study of low-resource dialogue generation by addressing
the problem of fast adaptation of a dialogue system to a new domain, at a greater scale of 47
information-seeking domains (i.e. a version of MetaLWOz which becomes our target dataset). As such,
we will explore alternative ways of using the support in-domain data other than fine-tuning the base
model on it.


\chapter{Dialogue Domain Adaptation} 

\label{Chapter5} 

\lhead{Chapter 5. \emph{Dialogue Domain Adaptation}} 


In this chapter, we take our research on few-shot dialogue modelling further and continue with the
problem of fast domain adaptation for dialogue systems. As argued in the previous chapters, domain
adaptation is the key approach to the development of data-efficient dialogue systems in the machine
learning framework; here we are going to explore this problem at a greater scale, i.e. through
the Eighth Dialog System Technology Challenge (\DSTC), Fast Domain Adaptation task. Specifically, we
propose the hybrid \textit{Generative-Retrieval\footnote{Following \cite{gpt2} and
\cite{zhang2019dialogpt}, we use the term ``generative'' in the sense of predicting (or generating)
the next word given the context, i.e. language generation.} Transformer, GRTr}\footnote{Code is
available at \colorhref{blue}{http://tiny.cc/grtr}}~--- a model leveraging knowledge transfer from a
large-scale pre-trained general-purpose language model and combining it with the response retrieval
logic. The model is able to maintain goal-oriented dialogue in a closed domain having only been
exposed to a small set of in-domain dialogues as the domain description. Our hybrid model is ranked
1st on the \metalwoz dataset as per human evaluation, and also performs competitively on automated
metrics when compared to other baselines~---both generation-only and retrieval-only models.

\section{Fast Domain Adaptation of a Dialogue System}
\label{ch5:problem}

\begin{figure}[t]
  \centering
  \includegraphics[width=1.0\linewidth]{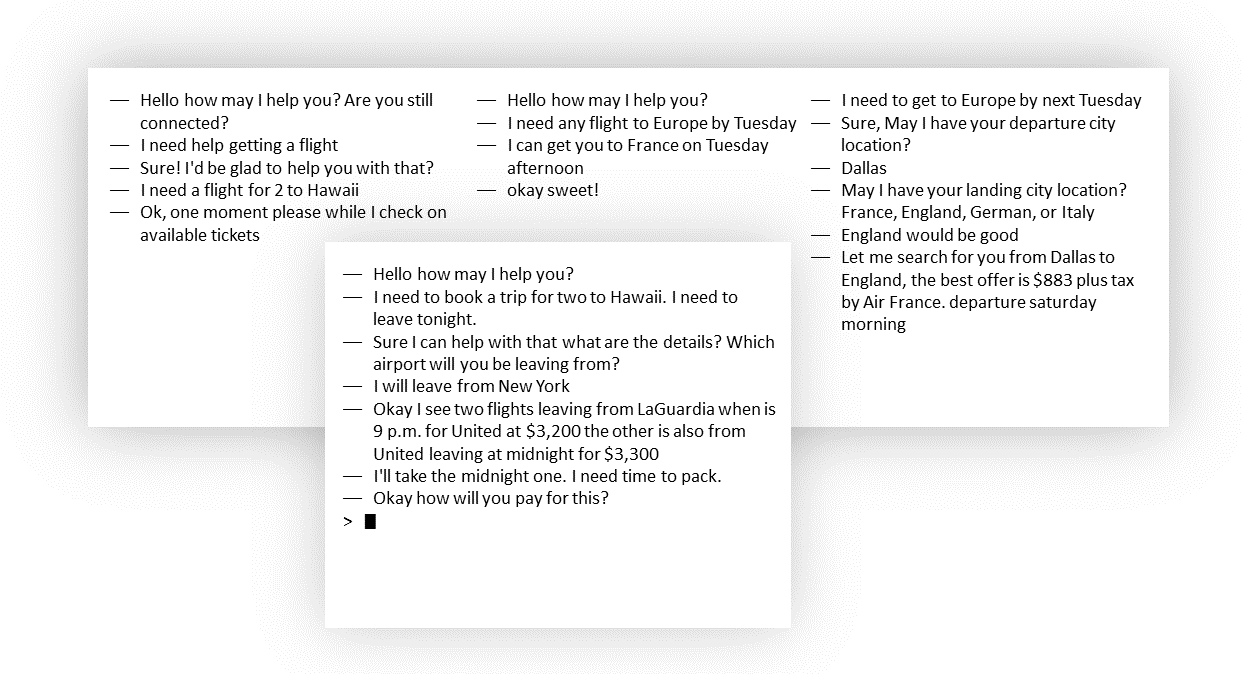}
  \caption[Example DSTC-8 support set and target dialogue in the travel domain]{Example \DSTC[8]
  support set (background) and target dialogue (foreground) in the travel domain}
  \label{fig:dstc8_data}
\end{figure}

As we saw in the previous chapter, few-shot knowledge transfer is a promising way to adapt a
goal-oriented dialogue system to a new domain. In this chapter, we are going to continue working in
the framework of \metalwoz dataset, but will increase the scale by making it our target dataset.
Since \metalwoz does not contain any goal-oriented annotations but overall represents cooperative
information-seeking dialogue between two humans, this task focuses on predicting the utterances on
the user's side. This can be considered a more challenging task since normally, user's utterances are
less predictable than those of the system, and a prospective successful dialogue model should have
a representation of the underlying user's goal in order to generate relevant queries to the system.
Our task takes place within the \DSTC[8] which we are going to describe next.


\subsection{\DSTC[8], Fast Domain Adaptation Task}
In the Eight Edition of \DSTC, its Domain Adaptation task focuses on building a model that predicts
user responses for a goal-oriented dialogue system for which only limited in-domain data is
available. The possible applications of an adaptive user-side dialogue model include Reinforcement
Learning-based setups which are highly dependent on the quality of the user simulator, as well as
data augmentation approaches for improving robustness and coverage of the target models.

The in-domain adaptation data could be collected from e.g. customer service transcripts, or written
by the developers themselves. From this in-domain data, the \emph{support set}, one would like to
extrapolate responses to novel dialogue contexts (the \emph{target})~--- see example in Figure
\ref{fig:dstc8_data}. However, the support set is typically too small to train a dialogue response
generation model. Instead, the approach assumed in the challenge is to adapt (or fine-tune) a
generic dialogue model trained on a large corpus of conversations over multiple \emph{source}
domains.

Technically, the problem setup is as follows: having trained the base model on the source domains,
the model is then fed with one target dialogue and a support set at a time. The model's task is to
predict the next user turn of the target dialogue, taking into account the support set before
producing a prediction. At prediction time, each target dialogue is processed in isolation from
other target dialogues, such that the model cannot use knowledge or state obtained from other
target/support data.

\section{Proposed Model}
\label{ch5:model}

\begin{figure}[t]
  \centering
  \includegraphics[width=0.95\linewidth]{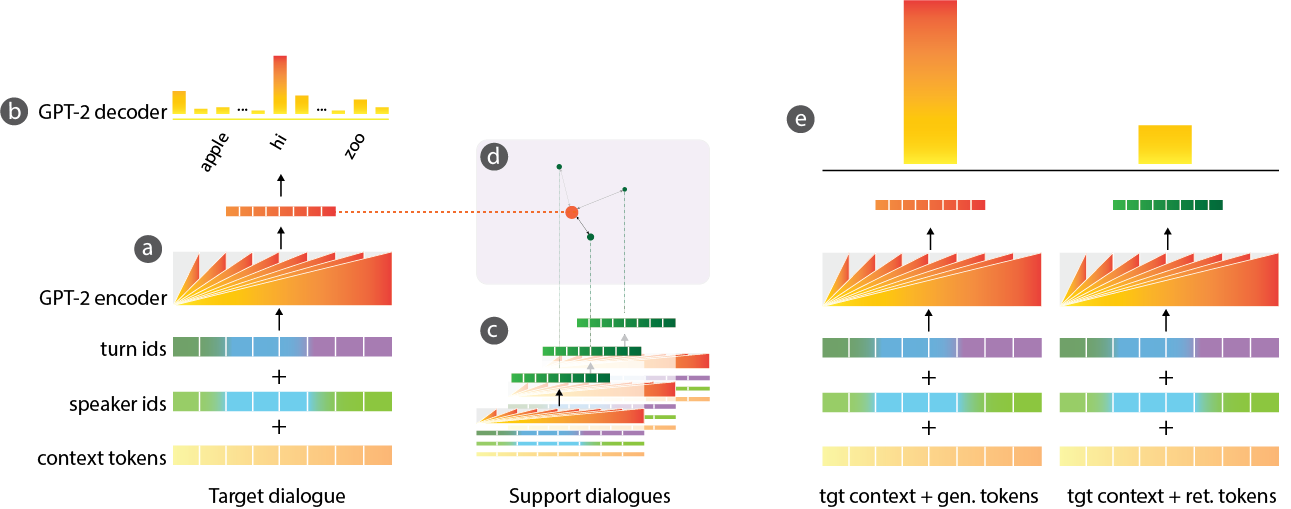}
  \caption[\GRTr model diagram]{\GRTr model diagram. We (a) encode the target dialogue context and (b)
  produce the `generated candidate'; next, we (c) encode support dialogue contexts in a similar way,
  then (d) find the nearest `support' neighbour and select its response as the `retrieved candidate';
  finally, we (e) rank the two candidates given the target context and produce the final result.}
  \label{fig:grtr_model}
\end{figure}

We use a language model pre-trained on a very large and diverse collection of textual data providing
a strong language prior and then adapt the model for our tasks in the form of fine-tuning. Our base
model is \GPT[2] \citep{DBLP:journals/corr/abs-1901-08149}, a transformer-based language model.
In order to adapt \GPT[2] for dialogue generation, we first augment the input embedding for each
token in the dialogue with (1) a speaker tag embedding identifying the speaker and (2) a turn
embedding, identifying the turn number in the current dialogue. These additional embedding matrices
are learned solely using the dialogue data. The input token embeddings are then obtained by summing
up these representations. We also add two task-specific output layers (or ``heads'') for our
purposes: a language modelling (\LM) head and a next-sentence prediction (\NSP) classification head,
both trained from randomly initialised parameters.

We fine-tune \GPT[2] for response generation by minimising the negative log-likelihood of response
tokens given the concatenation of dialogue context and the previous tokens in the response,
%
\begin{align}
  \label{eq:l_lm}
  \begin{split}
    \mathcal{L}_{\text{LM}} &= -\log P_{\text{LM}}(X \mid C) \\
    &= -\sum_{i=1}^{\left | X \right |}{\log P_{\text{LM}}(x_i \mid x_{i-1}, ..., x_{1}, C)},
  \end{split}
\end{align}
where $X$ is the response and $C$ is the dialogue context,~i.e. the concatenation of the tokens
in the previous utterances.

To predict the next sentence, we proceed as follows: given a context/response pair $(C, X)$, the
classification head is trained to produce a binary label $y$, which is $1$ if $X$ is the correct
response given the context $C$, and $0$ if $X$ is a distractor (a random utterance from the corpus).
We minimise the following binary cross-entropy:
\begin{gather}
  \label{eq:l_nsp}
  \begin{aligned}
    \begin{split}
      \mathcal{L}_{\text{NSP}} = - y \log P_{\text{NSP}}(y \mid X, C)
      - (1 - y) \log P_{\text{NSP}}(1 - y \mid X, C),
    \end{split}
  \end{aligned}\\
  \label{eq:nsp}
    P_{\text{NSP}}(y \mid X, C) = \text{softmax}(f_{\text{NSP}}(h_{X,C})),
\end{gather}
where $h_{X, C}$ is the last hidden state of the last \GPT[2] layer after having encoded the
concatenation of $X$ and $C$ and $f_{\text{NSP}}$ is the next-sentence prediction head (in our case
a simple linear transformation). In practice, for each $(C, X)$ pair in the corpus, we sample 1
distractor $\bar X$.


We obtain a suitable dialogue prior by fine-tuning the modified \GPT[2] model on the source domains
with both the language modelling and next-sentence prediction tasks as described above, therefore
minimising $\mathcal{L} = \mathcal{L}_{\text{NSP}} + \mathcal{L}_{\text{LM}}$.


\subsection{Fine-tuning on Target Domains and Prediction}
\label{ch5:pred}

As every test dialogue in the target domain/task is accompanied with a small support set of
dialogues from the same domain/task, we make use of this data by further fine-tuning the dialogue
model on the support dialogues. Crucially, we make sure not to accumulate any information between
test dialogues: after each fine-tuning on the support set, we reset the weights of the model to the
dialogue prior obtained by the fine-tuning stage described in the previous section.


In order to add diversity to the responses, \GPT[2] uses nucleus (top-$p$) sampling
\citep{holtzman.etal2020} during generation, i.e. the model's vocabulary $V$ is pruned to $V^{p}$,
the smallest set such that
\begin{align}
  \sum_{x \in V^{p}}{p(x \mid x_{1:i-1}, C)} \ge p,
\end{align}
and the final distribution from which the words are sampled is rescaled as follows:
\begin{align}
  P'(x \mid x_{1:i-1}) = 
  \begin{cases}
      \frac{P(x \mid x_{1:i-1}, C)}{\sum_{x \in V^{p}}{P(x \mid x_{1:i-1}, C)}}
      & \text{if } x \in V^{(p)}\\
      0,  & \text{otherwise.}
  \end{cases}
\end{align}

\subsection{Hybrid Generative-Retrieval Prediction}
\label{ch5:hybrid}

\IncMargin{1.5em}
\begin{algorithm}[H]
\DontPrintSemicolon
\SetAlgoLined
\LinesNumbered
  \Indm
  \KwIn{Enc~--- \GPT[2] encoder}
  \KwIn{Dec~--- \GPT[2] decoder (language modelling head)}
  \KwIn{NSP~--- \GPT[2] next sentence prediction head}
  \KwIn{$t$~--- turn number to predict}
  \KwIn{$X\_tgt$~--- target dialogue context of length $t - 1$}
  \KwIn{$X\_sup$~--- support dialogues (sequences of turns), each of length $\ge t$}
  \Indp
  $emb\_tgt \gets \text{Enc}(X\_tgt)$

  \ForEach{$i \in 1 \ldots \left| X\_{sup} \right| $}{
    $emb\_sup_i \gets \text{Enc}(X\_sup_{i, 1 \ldots t - 1})$
  }
  \tcc{Euclidean distance used}
  $j \gets \underset{i}{\operatorname{arg\,\min}} \, \braces*{~ \text{dist}(emb\_tgt, emb\_sup_i),%
    ~ i \in 1 \ldots \left | X\_{sup} \right | ~}$

  $y\_gen \gets \text{Dec}(emb\_tgt)$

  $y\_ret \gets X\_sup_{j,t}$

  $cands \gets \brackets{~ y\_gen,~y\_ret ~}$

  \tcc{$\oplus$ denotes concatenation}
  $k \gets \underset{i}{\operatorname{arg\,\max}} \, \braces*{~ \text{NSP}(\text{Enc}(X\_tgt \oplus cands_i)),%
    ~ i \in 1 \ldots \left | cands \right | ~}$

    \KwRet{$cands_k$}
  \caption{Hybrid generative-retrieval response prediction}
  \label{algo:grtr-prediction}
\end{algorithm}
\DecMargin{1.5em}

In our experiments, we found that retrieval baselines are quite effective in the automatic metrics
considered. Therefore, we combined retrieval techniques with our response generation model in a
hybrid approach~-- see Algorithm \ref{algo:grtr-prediction}.

The retrieval component is set up as follows: when predicting the $t$-th turn of the test dialogue,
the model embeds its context of length $t-1$ as well as all the support dialogue contexts of the
same length $t-1$ using the fine-tuned dialogue encoder. The encoding for the dialogue context is
the hidden state of the last layer of the Transformer model at the position corresponding to the
last token in the context. Then, it selects the nearest support context to the target context and
picks its $t$-th turn as the retrieved candidate response.

Finally, the model's own generated response and the best retrieved candidate response are ranked
using the \NSP classification head,~i.e.\ both responses are concatenated with the ground-truth
context and the one with the higher $P_{\text{NSP}}$ (Eq.~\ref{eq:nsp}) is selected. The above steps
are visualised in Figure~\ref{fig:grtr_model}.

\section{Baselines and Competing Models}
\label{ch5:baselines}


We compare our hybrid model to the retrieval baselines provided by the \DSTC[8] organisers. The
baselines ignore the training data and rely solely on the support sets: they embed each support
dialogue's context and find the one nearest to the target context using cosine distance as the
metric. They then return the turn following the identified context as the predicted response.
There are two retrieval-only baselines, which differ in their encoder:
(1) \BERT-based \citep{DBLP:conf/naacl/DevlinCLT19}, taken off-the-shelf, and
(2) SentencePiece/FastText-based~--- representing text as sequences of subword units (`pieces'),
with subword tokenisation logic trained in an unsupervised way, in our case on on the Reddit
Conversations corpus (the approach is modelled after \citealp{DBLP:conf/emnlp/GuWCLC18}).

Another baseline provided is a generation-only model, a bidirectional \LSTM-based \HRED
\citep{Serban:2016:BED:3016387.3016435} trained on \metalwoz.

All the submissions at the final stage of the challenge are as follows \citep{li2020results}:
\begin{itemize}[label=---]
  \item \textbf{Team A} trained a Bi\LSTM on the provided Reddit corpus, then fine-tuned the model at
  test-time using a mixture of \metalwoz and \multiwoz support dialogues, augmented to the context of
  the target dialogue, and dynamically-sampled Reddit threads,
  \item \textbf{Team B}~--- the work described in this chapter,
  \item \textbf{Team C} first fine-tuned \GPT[2] on the MetaLWOz training corpus, then fine-tuned it
  further on the support sets of the \metalwoz and \multiwoz test sets,
  \item \textbf{Team D} trained a Bi\LSTM encoder and attentional \LSTM decoder on both Reddit and
  \metalwoz training corpora, without any fine-tuning to the test sets.
\end{itemize}

\section{Datasets}
\label{ch5:data}

We use the main dataset for \DSTC[8] Track 2 ``Fast Domain Adaptation'' \metalwoz which we described
earlier in Section \ref{ch2:data_collection}. Example dialogues from \metalwoz can be found in
Appendix \ref{AppendixA}.

For evaluation purposes, the challenge organisers provide \multiwoz also described in Section
\ref{ch2:data_collection}. \multiwoz is not present at the base training stage, and a given dialogue
model only gets exposed to this data via support dialogues during the adaptation stage, therefore it
is used as a means for evaluating the adaptation performance to the data substantially different
from the main trainset. Dialogues in \multiwoz contain \NLU annotations, particularly for intent and
slots, which we use in order to to evaluate the systems' goal-oriented performance~--- see some
example utterances in Table \ref{tab:multiwoz_examples}.

\begin{table}[t]
  \center
  \footnotesize
  \begin{tabularx}{\linewidth}{@{}ll@{}}\toprule

  {\bf Utterance}&I am looking for a particular restaurant . It is called pizza hut city centre .\\
  {\bf Markup}&{\tt \{'Restaurant-Inform': [['Name', 'pizza hut city centre']]\}}\\
  \hline

  {\bf Utterance}&I am looking for a place to to stay that has cheap price range it should be in a type of hotel\\
  {\bf Markup}&{\tt \{{'Hotel-Inform': [['Type', 'hotel'], ['Price', 'cheap']]}\}}\\
  \hline

  {\bf Utterance}&I would like a taxi from Saint John 's college to Pizza Hut Fen Ditton .\\
  {\bf Markup}&{\tt \{'Taxi-Inform': [['Dest', 'pizza hut fen ditton'],}\\
  &\-\hspace{3cm} {\tt ['Depart', "saint john 's college"]]\}}\\
  \hline

  {\bf Utterance}&Okay that will work . Can you please tell me their phone number , postcode and the entrance fee ?\\
  {\bf Markup}&{\tt \{'Attraction-Request': [['Fee', '?'], ['Post', '?'], ['Phone', '?']]\}}\\
  \hline

  {\bf Utterance}&Just any time after 10:00 , can I get the train ID of one of them please ?\\
  {\bf Markup}&{\tt \{'Train-Inform': [['Leave', '10:00']], 'Train-Request': [['Id', '?']]\}}\\
  \hline

  \bottomrule
  \end{tabularx}
  \caption{Example annotated utterances from \multiwoz}
  \label{tab:multiwoz_examples}
\end{table}

\section{Experimental Setup and Evaluation}
\label{ch5:setup}

We perform training in two stages: training of the base model and fine-tuning it to the target
dialogue's support set. At the first stage, we train the model for the maximum of 5 epochs with
early stopping. The fine-tuning stage goes on for 1 epoch. \GPT[2] models use the context of 3
exchanges, or 5 turns: bot-user-bot-user-bot, predicting the next user's utterance. We mainly used
the `small' \GPT[2] checkpoint by HuggingFace~--- we also tried the `medium' one, but found no
improvement with it in our task.

\subsection{Human Evaluation}

\begin{table}[ht!]
  \centering
  \begin{tabularx}{0.6\linewidth}{@{}cXS{}}\toprule
  \textbf{Rank}&\multicolumn1l{\textbf{Submission}}&\multicolumn1c{\textbf{Win rate (\%)}}\\
  \midrule
  1&Gold response&62.32\\\cmidrule(r){1-2}
  \textbf{2}&\textbf{Team B (ours)}&\textbf{56.85}\\ 
  3&Team C&52.07\\
  4&Team A&47.35\\
  5&Baseline 1&44.18\\
  6&Team D&37.34\\\bottomrule
  \end{tabularx}
  \caption{Ranking from judges' pairwise comparisons}
  \label{tab:win-rate}
\end{table}

The main systems' goal is to generate appropriate responses towards maintaining a natural
cooperative dialogue on the user's side, so the main evaluation is performed involving human judges.
Specifically, Amazon Mechanical Turk workers were tasked to compare the candidate responses given
the dialogue context. Each comparison was pairwise between the results of two systems presented in
random order. Judges ranked the responses against the following criteria \citep{li2020results}:

\begin{itemize}[label=---]
  \item \textit{Usefulness}~--- whether the response is useful given the dialogue context and the
  user's overall final goal,
  \item \textit{Informativeness}~--- whether the response specifically contains information relevant
  to the conversation,
  \item \textit{Appropriateness}~--- whether the response is appropriate (on-topic, of a reasonable
  length, not repetitive) to the conversation,
  \item \textit{Easiness to answer}~--- given a hypothetical conversational bot on the system side,
  whether the response will be a valid input for it and presumably straightforward to process.
\end{itemize}

For each pairing, 3 independent comparisons were performed against each metric. 
The number of comparisons required was reduced by letting the \textsc{Multisort} algorithm
\citep{DBLP:conf/icml/MaystreG17} determine which responses to compare, causing more similar systems
with similar performance to be compared more often with each other. Bootstrapping over the 100
randomly chosen dialogue contexts was used to determine average ranks and assess the ranking
robustness \citep{hall2009using}.

\subsection{Automatic Evaluation}

\begin{figure}[t]
  \centering
  \includegraphics[width=\linewidth]{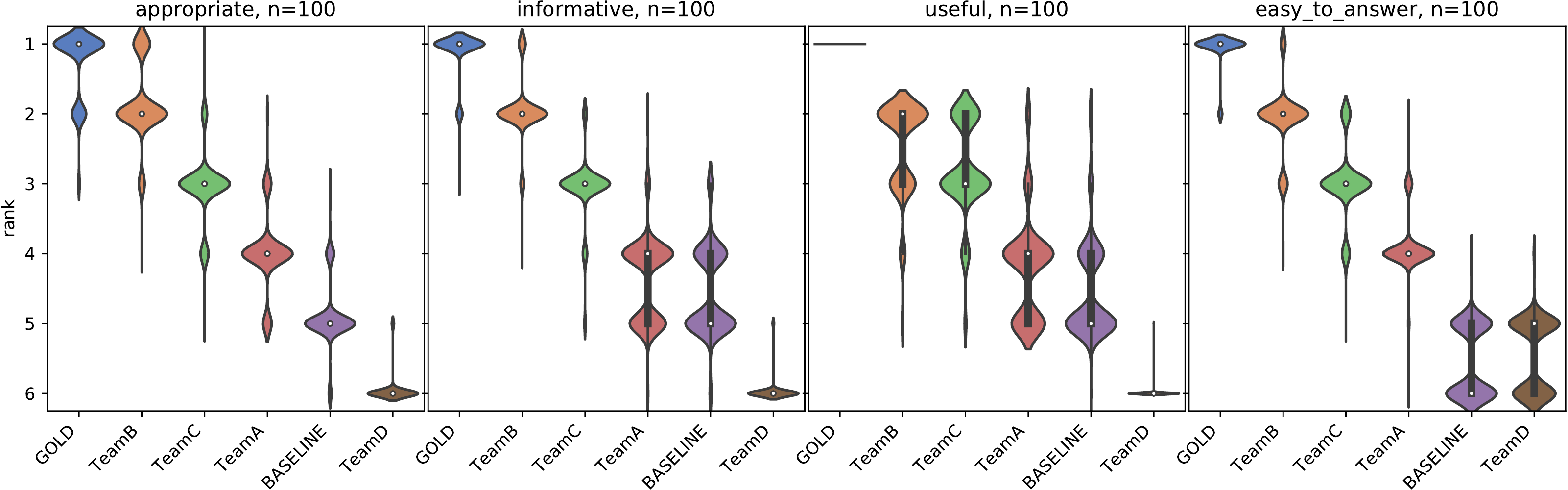}
  \caption{\DSTC[8] Fast Domain Adaptation~--- human evaluation}
  \label{fig:human-eval-by-metric}
\end{figure}

In addition to human evaluation, we also assess model performance using automatic metrics.
The models were evaluated on \metalwoz against word-overlap metrics such as \BLEU[1--3], \CIDEr,
\METEOR, \ROUGEL using the \NLGEval package \citep{sharma2017nlgeval}. Although not ideal for the
specifics of dialogue and spoken language in general (\citealp{adem17};
 \citealp{dziri19-evaluating}), such metrics approximate the overall quality of a response
 generation model and are especially useful for intermediate evaluation. We evaluate models in two
modes on \metalwoz: in \textit{pure task}, support dialogues are drawn from the same domain and task
as target dialogue; in \textit{cross-task}, support and target dialogues are from the same domain,
but different tasks.

We also perform additional evaluation of Entity/Intent F1 of the \multiwoz dataset in pure task mode
with pre-trained \NLU taggers from the \textsc{ConvLab} package \citep{lee2019convlab}. There is no
\multiwoz data available at the first stage (base model training), so all the exposure our model has
to this dataset is via support dialogues. Complementary to \metalwoz evaluation, this stage is
designed for assessing the models' goal-oriented performance.

\section{Results and Discussion}
\label{ch5:results}

\subsection{Human Evaluation}
Results of pairwise comparisons are shown in Table~\ref{tab:win-rate}. Our \GRTr system's responses
(Team~B) were preferred by the judges in \textbf{56\%} of direct comparisons. This surpasses the
next best system (Team C) performance by more than \textbf{4\%}, with only the gold human responses
being chosen more frequently.

Furthermore, from the bootstrap ranking distribution (Figure~\ref{fig:human-eval-by-metric}, lower
rank numbers are better), we see that, apart from the gold human responses (blue graphs), our model's
outputs (orange graphs) are consistently preferred over other submissions by the judges. Of all
metrics used, the most notable are `appropriateness' and `usefulness'. On the former, \GRTr
responses have the second visible peak at rank 1 competing with gold responses. On usefulness
however, rank~1 is held by the gold responses with no variation, and our model has the second
visible peak at rank~3, thus almost tying with Team C (green graphs).

\subsection{Automatic Evaluation}

\definecolor{chromeyellow}{rgb}{1.0, 0.65, 0.0}

\pgfplotsset{compat=1.7}

\begin{figure}[t]
  \small
  \centering
  \begin{subfigure}[b]{0.49\textwidth}
      \begin{tikzpicture}[thick,scale=0.89, every node/.style={transform shape}]
      \centering
      \begin{axis}[
            ybar, axis on top,
            title={\metalwoz pure task (\%)},
            bar width=0.1cm,
            ymajorgrids, tick align=inside,
            ymin=0, ymax=15,
            axis x line*=bottom,
            axis y line*=left,
            enlarge x limits=true,
            legend style={
                /tikz/every even column/.append style={column sep=0.4cm}
            },
            legend to name=named_grtr_metalwoz,
            legend columns=-1,
            legend entries={Ret. \BERT, Ret. \SPFT, \HRED, \GPT[2] base, \GPT[2] +sup, \GRTr},
            ylabel={Percentage},
            symbolic x coords={\BLEU[1], \BLEU[2], \BLEU[3], \ROUGEL},
           xtick=data
        ]
        \addplot [draw=none, fill=blue] coordinates {
          (\BLEU[1],7.93)
          (\BLEU[2],4.43) 
          (\BLEU[3],2.87)
          (\ROUGEL,6.91)};
        \addplot [draw=none, fill=electricgreen] coordinates {
          (\BLEU[1],9.57)
          (\BLEU[2],5.37) 
          (\BLEU[3],3.45)
          (\ROUGEL,7.19)};
       \addplot [draw=none, fill=yellow] coordinates {
          (\BLEU[1],8.66)
          (\BLEU[2],3.86) 
          (\BLEU[3],2.11)
          (\ROUGEL,7.75)};
        \addplot [draw=none, fill=chromeyellow] coordinates {
          (\BLEU[1],8.2)
          (\BLEU[2],3.95) 
          (\BLEU[3],2.22)
          (\ROUGEL,8.34)};
        \addplot [draw=none, fill=orange] coordinates {
          (\BLEU[1],11.33)
          (\BLEU[2],6.45) 
          (\BLEU[3],4.17)
          (\ROUGEL,10.74)};
        \addplot [draw=none, fill=red] coordinates {
          (\BLEU[1],12.73)
          (\BLEU[2],7.43) 
          (\BLEU[3],4.88)
          (\ROUGEL,11.77)};
      \end{axis}
      \end{tikzpicture}
  \end{subfigure}%
  \begin{subfigure}[b]{0.49\textwidth}
      \begin{tikzpicture}[thick,scale=0.89, every node/.style={transform shape}]
      \centering
      \begin{axis}[
            ybar, axis on top,
            title={\metalwoz cross-task (\%)},
            bar width=0.1cm,
            ymajorgrids, tick align=inside,
            ymin=0, ymax=15,
            axis x line*=bottom,
            axis y line*=left,
            enlarge x limits=true,
            symbolic x coords={\BLEU[1], \BLEU[2], \BLEU[3], \ROUGEL},
           xtick=data
        ]
        \addplot [draw=none, fill=blue] coordinates {
          (\BLEU[1],5.35)
          (\BLEU[2],2.16) 
          (\BLEU[3],1.05)
          (\ROUGEL,4.52)};
        \addplot [draw=none, fill=electricgreen] coordinates {
          (\BLEU[1],5.94)
          (\BLEU[2],2.25) 
          (\BLEU[3],0.93)
          (\ROUGEL,4.53)};
       \addplot [draw=none, fill=yellow] coordinates {
          (\BLEU[1],8.94)
          (\BLEU[2],3.87) 
          (\BLEU[3],2.02)
          (\ROUGEL,7.55)};
        \addplot [draw=none, fill=chromeyellow] coordinates {
          (\BLEU[1],8.37)
          (\BLEU[2],3.8) 
          (\BLEU[3],2.05)
          (\ROUGEL,8.55)};
        \addplot [draw=none, fill=orange] coordinates {
          (\BLEU[1],10.21)
          (\BLEU[2],5.26) 
          (\BLEU[3],2.95)
          (\ROUGEL,9.59)};
        \addplot [draw=none, fill=red] coordinates {
          (\BLEU[1],10.39)
          (\BLEU[2],5.31) 
          (\BLEU[3],2.95)
          (\ROUGEL,9.27)};
      \end{axis}
      \end{tikzpicture}
  \end{subfigure}%

  \ref*{named_grtr_metalwoz}
  \caption{Automatic evaluation on \metalwoz}
  \label{fig:results_grtr_metalwoz}
\end{figure}
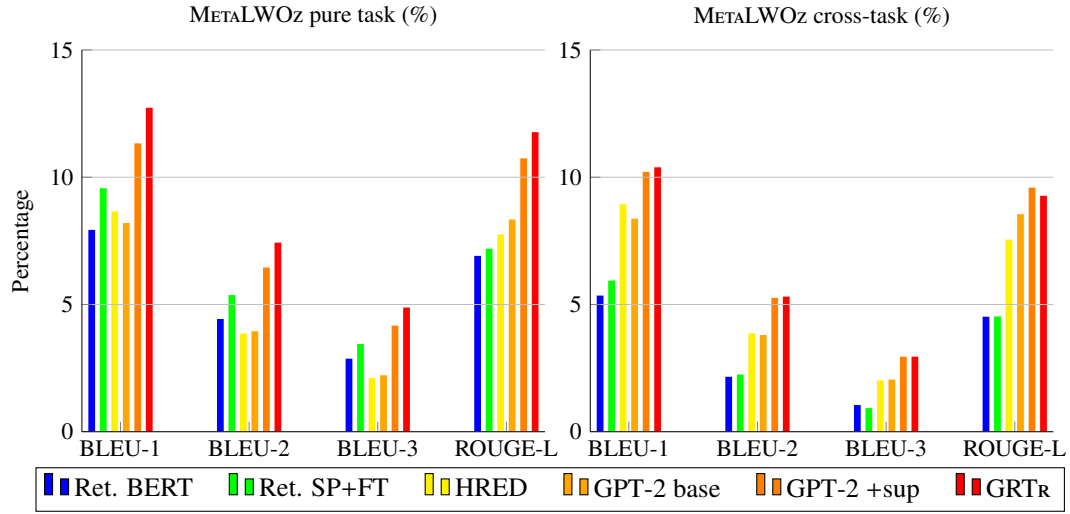

\definecolor{caputmortuum}{rgb}{0.35, 0.15, 0.13}
\definecolor{bulgarianrose}{rgb}{0.28, 0.02, 0.03}
\definecolor{burgundy}{rgb}{0.5, 0.0, 0.13}

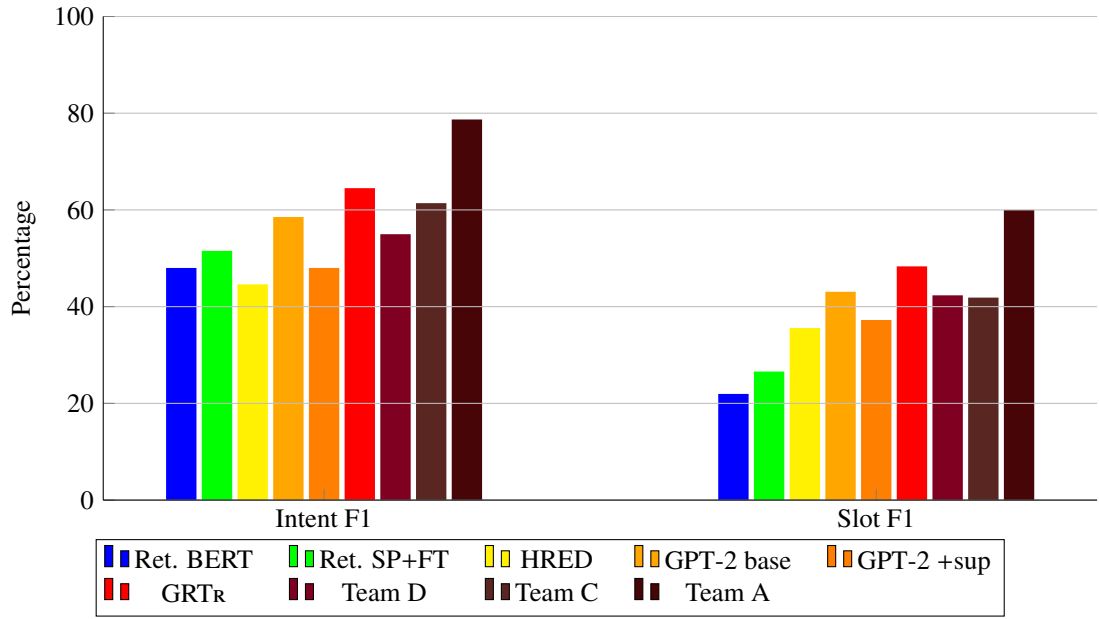
\begin{figure}[t]
  \small
  \centering
  \begin{tikzpicture}[thick,scale=1.0, every node/.style={transform shape}]
  \centering
  \begin{axis}[ybar,
        axis on top,
        bar width=0.4cm,
        height=8cm,
        width=\textwidth,
        ymajorgrids, tick align=inside,
        ymin=0, ymax=100,
        axis x line*=bottom,
        axis y line*=left,
        enlarge x limits=0.4,
        legend to name=named_grtr_multiwoz,
        legend columns=5,
        legend style={
          /tikz/every even column/.append style={column sep=0.4cm}
        },
        legend entries={Ret. \BERT, Ret. \SPFT, \HRED, \GPT[2] base, \GPT[2] +sup, \GRTr, Team D, Team C, Team A},
        ylabel={Percentage},
        symbolic x coords={Intent F1, Slot F1},
       xtick=data
    ]
    \addplot [draw=none, fill=blue] coordinates {
      (Intent F1,48)
      (Slot F1,21.95)};
    \addplot [draw=none, fill=electricgreen] coordinates {
      (Intent F1,51.53)
      (Slot F1,26.58)};
   \addplot [draw=none, fill=yellow] coordinates {
      (Intent F1,44.61)
      (Slot F1,35.57)};
    \addplot [draw=none, fill=chromeyellow] coordinates {
      (Intent F1,58.54)
      (Slot F1,43.07)};
    \addplot [draw=none, fill=orange] coordinates {
      (Intent F1,48)
      (Slot F1,37.24)};
    \addplot [draw=none, fill=red] coordinates {
      (Intent F1,64.5)
      (Slot F1,48.33)};
    \addplot [draw=none, fill=burgundy] coordinates {
      (Intent F1,54.98)
      (Slot F1,42.34)};
    \addplot [draw=none, fill=caputmortuum] coordinates {
      (Intent F1,61.40)
      (Slot F1,41.87)};
    \addplot [draw=none, fill=bulgarianrose] coordinates {
      (Intent F1,78.70)
      (Slot F1,60.00)};
  \end{axis}
  \end{tikzpicture}

  \ref*{named_grtr_multiwoz}
  \caption{Automatic evaluation on \multiwoz}
  \label{fig:results_grtr_multiwoz}
\end{figure}

Results on \metalwoz and \multiwoz against automatic evaluation metrics are shown in Figures
\ref{fig:results_grtr_metalwoz} and \ref{fig:results_grtr_multiwoz}, respectively (more detailed
\metalwoz evaluation is presented in Tables \ref{tab:results-metalwoz-pure} and
\ref{tab:results-metalwoz-cross} of Appendix \ref{AppendixB}). We observe that retrieval baselines
attain very competitive performance on both datasets, with FastText embeddings from Reddit leading
to overall better results than off-the-shelf \BERT, especially in the \textit{pure task} setting.

With \GRTr, we performed an ablation study to have a closer look into its performance. We evaluated
three versions:
\begin{itemize}[label=---]
  \item \GPT[2] base, a generation-only model trained on \metalwoz and not making use of the support
  data,
  \item \GPT[2] +sup, the base model fine-tuned to support data, also not using the retrieval logic,
  \item \GRTr, our full hybrid model.
\end{itemize}

As seen in Figure \ref{fig:results_grtr_metalwoz}, there is strong dependence on support dialogues
(`base' vs\@. `+sup') as the base model mostly struggles to compete with the baselines. Adding
retrieval logic (`GRTr' vs\@.\ `+sup') results in further performance gains. \HRED and
\GPT[2] base, the two models that did not use support dialogues, had comparable performance on
\metalwoz.

In goal-oriented metrics on \multiwoz (see Figure \ref{fig:results_grtr_multiwoz}), the same
performance pattern is observed with retrieval models, but \GPT[2] in the generation-only version
performs surprisingly better when not fine-tuned to support set (`base'). On the other hand, the
hybrid model experiences even more performance gain than on \metalwoz. Presumably, generating
responses for this dataset is harder due to the fact that it is not represented at the main training
stage, and there is not much utterance overlap with \metalwoz, so little knowledge transfer takes
place in this experiment. Compared to other submissions, we observe that \GRTr still outperforms most
of the competitors and only gives way to Team A's system. We hypothesise here the best \multiwoz
model (Team A) was fitted to the automatic evaluation metrics too tightly, with the negative side
effect observable in human evaluation results of Table \ref{tab:win-rate} and Figure
\ref{fig:human-eval-by-metric}, where this system was prevalently ranked 4${}^{\text{th}}$ and
5${}^{\text{th}}$.

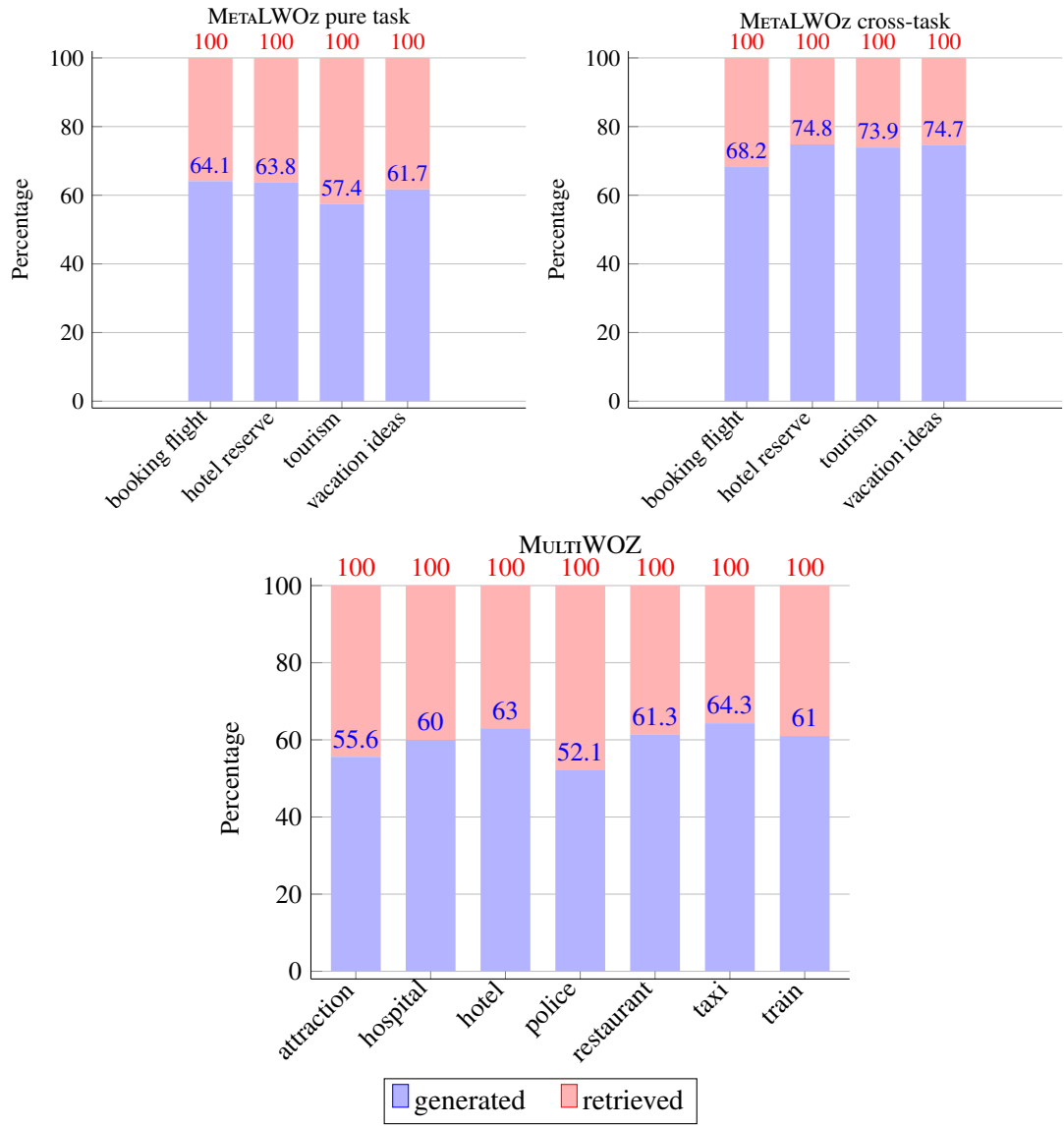
\begin{figure}[t]
  \centering
    \begin{subfigure}[b]{0.49\textwidth}
      \begin{tikzpicture}[thick,scale=0.89, every node/.style={transform shape}]
      \small
      \begin{axis}[
          ybar stacked,
          ymajorgrids,
          height=7cm,
          bar width=19pt,
          nodes near coords,
          ymin=0, ymax=100,
          title=\metalwoz pure task,
          enlarge y limits = 0.02,
          enlarge x limits=0.6,
          axis x line*=bottom,
          axis y line*=left,
          legend to name=named_grtr_gen_ret,
          legend columns=-1,
          legend style={
            /tikz/every even column/.append style={column sep=0.4cm}
          },
          legend entries={generated, retrieved},
          ylabel={Percentage},
          symbolic x coords={booking flight, hotel reserve, tourism, vacation ideas},
          xtick=data,
          x tick label style={rotate=45,anchor=east},
          ]
      \addplot+[draw=none] coordinates {(booking flight,64.1) (hotel reserve,63.8) 
        (tourism,57.4) (vacation ideas,61.7)};
      \addplot+[draw=none] coordinates {(booking flight,35.9) (hotel reserve,36.2) 
        (tourism,42.6) (vacation ideas,38.3)};
  
      \end{axis}
      \end{tikzpicture}
    \end{subfigure}%
    \begin{subfigure}[b]{0.49\textwidth}
      \begin{tikzpicture}[thick,scale=0.89, every node/.style={transform shape}]
      \small
      \begin{axis}[
          ybar stacked,
          ymajorgrids,
          height=7cm,
          bar width=19pt,
          nodes near coords,
          ymin=0, ymax=100,
          title=\metalwoz cross-task,
          enlarge y limits = 0.02,
          enlarge x limits=0.6,
          axis x line*=bottom,
          axis y line*=left,
          ylabel={Percentage},
          symbolic x coords={booking flight, hotel reserve, tourism, vacation ideas},
          xtick=data,
          x tick label style={rotate=45,anchor=east},
          ]
      \addplot+[draw=none] coordinates {
          (booking flight,68.2)
          (hotel reserve,74.8)
          (tourism,73.9)
          (vacation ideas,74.7)
      };
      \addplot+[draw=none] coordinates {
          (booking flight,31.8)
          (hotel reserve,25.2)
          (tourism,26.1)
          (vacation ideas,25.3)
      };
  
      \end{axis}
      \end{tikzpicture}
    \end{subfigure}
  
    \begin{subfigure}[b]{0.6\textwidth}
      \begin{tikzpicture}[thick,scale=1.0, every node/.style={transform shape}]
      \small
      \begin{axis}[
          ybar stacked,
          ymajorgrids,
          bar width=19pt,
          width=\textwidth,
          height=7cm,
          nodes near coords,
          ymin=0, ymax=100,
          title=\multiwoz,
          enlarge y limits = 0.02,
          enlarge x limits=0.1,
          axis x line*=bottom,
          axis y line*=left,
          ylabel={Percentage},
          symbolic x coords={attraction, hospital, hotel, police, restaurant, taxi, train},
          xtick=data,
          x tick label style={rotate=45,anchor=east},
          ]
      \addplot+[draw=none] coordinates {
          (attraction,55.6)
          (hospital,60.0)
          (hotel,63.0)
          (police,52.1)
          (restaurant,61.3)
          (taxi,64.3)
          (train,61.0)
      };
      \addplot+[draw=none] coordinates {
          (attraction,44.4)
          (hospital,40.0)
          (hotel,37.0)
          (police,47.9)
          (restaurant,38.7)
          (taxi,35.7)
          (train,39.0)
      };
  
      \end{axis}
      \end{tikzpicture}
    \end{subfigure}

    \ref*{named_grtr_gen_ret}

    \caption{\GRTr generated/retrieved response ratios}
    \label{fig:grtr-gen-ret-ratios}
\end{figure}

\subsection{Analysis of The Generated/Retrieved Responses}

\begin{table}[t]
  \center
  \small
  \begin{tabularx}{\linewidth}{@{}rll@{}}\toprule
  \textbf{Context}&\texttt{Wiz}&Hello how may I help you?\\
  &\texttt{Usr}&I need some advice about Rome.\\
  &\texttt{Wiz}&What do you want to know?\\
  \textbf{Gold response}&\texttt{Usr}&I like history, where should I go?\\
  \textbf{Predicted response}&\texttt{Usr}&Who was the founder of Rome?\\
  \hline
  
  \textbf{Context}&\texttt{Wiz}&Hello, how may I help you?\\
  &\texttt{Usr}&I'm looking for a train to broxbourne.\\
  &\texttt{Wiz}&Where would you like to leave from, and what day and time would you\\
  &&like to travel?\\
  &\texttt{Usr}&I'm going to be leaving from Cambridge. I need to leave on Tuesday\\
  &&by 13:00.\\
  &\texttt{Wiz}&Is there a time you would like to get there by?\\
  \textbf{Gold response}&\texttt{Usr}&I don't have an arrival preference.\\
  \textbf{Predicted response}&\texttt{Usr}&No I don't have a leave time preference.
  Just not too early\\
  &&if I don't have too.\\
  \hline
  
  \textbf{Context}&\texttt{Wiz}&Hello how may I help you?\\
  &\texttt{Usr}&I'd like to book a flight.\\
  &\texttt{Wiz}&Okay, any particular place in mind?\\
  &\texttt{Usr}&How am I supposed to book a flight to Greece? Me and my kawaii\\
  &&girlfriend were wondering. She says ``Ooooo, Greece. :3''\\
  &\texttt{Wiz}&I can book a flight for two if you want, There will be an evening flight\\
  &\texttt{Usr}&Hmm, wait, I don't really wanna book a flight. We were just curious!\\
  &&She says ``Hey! No bookies! :(''\\
  &\texttt{Wiz}&Oh, I was confused by that. Can I have your email address so I could\\
  &&send several flight options for you\\
  \textbf{Gold response}&\texttt{Usr}&Sure thing! My email address is ``weeabooking@otaku.corn''.\\
  &&She says ``I wanna watch my anime now! ;\_;''\\
  \textbf{Predicted response}&\texttt{Usr}&Well, I guess I'll just get back to you. Thanks!\\
  \bottomrule
  
  \end{tabularx}
  \caption{\GRTr example responses}
  \label{tab:grtr_examples}
\end{table}

In Figure~\ref{fig:grtr-gen-ret-ratios}, we show per-domain ratios of retrieved/generated responses
from the hybrid model. We find that the majority of the responses are generated, and the retrieval
logic works as the fallback option. On \metalwoz, which the model had more exposure to during the
training, generated responses ratio is generally slightly higher than that on \multiwoz which was
only seen by the model via support dialogues. Consequently, the model's overall confidence on this
dataset is lower, which results in more frequent fallbacks.

Generated candidates rarely duplicate the retrieved ones: we found that the percentage of
predictions with identical generated/retrieved candidates is $0.7\%$ (16 in total) for \metalwoz
pure, $0.6\%$ (15 in total) for \metalwoz cross, and $0.3\%$ (10 in total) for \multiwoz. Also, more
detailed information on the distribution of pairwise distances between \GRTr response candidates can
be found in Figures \ref{fig:gen_ret_distances_metalwoz_pure}~--- \ref{fig:gen_ret_distances_multiwoz}
of Appendix \ref{AppendixB}. Moreover, in Tables \ref{tab:grtr_closest_candiates_metalwoz_pure}~---
\ref{tab:grtr_distant_candiates_multiwoz} of Appendix \ref{AppendixB}, we show \GRTr example
predictions with the closest generated and retrieved candidates, as well as the most distant ones
for \metalwoz pure task, \metalwoz cross-task, and \multiwoz datasets. We observe that the
generated/retrieved candidates which were scored close to each other are either paraphrases of some
generic phrase (e.g. ``OK I'll go with that'', ``I like the idea'') or both work well in the
dialogue context (see examples for more detail). On the other hand, in cases with a higher
difference between the generated and the retrieved candidate's scores, most of the time we observe
that the retrieved one didn't match the context very well.

Overall, we observe in Table~\ref{tab:grtr_examples} that there are many cases in the data where the
gold response cannot possibly be inferred from the dialogue context. Specifically, the task was
posed in the way that no extra data, such as a  knowledge base or task description, was provided to
the system~--- therefore, the main goal intended for the hypothetical ideal system is to naturally
model human responses in a co-operative goal-oriented dialogue, and to do that in a data-efficient
way. This is reflected in the way human judges are asked about response quality.

\section{Conclusion}
\label{ch5:future}

We presented a hybrid generative-retrieval approach to goal-oriented dialogue with fast domain
adaptation via transfer learning. It attains robust and diverse language generation performance
across domains, and uses retrieval logic as a fallback mechanism in cases of low confidence.
Our method is ranked 1st by human judges of \DSTC[8] Fast Domain Adaptation task, and it attains
performance superior to a series of baselines in automated metrics on \metalwoz and \multiwoz
datasets. The future directions of this research mainly include incorporating a more principled
fine-tuning technique (i.e. `learning to learn') and will be discussed in detail in Chapter
\ref{Chapter9}.

\chapter{Spoken Disfluency Detection} 

\label{Chapter6} 

\lhead{Chapter 6. \emph{Spoken Disfluency Detection}} 

\newcommand{\mc}[2]{\multicolumn{#1}{c}{#2}}


Starting with this chapter, we will focus on practical aspects of data-efficient dialogue modelling.
The problem we are going to address here refers back to our study in Chapter \ref{Chapter3}. As we
saw, neural dialogue models lack robustness to certain aspects of spoken language such as high
surface variability and the presence of disfluencies (e.g. self-corrections, hesitations, restarts).
And although robustness can be improved using more representative training data, it will require
datasets of impractical sizes to account for all the linguistic phenomena of interest while keeping
the performance on the target downstream task the main priority. From a developer's point of view
though, it is highly desirable to be able to develop systems which can be trained from `clean'
examples while also able to generalise to the very diverse disfluent variations on the same data~---
thereby enhancing both data-efficiency and robustness. Therefore, it can be beneficial for the
dialogue systems research to develop a dedicated model that detects such variations in the user's
input and passes this information into the downstream dialogue pipeline~--- either modular or
end-to-end.

In this chapter, we present a multitask LSTM-based model\footnote{Tensorflow
\citep{tensorflow2015-whitepaper} and PyTorch \citep{paszke2017automatic} implementations together
with the trained models are available at \colorhref{blue}{http://bit.ly/multitask\_disfluency}} for
the incremental detection of disfluency structures, which can be hooked up to any component for
incremental interpretation (e.g. an incremental semantic parser), or else simply used to `clean up'
the current utterance as it is being produced.
We train the system on the Switchboard Dialog Acts (\SWDA) corpus and present its accuracy on this
dataset. Our model outperforms prior neural network-based incremental approaches by about 10
percentage points on \SWDA while employing a simpler architecture. To test the model's generalisation
potential to goal-oriented utterances with their characteristic types of disfluency patters, we
evaluate the same model on the \bAbIplus dataset (presented in Section \ref{ch3:babi+}), without any
additional training. This shows that our approach has good generalisation potential,
and sheds more light on which types of disfluency might be amenable to domain-general processing.

\section{Motivation}
\label{ch6:disfluency-intro}

As discussed in Section \ref{ch2:disfluency}, spontaneous spoken dialogue is often disfluent,
containing pauses, hesitations, self-corrections, and false starts. Processing such phenomena is
essential in understanding a speaker's intended meaning and controlling the flow of the conversation.
Furthermore, this processing needs to be \emph{word-by-word incremental} to allow further downstream
processing to begin as early as possible in order to handle real spontaneous human conversational
behaviour.

In this chapter, we build upon the previous best approaches to incremental disfluency detection, the
neural models of \cite{DBLP:conf/interspeech/HoughS15} and \cite{DBLP:conf/eacl/SchlangenH17}.
Our contributions are that: (1) we produce a new multitask \LSTM-based model with a simpler
architecture for incremental disfluency detection, with significantly improved  performance on the
\SWDA, a disfluency-tagged corpus of open-domain conversations; and (2) we perform a generalisation
experiment measuring how well the models perform on unseen data using the controlled environment
using \bAbIplus \citep{Shalyminov.etal17}, a dataset containing goal-oriented dialogue utterances
with the specific vocabulary and syntactic structures which make it fundamentally different from
\SWDA.

\section{A Multitask \LSTM-based Model for Spoken Disfluency Detection}
\label{ch6:lstm-disfluency-model}
Our approach to disfluency detection is a sequence tagging model which makes single-word predictions
given context words $w_{t-n+1}, ..., w_{t}$ of a maximum length $n$. We train it to perform two
tasks jointly \citep[c.f.][]{DBLP:conf/interspeech/HoughS15}:
\begin{enumerate}
  \item predicting the disfluency tag of
the current word, $P(y_t|w_{t-n+1}, ..., w_{t})$, and
  \item predicting the next word in the sequence in a language model way,
  $P(w_{t+1}|w_{t-n+1}, ..., w_{t})$.
\end{enumerate}

At training time, we optimise the two tasks jointly, but at test time we only look at the resulting
tags and ignore the \LM predictions.

\begin{figure}
\begin{center}
\includegraphics[width=10cm]{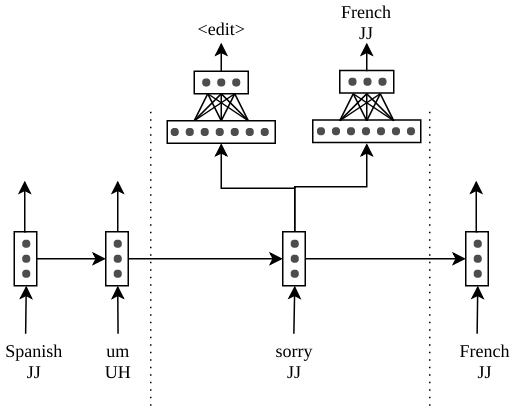}
\end{center}
\caption{Multitask \LSTM disfluency detector architecture}
\label{fig:disfluency_model_architecture}
\end{figure}
Our model uses a shared \LSTM encoder \cite{DBLP:journals/neco/HochreiterS97} with combined
`Word/Part-of-Speech tag' tokens which provides context embedding for two independent multilayer
perceptrons (\MLPs) making the predictions for the two tasks~--- see Figure
\ref{fig:disfluency_model_architecture}. The combined token vocabulary
(word+\POS) size for the \SWDA dataset is approximately 30\%~ larger than the original word-only
version~--- given this, concatenation is the simplest and most efficient way to pass \POS
information into the model.

The intuition behind adding an additional task to optimise for is that it \textit{serves as a
natural regulariser}: given an imbalanced distribution of a very few labels (see Section
\ref{ch6:disfluency-data} for the dataset description), only learning disfluency labels may lead to
a higher degree of overfitting, and introducing an additional task with a significantly wider output
space can help the model generalise better.

Other potential benefits of having the model work as an \LM is the possibility of unsupervised model
improvements, e.g. pre-training of the model's \LM part from larger text corpora or 1-shot
fine-tuning to new datasets with different word sequence patterns.

In order to address the problem of significantly imbalanced training data (the majority of the words
in the corpus are fluent), we use a weighted cross-entropy loss in which the weight of a data point
is inversely proportional to its label's frequency in the training set. Our overall loss function is
of the form:

\begin{equation}
  L = \mathit{WL}_{main} + \alpha L_{lm} + \frac{\lambda}{2}  \sum_{i} {w_i^2}
  \label{eq:multitask_lstm_loss}
\end{equation}

where $\mathit{WL}_{main}$ and $L_{lm}$ are respective losses for the disfluency tagging
(class-weighted) and language modelling tasks (\LM loss coefficient $\alpha$ is tuned empirically).
We use class weights in the main task's loss to deal with the highly imbalanced data (see Section
\ref{ch6:disfluency-data} for an overview od the data), so that the weight of the $k^{th}$ class is
calculated as $W_k=1/{(C_k)^\gamma}$, where $C_k$ is the number of $k^{th}$ class instances in the
training set, and $\gamma$ is a smoothing constant set empirically. The last term in Eq.
\ref{eq:multitask_lstm_loss} is the L2 regularisation which we apply to the model's weight
parameters $w_i$ (those of word embeddings, \LSTM gates, and \MLPs) leaving all the biases intact.
The L2 coefficient $\lambda$ is also tuned empirically.

\begin{figure}
\begin{center}
\includegraphics[width=0.85\textwidth]{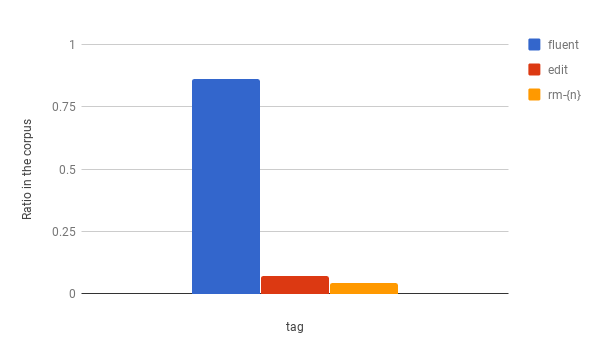}
\end{center}
\caption{Statistics of the \SWDA corpus}
\label{fig:swda_stats}
\end{figure}

\section{The Switchboard Dialog Acts Dataset}
\label{ch6:disfluency-data}

For training our model, we use the Switchboard Dialog Acts dataset (\SWDA) with manually annotated
disfluency tags \citep{swda}. We use a pre-processed version of the dataset by
\cite{DBLP:conf/interspeech/HoughS15} containing 90,497 utterances with transformed tagging:
following their convention, there are 27 tags in total consisting of: {\tt $<$f/$>$} tag for fluent
tokens; {\tt $<$e/$>$} for edit tokens; {\tt $<$rm-\{n\}/$>$} tags for repair tokens that determine
the start of the reparandum to be $n$ tokens/words back; and {\tt $<$rpSub$>$} \& {\tt $<$rpDel$>$}
tags which mark the end of the \texttt{repair} and classify whether the repair is a
\emph{substitution} or \emph{deletion} repair. The latter tokens can be combined with
{\tt $<$rm-\{n\}$>$} tokens, which explains the total of 27 tags~--- see (\ref{repair-rnn}) for an
example where the \texttt{repair} word, `Spanish', is tagged as \texttt{$<$rm-4$><$rpSub$>$} meaning
this is a substitution repair that retraces $4$ tokens back from the current token
(see Figure \ref{fig:swda_stats}).

\begin{equation}
    \underset{\langle f/\rangle}{\textrm{with}}
    \underbrace{[\underset{\langle f/\rangle}{\textrm{Italian}}}_{reparandum}+\underbrace{\{
        \underset{\langle e/\rangle}{\textrm{uh}}\underset{\langle e/\rangle}{\textrm{no}}
        \underset{\langle e/\rangle}{\textrm{uh}}\}}_{interregnum}
        \underbrace{\underset{\langle rpSub\rangle}{\underset{\langle rm-4\rangle
}{\textrm{Spanish}]}}}_{repair} \ \ \underset{\langle f/\rangle}{\textrm{cuisine}}
\label{repair-rnn}
\end{equation}

The distribution of different types of tokens is highly imbalanced: only about 4\% of all tokens are
involved in disfluency structures (the detailed statistics are shown in the tables in the end of
this chapter). See above, Section \ref{ch6:lstm-disfluency-model} for how our model deals with this.

\section{Disfluency Detection Generalisation to \bAbIplus}
To evaluate the out-of-dataset generalisation properties of our model and that of
\cite{DBLP:conf/interspeech/HoughS15}, we employ additional data which we generate using \bAbIplus
tools introduced in Chapter \ref{Chapter3}. \bAbIplus augmentations can be mixed in with complete
control over the syntactic and semantic contexts in which the phenomena appear, and therefore the
\bAbIplus environment allows controlled, focused experimentation of the effect of different
phenomena and their distributions on the performance of different models. Here, we use \bAbIplus
tools to generate new data for the controlled generalisation experiment%
\footnote{Data is available at\colorhref{blue}{http://bit.ly/babi\_plus\_disfluencies\_study}} of
what kinds of disfluency phenomena are captured better by each model.

We focus here on the following disfluency patterns:
\begin{itemize}[label=---]
  \item \textit{Hesitations}, e.g.\ as in ``we will be \textit{uhm} eight'' (mixed in are
  single edit tokens);
  \item \textit{Prepositional Phrase restarts (PP-restart)}, e.g. ``in a \textit{in a um in a}
  moderate price range'' (repair of a PP at its beginning with or without an interregnum);
  \item \textit{Clausal restarts (CL-restart)}, e.g.\ ``can you make a restaurant \textit{uhm yeah
  can you make a restaurant} reservation for four people with french cuisine in a moderate price
  range'' (repair of the utterance from the beginning starting at arbitrary positions);
  \item \textit{Corrections (NP and PP)}, e.g. ``with Italian \textit{sorry Spanish} cuisine'',
  as was initially discussed in Section \ref{ch6:disfluency-intro}.
\end{itemize}

We generated independent \bAbIplus datasets with each disfluency type. The disfluency phenomena above
were chosen to resemble disfluency patterns in the original \SWDA corpus (see Tables
\ref{tab:disfluency-common-repairs}, \ref{tab:disfluency-keyword-repair-patterns} for examples), as
well as intuitive considerations for the phenomena relevant for goal-oriented dialogue (namely,
corrections).

The intuition for a generalisation experiment with data like this is as follows: while having
similar disfluency patterns, our bAbI+ utterances differ from \SWDA in terms of the vocabulary and
the word sequences themselves as they are in the domain of goal-oriented human-computer dialogue~---
this property makes it possible to evaluate the generalisation capabilities of a model outside its
training domain.

\section{Evaluation and Experimental Setup}
\label{ch6:disfluency-eval}

\begin{table}[t]
\centering
\small
\begin{tabularx}{0.6\textwidth}{lllll}\toprule
\textbf{Model}&\textbf{F\textsubscript{e}}&\textbf{F\textsubscript{rm}}
&\textbf{F\textsubscript{rps}}\\
\midrule
\cite{DBLP:conf/interspeech/HoughS15}&0.902&0.711&0.689\\
\cite{DBLP:conf/eacl/SchlangenH17}&0.918&---&0.719\\
\LSTM&0.915&0.693&0.775\\
\ULMFiT \MTLSTM&0.889&0.735&0.795\\
\MTLSTM&\textbf{0.919}&\textbf{0.753}&\textbf{0.816}\\\bottomrule
\end{tabularx}
\caption{Evaluation of the disfluency tagging models on \SWDA}
\label{tab:disfluency-results}
\end{table}

\begin{table}[t]
  \centering
  \small
  \begin{tabularx}{\textwidth}{lccccccc}\toprule
  \textbf{Model}&\textbf{hesitations (F\textsubscript{e})}&\multicolumn{3}{c}{\textbf{PP restarts}}
  &\multicolumn{3}{c}{\textbf{CL-restarts}}\\
  &&\textbf{F\textsubscript{e}}&\textbf{F\textsubscript{rm}}&\textbf{F\textsubscript{rps}}
  &\textbf{F\textsubscript{e}}&\textbf{F\textsubscript{rm}}&\textbf{F\textsubscript{rps}}\\\midrule
  \cite{DBLP:conf/interspeech/HoughS15}&0.917&0.774&0.875&0.877&0.938&0.471&0.630\\
  \LSTM&\textbf{0.956}&\textbf{1.000}&0.982&0.993&0.948&0.360&0.495\\
  \ULMFiT \MTLSTM&0.902&0.917&0.937&0.963&0.958&0.456&0.637\\
  \MTLSTM&0.910&\textbf{1.000}&\textbf{0.993}&\textbf{0.997}&\textbf{0.991}&\textbf{0.484}
  &\textbf{0.659}\\
  \bottomrule
  \end{tabularx}
  \caption[Controlled generalisation evaluation on the 3 \bAbIplus datasets]{Controlled
  generalisation evaluation on the 3 \bAbIplus datasets (one per each disfluency phenomenon)}
  \label{tab:disfluency-controlled-generalisation}
\end{table}

We employ exactly the same evaluation criteria as \cite{DBLP:conf/interspeech/HoughS15}:
micro-averaged F1-scores for edit ($F_e$) and {\tt <rm-\{n\}/>} tokens ($F_{rm}$) as well as for
whole repair structures ($F_{rps}$). We compare our Multitask LSTM (or \MTLSTM) model to its
single-task version (disfluency tag predictions only) as well as to the system of
\cite{DBLP:conf/interspeech/HoughS15} and the joint disfluency tagging/utterance segmentation model
of \cite{DBLP:conf/eacl/SchlangenH17} on all of the applicable word-level metrics on dialogue
transcripts. These use a hand-crafted Markov Model for post-processing, whereas our model works in a
streamlined single-stage way.

Apart from that, we evaluate a version of our model that makes use of pre-trained contextual
embeddings~--- namely, AWD-\LSTM \citep{DBLP:conf/iclr/MerityKS18} trained on WikiText-103 dataset
\citep{DBLP:conf/iclr/MerityX0S17} following the \ULMFiT technique \citep{DBLP:conf/acl/RuderH18}%
\footnote{We used a pre-trained model from \texttt{fast.ai}~--- code available at
\colorhref{blue}{http://tiny.cc/ulmfit\_disfluency}}. Among the rest of widely-used pre-trained
contextual embeddings (e.g. \ELMo, \BERT, \GPT[2]), AWD-\LSTM is a strictly left-to-right
unidirectional model with the possibility of incremental word-by-word state updates which is an
essential feature of all the models in this experiment. In order to make use of the pre-trained
embeddings, we don't use combined word-POS tokens with this model and use separate \LSTM encoders
instead~--- the pre-trained one for words and a from-scratch one for POS-tags, respectively.

We train our model using the \SGD optimiser and monitor the $F_{rm}$ on the dev set as a stopping
criterion. The model's hyperparameters are tuned heuristically, the final values are listed in the
tables in the end of this chapter.

\section{Results}

The results are shown in Table \ref{tab:disfluency-results}. Both single- and multitask \LSTM are
able to outperform the \cite{DBLP:conf/interspeech/HoughS15} model on edit tokens and repair
structures, but the multitask one performs significantly better on {\tt <rm-\{n\}/>} tags and
surpasses both previous models. The reason $F_{rps}$ is higher than $F_{rm}$ in general is that due
to the tag conversion, fluent tokens inside reparandums and repairs are treated as part of repair,
and they contribute to the global positive and negative counters used in the micro-averaged F1.
Large-scale pre-training, while potentially useful for better coverage of the target dataset, did
not affect the final disfluency detection accuracy as \ULMFiT \MTLSTM's results show. The reason for
that could be the initially large size of the underlying pre-trained AWD-\LSTM: it contains 3 layers
with 1024, 1024, and 400 neurons, respectively, which is significantly more than what our best
performing model uses (we also tried only using the first layer of the AWD-\LSTM, which didn't
result in any significant accuracy improvement).

Controlled generalisation experiment results are shown in Table
\ref{tab:disfluency-controlled-generalisation}~--- note that we could only run the model of
\cite{DBLP:conf/interspeech/HoughS15} on bAbI+ data because that of
\cite{DBLP:conf/eacl/SchlangenH17} works in a setup different from ours. It can be seen that the
\LSTM tagger is somewhat overfitted to edit tokens on \SWDA. This is the reason it outperforms the
Multitask \LSTM on the hesitations dataset and has a tied 1.0 on edit tokens on PP restarts dataset.
In all other cases, Multitask LSTM demonstrates superior generalisation.

As for NP/PP self-corrections which are not present in Table
\ref{tab:disfluency-controlled-generalisation}: none of the systems tested were able to handle
these. Evaluation on this dataset revealed $0.0$ accuracy with all systems. We discuss these results
below.

\begin{table}[t]
    \footnotesize
    \begin{minipage}[t]{0.34\linewidth}
      \begin{tabularx}{0.95\textwidth}[t]{@{}lr}\toprule
      \textbf{Repair}&\textbf{Freq.}\\\midrule
      i i \textit{i}&139\\
      the the \textit{the}&33\\
      and and \textit{and}&31\\
      it it \textit{it}&29\\
      its its \textit{its}&26\\\hline
      it was \textit{it was}&67\\
      i dont \textit{i dont}&57\\
      i think \textit{i think}&44\\
      in the \textit{in the}&39\\
      do you \textit{do you}&23\\\hline
      a lot of \textit{a lot of}&7\\
      that was \textit{uh that was}&5\\
      it was \textit{uh it was}&5\\
      what do you \textit{what do you}&4\\
      i i dont \textit{i dont}&4\\
      \bottomrule
      \end{tabularx}
      \caption{Most common repairs in \SWDA of length 1---3}
      \label{tab:disfluency-common-repairs}
    \end{minipage}%
    \begin{minipage}[t]{0.65\linewidth}
      \begin{tabularx}{0.99\textwidth}[t]{@{}llc}\toprule
      \textbf{POS pattern}&\textbf{Examples}&\textbf{Freq., \%}\\\midrule
      DT NN DT NN&this woman this socialite&0.10\\
      &a can a garage&\\
      &the school that school&\\\hline
      JJ NN JJ NN&high school high school&0.03\\
      &good comedy good humor&\\
      &israeli situation palestinian situation&\\\hline
      DT UH DT NN&that uh that punishment&0.02\\
      &the uh the cauliflower&\\
      &that uh that adjustment&\\\hline
      DT NN UH DD NN&a friend uh a friend&0.01\\
      &a lot uh a lot&\\
      &a lot um a lot&\\\hline
      NN PRP VBP NN NN&ribbon you know hair ribbon&0.01\\
      &thing you know motion detector&\\
      \bottomrule
      \end{tabularx}
      \caption{\SWDA repairs by \POS-tag pattern}
      \label{tab:disfluency-pos-repair-patterns}
    \end{minipage}
\end{table}

\section{Conclusion}

\begin{table}[t]
  \footnotesize
  \centering
  \begin{tabularx}{0.9\textwidth}{@{}llS}\toprule
    \textbf{Keyword pattern}&\textbf{Examples}&\textbf{Freq., \%}\\\midrule
    sorry\texttt{$<$e/$>$} *&or \textit{im sorry} no&0.02\\
    &\textit{um im sorry} what&\\
    &thank you \textit{im sorry} i just got home from work&\\\hline
    sorry\texttt{$<$e/$>$} *\texttt{$<$rm-*/$>$}&and he told us theres two sixteen bit slots&0.009\\
    &\hspace{4pt}and two eight bit&\\
    &\hspace{4pt}\textit{sorry two four sixteen bit slots and two}&\\
    &\hspace{4pt}\textit{eight bit} slots available for the user&\\\hline
    i\texttt{$<$e/$>$} mean\texttt{$<$e/$>$} *&i mean&4\\
    &i mean yeah&\\
    &i mean uh&\\
    &i mean i&\\\hline
    i\texttt{$<$e/$>$} mean\texttt{$<$e/$>$} *\texttt{$<$rm-*/$>$}&i mean i i&0.5\\
    &but i mean whats whats happened here is is is&\\
    &i mean you youve&\\
    \bottomrule
  \end{tabularx}
  \caption{\SWDA repairs by interregnum}
  \label{tab:disfluency-keyword-repair-patterns}
\end{table}

We have presented a multitask \LSTM-based disfluency detection model which outperforms previous
neural network-based incremental models while being significantly simpler than them.

We have also demonstrated the generalisation potential of a disfluency detection model by
cross-dataset evaluation. As the results show, all models achieve reasonably high generalisation
level on the very local disfluency patterns such as hesitations and PP restarts.
However, the accuracy drops significantly on less restricted restarts spanning arbitrary regions of
utterances from the beginning. On the majority of the disfluency patterns, our model achieves a
superior generalisation level.

Interestingly, none of the models were able to detect NP or PP corrections such as those often
glossed in disfluency papers (e.g.\ ``A flight to Boston uh I mean to Denver''). The most likely
explanation for this could be the extreme sparsity of such disfluencies in the \SWDA dataset. 

We performed analysis of \SWDA disfluencies in order to explore this hypothesis and examined their
distribution based on length in tokens and \POS-tag sequence patterns of interest. As shown in
Tables \ref{tab:disfluency-common-repairs} and \ref{tab:disfluency-pos-repair-patterns}, the vast
majority of disfluencies found are just repetitions without speakers actually correcting themselves.
This observation is in line with prior studies, showing that the distribution of repair types varies
significantly across domains \citep{Colman.Healey11}, modalities \citep{Oviatt95}, and gender \& age
groups \citep{Bortfeld.etal01}~--- see \cite{Purver.etal18} for a nice discussion. While this is
very likely the correct explanation, we cannot rule out the possibility that such self-corrections
are inherently more difficult to process for particular models~---- that needs a separate training
dataset that holds frequency of particular repair structures constant.



Incremental phenomena of spontaneous spoken language are one aspect of natural input which dialogue
systems should process robustly. Another natural phenomenon that a dialogue system can be exposed to
is the presence of out-of-domain (\OOD) utterances at the input. With most of the neural
dialogue models' logic learned form data and not directly observable, it is often the case that the
system's behaviour on unseen input is unpredictable which significantly limits such system's
practical usability. In the next chapter, we will continue our work on dialogue systems' robustness
by addressing the problem of \OOD user's input detection and handling in a predictable way,
while keeping the overall training setup data-efficient.


\chapter{Improving Out-of-Domain Robustness of Dialogue Systems} 

\label{Chapter7} 

\lhead{Chapter 7. \emph{Out-of-Domain Robustness}} 


In this chapter, we are going to continue with the problem of the dialogue systems' robustness to
the natural and diverse user's input. In the previous chapter, we worked on improving robustness to
surface variations of the user's input represented as the disfluencies in the spoken language. Here,
we are going to cover another kind of input that can be considered anomalous, especially for the
systems trained in a data-efficient way from a few example dialogues~--- out-of-domain (\OOD) input,
i.e. user queries that the system is unable to interpret and process correctly. We explore the
problem of \OOD robustness of dialogue systems and the associated to it trade-off in accuracies on
seen and unseen data.

We present a new experimental testbed for studying the robustness of dialogue systems to \OOD input,
which is \bAbI Dialog Task 6 \citep{DBLP:conf/iclr/BordesBW17} augmented with \OOD content in a
controlled way. We then present turn dropout, a simple yet efficient technique for training
\OOD-robust dialogue systems based on automatic data augmentation and thus alleviating the problem of
the dependence on real \OOD data. With also propose a simple unified way to train the target systems
for \OOD input detection as well as for their main task at the same time.

Moreover, based on prior approaches to efficient \OOD input detection using autoencoders (see our
discussion in Section \ref{ch2:ood-robustness}), we propose a variant of Hybrid Code Network (\HCN)
dialogue management model \citep{DBLP:conf/acl/WilliamsAZ17} augmented with an autoencoder for
contextual \OOD input detection called AE-HCN(-\CNN). We experiment with this model, as well
as with a range of \HCN-family models: Variational \HCN (\VHCN), Hierarchical \HCN (\HHCN), the
original \HCN~--- adapted for training with turn dropout, and demonstrate the resulting
performance improvement as seen on \bAbI Dialog Task 6 and Google multi-domain dialogue datasets
\citep{shah2018building}.

\section{The Experimental Environment for \OOD Robustness Studies}
\label{ch7:data_augmentation}

\begin{table}[t]
    \centering
    \small
      \begin{tabular}{@{}cll@{}}\toprule
        1&\textit{sys:}&Hello, welcome to the Cambridge restaurant system.
        You can ask for restaurants by area, \\
        &&price range or food type. How may I help you?  \\
        2&\textit{usr:}&\textbf{I'd like to get away from Monday,
        August 15, 2016 to Wednesday, August 31, 2016.} \\
        &&\textbf{My budget is \$3200 and I would leave from Detroit.} \\
        3&\textit{sys:}&Sorry I didn't catch that. Could you please repeat? \\
        4&\textit{usr:}&\textbf{Will there be frost in Durham next week?}\\
        5&\textit{sys:}&Sorry I didn't catch that. Could you please repeat? \\
        6&\textit{usr:}&\textbf{\textit{So sorry man}}
        I want a moderately priced restaurant in the west part of town \\
        7&\textit{sys:}&$<$api call$>$\\
      \bottomrule
      \end{tabular}
      \caption[Augmented dialogue example]{Augmented dialogue example (\OOD content in bold,
      segment-level in italics)}
      \label{tab:augmentation_example}
\end{table}



In order to study the effect of \OOD input on end-to-end dialogue system's performance, we created
an experimental testbed based on (1) a dataset of real human-computer goal-oriented dialogues and
(2) the data augmentation technique~--- \bAbI tools (presented in Chapter \ref{Chapter3}) for mixing
real user utterances from other domains into the main dataset in a controlled way.

As our main dataset, we use \bAbI Dialog Task 6 \citep{DBLP:conf/iclr/BordesBW17} with real
human-computer conversations in the restaurant search domain initially collected for the
Dialog State Tracking Challenge 2 \citep{DBLP:conf/sigdial/HendersonTW14}. Our \OOD augmentations
are as follows:
\begin{itemize}[label=---]
  \item \textit{turn-level \OOD}: user requests from a foreign domain~--- the desired system
  behaviour for such input is the fallback action,
  \item \textit{segment-level \OOD}: interjections in the user in-domain requests~--- treated as
  valid user input and is supposed to be handled by the system in a regular way.
\end{itemize}

These two augmentation types reflect a specific dialogue pattern of interest (see Table
\ref{tab:augmentation_example}): first, the user utters a request from another domain at an
arbitrary point in the dialogue (each turn is augmented with the probability $p_{ood\_start}$),
and the system answers accordingly. This may go on for several turns in a row~--- each subsequent
turn is augmented with the probability $p_{ood\_cont}$. Eventually, the \OOD sequence ends up and the
dialogue continues as usual, with a segment-level \OOD of the user affirming their mistake. For this
study, we set $p_{ood\_start}$ to $0.2$ and $p_{ood\_cont}$ to $0.4$%
\footnote{We experimented with other values of $p_{ood\_start}$ and $p_{ood\_cont}$ but did not see
significant differences in the results. Further experiments for different domains are encouraged
using the tools provided}. 

While we introduce the \OOD augmentations in a controlled programmatic way, the actual \OOD content
is natural. The turn-level \OOD utterances are taken from conversational datasets in several foreign
domains:
\begin{itemize}[label=---]
    \item Frames dataset \citep{DBLP:conf/sigdial/AsriSSZHFMS17}~--- travel booking
    (1198 utterances),
    \item Stanford Multi-Domain (\SMD) Dialogues Dataset \citep{DBLP:conf/sigdial/EricKCM17}~---
    calendar scheduling, weather information retrieval, city navigation (3030 utterances),
    \item Dialog State Tracking Challenge 1 \citep{DBLP:conf/sigdial/WilliamsRRB13}~--- bus
    information (968 utterances).
\end{itemize}

In order to avoid incomplete/elliptical phrases, we only took the first user's utterances from the
dialogues.

For segment-level \OOD, we mined utterances with the explicit affirmation of a mistake from Twitter
and Reddit Conversations datasets (e.g. ``my mistake'', ``I'm so sorry'')~--- 701 and 500 utterances
respectively. Our datasets, as well as the tools for the OOD-augmentation of arbitrary datasets of
interest are openly available\footnote{See
\colorhref{blue}{https://github.com/ishalyminov/ood\_robust\_hcn}}.

\section{\OOD-Robust Dialogue Models}

\subsection{Hybrid Code Network Model Family}

\begin{figure}[t]
  \centering
  \includegraphics[width=0.9\linewidth]{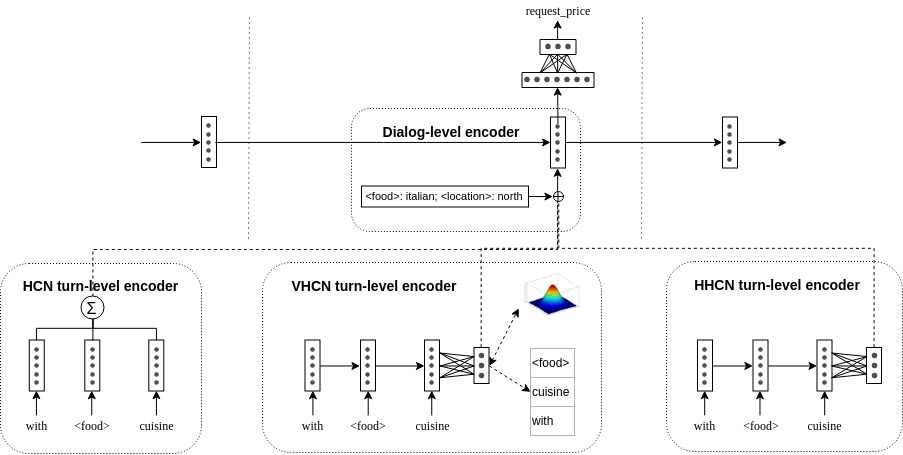}
  \caption{Hybrid Code Network model family}
  \label{fig:hcn_all}
\end{figure}

In this chapter, we experiment with the Hybrid Code Network family of models
\citep{DBLP:conf/acl/WilliamsAZ17} on \bAbI Dialog Task 6 data \citep{DBLP:conf/iclr/BordesBW17}.
\HCN is reported to be the best performing model to date for the original, \IND-only \bAbI Dialog
Task 6 data~--- thus, this is our primary experimental setup here.

As shown in Figure~\ref{fig:hcn_all}, HCN considers a dialogue as a sequence of turns. At each turn,
\HCN takes a tuple $(x_t, a_{t-1}, s_t)$ as the input to produce the next system action\footnote{A
system action can be either a text output or an api call.} $a_t$, where $x_t$ is a user utterance
consisting of $N$ tokens, i.e., $x_t = \{x_{t,1}, \ldots, x_{t,N}\}$, $a_{t-1}$ a one-hot vector
encoding the previous system action, and $s_t$ a contextual feature vector generated by
domain-specific code. The user utterance is encoded as a concatenation of a bag-of-words
representation and an average of word embeddings of the user utterance:

\begin{equation}
  u_t = \left[\text{bow}(x_t); \frac{1}{N}\sum_{i=1}^N{e(x_i)}\right]
  \label{eq:hcn_enc}
\end{equation}

where $e(\cdot)$ denotes a word embedding layer initialised with \GloVe~\citep{pennington2014glove}
or Google News-based \WordToVec \citep{DBLP:conf/nips/MikolovSCCD13}~--- frozen at the training time.
\HCN then considers the input tuple $(u_t, a_{t-1}, s_t)$ to update the dialogue state through an
\LSTM~\citep{DBLP:journals/neco/HochreiterS97}:

\begin{equation}
    h_t = \text{LSTM}\left(h_{t-1}, [u_t;a_{t-1};s_t]\right)
\end{equation}

Finally, a distribution over system actions is calculated by a dense layer with linear projection
parameters $W$ and $b$~--- the weight matrix and the bias vector, respectively~--- and with a
softmax activation:

\begin{equation}
    P(a_t) = \softmax(Wh_t + b)
\end{equation}

Thus, \HCN is a hierarchical dialogue management model with a turn-level and a dialogue-level
components (we will call them both encoders). The turn-level encoder produces a latent
representation of a single dialogue turn, and the dialogue-level encoder augments it with additional
dialogue-level features.

We will first explore a series of \HCN variations differing in their turn-level encoder (and the
corresponding optimisation objective)~--- see the illustration in Figure \ref{fig:hcn_all}.

The original \HCN as described above. Its optimisation objective for a single prediction is the
categorical cross-entropy with respect to the log-likelihood of the output dialogue action (here and
in Eq. \ref{eq:l_vhcn} we show maximisation objectives for simplicity. In the actual implementation,
they are minimised with their sign reversed):

\begin{equation}
\mathcal{L}_{\text{HCN}} = \log p\left(a_t \mid x_1, \ldots, x_t, a_{t-1}, s_t\right)
\label{eq:l_hcn}
\end{equation}
where $a_t$ is the dialogue action, $x_1, \ldots, x_t$ is the dialogue context ending
with the current user's turn, $a_{t-1}$ is the last system's action, and $s_t$ is the
domain-specific dialogue feature vector. Please note, losses of all our \HCN models are defined with
respect to the model parameters $\theta_{\text{HCN}}$, i.e. the parameters of the underlying \LSTMs,
embeddings, and projection layers, which we omit for visual simplicity.

Hierarchical \HCN (\HHCN) uses an \RNN (in our case an \LSTM cell) in order to encode each utterance:

\begin{equation}
u_t^{\text{HHCN}} = \text{LSTM}\left(e(x_t)\right)
\label{eq:hhcn}
\end{equation}
The optimisation objective is the same as of \HCN. Variants of this model were described by
\cite{DBLP:journals/corr/abs-1712-09943} and \cite{10.1007/978-3-319-95933-7_24}.
 
Variational \HCN (\VHCN) uses a Variational Autoencoder as the turn-level encoder, so that
the resulting turn encoding is \VAE's latent variable $z$ (see also our discussion of latent
variable models in Section \ref{ch2:latent_variable}):
\begin{equation}
u_t^{\text{VHCN}} = \mu\left(\text{LSTM}\left(e(x_t)\right)\right) + \sigma\left(\text{LSTM}\left(e(x_t)\right)\right) * N(0, 1)
\end{equation}
 
where $\mu$ and $\sigma$ are \MLPs for predicting $z$'s posterior distribution parameters,
and $N(0, 1)$ is a sample from its prior distribution, a standard Gaussian
\citep{DBLP:conf/conll/BowmanVVDJB16}.
 
This model differs from the previous two in the fact that it learns dialogue management and input
autoencoding jointly. In order to keep the user's input reconstruction task less complex than the
main one, we represent \VAE's reconstruction targets as bags of words (BoW) instead of sequences~---
in that, we follow \citep{DBLP:conf/acl/ZhaoZE17}. Thus, \VHCN optimisation objective is as follows:

\begin{equation}
  \mathcal{L}_{\text{VHCN}} = \mathbb{E}_{q(z)}\left[log(p(a_t \mid x_1, \ldots, x_t, a_{t-1}, s_t))\right]
  + \mathbb{E}_{q(z)}\left[p(x_t^{\text{BoW}} \mid z)\right]
  - KL\left(q(z\mid x_t) \mid\mid p(z)\right)
  \label{eq:l_vhcn}
\end{equation}
 
In the above formula, the first term is the main task's log-likelihood of the dialogue action $a_t$,
the second one is the \VAE's reconstruction term for the last user's input in the bag-of-words form
$x_t^{\text{BoW}}$, and the last turn is $KL$-divergence between the prior and posterior
distribution of the \VAE's latent variable $z$~--- following \cite{DBLP:conf/conll/BowmanVVDJB16},
we compute it in a closed form.

Another benefit of the BoW loss is, as reported in \cite{DBLP:conf/acl/ZhaoZE17}, it helps keep the
variational properties of the model (i.e. non-zero \KL-term) without the necessity of using the
\KL-term annealing trick \citep{DBLP:conf/conll/BowmanVVDJB16} which is itself challenging to control
in practice. Unlike the authors of the original BoW loss approximating the presence of each word in
the reconstructed bag with a feed-forward neural network and then summing up each word's
log-probability in the final loss, we use a simpler method and represent the BoW as a single
vocabulary-sized vector with k-hot values (1 for every word present in the bag, the rest of the
elements being 0's), and use a single sigmoid cross-entropy loss for it.

All the models above use the same dialogue-level \LSTM encoder with additional features concatenated
to the turn representations: BoW turn features, dialogue context features, and the previous system
action\footnote{Without the loss of the architecture generality, we have action mask vectors as the
additional features for the dialogue-level \LSTM \citep{DBLP:conf/acl/WilliamsAZ17}, but they do not
convey any information and are always set to 1's}.

\subsection{\AEHCN}
\label{ch7:aehcn}

\begin{figure}[t]
    \centering
    \includegraphics[width=0.9\linewidth]{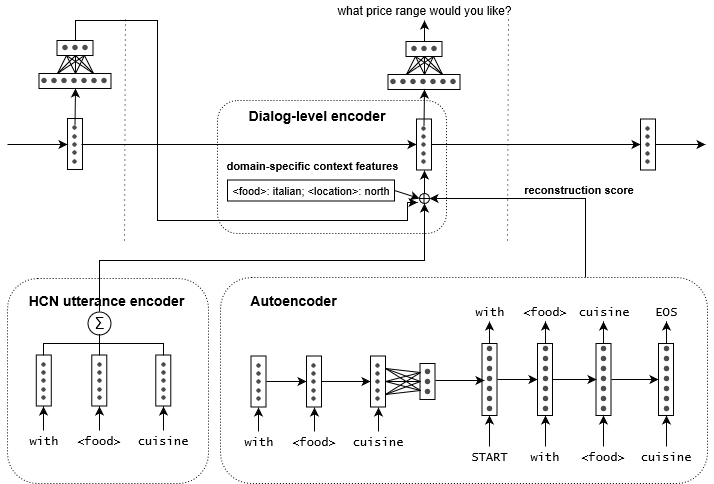}
    \caption{\AEHCN model architecture}
    \label{fig:ae-hcn}
\end{figure}

Finally, we introduce the two autoencoder-augmented architectures, \AEHCN \& \AEHCN-\CNN, where the
\HCN model is aware of the \AE's reconstruction score~--- together with the training method based on
automatic input augmentation.

\AEHCN is an \HCN whose dialogue-level encoder takes an additional input for dialogue state update~---
specifically, the autoencoder's reconstruction score $r_t$ for the user's utterance
(Figure~\ref{fig:ae-hcn}): 
\begin{equation}
    h_t = \text{LSTM}\left(h_{t-1}, [u_t;a_{t-1};s_t;r_t]\right)
\end{equation}
The autoencoder is a standard \SeqToSeq model which projects a user utterance into a latent vector
and reconstructs the user utterance. Specifically, the encoder reads $x_t$ using a \GRU
\citep{cho2014properties} to produce a 512-dimensional hidden vector $h_N^{\text{enc}}$ which in
turn gets linearly projected to a 200-dimensional latent vector $z$ with the corresponding weight
and bias parameters $W_z$ and $b_z$, respectively~--- see the formulas below:

\begin{gather}
    h_n^{\text{enc}} = \text{GRU}_{enc}\left(h_{n-1}^{enc}, e(x_n)\right), 1 < n < N\\
    z = W_{z}h_N^{enc} + b_z
\end{gather}

The output of the decoder $y$ at step $n$ is a distribution over words:
\begin{gather}
    P_{\text{dec}}(y_n) = \softmax\left(W_{\text{dec}}h_n^{\text{dec}} + b_{\text{dec}}\right)\\
    h_n^{\text{dec}} = \text{GRU}_{\text{dec}}\left(h_{n-1}^{\text{dec}}, e(y_{n-1})\right)\\
    h_0^{\text{dec}} = W_{\text{dec}}z + b_{\text{dec}}
\end{gather}
where $\text{GRU}_{\text{dec}}$ has 512 hidden units. 

The reconstruction score $r_t$ is the normalised generation probability of $x_t$ (which is both the
input and the output of the autoencoder):
\begin{equation}
    r_t = \frac{\sum_{n=0}^{N}\log P_{dec}(x_n)}{N}
\end{equation}
 
\subsection{\AEHCN-\CNN}
\AEHCN-\CNN is a variant of \AEHCN where user utterances are encoded using a \CNN layer with
max-pooling (following~\citealp{kim2014convolutional}) rather than equation~\ref{eq:hcn_enc}:

\begin{equation}
    x_t = \text{Pooling}_{max}\left(\text{CNN}(e(x_1),...,e(x_n))\right)
\end{equation}

The \CNN layer considers two kernel sizes (2 and 3) and has 100 filters for each kernel size.

\section{Training with Turn Dropout}

\begin{table}[t]
    \centering
    \begin{tabular}{@{}lccc@{}}\toprule
        \textbf{Hyperparameter}&\textbf{HCN}&\textbf{HHCN}&\textbf{VHCN}\\\midrule
        Embedding size&64&128&128\\
        Latent variable size&---&---&8\\
        Turn dropout ratio&0.4&0.6&0.3\\\midrule
        Learning rate&\multicolumn{3}{r}{0.001 (all models)}\\
        Early stopping threshold&\multicolumn{3}{r}{20 epochs (all models)}\\
        Optimiser&\multicolumn{3}{r}{Adam (all models)}\\
        Word dropout ratio&\multicolumn{3}{r}{0.2 (all models)}\\\bottomrule
    \end{tabular}
    \caption{Model hyperparameters}
    \label{tab:hcn_setup}
\end{table}

In order to train a system robust to \OOD in the absence of real \OOD examples, we employ a negative
sampling-based approach and generate them synthetically from the available \IND data with a
technique we call turn dropout. Namely, we replace random dialogue turns with the synthetic ones,
and assign them the fallback action.

\subsection{TD-\HCN}

\begin{table}[t]
    \centering
    \begin{tabular}{@{}lccccc@{}}\toprule
    &\textbf{bAbI Dialog Task 6}&\multicolumn{4}{c}{\textbf{bAbI Dialog Task 6 + OOD}}\\
    \cmidrule(r){2-2}\cmidrule(l){3-6}
    &Overall acc.&\multicolumn{1}{c}{Overall acc.}&\multicolumn{1}{c}{Seg. OOD acc.}&
    \multicolumn{1}{c}{OOD acc.}&OOD F1\\\midrule
    \HCN&0.557&0.438&\textbf{0.455}&0.0&0.0\\
    \HHCN&0.531&0.418&0.424&0.0&0.0\\
    \VHCN&0.533&0.413&0.413&0.0&0.0\\\cmidrule(r){1-1}
    TD-\HCN&0.563&\textbf{0.575}&0.257&\textbf{0.754}&\textbf{0.743}\\
    TD-\HHCN&0.505&0.455&0.435&0.274&0.418\\
    TD-\VHCN&\textbf{0.565}&0.545&0.407&0.530&0.667\\
    \bottomrule
    \end{tabular}
    \caption{Evaluation results}
    \label{tab:td_hcn_evaluation}
\end{table}

More formally, our dialogue features are as follows: {\tt $\left<f\_turn, f\_ctx, f\_mask, a\right>$}, i.e.
turn features (token sequences), dialogue context features, action masks, and target actions
respectively.

Under turn dropout, for a randomly selected dialogue $i$ and its turn $j$, we replace
{\tt $f\_turn_{ij}$} with a sequence of random vocabulary words (drawn from a uniform distribution
over the vocabulary) and UNK tokens, and corresponding {\tt $a_{ij}$} with the fallback action, and
leave all other features intact. In this way, we're simulating anomalous turns for the system given
usual contexts (as stored in the dialogue \RNN's state), and we put minimum assumptions on the
synthesised turns' structure (we only limit their lengths to be within the bounds of real
utterances).

\subsection{Training the AE-HCN(-\CNN)}

\begin{table}[t]
    \centering
    \begin{tabular}{@{}lccc@{}}\toprule
    \bf{Dataset} & \bf{\# Dialogues} & \bf{Avg. \#turns per dialogue} & \bf{\# actions}\\\midrule
    bAbI6 train	& 1618 & 20.08 & 58 \\
    \textcolor{white}{bAbI6} dev & 500 & 19.30 & 58 \\
    \textcolor{white}{bAbI6} test & 1117 & 22.07 & 58 \\
    \textcolor{white}{bAbI6} test-OOD & 1117 & 27.27 & 59 \\ \midrule

    GR train & 1116 & 9.07 & 247 \\
    \textcolor{white}{GR} dev & 349 & 6.53 & 247 \\
    \textcolor{white}{GR} test & 775 & 6.87 & 247 \\
    \textcolor{white}{GR} test-OOD & 775 & 9.01 & 248 \\ \midrule

    GM train & 362 & 8.78 & 194 \\
    \textcolor{white}{GM} dev & 111 & 9.14 & 194 \\
    \textcolor{white}{GM} test & 252 & 8.73 & 194 \\
    \textcolor{white}{GM} test-OOD & 252 & 11.25 & 195 \\ \bottomrule
    \end{tabular}
    \caption{bAbI6, GR, and GM dataset statistics}
    \label{tab:data_stats}
\end{table}

While essentially similar to the TD-\HCN algorithm described above, training these two architectures
with data augmentation involves providing reconstruction scores for the `dropped out' turns. We
describe the full procedure below.

To endow an AE-HCN(-\CNN) model with a capability of detecting \OOD utterances and producing fallback
actions without requiring real \OOD data, we augment training data with counterfeit turns.
We first select arbitrary turns in a dialogue at random according to a counterfeit \OOD probability
$\rho$, and insert counterfeit turns before the selected turns.
A counterfeit turn consists of a tuple $(x_t, a_{t-1}, s_t, r_t)$ as input and a fallback action
$a_t$ as output. We copy $a_{t-1}$ and $s_t$ of each selected turn to the corresponding counterfeit
turns since \OOD utterances do not affect previous system action and feature vectors generated by
domain-specific code.
Now we generate a counterfeit $x_t$ and $r_t$. Since we do not know \OOD utterances a priori, we
randomly choose one of the user utterances of the same dialogue to be $x_t$. This helps the model
learn to detect \OOD utterances because a random user utterance is contextually inappropriate just
like \OOD utterances are. We generate $r_t$ by drawing a sample from a uniform distribution,
$U[\alpha, \beta]$, where $\alpha$ is the maximum reconstruction score of training data and $\beta$
is an arbitrary large number. The rationale is that the reconstruction scores of \OOD utterances are
likely to be larger than $\alpha$ but we do not know what distribution the reconstruction scores of
\OOD turns would follow. Thus we choose the most uninformed distribution, i.e. a uniform distribution
so that the model may be encouraged to consider not only reconstruction score but also other
contextual features such as the appropriateness of the user utterance given the context, changes in
the domain-specific feature vector, and what action the system previously took.

\section{Experiment 1: \HHCN \& \VHCN}
\label{ch7:exp1}

We train our models only using the original \bAbI Dialog Task 6 dataset, and evaluate them on our
\OOD-augmented versions of it: we use the per-utterance accuracy as our main evaluation metric;
the models are trained with the same hyperparameters (where applicable) listed in Table
\ref{tab:hcn_setup}. The models use the common unified vocabulary including all words from our
datasets (including \OOD content). The intuition behind this is as follows: production dialogue
models often use word embedding matrices with vocabularies significantly exceeding that of the
training data in order to take advantage of additional generalisation power via relations like
synonymy, hyponymy, or hypernymy which are efficiently modelled by distributed word representations.
Therefore, simply mapping every unseen word to an `UNK' does not quite reflect that setting.

We tuned our models' hyperparameters using a 2-stage grid search, tracking the development set
accuracy. At the first stage, we adjusted the embedding dimensionality of our models (and the latent
variable size in case of \VHCN). Then, given the values found, at the second stage we adjusted turn
dropout ratio at the interval $[0.05 - 0.7]$. Exact hyperparameter values are detailed in
Table \ref{tab:hcn_setup}.  

The results are shown in Table \ref{tab:td_hcn_evaluation}~--- please note, apart from the
accuracies we report \OOD F1-measure, a metric showing the model's performance as a conventional
\OOD detector, with positive class being the fallback action, and the negative being all the \IND
classes actions.

Finally, given the stochastic nature of \VHCN, we reported its mean accuracy scores over 3 runs
(we used the same criterion for selecting the best model during the training procedure).

\subsection{Results and Discussion}

As our experiment showed, while learning to handle both \IND and \OOD input with access to \IND-only
data at the training time, there appears the following trade-off: a model performing better on the
`clean' test turns is prone to lower accuracy on \OOD~--- it can be said that it slightly overfits to
its devset. On the other hand, a model regularised with turn dropout during training naturally
performs better on unseen \OOD turns, but with not as high accuracy on its `clean', \IND test data.
Another side of the trade-off is the accuracy of \OOD detection vs robust handling of \IND input with
segment-level noise. As our results showed, the models specifically trained for \OOD detection all
demonstrate lower accuracy on the noisy \IND.

Among the models we evaluated, it is worth noting that the original \HCN demonstrated the best
performance as an \OOD detector (more than \textbf{74\%} F1-score) and thus overall \IND + \OOD
accuracy on the augmented dataset~--- more than \textbf{57\%}. While some parts of its architecture
(e.g. mean vector-based turn encoding or bag-of-words feature vector at the utterance level) may not
seem to be the most robust solution, the model demonstrate superior overall performance.
Averaging at the turn level instead of recurrent encoding (the case of \HHCN and \VHCN) makes the
model less dependent on actual word sequences seen during training but on the keywords themselves.

In turn, \VHCN demonstrated superior performance on \IND data when trained with turn dropout, more
than \textbf{56\%}~--- it benefited in terms of both overall accuracy and the absence of
false-positive {\OOD}s thus improving upon the original \HCN's results
\cite{DBLP:conf/acl/WilliamsAZ17}. An additional challenge was to train it while keeping its
variational properties (i.e. reasonably high \KL term)~--- the BoW reconstruction loss which we used
in order to simplify the secondary task, helped with this as well \citep{DBLP:conf/acl/ZhaoZE17}.
On the other hand, while achieving superior performance on clean data, \VHCN's properties did not
result in \OOD handling improvements.

\section{Experiment 2: AE-HCN(-\CNN)}
\label{ch7:exp2}

\begin{table}[t]
    \center
    \footnotesize
    \begin{tabular}{@{}lccccccccc@{}}\toprule
    & \multicolumn{3}{c}{\bf{bAbI6}} & \multicolumn{3}{c}{\bf{GR}} & \multicolumn{3}{c}{\bf{GM}} \\
    \cmidrule(r){2-4}\cmidrule(r){5-7}\cmidrule(r){8-10}
    & \bf{Test} & \multicolumn{2}{c}{\bf{Test-OOD}} & \bf{Test} & \multicolumn{2}{c}{\bf{Test-OOD}} & \bf{Test} & \multicolumn{2}{c}{\bf{Test-OOD}} \\
    \cmidrule(r){2-2}\cmidrule(r){3-4}\cmidrule(r){5-5}\cmidrule(r){6-7}\cmidrule(r){8-8}\cmidrule(r){9-10}
    & \bf{P@1} & \bf{P@1} & \bf{OOD F1} & \bf{P@3} & \bf{P@3} & \bf{OOD F1} & \bf{P@3} & \bf{P@3} & \bf{OOD F1}  \\
    \midrule
    \HCN & 53.41 & 41.95 & 0 & \bf{58.89} & 41.65 & 0 & 41.18 & 27.08 & 0 \\
    \AEHCN-Indep & 31.29 & 41.06 & 48.68 & 51.90 & 55.42 & 71.52 & 31.12 & 42.78 & 64.35 \\
    \AEHCN  & 53.58 & 55.04 & \bf{73.41} & 56.97 & 58.90 & 74.67 & 40.61 & 48.59 & \bf{69.31} \\
    \AEHCN-\CNN & \bf{55.04} & \bf{55.35} & 70.38 & 58.32 & \bf{64.51} & \bf{81.33} & \bf{45.12} & \bf{52.79} & 68.59 \\
    \bottomrule
    \end{tabular}
    \caption{AE-HCN(-\CNN) Evaluation results}
    \label{tab:res}
\end{table}

To study the effect of \OOD input on these two dialogue systems' performance, we use three
task-oriented dialogue datasets: bAbI6 as in the previous experiment, as well as GR and GM taken from
Google multi-domain dialogue datasets ~\citep{shah2018building}. Basic statistics of the datasets
are shown in Table~\ref{tab:data_stats}. bAbI6 deals with restaurant finding tasks, GM buying a
movie ticket, and GR reserving a restaurant table, respectively. We generated distinct action
templates by replacing entities with slot types and consolidating based on dialogue act annotations.
We augment test datasets (denoted as Test-\OOD in Table~\ref{tab:data_stats}) with the procedure
described in Section \ref{ch7:data_augmentation}.

\subsection{Experimental Setup and Evaluation}
\label{ch7:ae_hcn_evaluation}

We comparatively evaluate four different models:
\begin{enumerate}
    \item an \HCN model trained on in-domain training data;
    \item an \AEHCN-Indep model which is the same as the \HCN model except that it deals with \OOD
    utterances using an independent autoencoder-based rule to mimic~\cite{ryu2017neural}~---
    when the reconstruction score is greater than a threshold, the fallback action is chosen (we set
    the threshold to the maximum reconstruction score of training data);
    \item an AE-HCN(-\CNN) model trained on training data augmented with counterfeit \OOD turns~---
    the counterfeit \OOD probability $\rho$ is set to 15\% and $\beta$ to 30.
\end{enumerate}


\begin{table}[t]
    \center
    \begin{tabular}{@{}ccc@{}}\toprule
        \bf{Threshold} & \bf{Precision@1} & \bf{\OOD F1} \\
        \midrule
        6 & 40.39 & 48.38 \\
        7 & 42.56 & 50.46 \\
        8 & 43.69 & 51.08 \\
        9 & 52.21 & 63.86 \\
        10 & 47.27 & 44.44 \\
        \bottomrule
        \end{tabular}
    \caption{Performances of \AEHCN-Indep on bAbI6 Test-\OOD with different thresholds}
    \label{tab:thresh}
\end{table}

\begin{table}[t]
    \centering
    \begin{tabular}{@{}cccc@{}}\toprule
    & \bf{bAbI6 Test} & \multicolumn{2}{c}{\bf{bAbI6 Test-\OOD}} \\
    \cmidrule(r){2-2}\cmidrule(r){3-4}
    \bf{OOD Rate}& \bf{Precision@1} & \bf{Precision@1} & \bf{OOD F1} \\
    \midrule
    5\% & 55.25 & 55.48 & 69.72 \\
    10\% & 55.08 & 57.29 & 74.73 \\
    15\% & 55.04 & 55.35 & 70.38 \\
    20\% & 53.48 & 56.53 & 75.55 \\
    25\% & 53.72 & 56.66 & 73.13 \\
    30\% & 54.87 & 56.02 & 71.44 \\
    \bottomrule
    \end{tabular}
    \caption{\AEHCN-\CNN performance on bAbI6 with varying counterfeit \OOD rates}
    \label{tab:rate}
\end{table}

Our training setup overall follows that in Experiment 1 (see Table \ref{tab:hcn_setup}). Although
here, we pretrain the autoencoder on in-domain training data and keep it fixed while training the
main models.

The result is shown in Table~\ref{tab:res}. Since there are multiple actions that are appropriate
for a given dialogue context, we use per-utterance {\em Precision@K} as performance metric.
We also report F1-score for the \OOD detection to measure the balance between precision and recall.
The performances of \HCN on Test-\OOD are about 15 points down on average from those on Test, showing
the detrimental impact of \OOD utterances to such models only trained on in-domain training data. 
AE-HCN(-\CNN) outperforms \HCN on Test-\OOD by a large margin~--- about 17(20) points on average~---
while keeping the minimum performance trade-off compared to Test. Interestingly, \AEHCN-\CNN has
even better performance than \HCN on Test, indicating that, with the \CNN encoder, counterfeit \OOD
augmentation acts as an effective regularisation. In contrast, \AEHCN-Indep failed to robustly
detect \OOD utterances, resulting in much lower numbers for both metrics on Test-\OOD as well as
hurting the performance on Test. This result indicates two crucial points:

\begin{enumerate}
    \item the inherent difficulty of finding an appropriate threshold value without actually seeing
    \OOD data;
    \item the limitation of the models which do not consider context.
\end{enumerate}

\begin{figure}[t]
    \center
    \includegraphics[width=0.8\linewidth]{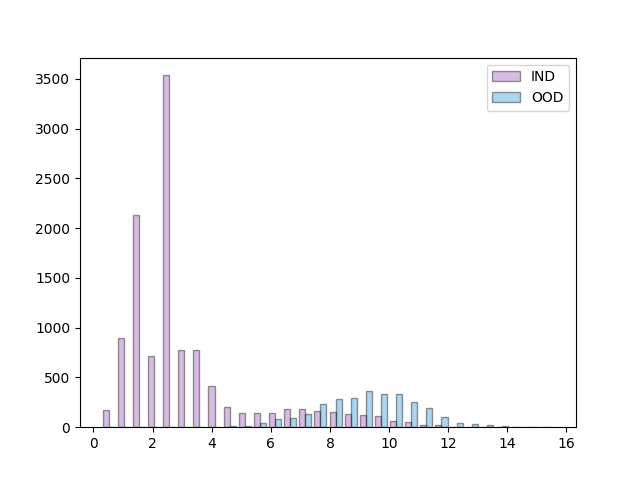}
    \caption{Histograms of \AE reconstruction scores for the bAbI6 test data}
    \label{fig:vis}
\end{figure}

For the first point, Figure~\ref{fig:vis} plots histograms of reconstruction scores for \IND and
\OOD utterances of bAbI6 Test-\OOD (the histograms for other datasets follow similar trends). If
\OOD utterances had been known a priori, the threshold should have been set to a much higher value
than the maximum reconstruction score of \IND training data ($6.16$ in this case).

For the second point, Table~\ref{tab:thresh} shows the search for the best threshold value for
\AEHCN-Indep on the bAbI6 task when given actual \OOD utterances (which is highly unrealistic for
the real-world scenario). Note that the best performance achieved at 9 is still not as good as that
of AE-HCN(-\CNN). This implies that we can perform better OOD detection by jointly considering other
context features.

Furthermore, we conduct an experiment of \AEHCN's sensitivity to the $\beta$ hyperparameter which is
the upper bound of the counterfeit reconstruction scores for turn dropout training. We vary $\beta$
from $30$ which is quite close to $\alpha$ (the corresponding lower reconstruction score bound
determined by the statistics of the trainset), up to $240$ which is more that  $10 \alpha$.
The results are shown in Table \ref{tab:alpha_beta}. As we observe in the table, there is no
dependence between the overall accuracy of the model and $\beta$ parameter, however the \OOD
detection F1-score kept increasing as the range of reconstruction scores fed into the model was
getting broader, and nearly stopped as $\beta$ neared $10 \alpha$. Therefore, it makes sense
to perform a grid search over $\beta$ while training the model.

Finally, we conduct a sensitivity analysis by varying counterfeit \OOD probabilities.
Table~\ref{tab:rate} shows performances of \AEHCN-\CNN on bAbI6 Test-\OOD with different $\rho$
values, ranging from 5\% to 30\%. The result indicates that our method manages to produce good
performance regardless of the $\rho$ value.
This superior stability nicely contrasts with the high sensitivity of \AEHCN-Indep with regard to
threshold values as shown in Table~\ref{tab:thresh}.

\begin{table}[t]
    \centering
    \begin{tabular}{@{}cccc@{}}\toprule
    $\bm{\alpha}$ & $\bm{\beta}$ & \bf{Precision@1} & \bf{OOD F1} \\
    \midrule
    23 & 30 & 59.3 & 75.3 \\
    23 & 60 & 57.8 & 75.5 \\
    23 & 90 & 56.9 & 75.6 \\
    23 & 120 & 59.3 & 75.9 \\
    23 & 150 & 58.3 & 76.1 \\
    23 & 180 & 58.0 & 76.2 \\
    23 & 210 & 57.0 & 75.7\\
    23 & 240 & 59.1 & 76.0\\
    \bottomrule
    \end{tabular}
    \caption{\AEHCN sensitivity to the $\beta$ hyperparameter on bAbI6 \OOD-augmented testset}
    \label{tab:alpha_beta}
\end{table}

\section{Conclusion}
\label{ch7:conclusion}

In this chapter, we explored the problem of robustness of neural dialogue systems to \OOD input.
Specifically, we presented a dataset for studying this problem along with a general procedure for
augmenting arbitrary datasets of interest for such purpose. Secondly, we introduced turn dropout,
a simple yet efficient technique for improving \OOD robustness of dialogue control models and
evaluated its effect on several Hybrid Code Network-family models.

We proposed a novel \OOD detection method that does not require OOD data without any restrictions by
utilising counterfeit \OOD turns in the context of a dialogue. In the presence of \OOD utterances,
our method outperforms the best performing dialogue models to date equipped with an \OOD detection
mechanism by a large margin~--- more than 17 points in Precision@K on average~--- while minimising
performance trade-off on in-domain test data. The detailed analysis sheds light on the difficulty of
optimising context-independent \OOD detection and justifies the necessity of context-aware \OOD
handling models.

This chapter concludes our contributions to data-efficiency of goal-oriented systems. In the next
chapter, we are going to explore the area of social, chat-oriented dialogue as well as the data
efficiency issues arising there. As such, in Chapters \ref{Chapter4} and \ref{Chapter5}, while
working on bootstrapping dialogue systems with minimum training data, we also cared about being able
to train our systems without the need for any annotations. In the next chapter, we are going to
pursue a similar goal in the chat-oriented dialogue area where the conversation models themselves
or their key components are dependent on the annotations in the form of the user feedback scores.


\chapter{Data-Efficiency in Social Dialogue} 

\label{Chapter8} 

\lhead{Chapter 8. \emph{Data-Efficiency in Social Dialogue}} 


In this final chapter, we are going to go beyond slot-filling goal-oriented conversations and look
at open-domain social dialogue. Chat-oriented dialogue differs from goal-oriented in the sheer fact
that there is no `goal' to pursue. That makes the objectives of such interaction harder to define,
and the conventional slot-value annotations widely used in goal-oriented dialogue become of little
relevance in the chat-oriented setup. That is why large-scale conversational models
\citep[e.g.][]{DBLP:journals/corr/VinyalsL15} were initially trained end-to-end from raw dialogue
transcripts essentially to mimic the responses seen in the contexts they occur.
However, in order to improve those models' base performance, especially in cases of user-faced
products, annotations were brought back \citep[e.g.][]{yu_strategy_2016}~--- their simplest form
being dialogue-level user ratings of the conversation, e.g. binary `good/bad' or Likert-scale 1---5
scores.

As was discussed in Chapter \ref{Chapter2}, both generation and retrieval-based models were widely
used for end-to-end chat-oriented dialogue. Under the prominent approach to such systems used in
practical applications (such as Amazon Alexa Prize)~--- the {\it bot ensemble}
(\citealp{mila}; \citealp{yu_strategy_2016}; \citealp{gen_ir_ensemble})~--- a collection, or
ensemble, of different bots is used, each of which proposes a candidate response to the user's
input, and a \emph{response ranker} selects the best response for the final system output to be
uttered to the user. In this chapter, we focus on the task of finding the best supervision signal
for training a response ranker for ensemble systems. Our contribution is twofold: first, we present
a neural ranker\footnote{Code and trained models available at
\colorhref{blue}{http://tiny.cc/alana\_ranker}} for ensemble-based dialogue systems and evaluate
its level of performance using an annotation type which was provided to the Alexa Prize 2017
participants by Amazon \citep{ram_conversational_2017}, namely per-dialogue user ratings.
Secondly and most importantly, we explore an alternative way of assessing social conversations
simply via their {\it length}, thus removing the need for any user-provided ratings.



\section{Data Efficiency in Open-Domain Dialogue}

Chatbots, or \textit{socialbots}, are dialogue systems aimed at maintaining an open-domain
conversation with the user spanning a wide range of topics, with the main objective  of being
engaging, entertaining, and natural.
Currently, social chat systems are in great demand in industry both as a means to increase user
retention with existing goal-oriented dialogue systems (e.g.\ Apple Siri, Google Assistant, or
Amazon Alexa) and as standalone conversational systems  for use e.g. in entertainment, robotics,
and healthcare.

With the objective described above, it is problematic to build an open-domain, social dialogue
system in a traditional, rule-based way \citep{Weizenbaum66} because hand-crafting such systems is
not practical~--- they are difficult to create and maintain, cannot be trained on new data for a new
setting, and cannot be automatically optimised. This is why many such systems are based on learning
dialogue behaviour from massive amounts of data such as Reddit conversations (%
\citealp{al-rfou_conversational_2016}; \citealp{liu_rubystar:_2017}), OpenSubtitles
\citep{DBLP:conf/lrec/LisonT16} or the Ubuntu Dialog Corpus \citep{ubuntu_corpus}.

There are two fundamental approaches to such systems. Generation-based models are predominantly
based on \SeqToSeq architecture (see Section \ref{ch2:generation}) and produce responses
word-by-word in a language model style given an encoded dialogue context representation.
Ranking (or retrieval) based models work similarly to an Information Retrieval engine (see Section
\ref{ch2:retrieval}): given a (preprocessed) user utterance as input, they first collect a pool of
candidate responses, and then use their own ranking function in order to select the best one.
Common sources of candidate responses are normally search engines over conversational corpora (e.g.\
the above-mentioned Reddit conversations and OpenSubtitles), rule-based systems, question answering
systems, or other response generation models. Ranking-based dialogue models are especially suitable
for the ensembles of bots (\citealp{mila}; \citealp{yu_strategy_2016}; \citealp{gen_ir_ensemble})
which we are focusing on here.

\subsection{The Need for Data Efficiency}

It is well known that deep learning models are highly data-dependent, but there are currently no
openly available data sources which can provide enough high-quality open-domain social dialogues
for building a production-level socialbot. Therefore, a common way to get the necessary data is to
collect it on a crowdsourcing platform \citep{edina}. Based on the model type and the development
stage, it may be necessary to collect either whole dialogues, or some form of human feedback on how
good a particular dialogue or turn is. However, both kinds of data are time-consuming and expensive
to collect.

The data efficiency of a dialogue model can be split into two parts accordingly:

\begin{itemize}[label=---]
\item \textit{sample efficiency}~--- the number of data points needed for the model to train.
As such, it is useful to specify an order of magnitude of the training set size for different types
of machine learning models;
\item \textit{annotation efficiency}~--- the amount of annotation effort needed. For instance,
traditional goal-oriented dialogue system architectures normally require intent,
slot-value, and dialogue state annotation \citep[e.g.][]{young_hidden_2010}, whereas end-to-end
conversational models work simply with raw text transcriptions
\citep[e.g.][]{DBLP:journals/corr/VinyalsL15}.
\end{itemize}

\subsection{Users' Ratings and Explicit Feedback}
\label{ch8:user-rating}

\begin{table}[t]
\begin{center}
\begin{tabular}{lc}\toprule
\bf Variables& \bf Pearson corr. coefficient\\\midrule
rating/length&$0.11$\\
rating/positive feedback&$0.11$\\
rating/negative feedback&$0.04$\\
length/positive feedback&$0.67$\\
length/negative feedback&$0.49$\\\bottomrule
\end{tabular}
\end{center}
\caption{Correlation study of key dialogue aspects}
\label{tab:correlation}
\end{table}

\begin{figure}[t]
    \begin{center}
    \includegraphics[width=\textwidth]{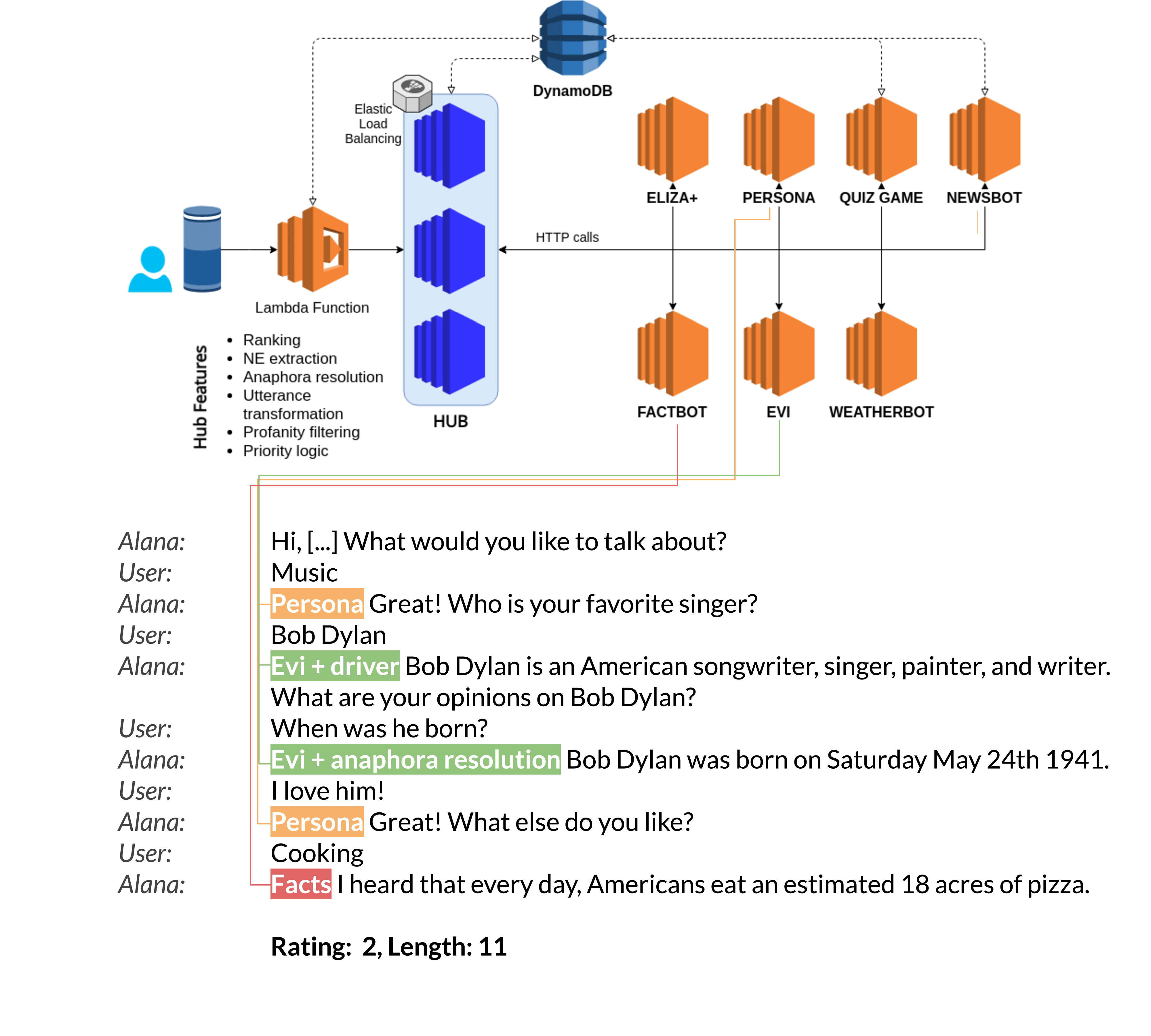}
    \end{center}
    \caption{Alana architecture, with an example chat}
    \label{fig:alana}
\end{figure}

The 2017 Alexa Prize challenge made it possible to collect large numbers of dialogues between real
users of Amazon Echo devices and various chatbots. The only annotation collected was per-dialogue
ratings elicited at the end of the conversations by asking the user {\it ``On a scale of 1 to 5, how
much would you like to speak with this bot again''}  \citep{venkatesh_evaluating_2017}.
Less than 50\% of conversations were actually rated; the rest were quit without the user giving a
score. In addition, note that a single rating is applied to an entire conversation (rather than
individual turns), which may consist of very many utterances. The conversations in the challenge
were about 2.5 minutes long on average, and about 10\% of conversations  were over 10 minutes long
\citep{ram_conversational_2017}~--- this makes the ratings very sparse. Finally, the ratings are
noisy~--- some dialogues which are clearly bad can get good ratings from some users, and vice-versa
(see a motivating example in Figure \ref{fig:alana}\footnote{Due to the restrictions on publicly
presenting real conversational data from the challenge, we provide a sketch of a dialogue which we
had with the system ourselves}).

Apart from users' ratings, we employed an additional metric of dialogue quality for our study~---
explicit user feedback. That is, we searched for dialogue turns containing positive or negative
user's sentiment. Additionally, we used a whitelist and a blacklist of hand-picked phrases to filter
out sentiments not addressed at the system directly. In total, we collected 605 unique utterances,
e.g. \textit{``that's pretty cool'', ``you're funny'', ``gee thanks'', ``interesting fact'',
``funny alexa you're funny''} and \textit{``weird'', ``sounds dumb'', ``strange''}. 

Given the main objective of social dialogue stated in the Alexa Prize rules as `long and engaging'
conversation, we tried to verify an assumption that user ratings reflect these properties of the
dialogue. Apart from our observations above, we performed a correlation analysis of user ratings
and aspects of dialogue directly reflecting the objective: dialogue length and explicit user
feedback (see Table \ref{tab:correlation}).
Although we have a significant number of dialogues which are both long and highly rated, the
correlation analysis was not able to show any relationship between dialogue length and rating.
Ratings aren't correlated with user feedback either (see Section~\ref{ch8:eval} for the details of
user feedback collection). On the other hand, we found a promising moderate correlation between the
conversation length and explicit positive feedback from the users (we counted the number of such
turns in a dialogue). The respective length/negative feedback relationship is slightly weaker.

Therefore, we experiment with conversation length for approximating user satisfaction and engagement
and use it as an alternative measure of dialogue quality. This allows us to take advantage of all
conversations, not just those rated by users, for training a ranker. While some conversations might
be long but not engaging (e.g.\ if there are a lot of misunderstandings, corrections, and speech
recognition errors), training a ranker only using length makes it extremely
{\it annotation-efficient}.

\section{A Neural Ranker for Open-Domain Conversation}
\label{ch8:neural}

\begin{figure}
\begin{center}
\includegraphics[width=0.7\columnwidth]{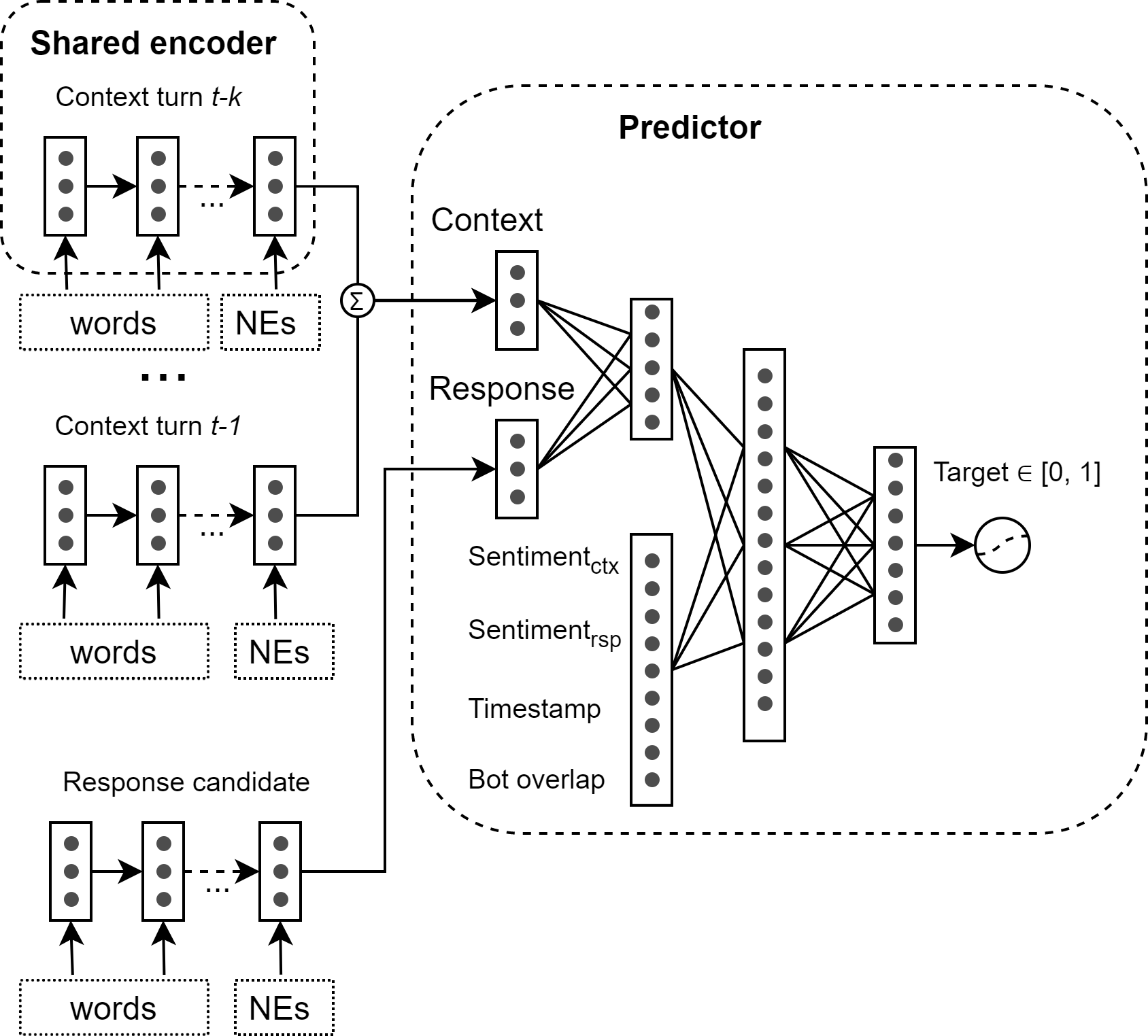}
\end{center}
\caption{Neural ranker architecture}
\label{fig:ranker_architecture}
\end{figure}


The ranker described here is part of Alana, Heriot-Watt University's Alexa Prize 2017 finalist
socialbot \citep{AlanaNIPS}~--- see the visualisation in Figure \ref{fig:alana}. Alana is an
ensemble-based model incorporating information-retrieval-based bots with news content and
information on a wide range of topics from Wikipedia, a question answering system, and rule-based
bots for various purposes, from amusing users with fun facts to providing a consistent persona.
The rule-based bots are also required to handle sensitive issues which can be raised by real users,
such as medical, financial, and legal advice, as well as profanities. 

\subsection{Ranker Architecture}
\label{ch8:architecture}

The architecture of our ranker\footnote{We would like to point out that technically, our model
produces a relevance \textit{score} for a response candidate given the dialogue context, trained
with a regression objective and only using training data with a partial order relation imposed (i.e.
$1.0$ for relevant, $0.0$ for non-relevant context-response pairs). Therefore, the model can as well
be considered a conversational response re-scorer. However, here we conform to the notation of
\cite{liu2011learning} under which our approach is categorised as point-wise ranking model with the
judgements in the form of relevance degrees (0/1).} is shown in Figure
\ref{fig:ranker_architecture}. The inputs to the model are 1-hot word-by-word vectors of a candidate
response and the current dialogue context (we use the 3 most recent system and user turns). They are
encoded into a latent representation using a single shared \RNN encoder based on \GRU cells
\citep{DBLP:conf/emnlp/ChoMGBBSB14}. The context embedding vectors are then summed up and
concatenated with the response embedding (Eq.~\ref{eq:encoder}):

\begin{equation}
\label{eq:encoder}
\text{Enc}(C, r) = \sum_{i}{\text{RNN}(C_i)} \oplus \text{RNN}(r)
\end{equation}

where $C$ is the dialogue context and $r$ is a response candidate.

The context and the response are represented using combined word-agent tokens (where agent is either
a specific bot from the ensemble or the user) and  are concatenated with the lists of named entities
extracted using Stanford \NER \citep{DBLP:conf/acl/FinkelGM05}. All the word-agent tokens and named
entities share the same unified vocabulary.

Encoder outputs, along with additional dialogue features such as context and response sentiment,
timestamp, and bot names in the context and the response, go into the \textit{Predictor},
a feed-forward neural network (\MLP) whose output is the resulting rating (Eq.~\ref{eq:predictor}):

\begin{equation}
\label{eq:predictor}
\text{Pred}(C, r) = \sigma\left(L\left(\text{Sem}(C, r) \oplus f(C, r)\right)\right)
\end{equation}
\begin{itemize}[label=---]
  \item[where:] $L(x) = \text{ReLU}(M x + b)$ is the layer used in the Predictor (the number of
  such layers is a model parameter),
  \item[] $\text{Sem} = L\left(\text{Enc}(C, r)\right)$ is the vector of semantic context-response
  features, and
  \item[] $f(C, r)$ is a vector of the additional dialogue features listed above.
\end{itemize}

We use \ReLU activation for the hidden layers because it is  known to be highly efficient with deep
architectures \citep{DBLP:conf/icml/NairH10}. Finally, we use sigmoid activation $\sigma$ for
generating the final prediction in the range $[0, 1]$.

\subsection{Training Method}
\label{ch8:train-method}

We use either dialogue rating or length as the prediction target scaled at $[0, 1]$ (as discussed in
Sections~\ref{ch8:dataset} and~\ref{ch8:eval}). The model is trained to minimise the Mean Squared
Error (\MSE) loss against the target using the Adagrad optimiser \citep{adagrad}.
In our training setup, the model learns to predict per-turn target values. However, since only
per-dialogue ones are available in the data, we use the following approximation: the target value of
a context-response pair is the target value of the dialogue containing it.
The intuition behind this is an assumption that the majority of turns in ``good" dialogues (either
length- or rating-wise) are ``good" in their local contexts as well~--- so that given a large number
of dialogues, the most successful and unsuccessful turns will emerge from the corresponding
dialogues.

\section{Baselines}
\label{ch8:baselines}

We compare our neural ranker to two other models also developed during the competition:
handcrafted and linear rankers~--- all three were deployed live in the Alana Alexa Prize 2017
finalist system \citep{AlanaNIPS}, and were therefore of sufficient quality for a production system
receiving thousands of calls per day. We also compare our model to a recently published
dual-encoder response selection model by \cite{lu_practical_2017} based on an approach conceptually
close to ours.

\subsection{Handcrafted Ranker}
In the handcrafted approach, several turn-level and dialogue-level features are calculated,
and a linear combination of those feature values with manually adjusted coefficients is used to
predict the final ranking. The list of features includes:

\begin{itemize}[label=---]
\item coherence, information flow, and dullness as defined by \cite{Li16_2};
\item overlap between the context and the response with regards to named entities and noun phrases;
\item topic divergence between the context turns and the response~--- topics are represented using
the Latent Dirichlet Allocation \citep[\LDA,][]{lda};
\item sentiment polarity, as computed by the NLTK Vader sentiment analyser
\citep{gilbert_vader:_2014}.\footnote{
\colorhref{blue}{http://www.nltk.org/howto/sentiment.html}}
\end{itemize}

\subsection{Linear Ranker}
\label{ch8:linear-ranker}

The linear ranker is based on the VowpalWabbit (\VW) linear model \citep{vowpalwabbit}.
\VW has a highly efficient implementation of stochastic gradient descent over various loss functions,
capable of handling a very large (and sparse) feature space and a high number of training examples.
We use the MSE loss function and the following features in our \VW ranker model:

\begin{itemize}[label=---]
\item bag-of-n-grams from the dialogue context (preceding 3 utterances) and the response,
\item position-specific n-grams at the beginning of the context and the response (first 5 positions),
\item dialogue flow features \citep{Li16_2}, the same as for the handcrafted ranker,
\item bot name, from the set of bots in the ensemble.
\end{itemize}

VW implements feature combinations (Cartesian product) out-of-the-box, which allows it to naturally
include a combination of n-gram features from the context and the response, as well as others~---
the detailed VW ranker configuration is shown in Table \ref{tab:vw_config} of Appendix
\ref{AppendixD}.

\subsection{Dual-Encoder Ranker}
The closest architecture to our neural ranker is that of \cite{lu_practical_2017}, who use a
dual-encoder \LSTM with a predictor \MLP for task-oriented dialogue in closed domains.
Unlike this work, they do not use named entities, sentiment, or other input features than basic word
embeddings. Dialogue context is not modelled explicitly either, and is limited to a single user
turn. We reproduced their architecture and set its parameters to the best ones reported in the
original paper.

\section{Training Data}
\label{ch8:dataset}


Our data is transcripts of conversations between our socialbot and real users of the Amazon Echo
collected over the challenge period, February -- December 2017.
The dataset consists of over 200,000 dialogues (5,000,000+ turns) from which over 100,000 dialogues
(totalling nearly 3,000,000 turns) are annotated with ratings. From this data, we sampled two
datasets of matching size for training our rankers, using the per-turn target value approximation
described in Section~\ref{ch8:train-method}~--- the \emph{Length} and \emph{Rating} datasets for the
respective versions of rankers.


The target values (length/rating) in both sets are normalised into the $[0,1]$ range, and the
Length set contains context-response pairs from long dialogues (target value above $0.7$) as
positive instances and context-response pairs from short dialogues (target value below $0.3$) as
negative ones. With the same selection criteria, the Rating set contains context-response pairs from
highly rated dialogues (ratings 4 and 5) as the positive instances and context-response pairs from
low-rated dialogues (ratings 1 and 2) as the negative ones. Each dataset contains 500,000 instances
in total, with equal proportion of positive and negative instances. We use a 8:1:1 split for
training, development, and  test sets.

Prior to creating both datasets, we filtered out of the dialogue transcripts all system turns which
cannot be treated as natural social interaction (e.g.\ a quiz game) as well as the outliers
(interaction length $\ge 95$th percentile or less than 3 turns long).%
\footnote{Some extremely long dialogues are due to users repeating themselves over and over, and so
this filter removes these bad dialogues from the dataset. Dialogues less than 3 turns long are often
where the user accidentally triggered the chatbot. These outliers amounted to about 14\% of our
data.}
Thresholds of $0.3$ and $0.7$ were set heuristically based on preliminary data analysis.
On the one hand, these values provide contrastive-enough ratings (e.g.\ we are not sure whether the
rating in the middle of the scale can be interpreted as negative or positive).
On the other hand, they allow us to get enough training data for both Length and Rating datasets.%
\footnote{Using more extreme thresholds did not produce enough data while less extreme ones did not
provide adequate training signal.}


\section{Evaluation and Experimental Setup}
\label{ch8:eval}

In order to tune the neural rankers, we performed a grid search over the shared encoder \GRU layer
size and the Predictor topology.\footnote{We tested \GRU sizes of 64, 128, 256 and Predictor layers
number/sizes of [128], [128, 64], [128, 32, 32].} The best configurations are determined by the loss
on the development sets. For evaluation, we used an independent dataset.


\subsection{Evaluation Based on Explicit User Feedback}
\label{ch8:user-feedback-dataset}

\begin{table}[t]
    \center
    \small
    \begin{tabularx}{\linewidth}{@{}rll@{}}\toprule
    \textbf{Context}&\texttt{Usr}&no\\
    &\texttt{Sys}&i would love to talk about the news or maybe science, how about you?\\
    &\texttt{Usr}&let's talk about the news\\
    \textbf{Good response}&\texttt{Sys}&I read this somewhere - When will Hillary Clinton break her silence on\\
    &&Rick and Morty's Szechuan sauce? Shall I say some more about this?\\
    \textbf{Bad response}&\texttt{Sys}&Hiya! What do you want to talk about?\\
    \hline

    \textbf{Context}&\texttt{Usr}&darth vader\\
    &\texttt{Sys}&anakin skywalker the character in star wars episode iii revenge of the sith\\
    &\texttt{Usr}&yeah that's him\\
    \textbf{Good response}&\texttt{Sys}&Oh, regarding darth vader, I read this - Star Wars: Darth Vader 7 Reveals\\
    &&Another Jedi Who Survived the Clone Wars. Shall I say some more about this?\\
    \textbf{Bad response}&\texttt{Sys}&Ok, I think you were talking about airplane, I heard that,\\
    &&An airplane or aeroplane is a powered, fixed-wing aircraft that is propelled\\
    &&forward by thrust from a jet engine or propeller. Do you want to know more\\
    &&about airplane?\\
    \hline

    \textbf{Context}&\texttt{Usr}&reading books same here\\
    &\texttt{Sys}&oh a book worm how nice i like reading books on the cloud let's see what else\\
    &\texttt{Usr}&harry potter harry potter\\
    \textbf{Good response}&\texttt{Sys}&I believe that Harry Potter is a series of fantasy novels written by British author\\
    &&J. K. Rowling.\\
    \textbf{Bad response}&\texttt{Sys}&So Are you more interested in sports or video games?\\
    \bottomrule

    \end{tabularx}
    \caption{Examples from the User feedback dataset for pairwise ranking}
    \label{tab:pairwise_ranking}
\end{table}

At the evaluation stage, we check how well the rankers can distinguish between the good responses
and the bad ones. The criterion for `goodness' that we use here is chosen to be independent from
both training signals. Specifically, we collected an evaluation set composed of dialogue turns
followed by explicit positive feedback from the users which was described earlier in Section
\ref{ch8:user-rating}. Our `bad' response candidates are randomly sampled across the dataset.
Please note: while we use $\langle$positive feedback based response, random response$\rangle$
pairs to evaluate the ranker, the opposite case, i.e. $\langle$negative feedback based response,
random response$\rangle$ is not useful for the evaluation. The reason for that is, having `gold'
bad system responses (followed by the negative user feedback), it is not feasible to find the good
alternatives to them in their exact contexts via random sampling. Due to the extremely high scarcity
of the contexts, all the randomly sampled responses will most likely be bad.

`Goodness' defined in this way allows us to evaluate how well our two approximated training signals
can optimise for the user's satisfaction as explicitly expressed at the turn level, thus leading to
our desired behaviour, i.e., producing long and engaging dialogues. The \emph{User feedback} dataset
contains 24,982 $\langle$ context, good response, bad response $\rangle$ tuples in total.

To evaluate the rankers on this dataset, we use \textit{precision@k}, which is commonly used for
information retrieval system evaluation (Eq.~\ref{eq:precision_at_k}).

\begin{equation}
P@k(c, R) = \frac{\sum_{i=1}^k{\text{Rel}\left(c, R_k\right)}}{k}
\label{eq:precision_at_k}
\end{equation}

where $c$ is dialogue context, $R$ is response candidates list, and $\text{Rel}$ is a binary
predicate indicating whether a particular response is relevant to the context.

Precision is typically used together with recall and F-measure. However, since our dialogue data is
extremely sparse so that it is hard to find multiple good responses for the same exact dialogue
context, recall and F-measure cannot be applied to this setting.
Therefore, since we only perform pairwise ranking, we use \textit{precision@1} to check that the
good answer is the top-ranked one.
Also due to data sparsity, we only perform this evaluation with \textit{gold positive} responses and
\textit{sampled negative} ones~--- it is typically not possible to find a good response with exactly
the same context as a given bad response.


\subsection{Interim Results}

\begin{table}[t]
    \centering
    \begin{tabular}{lcc}\toprule
    \textbf{Model}&\textbf{P@1 (eval set)}&\textbf{Loss (test set)}\\\midrule
    Handcrafted&0.478&---\\
    VowpalWabbit@length&0.742&0.199\\
    VowpalWabbit@rating&0.773&0.202\\
    DualEncoder@length&0.365&0.239\\
    DualEncoder@rating&0.584&0.247\\
    Neural@length&0.824&0.139\\
    Neural@rating&\textbf{0.847}&\textbf{0.138}\\\bottomrule
    \end{tabular}
    \caption{Ranking models evaluation}
    \label{tab:ranker_eval}
\end{table}

The results of our first experiment~--- namely, pairwise ranking precision on the independent
\emph{User feedback} dataset and loss on the Length/Rating test sets (Section~\ref{ch8:dataset}) for
the corresponding trainset sizes of 500,000~--- are shown in Table~\ref{tab:ranker_eval}. We can see
that the neural ranker trained with user ratings clearly outperforms all the alternative approaches
in terms of test set loss on its respective dataset as well as pairwise ranking precision on the
evaluation dataset. Also note that both versions of the neural ranker stand extremely close to each
other on both evaluation criteria, given a much greater gap between them and their
next-best-performing alternatives, the linear rankers.

The dual-encoder ranker turned out to be not an efficient model for our problem, partly because it
was originally optimised for a different task as reported by \cite{lu_practical_2017}.


\section{Training on Larger Amounts of Data}
\label{ch8:larger-data}

\begin{figure}
\centering
\pgfplotsset{scaled x ticks=false}
\begin{tikzpicture}[thick,scale=1.0, every node/.style={transform shape}]
\begin{axis}[
    width=0.7\textwidth,
    xlabel={Trainset size},
    ylabel={Precision@1},
    xmin=500000, xmax=1000000,
    ymin=0.7, ymax=1,
    xtick={500000, 600000, 700000, 800000, 900000, 1000000},
    legend style={font=\small},
    legend cell align={right},
    ymajorgrids=true,
    xmajorgrids=true,
    grid style=dotted,
]
 
\addplot[dashed, color=blue]
coordinates {
    (500000,0.847)(600000,0.847)(700000,0.847)(800000,0.847)(900000,0.847)(1000000,0.847)
};
\addplot[dashed, color=green]
coordinates {
    (500000,0.773)(600000,0.773)(700000,0.773)(800000,0.773)(900000,0.773)(1000000,0.773)
};
\addplot[color=red, mark=square]
coordinates {
    (500000,0.824)(600000,0.852)(700000,0.864)(800000,0.861)(900000,0.868)(1000000,0.859)
};
\addplot[color=orange, mark=triangle]
coordinates {
    (500000,0.742)(600000,0.752)(700000,0.761)(800000,0.757)(900000,0.751)(1000000,0.738)
};
\legend{Neural@rating (baseline), \VW@rating (baseline), Neural@length, \VW@length}
\end{axis}
\end{tikzpicture}
\caption{Comparison of rankers trained on extended datasets}
\label{fig:alana_extended_datasets}
\end{figure}
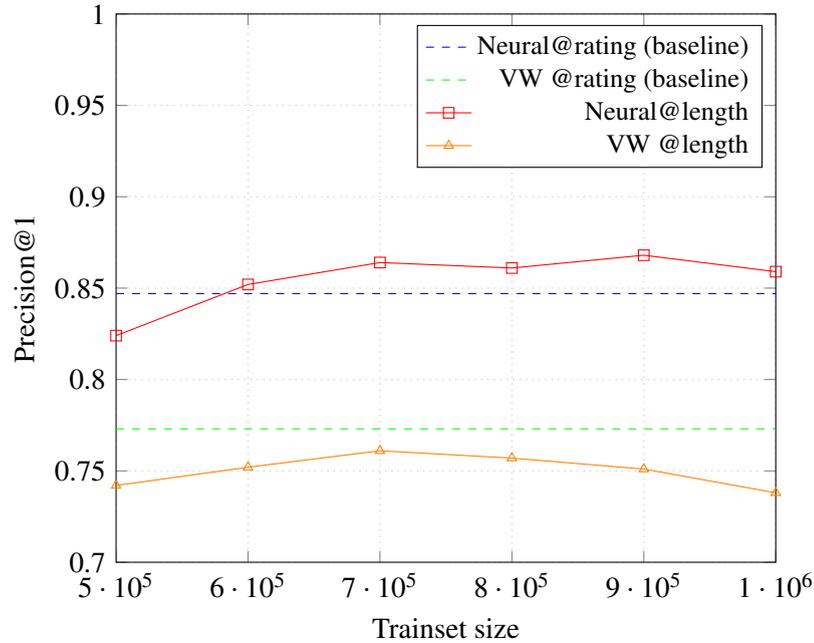

A major advantage of training on raw dialogue transcripts is data volume: in our case, we have
roughly twice as many raw dialogues as rated ones (cf.~Section~\ref{ch8:dataset}).
This situation is very common in data-driven development: since data annotation is a very expensive
and slow procedure, almost always there is significantly more raw data than annotated data of a high
quality. To illustrate this, we collected the extended training datasets of raw dialogues of up to
1,000,000 data points for training from the length signal. We trained our neural ranker and the VW
ranker using the same configuration as in Section~\ref{ch8:eval}.%

The results are shown in Figure \ref{fig:alana_extended_datasets}, where we see that the neural
ranker trained on the length signal consistently outperforms the ratings-based one. Its trend,
although fluctuating, is more stable than that of \VW~--- we believe that this is due to \VW's
inherent lower model capacity as well as its training setup, which is mainly optimised for speed.
The figure also shows that VW@length is worse than VW@rating, regardless of training data size.

\section{Discussion}
\label{ch8:discussion}

Our evaluation results show that the neural ranker presented above is an efficient approach to
response ranking for social conversation. On a medium-sized training set, the two versions of the
neural ranker, length and ratings-based, showed strongly superior performance to three alternative
ranking approaches, and performed competitively with each other. Furthermore, the experiment with
extended training sets shows that the accuracy of the length-based neural ranker grows steadily
given more unannotated training data, outperforming the rating-based ranker with only slightly
larger training sets.

The overall results of our experiments confirm that dialogue length, even approximated in quite a
straightforward way, provides a sufficient supervision signal for training a ranker for a social
conversation model.

\section{Related Work}
\label{ch8:related}

Work on response ranking for conversational systems has been been growing rapidly in recent years.
Some authors employ ranking based on heuristically defined measures:
\cite{yu_ticktock:_2015,yu_strategy_2016} use a heuristic based on keyword matching, part-of-speech
filters, and \WordToVec similarity. \cite{edina} apply standard information retrieval metrics
(\TFIDF) with importance weighting for named entities. However, most of the recent research attempts
to train the ranking function from large amounts of conversational data, as we do.
Some authors use task-based conversations, such as IT forums \citep{ubuntu_corpus} or customer
services (\citealp{lu_practical_2017}; \citealp{DBLP:conf/iwsds/KumarHCNN18}), while others focus on
online conversations on social media (e.g. \citealp{wu_ranking_2016};
\citealp{al-rfou_conversational_2016}).

The basic approach to learning the ranking function in most recent work is the same (e.g.
\citealp{ubuntu_corpus}; \citealp{al-rfou_conversational_2016}; \citealp{wu_ranking_2016}): the
predictor is taught to rank positive responses taken from real dialogue data higher than randomly
sampled negative examples. Some of the approaches do not even include rich dialogue contexts and use
only immediate context-response pairs for ranking (\citealp{ji_information_2014};
\citealp{yan_learning_2016}; \citealp{lu_practical_2017}).
Some authors improve upon this basic scenario: \cite{zhuang_ensemble_2018} take a desired emotion of
the response into account; \cite{liu_rubystar:_2017} focus on the engagement of responses based on
Reddit comments rating; \cite{fedorenko_avoiding_2017} train the ranking model in several
iterations, using highly ranked incorrect responses as negative examples for the next iteration.
Nevertheless, to our knowledge, none of the prior works attempt to optimise for long-term dialogue
quality; unlike in our work, their only ranking criterion is focused on the immediate response.


\section{Conclusion}
\label{ch8:conclusion}
We have presented a neural response ranker for open-domain `social' dialogue systems and described
two methods for training it using the common supervision signals coming from conversational data:
user-provided ratings and dialogue length. We demonstrated its efficiency by evaluating it using
explicit positive feedback as a measure for user engagement. Specifically, trained on ratings, our
neural ranker consistently outperforms several strong baselines; moreover, given larger amounts of
data and only using conversation length as the objective, the ranker performs better the
ratings-based one, reaching $0.87$ Precision@1.
This shows that conversation length can be used as an optimisation objective for generating engaging
social dialogues, which means that we no longer need the expensive and time-consuming procedure of
collecting per-dialogue user ratings, as was done for example in the Alexa Prize 2017 and is common
practice in conversational AI research.
Per-turn user ratings may still be valuable to collect for such systems, but these are even more
expensive and problematic to obtain.
Looking ahead, this advance will make data collection for social conversational agents simpler and
less expensive in the future.

\chapter{Conclusions and Future Work} 

\label{Chapter9} 

\lhead{Chapter 9. \emph{Conclusions and Future Work}} 


In this thesis, we have presented a series of techniques towards enabling the development of
data-efficient and robust dialogue systems in a data-driven way.

The core contributions of this thesis are the models for bootstrapping dialogue systems from
minimal data. Our dialogue knowledge transfer model \diktnet addresses the problem of training
dialogue response generation systems in a few-shot setup, and the hybrid \GRTr model is designed
for the adaptation to a new domain where there exists support data to retrieve responses from.

Our subsequent contributions presented in this thesis address potentially insufficient robustness of
the models trained from minimal data, addressing spoken disfluency detection and \OOD input detection
problems. The multitask \LSTM-based disfluency detector supports incremental word-by-word processing
and demonstrates generalisation potential beyond its main dataset~--- therefore, it can be used with
a wide range of dialogue models, potentially improving their coverage of naturally varied input data
with no extra training effort. In turn, the data-augmentation based training procedure for \OOD
handling is potentially applicable to the setups with the strictest training data limitations as it
does not require having real \OOD examples to adjust to and estimates the odds of encountering an
unusual utterance in usual contexts.

Finally, our study on data efficiency in social dialogue sheds light on a way to continuously
improve the performance of open-domain response rankers only using dialogue length as the main
supervision signal~--- thus avoiding the dependence on user ratings which are cumbersome and
expensive to collect and often times are overly noisy for the direct usage as a supervision signal.

\section{Directions for Future Work}
\label{ch9:future}

The findings we got from our studies in this thesis open up possibilities for further research~---
here, we will briefly outline the most promising directions of future work.

Firstly, the \diktnet presented in Chapter \ref{Chapter4}, while outperforming the previous best
model on the Stanford Multi-Domain dataset in both accuracy and the amount of in-domain data
required, still has got several areas of improvement. We see a potential benefit in bringing more
structure to the latent dialogue representation which would intuitively correspond better to the
actual structure of dialogue, e.g. as shown in \cite{DBLP:conf/naacl/ShiZY19}. Another potential
point of improvement is the system's goal-oriented performance: we assume that in a few-shot setup,
it highly depends on the way in which the system handles the KB entries: a study of the particular
copy mechanism employed in our setup and its optimality for the task might shed light on the ways to
attain higher Entity F1 scores. Finally, an evaluation beyond the word overlap-based metrics is
necessary for the adequate assessment of the system's performance in real-world settings.

Similarly, the absolute results of \GRTr, the winning entry at Dialog System technology Challenge 8
Fast Domain Adaptation task which we presented in Chapter \ref{Chapter5}, still suggest that domain
adaptation for response generation needs further research towards achieving production-level
performance. One promising direction that we are going to explore in our future work is the
meta-learning framework, e.g. as used by \cite{DBLP:conf/acl/QianY19}. Based on splitting the task
into multiple subtasks~--- with an independent copy of the base model for each one~--- and
subsequently incorporating the individual training progress back into the base model, meta-learning
approach will naturally fit the multi-domain, multi-task nature of the MetaLWOz dataset as well as
lead to a potentially better fine-tuning performance.

For the multitask \LSTM disfluency detector presented in Chapter \ref{Chapter6}, we see the main
issue being the modest out-of-dataset generalisation performance. Although it was a common issue for
all the models we evaluated, attaining a more practical level of generalisation is key for making
this model a truly reusable component for a wide range of dialogue system pipelines~--- therefore,
addressing this issue is our next step in this direction. As such, we will explore possibilities of
knowledge transfer to new closed domains in a 1-shot setting, both with regular supervised training
and unsupervised \LM fine-tuning.

Our study on robustness to out-of-domain input presented in Chapter \ref{Chapter7} of this thesis
also leaves space for further exploration. As such, we presented and evaluated a series of models
for the detection of \OOD utterances~--- all based on autoencoders of different types. It therefore
makes sense to explore other ways of scoring \OOD utterances than autoencoders: for example,
Generative Adversarial Networks (\GANs) have great potential, especially for their inherent
capabilities to produce realistic data samples. We are also interested in using traditional
generative models to produce more realistic counterfeit user utterances. Finally, it is worth noting
that our method is conceptually designed for improving robustness in extreme data-efficient settings
(i.e. 1-shot/few-shot training) by not requiring any other data than the available \IND dialogue
examples. However, in the interest of comparability with the original \HCN model, we performed our
experiments with full data~--- therefore, few-shot evaluation of our models is the immediate next
step in this research direction for us.

The final study of this thesis presented in Chapter \ref{Chapter8} addressed the problems of social,
or chat-oriented dialogue. As such, our neural model for conversational response ranking together
with the technique of training it from raw data confirmed that dialogue length, even approximated in
quite a straightforward way, provides a sufficient supervision signal for training a ranker for
a social conversation model. In our future work, we will attempt to further improve the model using
the same data in an adversarial setup following \cite{irgan}. We also plan to directly train our
model for pairwise ranking in the fashion of \cite{ranknet} instead of the current point-wise
approach. Finally, we are going to employ contextual sampling of negative responses using the
approximate nearest neighbour search \citep{JDH17} in order to perform a more efficient pairwise
training.

The above techniques have the potential to make an impact in industrial data-driven dialogue system
development. For that, they can be incorporated into a user-friendly `data efficiency toolkit' aimed
at usage by non-experts. As the main way to control the behaviour of the purely data-driven systems
assumes passing the corresponding training examples into the pipeline, there is the need for the
means to analyse and monitor the performance of the dialogue system's key components in a
transparent way for the end user. This can be done in terms of the specific decisions made by the
system and the training examples contributed to those. Although explainable machine learning was not
the scope of this work, it is important to point out that such means are vital for a data-driven
product targeted at non-expert users.

Crucially, the models and techniques presented here were evaluated in certain setups which are
mainly motivated by reproducibility and fair comparison to the previous models for the respective
tasks. These setups do not necessarily correspond to the designated settings of limited training
data. For example, the disfluency detection model as well as the \OOD detection models were trained
and evaluated on the full datasets, and therefore a few-shot evaluation with the possible subsequent
fine-tuning of the models is needed in order to make the them practically justified.

We see a strong potential for a significant impact~--- both in academia and industry~--- of the
data-efficient techniques presented in this thesis and will continue pursuing efforts to ensure wide
applicability of those in real-world scenarios following the steps outlined above.
We hope that the open-source resources released as part of this work will foster further research
in this direction across the dialogue research community.


\addtocontents{toc}{\vspace{2em}} 

\appendix 



\chapter{Dialogue Knowledge Transfer Networks~--- Supplementary Material} 

\label{AppendixA} 

\lhead{Appendix A. \emph{DiKTNet}} 

\begin{table}[ht!]
  \centering
  \footnotesize
    \begin{tabular}{lllllll}\toprule
      &\multicolumn{2}{c}{\textbf{Navigation}}&\multicolumn{2}{c}{\textbf{Weather}}&\multicolumn{2}{c}{\textbf{Schedule}}\\
      &\multicolumn{1}{c}{BLEU, \%}&\multicolumn{1}{c}{Entity F1, \%}&\multicolumn{1}{c}{BLEU, \%}&\multicolumn{1}{c}{Entity F1, \%}&\multicolumn{1}{c}{BLEU, \%}&\multicolumn{1}{c}{Entity F1, \%}\\\midrule
      ZSDG&5.9&14.0&8.1&31&7.9&36.9\\
      NLU\_ZSDG&$6.1\pm2.2$&$12.7\pm3.3$&$5.0\pm1.6$&$16.8\pm6.7$&$6.0\pm1.7$&$26.5\pm5.4$\\
      NLU\_ZSDG+Stage1&$7.9\pm1$&$12.3\pm2.9$&$8.7\pm0.6$&$21.5\pm6.2$&$8.3\pm1$&$20.7\pm4.8$\\
      \hline
      HRED@1\%&$6.0\pm1.8$&$9.8\pm4.8$&$6.9\pm1.1$&$22.2\pm 10.7$&$5.5\pm0.8$&$25.6\pm8.2$\\
      HRED@3\%&$7.9\pm0.7$&$11.8\pm4.4$&$9.6\pm1.8$&$39.8\pm 7$&$8.2\pm1.1$&$34.8\pm4.4$\\
      HRED@5\%&$8.3\pm1.3$&$15.3\pm6.3$&$11.5\pm 1.6$&$38.0\pm10.5$&$9.7\pm1.4$&$37.6\pm8.0$\\
      HRED@10\%&$9.8\pm0.8$&$19.2\pm3.2$&$12.9\pm2.4$&$40.4\pm11.0$&$12.0\pm1.0$&$38.2\pm4.2$\\
      \hline
      HRED+VAE@1\%&$3.6\pm2.6$&$9.3\pm4.1$&$6.8\pm1.3$&$23.2\pm10.1$&$4.6\pm1.6$&$28.9\pm7.3$\\
      HRED+VAE@3\%&$6.9\pm1.9$&$15.6\pm5.8$&$9.5\pm2.6$&$32.2\pm11.8$&$6.6\pm1.7$&$34.8\pm7.7$\\
      HRED+VAE@5\%&$7.8\pm1.9$&$12.7\pm4.2$&$10.1\pm2.1$&$40.3\pm10.4$&$8.2\pm1.7$	&$34.2\pm8.7$\\
      HRED+VAE@10\%&$9.0\pm2.0$&$18.0\pm5.8$&$12.9\pm2.2$&$40.1\pm7.6$&$11.6\pm1.5$&	$39.9\pm6.9$\\\hline
      HRED+Stage1@1\%&$7.1\pm0.8$&$10.1\pm4.5$&$10.6\pm2.1$&$31.4\pm8.1$&$7.4\pm1.2$&$29.1\pm6.6$\\
      HRED+Stage1@3\%&$9.2\pm0.8$&$14.5\pm4.8$&$13.1\pm1.7$&$40.8\pm6.1$&$9.2\pm1.2$&$32.7\pm6.1$\\
      HRED+Stage1@5\%&$10.3\pm1.2$&$15.6\pm4.5$&$14.5\pm2.2$&$40.9\pm8.6$&$11.8\pm1.9$&$37.6\pm6.1$\\
      HRED+Stage1@10\%&$12.3\pm0.9$&$17.3\pm4.5$&$17.6\pm1.9$&$47.5\pm6.0$&$15.2\pm1.6$	&$38.7\pm8.4$\\\hline
      HRED+ELMo@1\%&$5.8\pm1.9$&$18.2\pm3.8^\mathbf{\star}$&$7.3\pm2.6$&$38.5\pm11.1$&$6.3\pm2.6$&$36.3\pm9.2$\\
      HRED+ELMo@3\%&$8.0\pm1.3$&$17.2\pm4.2$&$10.6\pm1.1$&$42.0\pm11.0$&$9.5\pm2.0$&$39.6\pm9.2$\\
      HRED+ELMo@5\%&$9.4\pm0.8$&$21.5\pm7.3$&$12.1\pm2.0$&$39.0\pm12.8$&$11.3\pm2.1$&$40.0\pm5.6$\\
      HRED+ELMo@10\%&$9.9\pm1.1$&$24.3\pm5.7$&$14.9\pm2.7$&$41.4\pm12.0$&$14.5\pm1.4$&$43.4\pm3.9$\\\hline
      \textbf{DiKTNet@1\%}&$\mathbf{8.4\pm0.7^\star}$&$\mathbf{15.2\pm4.0}$&$\mathbf{11.5\pm1.7^\star}$&$\mathbf{43.0\pm10.5^\star}$&$\mathbf{8.1\pm0.8^\star}$&$\mathbf{40.5\pm6.3^\star}$\\
      DiKTNet@3\%&$10.4\pm1.2$&$19.2\pm4.8$&$15.7\pm2.1$&$44.0\pm11.7$&$11.1\pm1.3$&$38.2\pm5.8$\\
      DiKTNet@5\%&$11.5\pm1.1$&$23.9\pm2.9$&$15.5\pm2.1$&$39.5\pm14.8$&$13.7\pm2.0$&$41.1\pm3.8$\\
      DiKTNet@10\%&$12.9\pm1.0$&$26.8\pm4.2$&$20.4\pm1.2$&$48.0\pm5.6$&$17.5\pm1.3$&$42.8\pm2.6$\\
    \bottomrule
    \end{tabular}
    \caption[\diktnet evaluation results]{Evaluation results. Marked with asterisks are individual
    results higher than \ZSDG's performance and which are achieved with the minimum amount of
    training data. In bold is the model consistently outperforming \ZSDG in all domains and metrics
    with minimum data.}
    \label{tab:diktnet_results}
\end{table}

\begin{table}[ht]
\small
\center
\begin{tabularx}{\textwidth}{lSlS}\toprule
\textbf{Domain}&\textbf{\#Dialogues}&\textbf{Domain}&\textbf{\#Dialogues}\\\midrule
UPDATE CALENDAR&1991&GUINESS CHECK&1886\\
ALARM SET&1681&SCAM LOOKUP&1658\\
PLAY TIMES&1601&GAME RULES&1590\\
CONTACT MANAGER&1483&LIBRARY REQUEST&1339\\\midrule
INSURANCE&1299&HOME BOT&1210\\
HOW TO BASIC&1086&CITY INFO&965\\
TIME ZONE&951&TOURISM&935\\
SHOPPING&903&BUS SCHEDULE BOT&898\\\midrule
CHECK STATUS&784&WHAT IS IT&776\\
STORE DETAILS&737&APPOINTMENT REMINDER&668\\
PRESENT IDEAS&664&GEOGRAPHY&653\\
SKI BOT&607&MOVIE LISTINGS&607\\\midrule
UPDATE CONTACT&581&ORDER PIZZA&577\\
EDIT PLAYLIST&574&SPORTS INFO&561\\
BOOKING FLIGHT&555&WEATHER CHECK&551\\
EVENT RESERVE&539&RESTAURANT PICKER&535\\\midrule
LOOK UP INFO&533&AUTO SORT&514\\
QUOTE OF THE DAY BOT&513&WEDDING PLANNER&510\\
MAKE REST. RESERVATIONS&510&AGREEMENT BOT&507\\
NAME SUGGESTER&499&APARTMENT FINDER&499\\\midrule
HOTEL RESERVE&497&PHONE PLAN BOT&496\\
DECIDER BOT&487&VACATION IDEAS&486\\
PHONE SETTINGS&473&POLICY BOT&447\\
PROMPT GENERATOR&446&MUSIC SUGGESTER&445\\\midrule
PET ADVICE&426&BANK BOT&367\\
CATALOGUE BOT&288&&\\
\bottomrule

\end{tabularx}
\caption{\metalwoz domains}
\label{tab:maluuba_domains}
\end{table}

\section*{\metalwoz~--- dialogue examples}
\label{a:metalwoz_examples}
\textbf{Domain}: Ski \\
\textbf{User Task}: You want to know if there are good ski hills an hour's drive from your current location. \\
\textbf{Bot Task}: Tell the user that there are no ski hills in their immediate location. \\
\textcolor{red}{Bot} Hello how may I help you? \\
\textcolor{blue}{User} Is there any good ski hills an hour's drive from my current location? \\
\textcolor{red}{Bot} I'm sorry to inform you that there are no ski hills in your immediate location \\
\textcolor{blue}{User} Can you help me find the nearest? \\
\textcolor{red}{Bot} Absolutely! It looks like you're about 3 hours away from Bear Mountain. That seems to be the closest.\\
\textcolor{blue}{User} Hmm.. sounds good \\
\textcolor{red}{Bot} Alright! I can help you get your lift tickets now!When will you be going? \\
\textcolor{blue}{User} Awesome! please get me a ticket for 10pax \\
\textcolor{red}{Bot} You've got it. Anything else I can help you with? \\
\textcolor{blue}{User} None. Thanks again! \\
\textcolor{red}{Bot} No problem! \\
\\
\textbf{Domain}: Game rules \\
\textbf{User Task}: Ask how to win at the text adventure Zork. If the Bot cannot help you, ask them what games it can help you with.\\
\textbf{Bot Task}: Tell the user that you are programmed to help them play games, not win them. If the user starts talking about other games, go along with them and try to help.  \\ 
\textcolor{red}{Bot} Hello how may I help you? \\
\textcolor{blue}{User} I want to know how I can win playing Zork? \\
\textcolor{red}{Bot} I am programmed to help you play games, not win them \\
\textcolor{blue}{User} What games can you help me with? \\
\textcolor{red}{Bot} Any game \\
\textcolor{blue}{User} Can you specify a certain game? \\
\textcolor{red}{Bot} Zork \\
\textcolor{blue}{User} What are the rules of Zork? \\
\textcolor{red}{Bot} You have to follow through a cave adventure \\
\textcolor{blue}{User} What is the ending of the game? \\
\textcolor{red}{Bot} The user comes out of the cave \\
\\
\textbf{Domain}: Music suggestion \\
\textbf{User Task}: Tell the bot that you like a certain genre of music, but don't prompt it to give you suggestions \\
\textbf{Bot Task}: Provide suggestions, even if the user doesn't ask for them (or tells you they don't need any) \\ 
\textcolor{red}{Bot} Hello how may I help you? \\
\textcolor{blue}{User} Hello.  I like classic rock music. \\
\textcolor{red}{Bot} Dream on Aerosmith \\
\textcolor{blue}{User} No, I'm not a huge fan of that song.  Is there anything else available? \\
\textcolor{red}{Bot} Sweet Home Alabama \\
\textcolor{blue}{User} Well, I do like Lynyrd Skynyrd, but that song is so cliche.  Do you have anything else by the same band? \\ 
\textcolor{red}{Bot} Free bird \\
\textcolor{blue}{User} Yes, that is the perfect song for how I am feeling right now! \\ 
\textcolor{red}{Bot} Great, Is there anything else? \\
\textcolor{blue}{User} No, but I really appreciate your assistance.  Thank you so much! \\
\textcolor{red}{Bot} Youre welcome \\
\\
\textbf{Domain}: Information retrieval \\
\textbf{User Task}: Describe a movie where these guys have to climb a mountain and then throw a thing into the mountain and ask the bot if they know what the movie is called. \\
\textbf{Bot Task}: Ask the user to elaborate as much as possible about the movie they are asking about, and then suggest they're thinking of Lord of the Rings. \\ 
\textcolor{red}{Bot} Hello how may I help you? \\
\textcolor{blue}{User} I'm trying to remember this movie. Can you help? \\
\textcolor{red}{Bot} I would be happy to help.  Can you tell me what it is about? \\
\textcolor{blue}{User} In it these guys have to climb up a mountain. Then they throw something into the mountain. \\
\textcolor{red}{Bot} Do you know what genre the movie is, such as comedy, horror, sci-fi, or fantasy? \\
\textcolor{blue}{User} Not sure about that. \\
\textcolor{red}{Bot} Do you any of the actor or character names in the movie? \\
\textcolor{blue}{User} No... All I can remember is the stuff about the mountain. \\
\textcolor{red}{Bot} Is it a new movie, or an older movie? \\
\textcolor{blue}{User} I don't know. \\
\textcolor{red}{Bot} It sounds like the movie you are talking about could be Lord of the Rings.  Does that sound right? \\
\textcolor{blue}{User} It could be that. Thanks for the help. \\

\chapter{Generative-Retrieval Transformers~--- Supplementary Material} 

\label{AppendixB} 

\lhead{Appendix B. \emph{\GRTr}} 

\begin{table}[h!]
    \begin{tabular}{lSSSSSS}
        \toprule
        &\textbf{\BLEU[1]}&\textbf{\BLEU[2]}&\textbf{\BLEU[3]}&\textbf{\CIDEr}&\textbf{\METEOR}&\textbf{\ROUGEL}\\
        \midrule
        Retrieval,\ \BERT&7.93&4.43&2.87&12.56&7.38&6.91\\
        Retrieval,\ \SPFT&9.57&5.37&3.45&14.32&6.98&7.19\\
        \HRED&8.66&3.86&2.11&13.73&6.02&7.75\\
        \cmidrule(r){1-1}

        \GPT[2] base\footnotemark[1]&8.2\phantom0&3.95&2.22&16.41&6.1&8.34\\
        \GPT[2] +sup\footnotemark[2]&11.33&6.45&4.17&23.38&8.23&10.74\\
        \textbf{\GRTr}&\bf{12.73}&\bf{7.43}&\bf{4.88}&\bf{28.74}&\bf{9.18}&\bf{11.77}\\
        \bottomrule
    \end{tabular}

    \small{\footnotemark[1] does not use support set.\footnotemark[2] fine-tuned to support set, but does not use retrieval logic}
    \caption{Automatic evaluation results on \metalwoz pure task} 
    \label{tab:results-metalwoz-pure}
\end{table}

\begin{table}[h!]
    \begin{tabular}{lSSSSSS}
        \toprule
        &\textbf{\BLEU[1]}&\textbf{\BLEU[2]}&\textbf{\BLEU[3]}&\textbf{\CIDEr}&\textbf{\METEOR}&\textbf{\ROUGEL}\\
        \midrule
        Retrieval,\ \BERT&5.35&2.16&1.05&4.98&4.56&4.52\\
        Retrieval,\ \SPFT&5.94&2.25&0.93&4.69&4.29&4.53\\
        \HRED&8.94&3.87&2.02&12.65&6.05&7.55\\
        \cmidrule(r){1-1}

        \GPT[2] base&8.37&3.8&2.05&15.6&6.17&8.55\\
        \GPT[2] +sup&10.21&5.26&2.95&18.06&7.06&\bf{9.59}\\
        \textbf{\GRTr}&\bf{10.39}&\bf{5.31}&\bf{2.95}&\bf{18.26}&\bf{7.1}&9.27\\
        \bottomrule
    \end{tabular}
    \caption{Automatic evaluation results on \metalwoz cross-task} 
    \label{tab:results-metalwoz-cross}
\end{table}


\begin{table}[h!]
    \center
    \footnotesize
    \begin{tabularx}{\linewidth}{@{}rll@{}}\toprule
    \textbf{Context}&\texttt{Wiz}&Hello how may I help you?\\
    &\texttt{Usr}&I'm trying to book rooms\\
    &\texttt{Wiz}&For where?\\
    &\texttt{Usr}&I need a few hotel rooms in Tucson\\
    &\texttt{Wiz}&how many total rooms?\\
    \textbf{Gold response}&\texttt{Usr}&I need ten rooms\\
    \textbf{Generated candidate}&\texttt{Usr}&Five rooms [$\mathbf{2.794}$]\\
    \textbf{Retrieved candidate}&\texttt{Usr}&I need 4 rooms on the same floor [$2.793$]\\
    \hline

    \textbf{Context}&\texttt{Wiz}&Hello how may I help you?\\
    &\texttt{Usr}&Want some info about Cyprus.\\
    &\texttt{Wiz}&What would you like to know about Cyprus?\\
    \textbf{Gold response}&\texttt{Usr}&What's best time to visit there?\\
    \textbf{Generated candidate}&\texttt{Usr}&What is the best time to visit? [$\mathbf{1.088}$]\\
    \textbf{Retrieved candidate}&\texttt{Usr}&What is the best time to visit Cyprus? [$1.087$]\\
    \hline

    \textbf{Context}&\texttt{Wiz}&Hello how may I help you?\\
    &\texttt{Usr}&I need some help.\\
    &\texttt{Wiz}&yes, how can i help you?\\
    &\texttt{Usr}&I'm going to Montreal, and good festivals?\\
    &\texttt{Wiz}&yes, there will have music festivals and dancing party. and wine testing party.\\
    &\texttt{Usr}&In the summer?\\
    &\texttt{Wiz}&yes, it will in the august.\\
    &\texttt{Usr}&September?\\
    &\texttt{Wiz}&there only have wine testing party for september.\\
    \textbf{Gold response}&\texttt{Usr}&Bummer\\
    \textbf{Generated candidate}&\texttt{Usr}&ok thanks [-1.282]\\
    \textbf{Retrieved candidate}&\texttt{Usr}&That sounds interesting. Can you book me for that day? [$\mathbf{-1.274}$]\\
    \hline

    \textbf{Context}&\texttt{Wiz}&Hello how may I help you?\\
    &\texttt{Usr}&I need some ideas\\
    &\texttt{Wiz}&Sure About what?\\
    &\texttt{Usr}&Where's a good place to bring kids for vacation\\
    &\texttt{Wiz}&I highly suggest Disney Land\\
    &\texttt{Usr}&That's a good idea. When is a good time for us to go? When it's not so hot?\\
    &\texttt{Wiz}&But the place is vey expensive\\
    \textbf{Gold response}&\texttt{Usr}&Yeah, I'm okay with that\\
    \textbf{Generated candidate}&\texttt{Usr}&I don't think that's a good idea. What are the best places to bring kids? [$0.375$]\\
    \textbf{Retrieved candidate}&\texttt{Usr}&Oh I know we'd never go if we didn't get a Florida discount on tickets [$\mathbf{0.384}$]\\
    \hline

    \textbf{Context}&\texttt{Wiz}&Hello how may I help you?\\
    &\texttt{Usr}&I need to book a flight and i need it by tonight\\
    &\texttt{Wiz}&Ok, book a flight, it will be expensive at this late date\\
    &\texttt{Usr}&Book me a flight to hawaii for 2 people leaving tonight\\
    &\texttt{Wiz}&Ok, let me see what I have, I have two tickets for 100,000 dollars\\
    \textbf{Gold response}&\texttt{Usr}&Book it. Money is no limit\\
    \textbf{Generated candidate}&\texttt{Usr}&Money is not a problem for us [$0.805$]\\
    \textbf{Retrieved candidate}&\texttt{Usr}&Sounds good I'll take them [$\mathbf{0.819}$]\\

    \bottomrule
    \end{tabularx}
    \caption[\metalwoz pure task: \GRTr predictions with the closest generated/retrieved candidates]
    {\GRTr predictions with the closest generated/retrieved candidates~--- \metalwoz pure
    task (in bold is the model's final response)}
    \label{tab:grtr_closest_candiates_metalwoz_pure}
\end{table}

\begin{table}[h!]
    \center
    \footnotesize
    \begin{tabularx}{\linewidth}{@{}rll@{}}\toprule

    \textbf{Context}&\texttt{Wiz}&Hello how may I help you?\\
    &\texttt{Usr}&Hello. I need to book a flight for two\\
    &\texttt{Wiz}&ok where are you going?\\
    &\texttt{Usr}&I will be heading to Hawaii and I need to leave tonight\\
    &\texttt{Wiz}&will you be bringing extra luggage?\\
    &\texttt{Usr}&No, I will need first class seats. Money is no object\\
    &\texttt{Wiz}&i have 2 tickets for \$50,000\\
    &\texttt{Usr}&Okay, that will be fine. Please purchase with my credit card on file\\
    &\texttt{Wiz}&just to verify, what is the last 2 digits of your credit card?\\
    \textbf{Gold response}&\texttt{Usr}&The last two digits of my credit card are 00\\
    \textbf{Generated candidate}&\texttt{Usr}&The last 2 digits are: 5555 [$\mathbf{5.081}$]\\
    \textbf{Retrieved candidate}&\texttt{Usr}&Sure. Just sent it to my email. Thanks! [$-9.761$]\\
    \hline

    \textbf{Context}&\texttt{Wiz}&Hello how may I help you?\\
    &\texttt{Usr}&I need an idea.\\
    &\texttt{Wiz}&Sure, what's up?\\
    &\texttt{Usr}&Can use advise me on a vacation spot?\\
    &\texttt{Wiz}&Yes. Disney World and Disney Land are always fun for children and adults,\\
    &&but it's really expensive\\
    &\texttt{Usr}&Anything closer?\\
    &\texttt{Wiz}&Have you ever been to Niagara Falls?\\
    &\texttt{Usr}&Yes,but that's too far away.\\
    &\texttt{Wiz}&Where is your location?\\
    \textbf{Gold response}&\texttt{Usr}&Nebraska\\
    \textbf{Generated candidate}&\texttt{Usr}&I'm in New York. [$\mathbf{3.711}$]\\
    \textbf{Retrieved candidate}&\texttt{Usr}&ok, thanks [$-9.605$]\\
    \hline

    \textbf{Context}&\texttt{Wiz}&Hello how may I help you?\\
    &\texttt{Usr}&I need to book a room for me\\
    &\texttt{Wiz}&OK. Where do you want to stay?\\
    &\texttt{Usr}&In downtown Toronto\\
    &\texttt{Wiz}&And for what dates would you like the room?\\
    &\texttt{Usr}&I need a 2 bedroom on the second floor I'd like to stay for this weekend\\
    &\texttt{Wiz}&And how many people will be in your party?\\
    \textbf{Gold response}&\texttt{Usr}&2\\
    \textbf{Generated candidate}&\texttt{Usr}&Just myself and a few friends please [$\mathbf{3.82}$]\\
    \textbf{Retrieved candidate}&\texttt{Usr}&Price isn't a concern. I would like a room on the second floor\\
    &&with 2 beds if that helps. [$-9.183$]\\
    \hline

    \textbf{Context}&\texttt{Wiz}&Hello how may I help you?\\
    &\texttt{Usr}&I need some ideas.\\
    &\texttt{Wiz}&Okay, what about?\\
    &\texttt{Usr}&I'm going to be taking my vacation soon\\
    &\texttt{Wiz}&Where will you like to go?\\
    \textbf{Gold response}&\texttt{Usr}&I need someone good for the kids. somewhere*\\
    \textbf{Generated candidate}&\texttt{Usr}&Well, I'm going to be in Miami [$\mathbf{2.13}$]\\
    \textbf{Retrieved candidate}&\texttt{Usr}&yes please. [$-10.672$]\\
    \hline

    \textbf{Context}&\texttt{Wiz}&Hello how may I help you?\\
    &\texttt{Usr}&I have 5 days of vacation next month I would like to go somewhere\\
    &\texttt{Wiz}&Do you have any favorite places you have in mind?\\
    &\texttt{Usr}&Anywhere in colorado, wyoming, montana\\
    &\texttt{Wiz}&Let me look that up for you! If you're planning for Denver, Colorado.\\
    &&I see a few popular tourist attractions: Denver Art Museum and Elitch Gardens\\
    &&Theme Park Would you like to vacation there, or would you like me to continue\\
    &&based on your list of favorite places?\\
    \textbf{Gold response}&\texttt{Usr}&That's a good idea. When is a good time for us to go? When it's not so hot?\\
    \textbf{Generated candidate}&\texttt{Usr}&Sure, I'll take it! Thanks for the suggestions [$\textbf{-0.11}$]\\
    \textbf{Retrieved candidate}&\texttt{Usr}&What are some cheap Florida places to go? [$-11.762$]\\

    \bottomrule
    \end{tabularx}
    \caption[\metalwoz pure task: \GRTr predictions with the most distant generated/retrieved candidates]{\GRTr predictions with the most distant generated/retrieved candidates~--- \metalwoz
    pure task (in bold is the model's final response)}
    \label{tab:grtr_distant_candiates_metalwoz_pure}
\end{table}

\begin{table}[h!]
    \center
    \footnotesize
    \begin{tabularx}{\linewidth}{@{}rll@{}}\toprule

    \textbf{Context}&\texttt{Wiz}&Hello how may I help you?\\
    &\texttt{Usr}&help me book a flight\\
    &\texttt{Wiz}&do you have a particular place in mind?\\
    &\texttt{Usr}&i hear Greece is nice this time of year. how do i get a flight there?\\
    &\texttt{Wiz}&yes, it is; i can book your flights now, if you like\\
    &\textit{Usr}&wait don't! i am only curious on how to fly there\\
    &\texttt{Wiz}&AA Airlines fly to Greece everyday\\
    &\textit{Usr}&how much is the ticket?\\
    &\textit{Wiz}&we have specials this month for \$350 roundtrip\\
    \textbf{Gold response}&\texttt{Usr}&i don't have enough funds. i'll save up\\
    \textbf{Generated candidate}&\texttt{Usr}&how much is the flight to Germany? [$\mathbf{1.569}$]\\
    \textbf{Retrieved candidate}&\texttt{Usr}&thanks [$1.57$]\\
    \hline

    \textbf{Context}&\texttt{Wiz}&Hello how may I help you?\\
    &\texttt{Usr}&Hello bot!\\
    &\texttt{Wiz}&Hi, how can I help you?\\
    &\texttt{Usr}&Are there any good festivals in Montreal this summer?\\
    &\texttt{Wiz}&Yes there are a few, would you like me to list them?\\
    &\textit{Usr}&List the best one.\\
    &\texttt{Wiz}&In my opinion the best one is the beer festival!\\
    \textbf{Gold response}&\texttt{Usr}&Awesome, I love to drink. How much are tickets.\\
    \textbf{Generated candidate}&\texttt{Usr}&Okay. I'll go with that. [$\mathbf{1.754}$]\\
    \textbf{Retrieved candidate}&\texttt{Usr}&I like that idea. [$1.752$]\\
    \hline

    \textbf{Context}&\texttt{Wiz}&Hello how may I help you?\\
    &\texttt{Usr}&Hi! I need help with my travelling.\\
    &\texttt{Wiz}&How can I help you? Okay good. I can offer you a flight to Franc, England, Germany, and Italy\\
    &\texttt{Usr}&Sure. Can you get me a flight to Germany by next tuesday?\\
    &\texttt{Wiz}&Alright done. Flight to Gemany.\\
    \textbf{Gold response}&\texttt{Usr}&How much does it cost?\\
    \textbf{Generated candidate}&\texttt{Usr}&Okay. I need to book a flight to Germany on Monday. [$\mathbf{-1.916}$]\\
    \textbf{Retrieved candidate}&\texttt{Usr}&Perfect, what's the price tag? [$-1.921$]\\
    \hline

    \textbf{Context}&\texttt{Wiz}&Hello how may I help you?\\
    &\texttt{Usr}&I am heading to Montreal this Summer, any good festivals around that time?\\
    &\texttt{Wiz}&Montreal Pride\\
    &\texttt{Usr}&What will I do there?\\
    &\texttt{Wiz}&It's a music fest\\
    \textbf{Gold response}&\texttt{Usr}&Great! I like that bot\\
    \textbf{Generated candidate}&\texttt{Usr}&Is there anything else? [$-0.309$]\\
    \textbf{Retrieved candidate}&\texttt{Usr}&Other suggestions? [$\mathbf{-0.3}$]\\
    \hline

    \textbf{Context}&\texttt{Wiz}&Hello how may I help you?\\
    &\texttt{Usr}&i need help with vacations\\
    &\texttt{Wiz}&How can I help you with that?\\
    &\texttt{Usr}&whats a good place to bring kids\\
    &\texttt{Wiz}&Disney World or Disney Land is always fun! A little pricey but fun.\\
    \textbf{Gold response}&\texttt{Usr}&anything else?\\
    \textbf{Generated candidate}&\texttt{Usr}&ok i will go there [$0.799$]\\
    \textbf{Retrieved candidate}&\texttt{Usr}&sounds good, book tickets [$\mathbf{0.812}$]\\

    \bottomrule
    \end{tabularx}
    \caption[\metalwoz cross-task: \GRTr predictions with the closest generated/retrieved candidates]{\GRTr predictions with the closest generated/retrieved candidates~--- \metalwoz
    cross-task (in bold is the model's final response)}
    \label{tab:grtr_closest_candiates_metalwoz_cross}
\end{table}

\begin{table}[h!]
    \center
    \footnotesize
    \begin{tabularx}{\linewidth}{@{}rll@{}}\toprule

    \textbf{Context}&\texttt{Wiz}&Hello how may I help you?\\
    &\texttt{Usr}&i need a room\\
    &\texttt{Wiz}&Sure, where are you looking for a room at?\\
    &\texttt{Usr}&with 2 bedrooms\\
    &\texttt{Wiz}&Okay, I can help with that.\\
    &\texttt{Usr}&and it is located on the second floor of a hotel\\
    &\texttt{Wiz}&Okay, what city are you interested in?\\
    &\texttt{Usr}&te specific place is in toronto\\
    &\texttt{Wiz}&What date would you like to book the room for?\\
    \textbf{Gold response}&\texttt{Usr}&Next monday\\
    \textbf{Generated candidate}&\texttt{Usr}&May 16th [$\mathbf{2.642}$]\\
    \textbf{Retrieved candidate}&\texttt{Usr}&Thanks a lot! [$-13.151$]\\
    \hline

    \textbf{Context}&\texttt{Wiz}&Hello how may I help you?\\
    &\texttt{Usr}&Hey, I've got a question about tourism\\
    &\texttt{Wiz}&Sure, I can help with that.\\
    &\texttt{Usr}&I'm curious if there are any bars around here\\
    &\texttt{Wiz}&Could you specify your location for me?\\
    \textbf{Gold response}&\texttt{Usr}&Where I am?\\
    \textbf{Generated candidate}&\texttt{Usr}&I'm in Tampa, I need some info on the best time to visit the city [$\mathbf{3.008}$]\\
    \textbf{Retrieved candidate}&\texttt{Usr}&I need some information about Rome and where to go. [$-12.744$]\\
    \hline

    \textbf{Context}&\texttt{Wiz}&Hello how may I help you?\\
    &\texttt{Usr}&I'm here visiting, are there any good bars in the area?\\
    &\texttt{Wiz}&I'm sorry I can't help unless you clarify your question\\
    &\texttt{Usr}&I'm looking for good bars in the area, craft beer or wine?\\
    &\texttt{Wiz}&What area?\\
    &\texttt{Usr}&Downtown in the city\\
    &\texttt{Wiz}&What city?\\
    \textbf{Gold response}&\texttt{Usr}&Tampa\\
    \textbf{Generated candidate}&\texttt{Usr}&Montreal [$\textbf{4.48}$]\\
    \textbf{Retrieved candidate}&\texttt{Usr}&Ok thanks for the tip [$-10.164$]\\
    \hline

    \textbf{Context}&\texttt{Wiz}&Hello how may I help you?\\
    &\texttt{Usr}&I am a history buff. Do you know if there are any great historical sites\\
    &&to visit in Rome?\\
    &\texttt{Wiz}&Leonardo da Vinci was one of the greatest inventors of all time and they have\\
    &&a museum for him there\\
    &\texttt{Usr}&Oh that is so cool. I want to see that. Are there any other places to visit?\\
    &\texttt{Wiz}&Well if you want to go look at a bunch of old bones of dead people that have died\\
    &&throughout the ages you can always go to the catacombs\\
    \textbf{Gold response}&\texttt{Usr}&Okay that sounds like a great place to go also. Keep the ideas coming\\
    \textbf{Generated candidate}&\texttt{Usr}&Oh, that sounds fun. I'll do that. [$\mathbf{2.109}$]\\
    \textbf{Retrieved candidate}&\texttt{Usr}&I would like to know more about the air-balloon rides [$-12.116$]\\
    \hline

    \textbf{Context}&\texttt{Wiz}&Hello how may I help you?\\
    &\texttt{Usr}&Can you help me with information on Montreal?\\
    &\texttt{Wiz}&What would you like to know?\\
    \textbf{Gold response}&\texttt{Usr}&Are there any festivals there in the summer?\\
    \textbf{Generated candidate}&\texttt{Usr}&What is the tallest building there? [$\mathbf{1.905}$]\\
    \textbf{Retrieved candidate}&\texttt{Usr}&Ok, thank you. I was wondering where in Rome I can visit\\
    &&to learn about it's history. [$-12.161$]\\

    \bottomrule
    \end{tabularx}
    \caption[\metalwoz cross-task: \GRTr predictions with the most distant generated/retrieved candidates]{\GRTr predictions with the most distant generated/retrieved candidates~--- \metalwoz
    cross-task (in bold is the model's final response)}
    \label{tab:grtr_distant_candiates_metalwoz_cross}
\end{table}

\begin{table}[h!]
    \center
    \footnotesize
    \begin{tabularx}{\linewidth}{@{}rll@{}}\toprule

    \textbf{Context}&\texttt{Wiz}&Hello how may I help you?\\
    &\texttt{Usr}&I am looking for a place to dine. The restaurant should be expensive and should\\
    &&serve Chinese food.\\
    &\texttt{Wiz}&Would you like the restaurant in a specific place?\\
    \textbf{Gold response}&\texttt{Usr}&Not in any specific area please. I need to book a table for 3 at 14:00 on Saturday.\\
    \textbf{Generated candidate}&\texttt{Usr}&Do you have the address? [$\mathbf{-1.633}$]\\
    \textbf{Retrieved candidate}&\texttt{Usr}&I am looking for a restaurant in the centre of town, please. [$-1.635$]\\
    \hline

    \textbf{Context}&\texttt{Wiz}&Hello how may I help you?\\
    &\texttt{Usr}&I need a hotel in the south of the city, it is ok if the hotel doesn't has free parking.\\
    &\texttt{Wiz}&I have one hotel that meets your needs . It 's the Lensfield Hotel that offers free \\
    &&wifi and parking . They have a 3 star rating . Would you like more info on it ?\\
    &\texttt{Usr}&What is the price of the Lensfield?\\
    &\texttt{Wiz}&The Lensfield Hotel is expensive.\\
    &\texttt{Usr}&Is there a moderate place in the in the south that has free parking?\\
    &\texttt{Wiz}&No, no hotels meet your criteria.\\
    &\texttt{Usr}&Are there any moderately priced hotels that don't have free parking, but have wifi?\\
    &\texttt{Wiz}&We have 2 locations that are in the south , moderately priced , and have free wifi .\\
    &&They are both guesthouses , would one of those meet your needs ?\\
    \textbf{Gold response}&\texttt{Usr}&Yes, can you give me the address of your favourite one?\\
    \textbf{Generated candidate}&\texttt{Usr}&Can you book me for 2 nights for 2 nights starting from Wednesday. [$-0.013$]\\
    \textbf{Retrieved candidate}&\texttt{Usr}&Thanks for the information! [$\mathbf{-0.01}$]\\
    \hline

    \textbf{Context}&\texttt{Wiz}&Hello how may I help you?\\
    &\texttt{Usr}&I am hoping you might be able to help me find a place to stay last minute . I am on\\
    &&a bit of a budget so I need something moderate .\\
    &\texttt{Wiz}&I have three hotels available. The Ashley Hotel, Cityroomz, and Lovell Lodge.\\
    &&Do you need parking?\\
    &\texttt{Usr}&I need one in the centre and has free wifi.\\
    &\texttt{Wiz}&Then Cityroomz is just for you! It is located at sleeperz hotel, station road.\\
    &&Do you need more information or do you want to book a room ?\\
    \textbf{Gold response}&\texttt{Usr}&That is all that I need. Thank you so much.\\
    \textbf{Generated candidate}&\texttt{Usr}&I would like to book it for 4 people and 1 night starting on Saturday. [$\mathbf{-0.624}$]\\
    \textbf{Retrieved candidate}&\texttt{Usr}&Okay, I 'd like to book a room at the Gonville Hotel for 4 nights.\\
    &&There will be 6 people and we will be arriving on Saturday. [$\mathbf{-0.628}$]\\
    \hline

    \textbf{Context}&\texttt{Wiz}&Hello how may I help you?\\
    &\texttt{Usr}&I am looking for information about the Parkside Police Station.\\
    &\texttt{Wiz}&Sure, Parkside Police Station Telephone 01223358966, Address Parkside, \\
    &&Cambridge, Postcode CB11JG. Is there anything else I can do for you today ?\\
    \textbf{Gold response}&\texttt{Usr}&No, that's everything I needed, thanks.\\
    \textbf{Generated candidate}&\texttt{Usr}&No, that is all, thank you. [$\mathbf{2.783}$]\\
    \textbf{Retrieved candidate}&\texttt{Usr}&Yes, I 'll also need their phone number. [$2.779$]\\
    \hline

    \textbf{Context}&\texttt{Wiz}&Hello how may I help you?\\
    &\texttt{Usr}&I'm looking for a place to dine that serves British food. It should be in the \\
    &&moderate price range.\\
    &\texttt{Wiz}&How about The Oak Bistro? It's located in the Centre. Would you like more \\
    &&information or to book a table?\\
    \textbf{Gold response}&\texttt{Usr}&The Oak Bistro sounds fabulous. Can I get a table there on Sunday at 18:30?\\
    &&There will be 6 of us.\\
    \textbf{Generated candidate}&\texttt{Usr}&Yes, I need to book a table for 5 people at 15:30 on saturday. [$\mathbf{1.259}$]\\
    \textbf{Retrieved candidate}&\texttt{Usr}&Yes, I 'd like a table for a party of 8 at 11:15 on Thursday. [$1.255$]\\

    \bottomrule
    \end{tabularx}
    \caption[\multiwoz: \GRTr predictions with the closest generated/retrieved candidates]{\GRTr
    predictions with the closest generated/retrieved candidates~--- \multiwoz (in bold is the
    model's final response)}
    \label{tab:grtr_closest_candiates_multiwoz}
\end{table}

\begin{table}[h!]
    \center
    \footnotesize
    \begin{tabularx}{\linewidth}{@{}rll@{}}\toprule

    \textbf{Context}&\texttt{Wiz}&Hello how may I help you?\\
    &\texttt{Usr}&i am looking for a train. The train should leave on Monday and should arrive by 12:45.\\
    &\texttt{Wiz}&Where would you like to go?\\
    \textbf{Gold response}&\texttt{Usr}&I would like to depart from Cambridge, and travel to Bishops Stortford.\\
    \textbf{Generated candidate}&\texttt{Usr}&i would like to go to Cambridge. [$\mathbf{5.253}$]\\
    \textbf{Retrieved candidate}&\texttt{Usr}&I would like to leave after 09:00 on Sunday. [$-11.019$]\\
    \hline

    \textbf{Context}&\texttt{Wiz}&Hello how may I help you?\\
    &\texttt{Usr}&I'm looking for a restaurant in the west side that's in the moderate price range.\\
    &\texttt{Wiz}&I have three listings, one for british, indian, italian. Are any of those types of food\\
    &&you are looking for?\\
    &\texttt{Usr}&Are any of them available for 8 people at 19:00 on Wednesday?\\
    &\texttt{Wiz}&There are two restaurants available the Indian restaurant Meghna and the Italian \\
    &&restaurant Prezzo 8 people at 19:00 on Wednesday . Which one do you prefer ?\\
    \textbf{Gold response}&\texttt{Usr}&The Italian restaurant Prezzo sounds good.\\
    \textbf{Generated candidate}&\texttt{Usr}&Prezzo 8 will be great, how about Meghna? [$\mathbf{7.966}$]\\
    \textbf{Retrieved candidate}&\texttt{Usr}&Let's try the Brasserie. [$-6.55$]\\
    \hline

    \textbf{Context}&\texttt{Wiz}&Hello how may I help you?\\
    &\texttt{Usr}&Hello there. I am trying to locate a train for Monday.\\
    &\texttt{Wiz}&Sure. Where will you be going?\\
    &\texttt{Usr}&I am going to Cambridge.\\
    &\texttt{Wiz}&All right, and where would you like to depart from?\\
    &\texttt{Usr}&I 'll will be leaving from London Liverpool Street.\\
    &\texttt{Wiz}&I've got 10 trains that are on that route. Do you have a preferred departure or arrival time?\\
    &\texttt{Usr}&I need on that will leave London after 21:00.\\
    &\texttt{Wiz}&I have two trains that will leave after 21:00 . One leaves at 21:39 and one leaves \\
    &&at 23:39 . Which would you prefer ?\\
    \textbf{Gold response}&\texttt{Usr}&What time does the one that leaves at 21:39 arrive in Cambridge?\\
    &&Also, what is the train ID?\\
    \textbf{Generated candidate}&\texttt{Usr}&I 'd like to leave after 21:00 and leave the train that leaves after 21:00 [$-12.611$]\\
    \textbf{Retrieved candidate}&\texttt{Usr}&I don't actually need to book at the moment. I just need to know the departure time\\
    &&and travel time for this train, if you don't [$\mathbf{0.699}$]\\
    \hline

    \textbf{Context}&\texttt{Wiz}&Hello how may I help you?\\
    &\texttt{Usr}&I am looking for city centre north b and b\\
    &\texttt{Wiz}&I have found the guesthouse you were wanting. Would you like me to book this for you?\\
    \textbf{Gold response}&\texttt{Usr}&Yes, please book it for 1 person and for 5 nights starting Friday.\\
    \textbf{Generated candidate}&\texttt{Usr}&Yes please book it for 5 people. [$\mathbf{0.706}$]\\
    \textbf{Retrieved candidate}&\texttt{Usr}&Do any of those include free parking? [$-11.545$]\\
    \hline

    \textbf{Context}&\texttt{Wiz}&Hello how may I help you?\\
    &\texttt{Usr}&Hi, I'm looking for some train information. Could you tell me what trains leave\\
    &&on Wednesday for Norwich?\\
    &\texttt{Wiz}&There are 19 entries found. Where would you be coming from?\\
    &\texttt{Usr}&I 'll be departing from Cambridge and I need to arrive by 12:00.\\
    &\texttt{Wiz}&There is a train that arrives at 11:55. The trainID is TR9635.\\
    &&Would you like me to book that?\\
    \textbf{Gold response}&\texttt{Usr}&Sure, that sounds great.\\
    \textbf{Generated candidate}&\texttt{Usr}&Yes, that would be great. [$\mathbf{4.836}$]\\
    \textbf{Retrieved candidate}&\texttt{Usr}&Great can I get TR5173 booked for 3 people please? [$-7.307$]\\

    \bottomrule
    \end{tabularx}
    \caption[\multiwoz: \GRTr predictions with the most distant generated/retrieved candidates]{
    \GRTr predictions with the most distant generated/retrieved candidates~--- \multiwoz (in bold is
    the model's final response)}
    \label{tab:grtr_distant_candiates_multiwoz}
\end{table}

\begin{figure}[t]
    \centering
    \begin{tikzpicture}[thick,scale=0.79, every node/.style={transform shape}]
    \begin{axis}[
        axis on top,
        area style,
        minor tick num=1,
        scale only axis,
        ymajorgrids,
        enlarge y limits=false,
        ymin=0, ymax=1300,
        xlabel={Distance},
        ylabel={Count}
    ]
    \addplot+[ybar interval,mark=no] plot coordinates { 
        (1.00000e-03, 1234.)
        (1.61520e+00, 583.)
        (3.22940e+00, 250.)
        (4.84360e+00, 119.)
        (6.45780e+00, 61.)
        (8.07200e+00, 20.)
        (9.68620e+00, 19.)
        (1.13004e+01, 4.)
        (1.29146e+01, 4.)
        (1.45288e+01, 3.)
        (1.61430e+01, 0)
    };
    \end{axis}
\end{tikzpicture}
\caption{\metalwoz pure task: a histogram of pairwise distances between generated and retrieved \GRTr candidates}
\label{fig:gen_ret_distances_metalwoz_pure}
\end{figure}

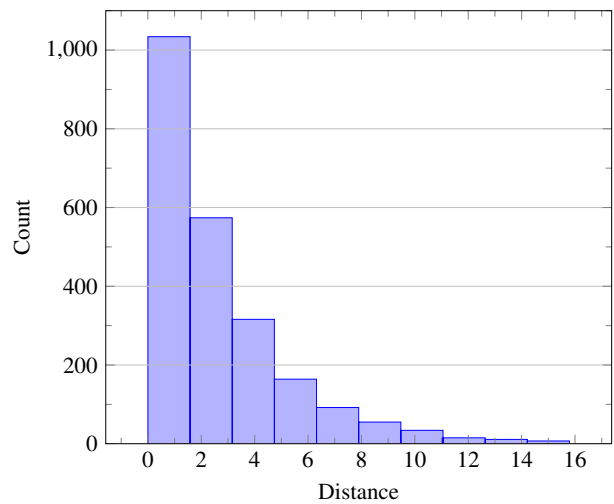
\begin{figure}[t]
    \centering
    \begin{tikzpicture}[thick,scale=0.79, every node/.style={transform shape}]
    \begin{axis}[
        axis on top,
        area style,
        minor tick num=1,
        scale only axis,
        ymajorgrids,
        enlarge y limits=false,
        ymin=0, ymax=1100,
        xlabel={Distance},
        ylabel={Count}
    ]
    \addplot+[ybar interval,mark=no] plot coordinates { 
        (1.00000e-03, 1034.)
        (1.58020e+00, 574.)
        (3.15940e+00, 316.)
        (4.73860e+00, 164.)
        (6.31780e+00, 92.)
        (7.89700e+00, 55.)
        (9.47620e+00, 34.)
        (1.10554e+01, 15.)
        (1.26346e+01, 11.)
        (1.42138e+01, 7.)
        (1.57930e+01, 0)
    };
    \end{axis}
\end{tikzpicture}
\caption{\metalwoz cross-task: a histogram of pairwise distances between generated and retrieved \GRTr candidates}
\label{fig:gen_ret_distances_metalwoz_cross}
\end{figure}

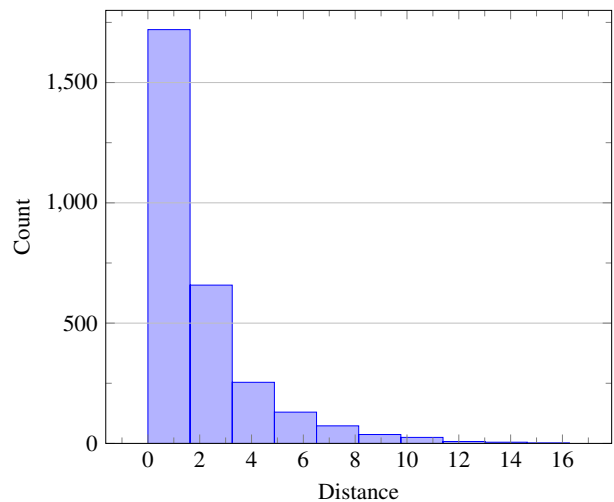
\begin{figure}[t]
    \centering
    \begin{tikzpicture}[thick,scale=0.79, every node/.style={transform shape}]
    \begin{axis}[
        axis on top,
        area style,
        minor tick num=1,
        scale only axis,
        ymajorgrids,
        enlarge y limits=false,
        ymin=0, ymax=1800,
        xlabel={Distance},
        ylabel={Count}
    ]
    \addplot+[ybar interval,mark=no] plot coordinates {
        (1.00000e-03, 1720.)
        (1.62810e+00, 658.)
        (3.25520e+00, 254.)
        (4.88230e+00, 130.)
        (6.50940e+00, 73.)
        (8.13650e+00, 37.)
        (9.76360e+00, 25.)
        (1.13907e+01, 8.)
        (1.30178e+01, 5.)
        (1.46449e+01, 2.)
        (1.62720e+01, 0)
    };
    \end{axis}
\end{tikzpicture}
\caption{\multiwoz: a histogram of pairwise distances between generated and retrieved \GRTr candidates}
\label{fig:gen_ret_distances_multiwoz}
\end{figure}


\chapter{Disfluency Detection~--- Supplementary Material} 

\label{AppendixC} 

\lhead{Appendix C. \emph{Disfluency Detection}} 

\begin{table}[h]
    \centering
    \begin{tabular}{ll}
    \toprule
    \textbf{Parameter}&\textbf{Value}\\
    \midrule
    Optimiser&stochastic gradient descent\\
    Loss function&weighted cross-entropy\\
    Vocabulary size&6157\\
    Embedding size&128\\
    \MLP layer sizes&[128]\\
    Learning rate&0.01\\
    Learning rate decay&0.9\\
    Batch size&32\\
    $\alpha$&0.1\\
    $\lambda$&0.001\\
    $\gamma$&1.05\\
    \bottomrule
    \end{tabular}
    \caption{Multi-task \LSTM training setup}
    \label{tab:lstm_setup}
\end{table}

\begin{table}[h]
    \centering
    \begin{tabular}{llS}
    \toprule
    \textbf{Label type}&\textbf{Label}&\textbf{Frequency}\\
    \midrule
    Fluent token&\texttt{<f/>}&574771\\
    Edit token&\texttt{<e/>}&45729\\
    Single-token substitution&\texttt{<rm-\textit{\{1-8\}}/><rpEndSub/>}&13003\\
    Single-token deletion&\texttt{<rm-\textit{\{1-8\}}/><rpEndDel/>}&1011\\
    Multi-token substitution start&\texttt{<rm-\textit{\{1-8\}}/><rpMid/>}&6976\\
    Multi-token substitution end&\texttt{<rpEndSub>}&6818\\
    \bottomrule
    \end{tabular}
    \caption{\SWDA labels}
    \label{tab:swda_labels}
\end{table}

\chapter{Data-Efficiency in Social Dialogue~--- Supplementary Material} 

\label{AppendixD} 

\lhead{Appendix D. \emph{Social Dialogue}} 

\begin{table}[ht!]
    \centering
    \begin{tabular}{ll}\toprule
    \textbf{Parameter}&\textbf{Value}\\\midrule
    vocabulary size&60000\\
    learning rate&0.01\\
    embedding size&256\\
    \RNN cell type&GRU\\
    optimiser&Adagrad\\
    loss function&\MSE\\
    dropout ratio&0.4\\
    predictor layer sizes&$[256]$ (length), $[128, 32, 32]$ (rating)\\
    batch size&8\\
    max utterance length&30 tokens\\
    \bottomrule
    \end{tabular}
    \caption{Neural rankers training setup}
    \label{tab:neural_ranker_config}
\end{table}

\begin{table}[h!]
    \begin{center}
    \begin{tabular}{ll}\toprule
    \textbf{Parameter}&\textbf{Value}\\\midrule
    feature bit length&16\\
    loss function&squared\\
    features&context ngrams, response ngrams,\\
    &turn number, bot, utterance length,\\
    &handcrafted features\\
    quadratic features&response ngrams $\times$ response ngrams,\\
    &context ngrams $\times$ response ngrams,\\
    &bot name $\times$ response ngrams,\\
    &bot name $\times$ context ngrams,\\
    &bot name $\times$ handcrafted features\\
    cubic features&bot name $\times$ context ngrams $\times$ response ngrams\\
    holdout set&off\\
    passes number&1\\
    \bottomrule
    \end{tabular}
    \end{center}
    \caption{VowpalWabbit ranker training setup}
    \label{tab:vw_config}
\end{table}

\addtocontents{toc}{\vspace{2em}} 

\backmatter


\label{Bibliography}

\lhead{\emph{Bibliography}} 

\bibliographystyle{apalike} 

\bibliography{intro,ch2,ch3,ch4,ch5,ch6,ch7,ch8} 

\end{document}